\documentclass[11pt]{llncs}
\usepackage[margin=0.5cm, includefoot, footskip=16pt]{geometry}
\usepackage[english]{babel}
\usepackage[utf8]{inputenc}
\usepackage{url}
\usepackage{amsmath, amsfonts}
\DeclareMathOperator*{\argmax}{argmax} 
\usepackage{eqparbox}
\usepackage[T1]{fontenc}
\usepackage{dsfont}
\usepackage{graphicx}
\usepackage{subcaption}
\usepackage{color}
\usepackage{xcolor}
\usepackage{float}
\usepackage{multirow}
\usepackage{diagbox}
\usepackage[switch]{lineno}
\usepackage{makecell}
\usepackage[disable]{todonotes}
\usepackage{subfiles} 
\usepackage{lmodern}
\usepackage{algpseudocode}
\usepackage[]{algorithm}

\usepackage[font=small,labelfont=bf]{caption}
\usepackage{wrapfig}

\newcommand{\todocwyf}[1]{\todo{@CW,YF: {\small #1}}}

\newcommand{\figref}[1]{Fig.~\ref{#1}}
\newcommand{\secref}[1]{Sec.~\ref{#1}}

\DeclareRobustCommand
      \Compactcdots{\mathinner{\cdot\mkern-3mu\cdot\mkern-3mu\cdot}}

\usepackage{enumitem}
\setitemize{noitemsep,topsep=0pt,parsep=0pt,partopsep=0pt}
\setenumerate{noitemsep,topsep=0pt,parsep=0pt,partopsep=0pt}

\newcommand{\setof}[1]{\ensuremath{ \left\{ #1 \right\}}}
\newcommand{\defas}[0]{\ensuremath{\stackrel{\mathrm{\scriptscriptstyle{def}}}{=}}}
\newcommand{\nat}[0]{\ensuremath{\mathbb{N}}}
\newcommand{\natpos}[0]{\ensuremath{\mathbb{N} \setminus \{ 0 \}}}
\newcommand{\real}[0]{\ensuremath{\mathbb{R}}}
\newcommand{\intinterval}[2]{\ensuremath{[{#1} \Compactcdots {#2}]}}
\newcommand{\realinterval}[2]{[{#1}, {#2}]}
\newcommand{\vvector}[1]{\mathbf{#1}}
\newcommand{\boxcoveredspace}[1]{\ensuremath{\mathit{CovB(#1)}}}

\newcommand{\boxcoverage}[1]{\ensuremath{\mathit{coverage}_X\left( #1 \right)}}
\title{Customizable Reference Runtime Monitoring of Neural Networks using Resolution Boxes\thanks{This paper is supported by the European Union’s Horizon 2020 research and innovation programme under grant agreement No. 956123 - FOCETA and the French national program ``Programme Investissements d’Avenir IRT Nanoelec" (ANR-10-AIRT-05).}}
\author{Changshun Wu\inst{1}, Yli\`{e}s Falcone\inst{2}, and Saddek Bensalem\inst{1}}
\institute{
Univ. Grenoble Alpes, CNRS, Grenoble INP, Verimag,
38000 Grenoble, France\\
\and
Univ. Grenoble Alpes, Inria, CNRS, Grenoble INP, LIG,
38000 Grenoble, France\\
}
\date{}

\begin{document}
\maketitle
\begin{abstract}
Classification neural networks fail to detect inputs that do not fall inside the classes they have been trained for.
Runtime monitoring techniques on the neuron activation pattern can be used to detect such inputs.
We present an approach for the runtime verification of classification systems via data abstraction.
Data abstraction relies on the notion of box with a resolution.
Box-based abstraction consists in representing a set of values by its minimal and maximal values in each dimension.
We augment boxes with a notion of resolution; this allows to define the notion of clustering coverage, which is intuitively a quantitative metric over boxes that indicates the quality of the abstraction.
This allows studying the effect of different clustering parameters on the constructed boxes and estimating an interval of sub-optimal parameters.
Moreover, we show how to automatically construct monitors that make use of both the correct and incorrect behaviors of a classification system.
This allows checking the size of the monitor abstractions and analysing the separability of the network.
Monitors are obtained by combining the sub-monitors of each class of the system placed at some selected layers.
Our experiments demonstrate the effectiveness of our clustering coverage estimation and show how to assess the effectiveness and precision of monitors according to the selected clustering parameter and the chosen monitored layers.
%
%
\end{abstract}

\section{Introduction}

During the past decade, the growth of computing power and the abundance of data from all kinds of sources have changed many fields of science and technology, from biology and medicine to economy and social sciences.
The so-called (new) artificial intelligence (AI)\footnote{AI is used in a broad sense, to include many related fields and buzzwords, such as machine learning, big data, data science, and so on.} revolution is also reshaping our entire industry, our society, and our day-to-day lives.
In fact, it is quite reasonable to say that after a long ``winter sleep", AI is once again flourishing thanks to recent advances in machine learning (ML) algorithms.
Today, many systems already incorporate an increasing amount of intelligence and autonomy (face recognition, industrial robotics, data processing, etc.).
More specifically, the AI System technologies are expected to bring large-scale improvements through new products and services across a myriad of applications ranging from healthcare to logistics through manufacturing, transport and more.

Today's AI-based systems rely on so-called learning-enabled components (LECs)~\cite{Neema10}.
For example, a learning-enabled autonomous system may rely on a LEC performing visual perception and recognition.
To trust the AI system we must also trust its LECs.
However, such systems (especially deep recurring neural networks) are highly unpredictable~\cite{SzegedyZSBEGF13}.
Prototyping such systems may seem quick and easy, but prototypes are not safe and incur a high cost, referred to as technical debt~\cite{SculleyHGDPECYC15}.
These problems are not theoretical. 
Traffic violations, accidents, and even human casualties have resulted from faults of LECs.
It is therefore correctly and widely recognized that AI-based systems are not often trustworthy~\cite{ResearchChallenges}.
The safety-critical nature of such systems involving LECs raises the need for formal methods~\cite{SeshiaS16}.
In particular, how do we systematically find bugs in such systems?
We believe that we need formal verification techniques, in order to i) model AI systems, and in particular their LECs; ii) specify what properties these systems have, or should have; iii) reason about such properties and ultimately verify that they are satisfied. Indeed, these research areas are currently emerging, see for instance~\cite{HuangKWW17}.

Unfortunately, traditional verification techniques are not usable for such purpose for several reasons.
First, LECs are generally not specified; they are essentially obtained from examples and the explainability of their synthesis and implementation is not well understood.
Second, such components take as input complex data, typically highly dimensional vectors with immense state space.
Third, the abnormal behaviors are often discovered when the system is in operation. Some of the essential properties of such systems cannot be guaranteed statically at design time.
These properties should be enforced at runtime by using monitoring and control-based techniques.

Verification techniques for data-driven learning components have been actively developed in the past 3-4 years~\cite{DuttaJST18,Ehlers17,gehr2018ai2,KatzBDJK17,LomuscioM17,NarodytskaKRSW18,PulinaT10,RuanHK18,WengZCSHDBD18,LiLYCHZ19}.
However, their scalability is required to be significantly improved to meet industrial needs.
Safety-critical systems involving LECs like self-driving cars, extensively use data-based techniques, for which we do not have so far a theory allowing behavioural predictability.
To favor scalability, some research efforts in the last few years have focused on using dynamic verification techniques such as testing~\cite{pei2017deepxplore,sun2018testing,tian2018deeptest,wicker2018feature,xie2018coverage} and runtime verification~\cite{cheng2019runtime,henzinger2019outside,lukina2020into,cheng2020provably}.

In this paper, we contribute to the research efforts on runtime verification for learning-enabled components.
In designing a runtime verification approach for leaning-enabled components, one of the main challenges is the absence of a clear behavioral specification of the component to be checked at runtime.
This implies a fundamental shift in the paradigm: to move from behavior-based verification to data-based verification.
Henceforth, existing approaches essentially proceed in two steps.
First, one characterizes the seen data (e.g., via probability distribution) or the compact patterns generated from them (via Boolean or high-dimension geometry abstraction).
Then, one uses the established characterizations as references to check the system decisions on new inputs by checking the similarity between the produced patterns at runtime and the reference patterns.
The above pioneering approaches can be split in two categories depending on the abstractions used to record the reference patterns: Boolean abstraction~\cite{cheng2019runtime,cheng2020provably} and geometrical-shape abstraction~\cite{henzinger2019outside,lukina2020into,cheng2020provably}.
\paragraph{Approach and contributions.}
%
We focus on the monitoring approaches based on geometrical shape abstraction.
In particular, we extend the work in~\cite{henzinger2019outside} to address some of its limitations.
The approach in~\cite{henzinger2019outside} only leverages the good reference behavior of the network.
Ignoring bad reference behavior discards some useful information.
The geometrical shape abstractions of the good and bad reference behaviour may intersect or not.
Thus, a new generated pattern can belong to both abstractions.
Hence, the verdicts produced by~\cite{henzinger2019outside} based only on behaviors may be partial.
We use both the good and bad network behaviours correct and incorrect decisions respectively) as references to build box abstractions.
Using these references, at an abstract level, a runtime monitor assigns verdicts to a new input as follows:
\begin{itemize}
\item
if the input generates patterns (e.g., output values at close-to-output layers) that fall only within the good references, then the input is accepted;
\item
if the input generates patterns captured by both the good and bad references, it marks the input as uncertain;
\item otherwise, the input is rejected.
\end{itemize}
Introducing uncertainty verdicts allows identifying suspicious regions when the abstractions of good and bad references overlap.
By reducing the abstraction size, one may remove the overlapping regions and obtain suitable abstraction size.
Otherwise, it indicates that the network the network does not have a good separability: the positive and negative samples are tangled.
This provides feedback to the network designer.
It also permits comparing the regions of patterns and thus enables the study of the relationship between good and bad behavior patterns of the network.
Moreover, we introduce the notion of \emph{box with resolution} which consists intuitively in tiling the space of a box.
By doing this, one can measure the clustering coverage which is an indicator of the coarseness of the abstraction.
Using the clustering coverage, we can assess the precision of the boxes.
In choosing the box resolution, there is a tradeoff between the precision of the related abstractions and the related overhead (i.e., augmenting the precision augments the overhead).
To control the precision, we discuss how to tune the clustering parameters by observing the clustering coverage and the number of uncertainties.
We achieve better precision and recall in all cases.

\paragraph{Paper organization.}
The rest of this paper is structured as follows.
Section~\ref{sec:preliminaries} introduces preliminary concepts and notation.
Section~\ref{sec:boxes_resolution} defines boxes with a resolution.
Section~\ref{sec:framework} presents our monitoring framework.
Section~\ref{sec:experiments} presents the results of our experimental evaluations.
Section~\ref{sec:rw} positions our work and details the improvements over state-of-the-art approaches.
Section~\ref{sec:conclusion} concludes.

\section{Preliminaries}
\label{sec:preliminaries}
%
For a set $E$, $|E|$ denotes the cardinality of $E$.
Let $\nat$ and $\real$ be the sets of natural and real numbers, respectively.
For $x \in \real$, $\lceil x \rceil$ denotes the ceiling of $x$, that is the least integer greater than $x$.
To refer to intervals of integers, we use $\intinterval{a}{b}$ with $a, b \in \nat$ and $a \leq b$.
To refer to intervals of real numbers, we use $\realinterval{a}{b}$ with $a, b \in \real \cup \{{-\infty}, \infty\}$ and if $a, b \in \real$, then $a \leq b$.
For $n \in \natpos$, $\real^n \defas \underbrace{ \real \times \cdots \times \real}_{n\ \text{times}}$ is the space of real coordinates of dimension $n$ and its elements are called $n$-dimensional vectors.
We use $\vvector{x} = (x_1, \ldots, x_n)$ to denote an $n$-dimensional vector and $\theta_i: \mathbb{R}^n \rightarrow \mathbb{R}$ the projection function on the $i$-th dimension for $i \in \intinterval{1}{n}$, i.e., $\theta_i(\vvector{x}) = x_i$.
%
\subsection{Feedforward Neural Networks}
%
A neuron is an elementary unit mathematical function.
A \emph{(forward) neural network} (NN) is a sequential structure of $L \in \natpos$ layers, where, for $i \in \intinterval{1}{L}$, the $i$-th layer comprises $d_{i}$ neurons and implements a function $g^{(i)} : \mathbb{R}^{d_{i-1}} \rightarrow \mathbb{R}^{d_{i}}$.
In a network, the inputs of neurons at layer $i$ comprise (1) the outputs of neurons at layer $(i-1)$ and (2) a bias.
Moreover, the outputs of neurons at layer $i$ are inputs for neurons at layer $i+1$.

Given a network input $\vvector{x} \in \real^{d_{0}}$, the output at the $i$-th layer is computed by the function composition $f^{(i)}(\vvector{x}) \defas g^{(i)}(\cdots g^{(2)}(g^{(1)}(\vvector{x})))$.
For networks used for classification tasks, aka classification networks, \emph{the decision} $\mathsf{dec}(\vvector{x})$ of classifying input $\vvector{x}$ into a certain class is given by the index of the neuron of the output layer whose value is maximum, i.e., $\mathsf{dec}(\vvector{x}) \defas \argmax_{1 \leq i \leq d_{L}} \theta_i(f^{(L)}(\vvector{x}))$.
In this paper, we only consider well-trained networks, i.e., networks in which the weights and bias related to neurons are fixed.
The method developed in what follows is applicable to networks with various neuron activation functions.
%
\subsection{Abstraction for Neural Networks}
%
To construct runtime monitors, we need to use and represent the high-level features from NNs, which are large sets of vectors.
For this, the classical approach in verification is to use \emph{abstraction} techniques (e.g., abstract interpretation~\cite{cousot1992abstract}) to over-approximate a given set of finite vectors into a set of mathematical constraints.
Abstraction techniques have recently also been used for NNs, e.g., \emph{abstract interpretation} and \emph{intervals bound} are used to verify the network's safety and robustness~\cite{gehr2018ai2,gowal2018effectiveness}, along with \emph{boolean}~\cite{cheng2019runtime} and \emph{box}~\cite{henzinger2019outside} abstraction for runtime monitoring of NNs.

As runtime monitoring of NNs require intensive usage of the high-level features, affordable computational complexity of \emph{storage}, \emph{construction}, \emph{membership query}, and \emph{coarseness control} is paramount.
While there are other candidate abstractions using different geometry shapes (such as \emph{zonotope}, and \emph{polyhedra}), our preliminary study of these alternative shapes along with the complexity considerations, led to choose to  follow~\cite{henzinger2019outside} and use box abstraction for the purpose of this paper.
%
\subsection{Box Abstraction~\cite{henzinger2019outside}}
%
We briefly review box abstraction.
A box is essentially a set of contiguous $n$-dimensional vectors constrained by real intervals.
\begin{definition}[(tight) Box abstraction]
For a set $X \subseteq \real^n$ of $n$-dimensional vectors, its (tight) \emph{box abstraction} is defined as $B(X) \defas \setof{ (x_1, \ldots,x_n) \in \real^n \mid \bigwedge\limits_{i\in \intinterval{1}{n}} a_i \leq x_i \leq b_i }$, where $a_i = \mathsf{min}( \setof{\theta_i(\vvector{x})})$ and $b_i = \mathsf{max}(\setof{\theta_i(\vvector{x})})$, for $\vvector{x} \in X$ and $i \in\intinterval{1}{n}$.
\end{definition}
The box is said to be tight because the interval built on each dimension is the minimum one containing all the values of the given points on the corresponding dimension.
We only consider tight boxes and refer to them as boxes.

A box is equivalently encoded as the list of intervals of its bounds on each dimension, i.e., and we equivalently note $B(X) = \big[ \realinterval{a_1}{b_1}, \cdots, \realinterval{a_n}{b_n} \big]$.
Moreover, given two n-dimensional boxes $B(X')=[ \realinterval{a'_1}{b'_1}, \ldots, \realinterval{a'_n}{b'_n}]$ and $B(X) = [\realinterval{a_1}{b_1}, \cdots, \realinterval{a_n}{b_n}]$, $B(X')$ is said to be a \emph{sub-box} of $B(X)$ if the vectors of $B(X')$ are all in $B(X)$, i.e., if $\forall i \in \intinterval{1}{n}: a_i \leq a'_i \wedge b'_i \leq b_i$.
For two datasets $X'$ and $X$, if $X' \subseteq X$, then $B(X')$ is a sub-box of $B(X)$.
Furthermore, the \emph{emptiness check} of a box as well as the \emph{intersection} between two $n$-dimensional boxes is a box and can be easily computed using their lower and upper bounds on each dimension.
\begin{example}[(tight) Box abstraction]
\label{ex:boxDefinition}
Considering a set of vectors (dataset) $X = \{ (0.1, 0.5),$ $ (0.1, 1.0), (0.2, 0.8), (0.6, 0.2), (1.0, 0.3) \}$, its box abstraction $B(X)$ is the set
\[
\setof{ (x_1, x_2) \in \real^2 \mid x_1 \in \realinterval{0.1}{1.0}, x_2 \in \realinterval{0.2}{1.0} },
\]
which can be encoded as $[\realinterval{0.1}{1.0}, \realinterval{0.2}{1.0}]$.
\end{example}
\begin{remark}
A box abstracting a set of $n$-dimensional vectors can be represented by $2\times n$ bounds.
The complexity of building a box for a set of $n$-dimensional vectors of cardinal $m$ is $O(m \times n)$, while the membership query of a vector in a box is $O(n)$.
The control of the coarseness or size of a box can be easily realized by enlarging or reducing the bounds.
\end{remark}
%
\subsection{Clustering}
%
\begin{figure}[t]
  \centering
  \begin{subfigure}[b]{.49\linewidth}
  	\centering
  	\includegraphics[width=0.65\textwidth]{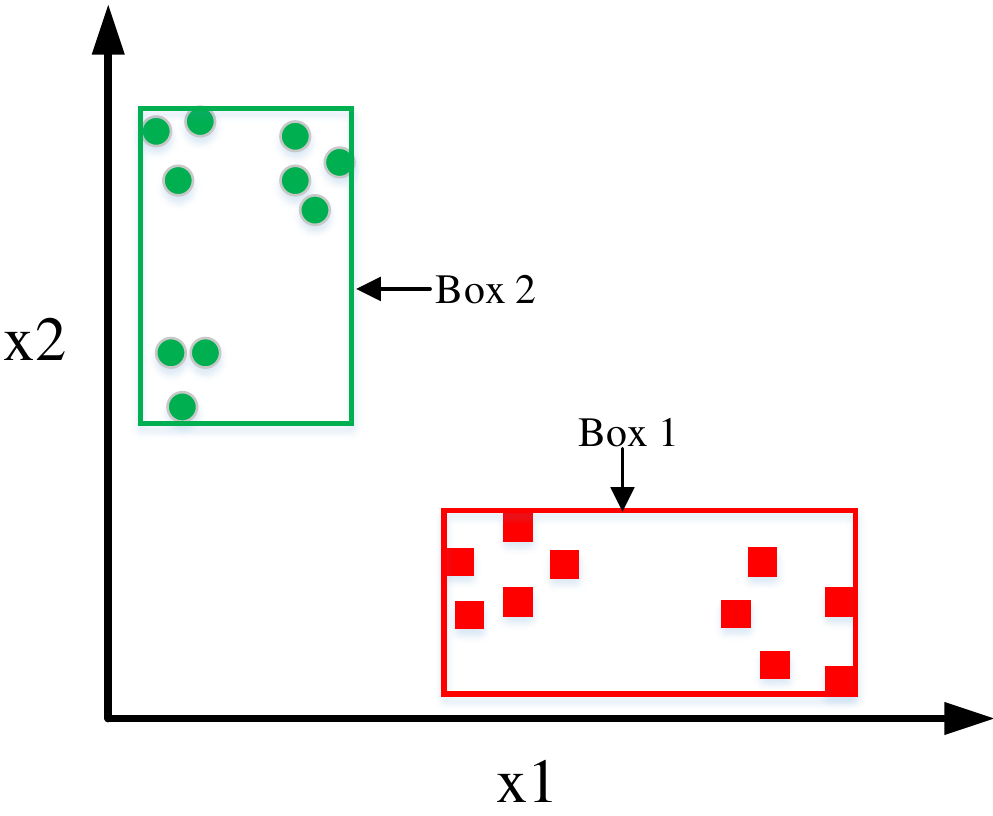}
  	\caption{Without clustering.}
  	\label{fig:boxClustering:without}
  \end{subfigure}
  \begin{subfigure}[b]{.49\linewidth}
   \centering
    \includegraphics[width=0.65\textwidth]{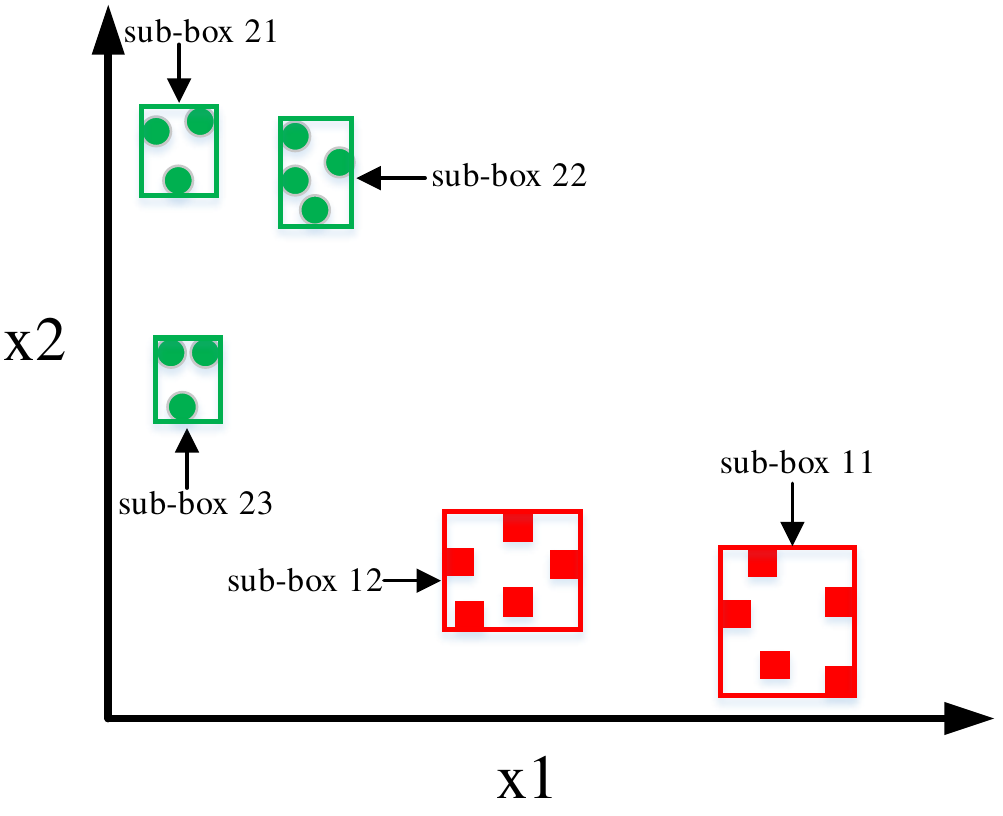}
    \caption{With clustering.}
    \label{fig:boxClustering:with}
  \end{subfigure}
  \caption{Box abstractions for two sets of points (the green and red one) built without and with clustering.}
  \label{fig:boxClustering}
\end{figure}
In certain cases, the box abstraction of a set $X$ is such that most of the elements of $X$ end up close to the boundaries of the box (there is ``white space").
For instance, this is the case with the two boxes in \figref{fig:boxClustering:without}.
Consequently, more elements not originally present in the set end up in the box abstraction.
This situation is not desirable for our monitoring purpose.

To remedy this, one can apply a clustering algorithm to determine a partition of the set based on an appropriate similarity measure such as distance (e.g., $k$-means clustering~\cite{lloyd1982least}), density (DBSCAN~\cite{schubert2017dbscan}), connectivity (hierarchical clustering~\cite{murtagh2012algorithms}), etc.
Finding the best clustering algorithms for monitoring remains an open question out of the scope of this paper.
We choose adopt $k$-means even though other clustering algorithms are certainly also usable  with our monitoring framework.

By grouping the points that are close to each others in a cluster and computing the box abstractions on each cluster separately permits abstracting these sets more precisely as the union of the boxes computed for each cluster.
Consequently, by using clustering first and then abstraction, we end up with a set of boxes based on the partitions determined by the clustering algorithms.
Clustering and its benefit in terms of precision are illustrated in \figref{fig:boxClustering}b.

\section{Boxes with a Resolution}
\label{sec:boxes_resolution}

Applying clustering algorithm before computing abstractions for a set of points was first proposed in~\cite{henzinger2019outside}.
Clustering is essential to construct the monitor therein.
The clustering parameters directly influence the monitor performance.
However, the effect of using clustering algorithms in monitoring nor the relationship between the clustering parameters and the monitor performance were not discussed and studied in~\cite{henzinger2019outside}.
Moreover, going back to \figref{fig:boxClustering}, with boxes, we observe that it is difficult to \emph{quantify} the precision of the abstraction provided by boxes.

To address the aforementioned problems, we introduce the notion of \emph{box with a resolution} providing insights into the relationship between the clustering parameters and the performance of the monitor.
Specifically, a box with a resolution computed for a set of points is a box divided into a certain number of cells of the same size.
Moreover, the ratio between the number of cells covered by the set of boxes computed for the partition of the set of points (obtained by clustering) to the total number of cells can be used to measure the relative coarseness of the box abstractions computed with and without clustering.
We refer to this metric as the \emph{clustering coverage}.
The purpose of this section is to compute efficiently (an approximation) of the clustering coverage using box abstraction.

In the following, we consider a set $X$ of $n$-dimensional vectors, its $n$-dimensional box (abstraction) $B(X) = [\realinterval{a_1}{b_1}, \cdots, \realinterval{a_n}{b_n}]$, its \emph{partition} $\pi(X) = \setof{X^1,\ldots, X^k }$ obtained by a clustering algorithm, and the set of boxes computed for the partitioned dataset $ \mathcal{B}_X = \setof{B^1, \ldots, B^k}$ where $B^i = B(X^i)$ for $i \in \intinterval{1}{k}$.
We refer informally to $B(X)$ as the global box and to $B^1, \ldots, B^k$ as the local boxes.
Our goal is to measure the relative sizes of the local boxes to the global one.
When computing the local boxes, we observe that they often exist on different dimensions, i.e., some of their intervals is of length $0$.
\paragraph{Box covered space.}
We start by introducing the notion of \emph{box covered space}.
Intuitively, the box covered space associated with $B(X)$ is a box of (possibly) lower dimension used for measuring the spaced covered.
\begin{definition}[Box covered space]
The \emph{covered space} of $B(X)$ is defined as $\boxcoveredspace{X} \defas [\realinterval{a_i}{b_i} \text{ if } a_i \neq b_i \mid \realinterval{a_i}{b_i} \in B(X)]$.
\end{definition}
That is, the covered space of $B(X)$ is the box encoded as the list of non-zero-length intervals of $B(X)$.
\begin{example}[Box covered space]
Consider dataset $X = \setof{(1.5, 2.0, 1), (1.8, 2.3, 1)}$ and its box $B(X) = [ \realinterval{1.5}{1.8}, \realinterval{2.0}{2.3},$ $\realinterval{1}{1}]$.
Its box covered space is $\boxcoveredspace{X} = [\realinterval{1.5}{1.8},$ $\realinterval{2.0}{2.3}]$.
\end{example}
\paragraph{Adding resolution.}
Let $|\boxcoveredspace{X}|$ denote the length of $\boxcoveredspace{X}$ -- it is the number of dimensions on which the box abstraction of $X$ exists.
We divide each interval/dimension of $\boxcoveredspace{X}$ into $|X|$ subintervals of the same length.
Consequently, the space enclosed in $\boxcoveredspace{X}$ is equally divided into $|X|^{|\boxcoveredspace{X}|}$ subspaces.
Each such a subspace is called a \emph{box cell}.
Each box cell can be encoded (as a box) by $|\boxcoveredspace{X}|$ intervals and can be indexed by coordinates of size $|\boxcoveredspace{X}|$.
We can hence reuse the notions and notations related to boxes.
We use $\mathcal{C}_X$ to denote the set of cells obtained from $\boxcoveredspace{X}$.
\newcommand{\covcell}[1]{\mathit{CovCell}\left(#1\right)}
\begin{definition}[Covered cells]
A box cell $c \in \mathcal{C}_X$ is said to be \emph{covered} by a box $b \in \mathcal{B}_X$ if $c\ \cap\ b \neq \emptyset$.
The set of covered cells by a box $b$, $\setof{c \in \mathcal{C}_X \mid c \cap b \neq \emptyset}$, is denoted by $\covcell{b}$.
\end{definition}
Using the intervals defining a local box $b$ and the global box $B(X)$, we can compute the number of covered cells in $b \in \mathcal{B}$.
\begin{property}[Number of covered cells]
For $b \in \mathcal{B}$, let $b = [ \realinterval{a'_1}{b'_1}, \ldots, \realinterval{a'_n}{b'_n}]$, we have:
$|\covcell{b}| = \prod_{i=1}^{n} n_i$, with:
\[
\begin{array}{ll}
	n_i & =
	\left\{
	\begin{array}{ll}
	\left\lceil \frac{|X| \times (a'_i - a_i)}{(b_i - a_i)} \right\rceil - \left\lceil \frac{|X| \times (b'_i - a_i)}{(b_i - a_i)} \right\rceil + 1 & \text{if } b'_i \neq a'_i, \\
	1 & \text{otherwise}.
	\end{array}
	\right.
\end{array}
\]	
\end{property}
With the number of covered cells, we can define the coverage of a sub-box to a (larger) box.
\begin{definition}[Sub-box coverage]
\label{def:coverage}
The sub-box coverage of $b \in \mathcal{B}_X$ to $B(X)$ is defined as: $\boxcoverage{b} = \frac{|\covcell{b}|}{|X|^{|\boxcoveredspace{X}|}}$.
\end{definition}
The sub-box coverage of $b$ to $B(X)$ is the ratio of the number of cells covered by $b$ to the total number of cells in $\boxcoveredspace{X}$.
\begin{example}[Sub-box coverage]
Considering the set $X$ in Example~\ref{ex:boxDefinition} and its subset $X^1 = \setof{ (0.1, 0.5), (0.1, 1.0), (0.2, 0.8)}$ along with the corresponding covered space $\boxcoveredspace{X} = [ \realinterval{0.1}{1.0}, \realinterval{0.2}{1.0} ]$, the coverage of sub-box $B(X^1) = [ \realinterval{0.1}{0.2}, \realinterval{0.5}{1.0}]$ is:
\[
\boxcoverage{B(X^1)} = \frac{ |\covcell{B(X^1)}|}{|X|^2} = 4/25.
\]
\end{example}
\paragraph{Clustering coverage estimation.}
We extend the notion of sub-box coverage to set $\mathcal{B}_X$ of local boxes.
We note it $\boxcoverage{\mathcal{B}_X}$ and define it as the ratio between the total number of cells covered by the union of boxes in $\mathcal{B}_X$ to the whole number of cells in $\boxcoveredspace{X}$: $\boxcoverage{\mathcal{B}_X} = \boxcoverage{\cup_{b \in \mathcal{B}_X}}$.

The exact value of $\boxcoverage{\mathcal{B}_X}$ can, in theory, be easily computed.
However, in practice the computation may be very expensive with high dimensionality due to the large number of cells and intersections between boxes.

Henceforth, we only estimate the coverage value by only considering the pair-wise intersections of boxes.
This is a reasonable approximation because the set of sub-boxes considered is built from a partition of the input dataset after applying a clustering algorithm: in principle a good clustering implies few elements in the intersections between the clusters, especially if the number of clusters is important.

For $\mathcal{B}_X$, we define $\mathcal{B}_X^{\rm int} = \setof{ b_i \cap b_j \mid i \in \intinterval{1}{k-1}, j \in \intinterval{i+1}{k} }$ as the set of pair-wise intersections of boxes in $\mathcal{B}_X$.
\begin{proposition}[Clustering coverage estimation]
The clustering coverage is lower and upper bounded by $r_{l}$ and $r_{u}$, respectively, where:
\[
\begin{array}{ccc}
r_{u} = \sum_{b \in \mathcal{B}_X}\limits \boxcoverage{b}
&
\text{ and }
& r_{l} = r_{u} - \sum_{b \in \mathcal{B}_X^{\rm int}}\limits \boxcoverage{b}.
\end{array}
\]
\end{proposition}
\begin{example}[Clustering coverage]
Considering the set $X$ in Example~\ref{ex:boxDefinition}, we assume that it is partitioned into two clusters $X^1 = \setof{(0.1, 0.5), (0.1, 1.0), (0.2, 0.8)}$ and $X^2 = \{(0.6, 0.2),$ $(1.0, 0.3)\}$ along with two smaller boxes $B(X^1) = [\realinterval{0.1}{0.2}, \realinterval{0.5}{1.0} ]$ and $B(X^2) = [\realinterval{0.6}{1.0}, \realinterval{0.2}{0.3}]$, then the clustering coverage is $0.28$ because:
\[
r_{l} = r_{u} = \frac{ |\covcell{ B(X^1 }| + |\covcell{ B(X^2}| }{ 5^2 } = \frac{4 + 3}{25} = 0.28.
\]
\end{example}
\begin{remark}
The clustering coverage allows to better assess the amount of ``blank space" between the points in a given set.
The lower the value is, the more blank space there exits.
On the one hand, obtaining smaller clusters (equivalence classes) before applying abstraction technique augments the precision of the abstraction.
On the other hand, having too small clusters has drawbacks: (1) it augments the computational overhead, and (2) it induces some form of overfitting for the monitor.
We further demonstrate the effect of clustering in our experiments.
\end{remark}

\section{Runtime Monitoring of NNs using Resolution Boxes}
\label{sec:framework}

The frameworks defined in~\cite{cheng2019runtime,henzinger2019outside} \emph{only} utilize the high-level features obtained from the layers close to the output layer (aka, close-to-output layers).
Moreover, to build the monitor, \cite{cheng2019runtime,henzinger2019outside} only consider ``good behaviors", i.e., the features of correctly classified inputs as reference.
Our framework inspires from \cite{cheng2019runtime,henzinger2019outside} but additionally makes use of the ``bad behaviors", i.e. the ones of misclassified inputs.
This has two advantages.
First, it allows refining the monitor output by adding a notion of \emph{uncertainty} to the previous monitor verdicts, i.e., \emph{accept} and \emph{reject}.
Second, when the monitor produces "uncertainty" as output, it avoids false negatives.

\begin{figure}[t]
  \centering
  \includegraphics[width=0.9\textwidth]{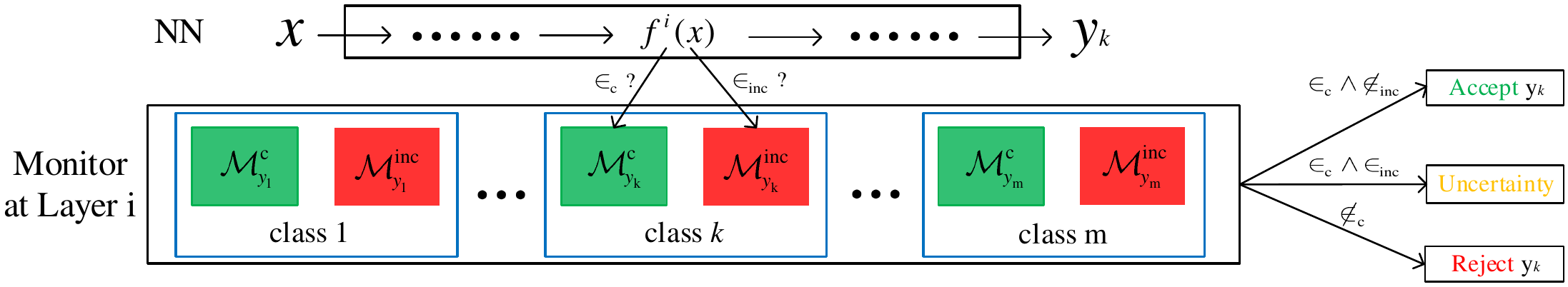}
  \caption{Framework of runtime monitoring of neural networks.}\label{fig:rvNN}
\end{figure}
%
\subsection{Clustering Parameter Selection using Resolution}\label{sec:clusterParameter}
%
In this subsection, using the clustering coverage estimation from the previous section, we show how we can adjust the clustering parameter of the $k$-means clustering algorithm.
Recall that the $k$-means algorithm divides a set of $N$ samples from a set $X$ into $k$ disjoint clusters $C = \setof{C^1,\ldots, C^k }$~\cite{lloyd1982least}.
Each cluster $C_j$ is represented by the mean $\mu_j$ of the samples in the cluster.
The \emph{clustering parameter} $\tau$ serves as a threshold for determining the number $k$ of clusters as follows.
Essentially, the algorithm starts by grouping the inputs into one cluster and iteratively increment the number of cluster.
At each step it computes the so called inertia, where at step $k$, $\mathit{inertia}^k = \sum\limits_{j=1}^{k}\sum\limits_{i=1}^{|C^j|} \|x_i - \mu_j\|^2$ and computes the improvement over the previous step and compares it to the threshold $\tau$.
That is, it checks whether $1 - \frac{\mathit{inertia}^{k+1}}{\mathit{inertia}^{k}} < \tau$ and stops if the answer is positive\footnote{In practice, for a given $\tau$, searching the fine number of clusters from $k=1$ may lead to waste of computational resource.
To avoid this, one can store the results of fine number of clusters for the tried values of clustering parameter into a dictionary.
When a new $\tau$ is input, one can start the search from the fine number of clusters corresponding to the least existing value of clustering parameter less than the new input $\tau$.
If all existing values of $\tau$ are less than the new one, one starts the search from $k=1$.}.

In the following, for a given set of n-dimensional vectors $V$, let $\pi_{\tau}(V)$ denote the partition of $V$ by using $\tau$ for the $k$-means clustering algorithm and $\overline{\boxcoverage{\pi_{\tau}(V)}}$ be the mean value of the estimated bounds of clustering coverage using the method from \secref{sec:boxes_resolution}.

Our experiments suggested (see \secref{sec:experiment}) that the difference of clustering coverage between the parameters $\tau$ in the regions close to two endpoints is very small, that is the higher variations of  clustering coverage with $\tau$ mainly happen in an intermediate interval of $\realinterval{0}{1}$.
Thus, it is of interest to identify such regions by determining the maximum value $\tau_{\rm max}$ (resp. the minimum $\tau_{\rm min}$) whose corresponding clustering coverage is close enough to the one for which $\tau = 1$ (resp. $\tau = 0$).
For identifying such regions, we use binary searches of the values of $\tau_{\rm min}$ and $\tau_{\rm max}$ with two user-specified thresholds: $\epsilon_{\rm cov}$ for coverage difference and $\epsilon_{\rm ival}$ for interval length.
Since searching procedures for $\tau_{\rm min}$ and $\tau_{\rm max}$ are similar, we only present the one for $\tau_{\rm max}$ in Algorithm~\ref{alg:clusterPara}.
The search starts from boundaries of interval $[0, 1]$ (line $1$) and iterate to update one of $\tau_{\rm min_u}$ and $\tau_{\rm max_u}$ by its mean $\tau_{\rm mean}$ as follows (line $4-7$): if $\overline{\boxcoverage{\pi_{1}(V)}} - \overline{\boxcoverage{\pi_{\tau_{\rm mean}}(V)}}$ $>$ $\epsilon_{\rm cov}$, $\tau_{\rm min_u}$ = $\tau_{\rm mean}$; otherwise $\tau_{\rm max_u}$ = $\tau_{\rm mean}$.
The iteration will stop until the length of interval $[\tau_{\rm min_u}, \tau_{\rm max_u}]$ is lesser than or equal to $\epsilon_{\rm ival}$.
The other algorithm for $\tau_{\rm min}$ can be similarly realized by replacing the condition in line $4$ by $\overline{\boxcoverage{\pi_{\tau_{\rm mean}}(V)}}$ - $\overline{\boxcoverage{\pi_{0}(V)}}$ $>$ $\epsilon_{\rm cov}$, inverting instructions $5$ and $7$, and returning $\tau_{\rm min}$.

Then, one can roughly divide the domain of $\tau$ into three parts: $[0, \tau_{\rm min}]$, $[\tau_{\rm min}, \tau_{\rm max}]$, and $[\tau_{\rm max}, 1]$.
Each part demonstrates the correlated effect on partitioning the dataset in terms of space coverage.
Based on this, one can fine-tune the value of $\tau$ according to the monitor performance (see~\secref{sec:experiment:monitor_precision}).
We believe that this can also be used with other geometrical shape abstractions for initial clustering parameter selection, especially when the enclosed space is hard to calculate in high-dimensional space, e.g., zonotope and polyhedra.

%
\begin{algorithm}
\caption{Search of clustering parameter $\tau_{\rm max}$}
\label{alg:clusterPara}
	\begin{algorithmic}[1]	
		\Require  $V$ (a set of n-dimensional vectors), $\epsilon_{cov}$ (threshold of coverage difference), $\epsilon_{ival}$ (threshold of interval length)
		\Ensure clustering parameter $\tau_{\rm max}$
		\State $\tau_{\rm min_u}, \tau_{\rm max_u} \leftarrow 0, 1$
		\While{$\tau_{\rm max_u} - \tau_{\rm min_u}$ > $\epsilon_{ival}$}
		    \State $\tau_{\rm mean} = (\tau_{\rm max_u} + \tau_{\rm min_u}) / 2$;
            \If{$\overline{\boxcoverage{\pi_{1}(V)}} - \overline{\boxcoverage{\pi_{\tau_{\rm mean}}(V)}}$ $>$ $\epsilon_{cov}$}
                \State $\tau_{\rm min_u}$ = $\tau_{\rm mean}$
            \Else
                \State $\tau_{\rm max_u}$ = $\tau_{\rm mean}$
            \EndIf
		\EndWhile
		\State \Return $\tau_{\rm max_u}$
	\end{algorithmic}
\end{algorithm}
\subsection{Monitor Construction}
%
We construct monitors and attach them to some specific chosen layers.
For a given layer $\ell$ and each output class $y \in \mathcal{Y}$, we construct a monitor $\mathcal{M}_{y, \ell}$.
A monitor
comprises two parts, $\mathcal{M}_{y, \ell}^{\rm c}$ and $\mathcal{M}_{y, \ell}^{\rm inc}$, both of which are sets of abstractions used as references of high-level features for inputs correctly and incorrectly classified as $y$, respectively.

Algorithm~\ref{alg:buildingMonitor} constructs the monitor in three steps: i) extract the values of high-level features at monitored layer $\ell$ (line $1$ and $2$);
ii) apply clustering algorithm  to the obtained features and partition them into local distributed clusters (line $3$), see \secref{sec:clusterParameter};
iii) construct an abstraction for each cluster. The union of abstractions computed as such forms $\mathcal{M}_{y, \ell}^{\rm c}$.

\begin{algorithm}[t]
        \caption{\small Construct abstraction for class $y$ at layer $\ell$}
        \label{alg:buildingMonitor}
        \begin{algorithmic}[1]
        \Require $y \in \mathcal{Y}$ (output class), $\ell$ (monitored layer), $D = \setof{(\vvector{x}^1, y^1), \ldots, ( \vvector{x}^m, y^m) } $ (training data), $\tau$ (clustering parameter)
        \Ensure $\mathcal{M}_{y, \ell} = (\mathcal{M}^{\rm c}_{y, \ell},\ \mathcal{M}^{\rm inc}_{y, \ell})$ (a pair of two sets of abstractions) \vspace{1mm}

        \State $V^{\rm c}_{y, \ell} \leftarrow \setof{ f^{\ell}(\vvector{x}) \mid (\vvector{x}, y') \in D \wedge y' = y \wedge y = \mathbf{dec(x)} }$ \Comment{collect the neuron values at layer $\ell$ for inputs correctly classified in $y$}
        \State $V^{\rm inc}_{y, \ell} \leftarrow \{f^{\ell}(\vvector{x})\ |\ (\vvector{x}, y' ) \in D \wedge y'\neq y \wedge y = \textbf{dec($\vvector{x}$)} \}$ \Comment{collect the neuron values at layer $\ell$ for inputs incorrectly classified as class $y$}
        \State $\mathbb{C}^{\rm c}_{y, \ell}$, $\mathbb{C}^{\rm inc}_{y, \ell}$ $\leftarrow$ \textbf{cluster}($V^{\rm c}_{y, \ell}, \tau$), \textbf{cluster}($V^{\rm inc}_{y, \ell}, \tau$) \Comment{divide collected vectors into clusters}
        \State $\mathcal{M}^{\rm c}_{y, \ell}$, $\mathcal{M}^{\rm inc}_{y, \ell}$ $\leftarrow$ $\emptyset$, $\emptyset$ \Comment{sets of abstractions for class $y$}
        \For{$C \in \mathbb{C}^{\rm c}_{y, \ell}$, $C'\in \mathbb{C}^{\rm inc}_{y, \ell}$}
            \State $A^C_{y, \ell},\ A^{C'}_{y, \ell} \leftarrow$ \textbf{abstract}($C$), \textbf{abstract}($C'$) \Comment{construct abstractions for vectors in cluster $C$ and $C'$}
            \State $\mathcal{M}^{\rm c}_{y, \ell}$, $\mathcal{M}^{\rm inc}_{y, \ell}$  $\leftarrow$ $\mathcal{M}^{\rm c}_{y, \ell} \cup \{A^C_{y, \ell}\}$, $\mathcal{M}^{\rm inc}_{y, \ell} \cup \{A^{C'}_{y, \ell}\}$
        \EndFor
        \State \Return $\mathcal{M}_{y, \ell} = (\mathcal{M}^{\rm c}_{y, \ell},\ \mathcal{M}^{\rm inc}_{y, \ell})$
        \end{algorithmic}
\end{algorithm}

\begin{minipage}{0.965\textwidth}
    \begin{minipage}[b]{0.38\textwidth}
        \centering
      \includegraphics[width=1.02\textwidth]{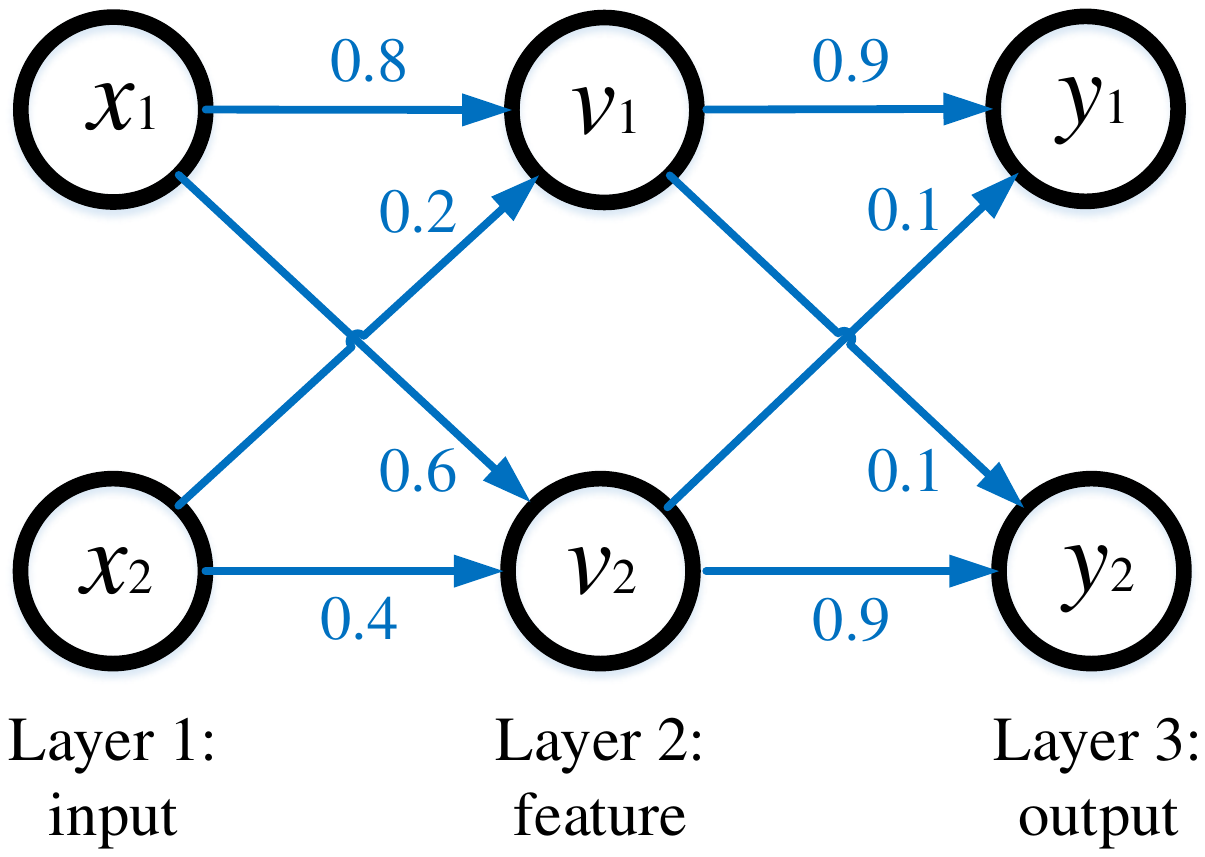}
        \captionof{figure}{An example of neural network.}
        \label{fig:neurNetwork}
    \end{minipage}
    \hfill
    \begin{minipage}[b]{0.59\textwidth}
    	\centering
    	\resizebox{\linewidth}{!}{%
    \begin{tabular}{cccccccc}
    \hline
    true label & $x_1$ & $x_2$ & $v_1$ & $v_2$ & $y_1$ & $y_2$ & prediction \\ \hline
    class 1 & 0.1 & 0 & 0.08 & 0.06 & \textbf{0.078} & 0.062 & class 1 \\ \hline
    class 1 & 0.2 & 0.1 & 0.18 & 0.16 & \textbf{0.222} & 0.162 & class 1 \\ \hline
    class 1 & 0.8 & 0.3 & 0.7 & 0.6 & \textbf{0.69} & 0.61 & class 1 \\ \hline
    class 1 & 0.9 & 0.4 & 0.8 & 0.7 & \textbf{0.79} & 0.71 & class 1 \\ \hline
    class 2 & 0.3 & 0.25 & 0.29 & 0.28 & \textbf{0.289} & 0.281 & {\boxed{class 1}} \\ \hline
    class 2 & 0.4 & 0.35 & 0.39 & 0.38 & \textbf{0.389} & 0.381 & {\boxed{class 1}} \\ \hline
    class 2 & 0.5 & 0.8 & 0.56 & 0.62 & 0.566 & \textbf{0.614} & class 2 \\ \hline
    class 2 & 0.6 & 0.9 & 0.66 & 0.72 & 0.666 & \textbf{0.714} & class 2 \\ \hline
    \end{tabular}
    }
    \captionof{table}{Data and patterns related to the network in Fig.~\ref{fig:neurNetwork}.}
    \label{table:monitor}
    \end{minipage}
\end{minipage}

Below is an example for illustrating the algorithm above.
\begin{example}\label{ex:buildMonitor}
Consider the network in Fig.~\ref{fig:neurNetwork} and the data used to build its monitor are given by columns 1-3 in Table~\ref{table:monitor}.
We consider output class $1$ at layer $2$.
Algorithm~\ref{alg:buildingMonitor}:
\begin{enumerate}
  \item extracts the features (values of $v_1$ and $v_2$ in Table~\ref{table:monitor}) generated at layer 2 and classify them into two sets $V_{1, 2}^{\rm c} = \setof{(0.078, 0.062), (0.222, 0.162), (0.69, 0.61), (0.79, 0.71) }$ and $V_{1, 2}^{\rm inc} = \setof{(0.289, 0.281), (0.389, 0.381)}$ according to whether their inputs are correctly classified or not.
  \item partitions set $V_{1, 2}^{\rm c}$ into a set of two clusters $\mathbb{C}_{1, 2}^{\rm c}$ = $\{C_1, C_2\} \}$ with:
  \begin{itemize}
  	\item $C_1 = \{(0.078, 0.062), (0.222, 0.162)\}$,
  	\item $C_2 = \{(0.69, 0.61), (0.79, 0.71)$,
  \end{itemize}
  and keep $V_{1,2}^{\rm inc}$ as a single cluster $\mathbb{C}_{1, 2}^{\rm inc}$ = $\setof{ C_{3} }$ with $C_3 = \setof{(0.289, 0.281), (0.389, 0.381)}$;
  \item builds the box abstractions for each cluster obtained at step 2: $A_{1, 2}^{C_1}$ = $B(C_1)$ = $[\realinterval{0.078}{0.222}, \realinterval{0.062}{0.162}]$, $A_{1, 2}^{C_2}$ = $B(C_2)$ = $[\realinterval{0.69}{0.79}$, $[0.61, 0.71]]$, and $A_{1, 2}^{C_{3}}$ = $B(C_{3})$ = $[\realinterval{0.289}{0.389}, \realinterval{0.281}{0.381}]$.
\end{enumerate}
The monitor for class 1 is the pair of sets: $\mathcal{M}_{1, 2} =(\mathcal{M}_{1, 2}^{\rm c}, \mathcal{M}_{1, 2}^{\rm inc})$, where $\mathcal{M}_{1, 2}^{\rm c} = \{A_{1, 2}^{C_1}, A_{1, 2}^{C_2}\}$ and $\mathcal{M}_{1, 2}^{\rm c} = \{A_{1, 2}^{C_{3}}\}$.
\end{example}
%
\subsection{Monitor Execution}
%

\begin{algorithm}[H]
        \caption{\small Monitoring at Layer $\ell$}
        \label{alg:useMonitor}
        \begin{algorithmic}[1]
        \Require network input $\vvector{x}$, pairs of two sets of abstractions: $\mathcal{M}_{1, \ell} = (\mathcal{M}^{\rm c}_{1,\ell}, \ \mathcal{M}^{\rm inc}_{1, \ell}), \ldots, \mathcal{M}_{|\mathcal{Y}|, \ell} = (\mathcal{M}^{\rm c}_{|\mathcal{Y}|, \ell},\ \mathcal{M}^{\rm inc}_{|\mathcal{Y}|, \ell})$
        \Ensure answer ``accept'', ``reject'', or ``uncertainty'' \vspace{1mm}
        \State $S\_correct$, $S\_incorrect$ $\leftarrow$ $False$, $False$ \Comment{flags used to record if the new produced output is similar to correct or incorrect behavior }
        \State $\vvector{v} \leftarrow$ $f^{\ell}(\vvector{x}$) \Comment{collect output at layer $\ell$}
        \State $y \leftarrow$ \textbf{dec($\vvector{x}$)}) \Comment{predict class of $\vvector{x}$}
        \If{$\vvector{v} \in \mathcal{M}^{\rm c}_{y, \ell}$}
            \State $S\_correct$ $\leftarrow$ $True$  \Comment{found a correct-classification abstraction containing $\vvector{v}$}
        \EndIf
        \If{$\vvector{v} \in \mathcal{M}^{\rm inc}_{y, \ell}$}
            \State $S\_incorrect$ $\leftarrow$ $True$  \Comment{found an incorrect-classification abstraction containing $\vvector{v}$}
        \EndIf
        \If{$S\_correct$ $\wedge$ $S\_incorrect$}
            \State \Return ``uncertainty'' \Comment{$\vvector{v}$ is contained by both $\mathcal{M}^{\rm c}_y$ and $\mathcal{M}^{\rm inc}_y$}
        \ElsIf{$S\_correct$ $\wedge$ $\neg S\_incorrect$}
            \State \Return ``accept'' \Comment{$\vvector{v}$ is only contained by $\mathcal{M}^{\rm c}_y$}
        \Else
            \State \Return ``reject'' \Comment{$\vvector{v}$ is not contained by $\mathcal{M}^{\rm c}_y$}
        \EndIf
        \end{algorithmic}
\end{algorithm}
After constructing the monitor, it will be deployed with the network in parallel and work according to the procedure presented in Algorithm~\ref{alg:useMonitor}.
Note that for simplicity in line $4$ and $7$, we use $\vvector{v} \in \mathcal{M}^{\rm c}_{y, \ell}$ ($\mathcal{M}^{\rm inc}_{y, \ell}$) to denote a vector $\vvector{v}$ is contained in some element of set $\mathcal{M}^{\rm c}_{y, \ell}$ ($\mathcal{M}^{\rm inc}_{y, \ell}$).
For each new input, the value at monitored layer $\ell$ will first be produced (line $2$) and then checked (line $4-13$) whether it is contained by one of abstractions from $\mathcal{M}_{y, \ell}^{\rm c}$ and $\mathcal{M}_{y, \ell}^{\rm inc}$ according the network prediction of class $y$ (line $3$).
Out of the four possibilities, we distinguish three outcomes: i) ``uncertainty'', if $f(\vvector{x})$ is contained by both some abstraction from $\mathcal{M}_{y, \ell}^{\rm c}$ and $\mathcal{M}_{y, \ell}^{\rm inc}$; ii) ``accept'', if $f(\vvector{x})$ is only contained by some abstraction from $\mathcal{M}_{y, \ell}^{\rm c}$; iii) ``reject'', otherwise.

\begin{example}
Consider the network in Fig.~\ref{fig:neurNetwork} and the monitor constructed in Example~\ref{ex:buildMonitor} in parallel.
Since there exists no overlap between the built abstractions for each class, the monitor has only two possible outcomes: accept and reject.
Assume now $\vvector{x}^1 = (0.15, 0.1)$ and $\vvector{x}^2 = (0.6, 0.5)$ are input to the network.
We first collect its output at watched layer 2: $f^2(\vvector{x}^1) = (0.14, 0.13)$, and $f^2(\vvector{x}^1) = (0.58, 0.56)$.
Then, the network outputs the predictions: \textbf{dec($\vvector{x}^1$)}=1, \textbf{dec($\vvector{x}^2$)}=1.
Based on the predicted class of each input, the monitor checks if its produced feature is in some corresponding abstractions or not. The determination is that $f^2(\vvector{x}^1)$ is inside abstraction $A_{1, 2}^{C_1}$, while $f^2(\vvector{x}^2)$ is outside any abstraction in $\mathcal{M}_{1, 2}^{\rm c}$.
Finally, the monitor accepts $\vvector{x}^{1}$ and rejects $\vvector{x}^{2}$.
\end{example}
\begin{figure}[t]
  \centering
  \begin{subfigure}[b]{.49\linewidth}
  	\centering
  	\includegraphics[width=0.36\textwidth]{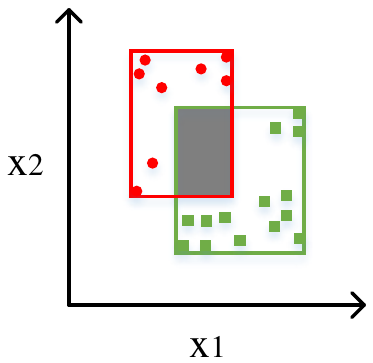}
  	\caption{Too coarse abstractions.}
  	\label{fig:uncertainty:abstraction}
  \end{subfigure}
  \begin{subfigure}[b]{.49\linewidth}
    \centering
    \includegraphics[width=0.36\textwidth]{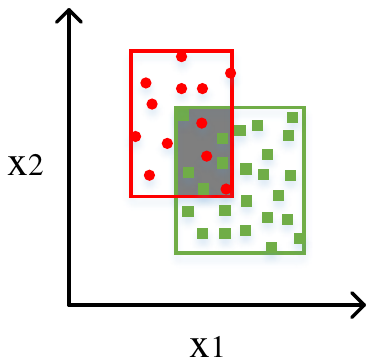}
    \caption{Bad separability of classification.}
    \label{fig:uncertainty:network}
  \end{subfigure}
  \caption{An illustration of the source of uncertainty.}
  \label{fig:uncertainty}
\end{figure}

\begin{remark}[About uncertainty]
Using ``uncertainty'' provides a new dimension to verify the quality of the built monitor for a given network, since it measures the ``overlap'' between the abstractions of correct and incorrect behaviors.
The more ``uncertainty'' a monitor produces as verdict, the worse the abstraction is.
The reason for a high level of uncertainty can be twofold, as illustrated in Fig~\ref{fig:uncertainty}: i) the abstraction built is too coarse; ii) the network intrinsically has a bad separability of classification.
\end{remark}

\section{Experimental Evaluation}

The objective of this section is twofold: i) to show that our method for estimating clustering coverage produces precise approximations in practice, i.e., the difference between the estimated lower and upper bounds is zero or negligible; ii) to assess the monitor performance under different settings of clustering parameter $\tau$ and chosen monitoring layers.

We use the MNIST~\cite{lecun1998gradient} benchmark, which is a dataset for image classification of handwritten digits ($0$ to $9$).
MNIST consists of a training set of $60,000$ samples and a test set of $10,000$ samples.
We use the test set of F\_MNIST~\cite{xiao2017fashion} to simulate abnormal inputs, that is $10,000$ samples from $10$ classes of Zalando article images.
For consistency, we used the common network in~\cite{cheng2019runtime} and~\cite{henzinger2019outside}, whose accuracies on training and test sets are $99.76\%$ and $99.24\%$, respectively.
We use the last four layers of the network to build the monitors (denoted as layer 6, 7, 8, and 9 in what follows).
We extract the high-level features from these layers to create the monitors' references.
For clustering parameter, we tried out the $12$ values in $[1.0, 0.9, 0.8, 0.7, 0.6, 0.5, 0.4, 0.3, 0.2, 0.1, 0.05, 0.01]$.
Since a monitor is built for each output class at each monitored layer, $480$ monitors in total were constructed and tested during the experimentation.

To reduce the computation cost of the experiments (and facilitate its reproducibility), we divided the experiments in $3$ steps:
\begin{enumerate}
  \item high-level features extraction, which can be one-time generated in seconds and used for multiple times afterwards;
  \item partition of the features with many options of clustering parameter, during the experimentation, which took the most time due to the search of the fine $k$. But, this can be very efficient if the options are tried out in descending order;
  \item monitor creation and test, which can be done immediately.
\end{enumerate}
%
\subsection{Clustering Coverage Estimation}\label{sec:experiment}
%

\begin{figure*}[htbp]
    \centering
    \begin{subfigure}[htbp]{0.23\textwidth}
        \centering
        \includegraphics[width=\textwidth]{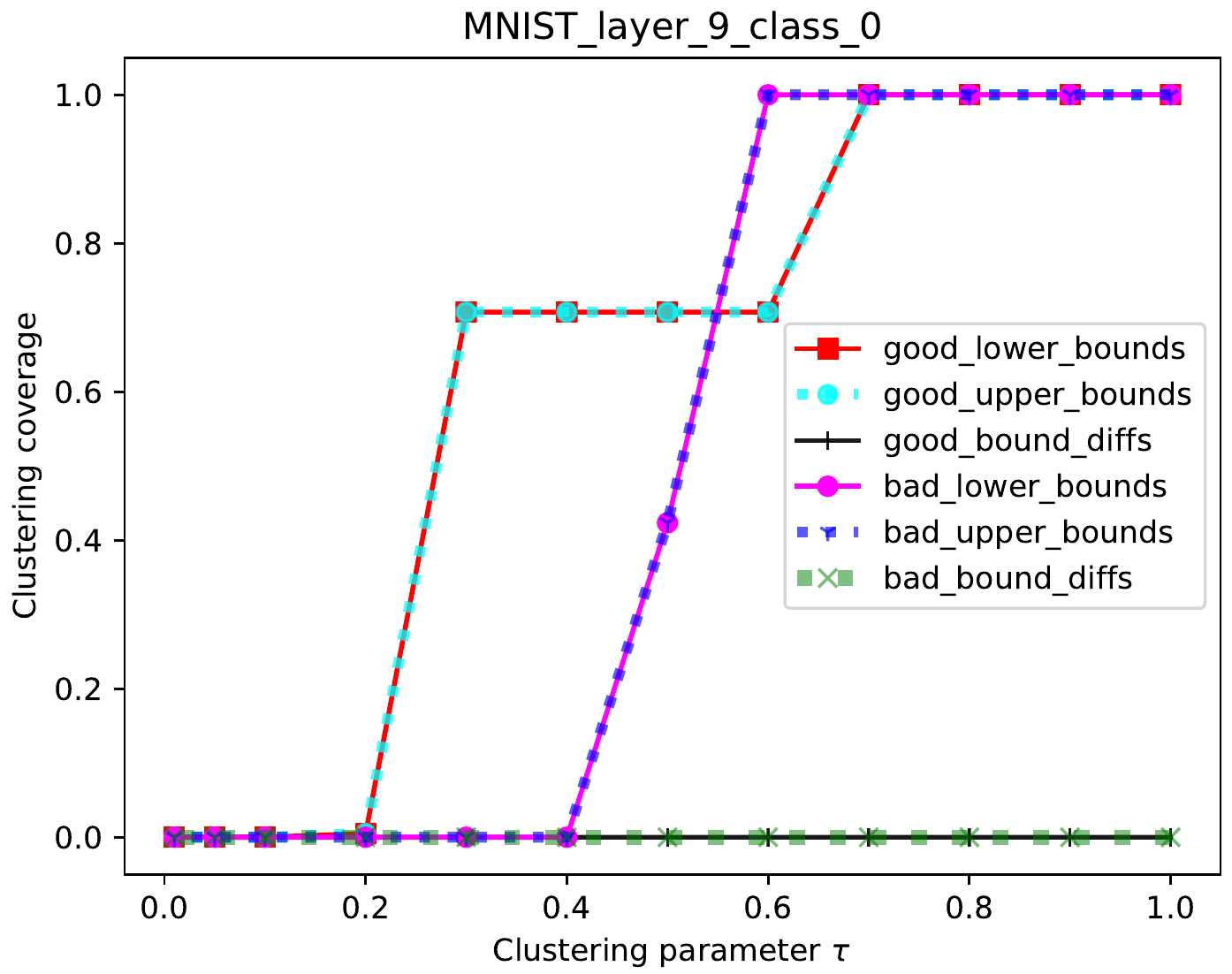}
    \end{subfigure}
    \hfill
    \begin{subfigure}[htbp]{0.23\textwidth}
        \centering
        \includegraphics[width=\textwidth]{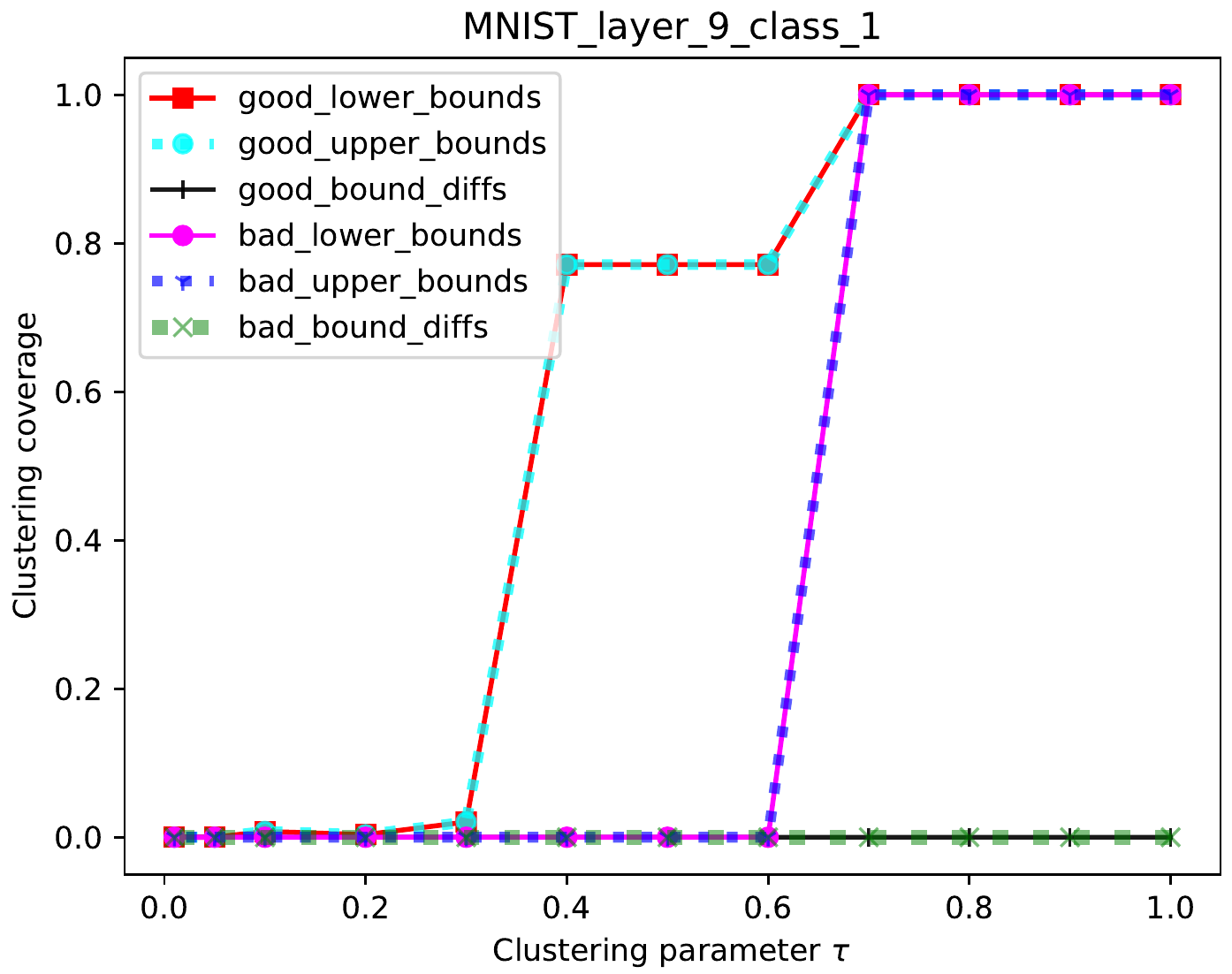}
    \end{subfigure}
    \hfill
    \begin{subfigure}[htbp]{0.23\textwidth}
        \centering
        \includegraphics[width=\textwidth]{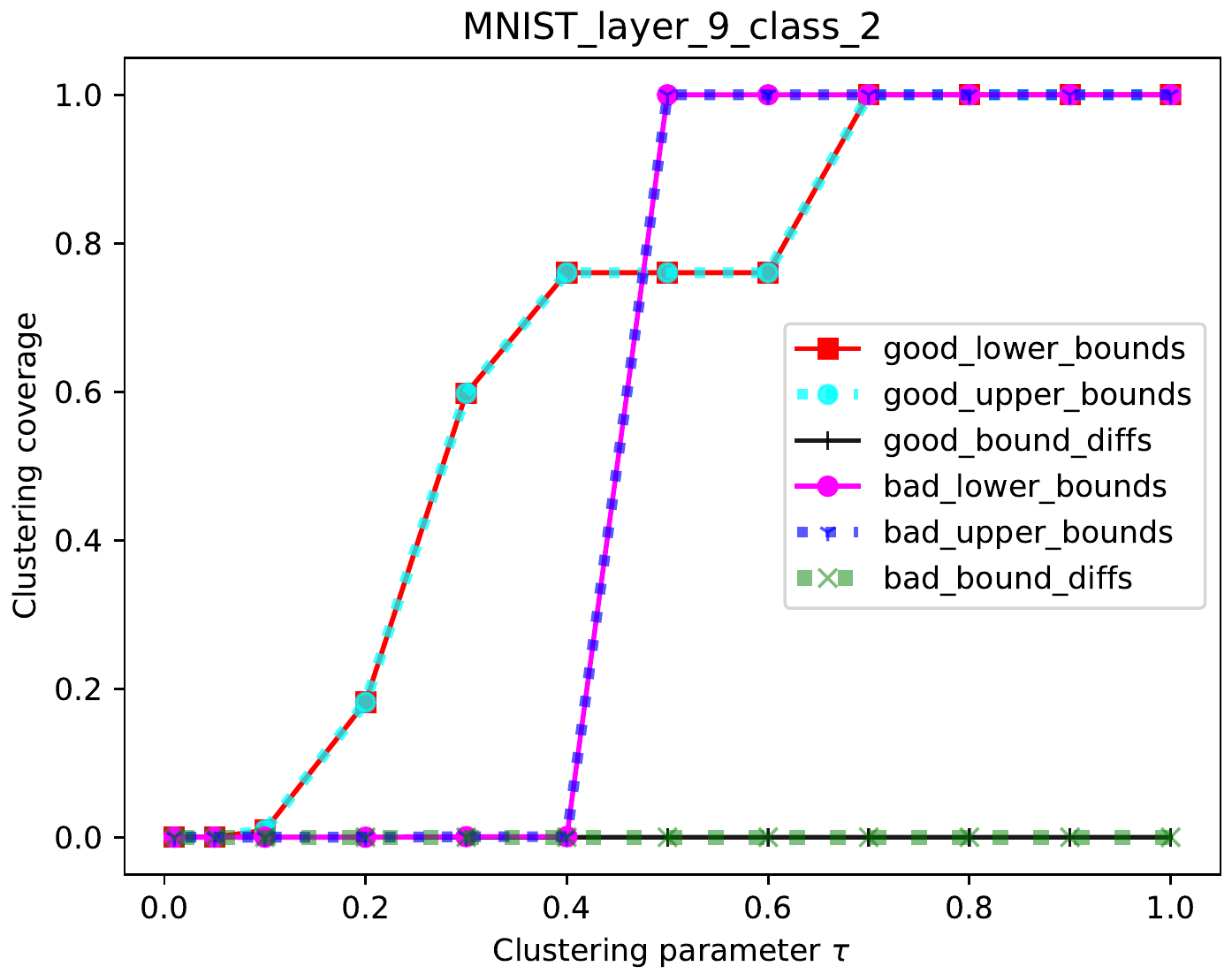}
    \end{subfigure}
    \hfill
    \begin{subfigure}[htbp]{0.23\textwidth}
        \centering
        \includegraphics[width=\textwidth]{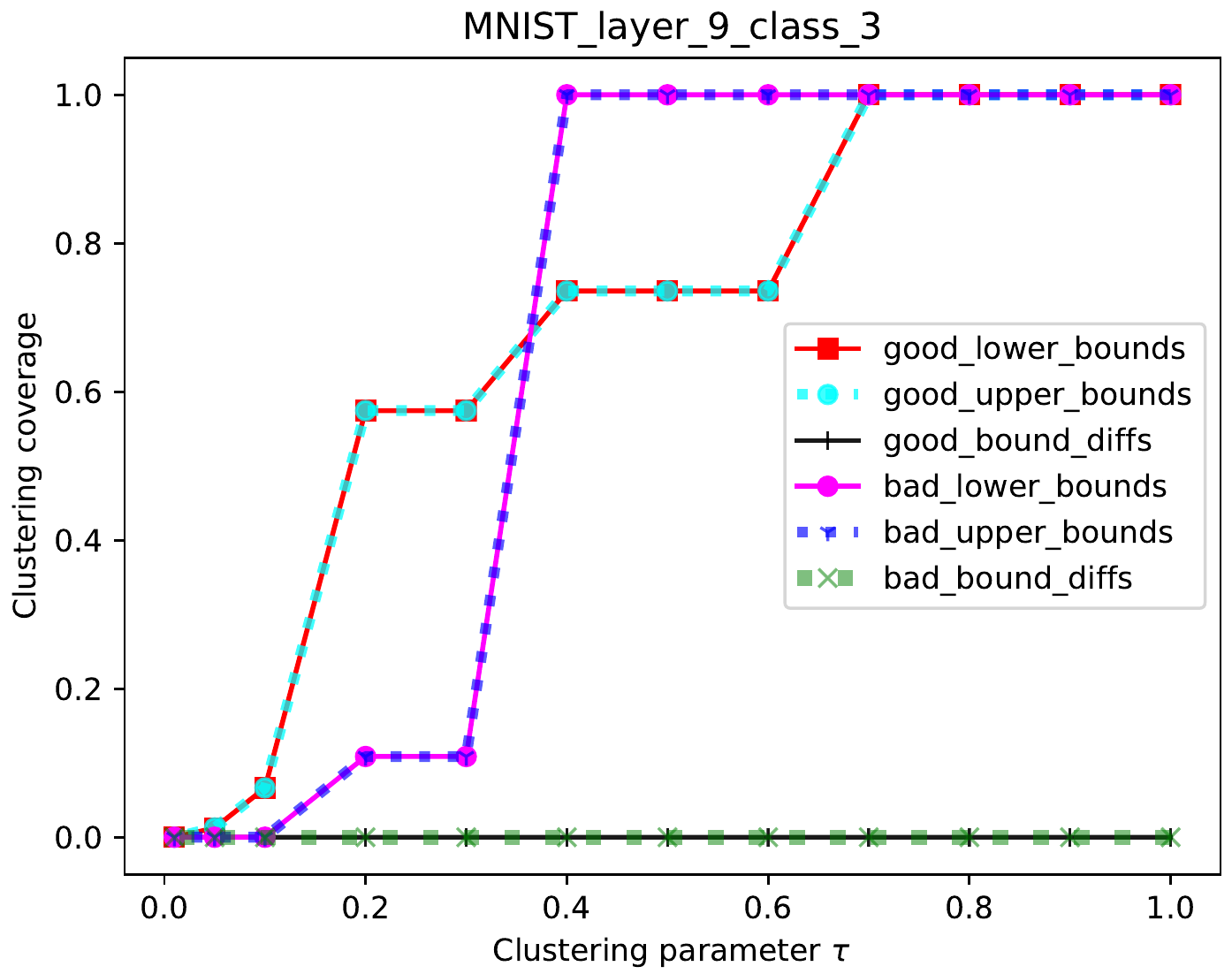}
    \end{subfigure}

    \begin{subfigure}[htbp]{0.23\textwidth}
        \centering
        \includegraphics[width=\textwidth]{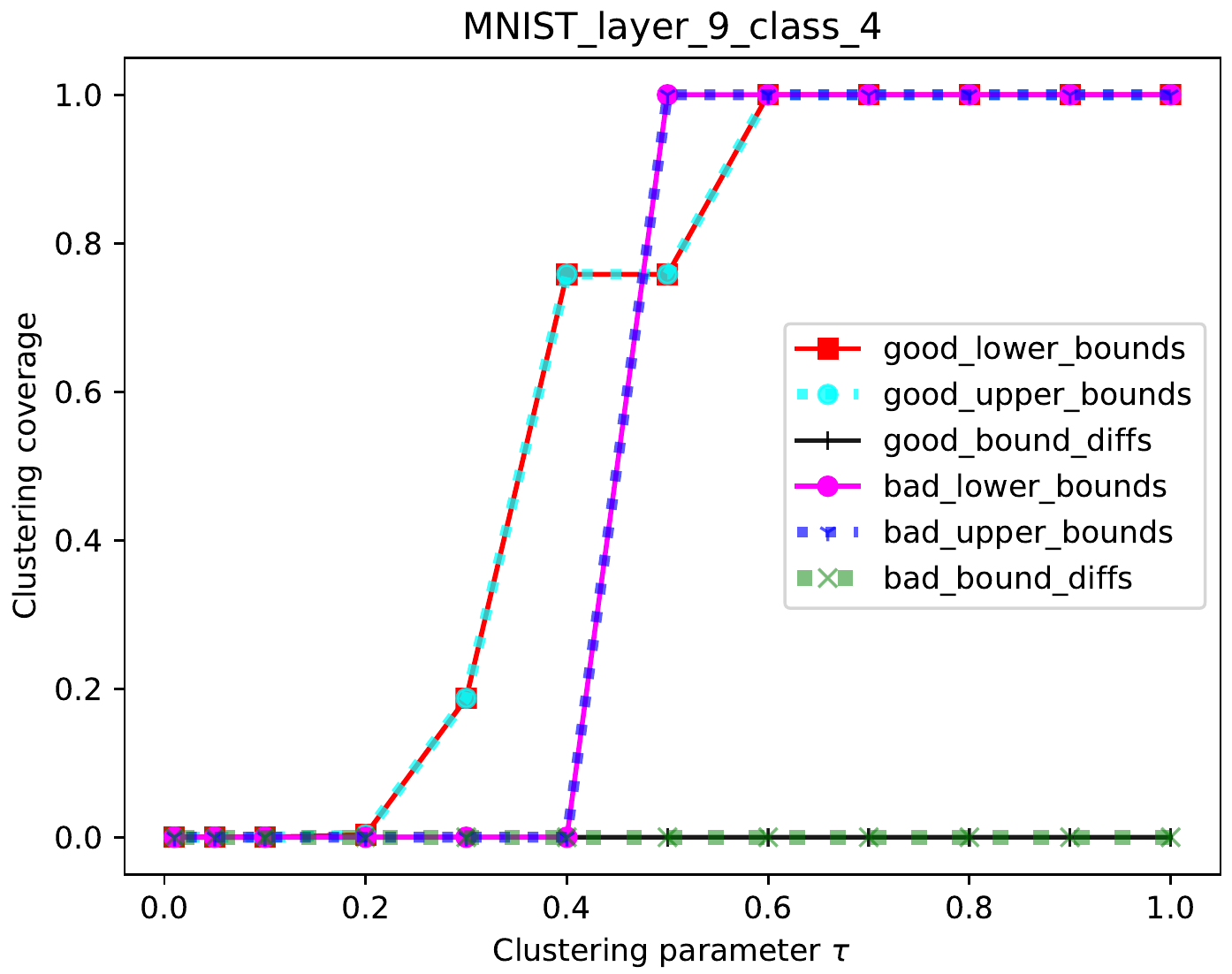}
    \end{subfigure}
    \hfill
    \begin{subfigure}[htbp]{0.23\textwidth}
        \centering
        \includegraphics[width=\textwidth]{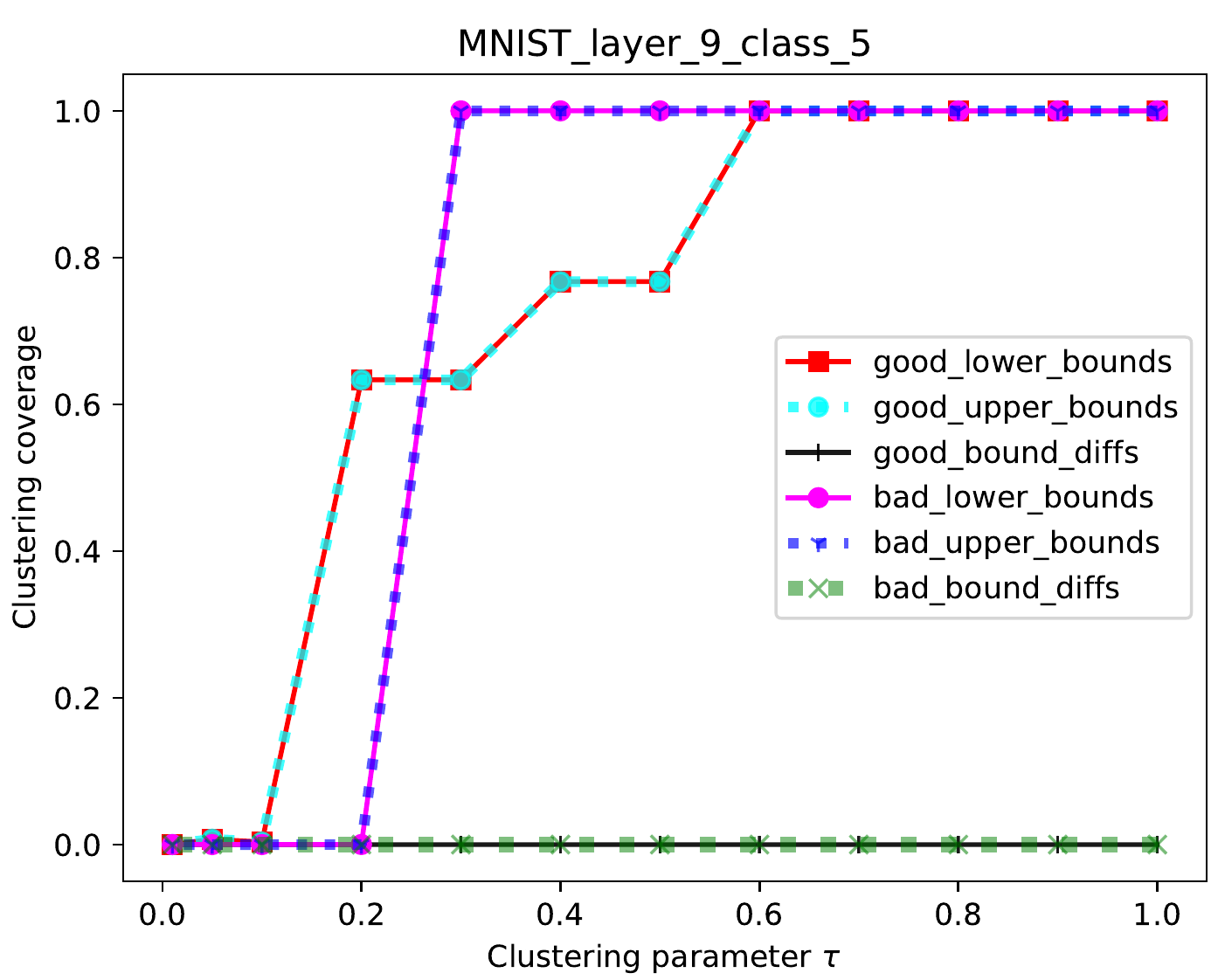}
    \end{subfigure}
    \hfill
    \begin{subfigure}[htbp]{0.23\textwidth}
        \centering
        \includegraphics[width=\textwidth]{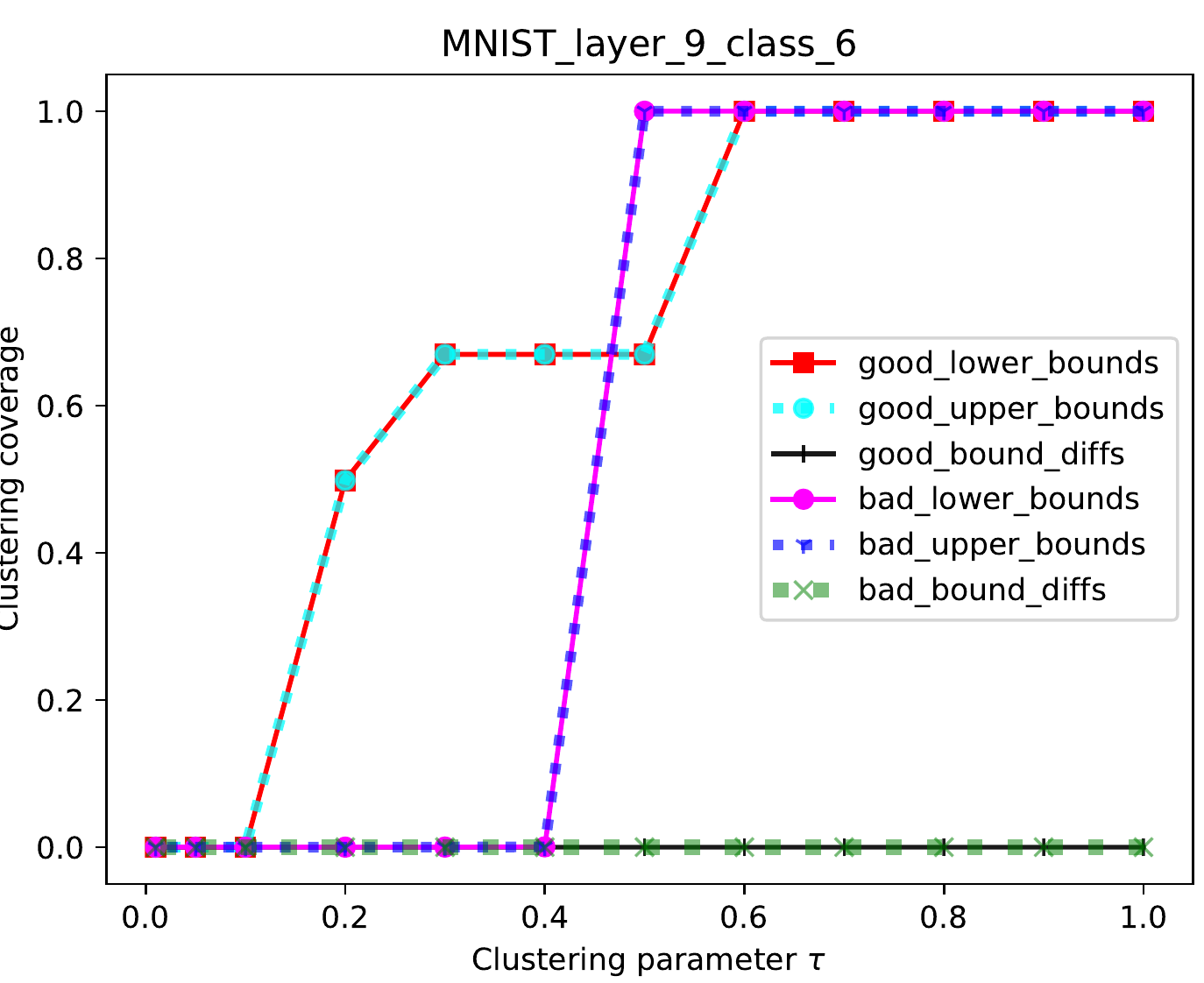}
    \end{subfigure}
    \hfill
	\begin{subfigure}[htbp]{0.23\textwidth}
        \centering
        \includegraphics[width=\textwidth]{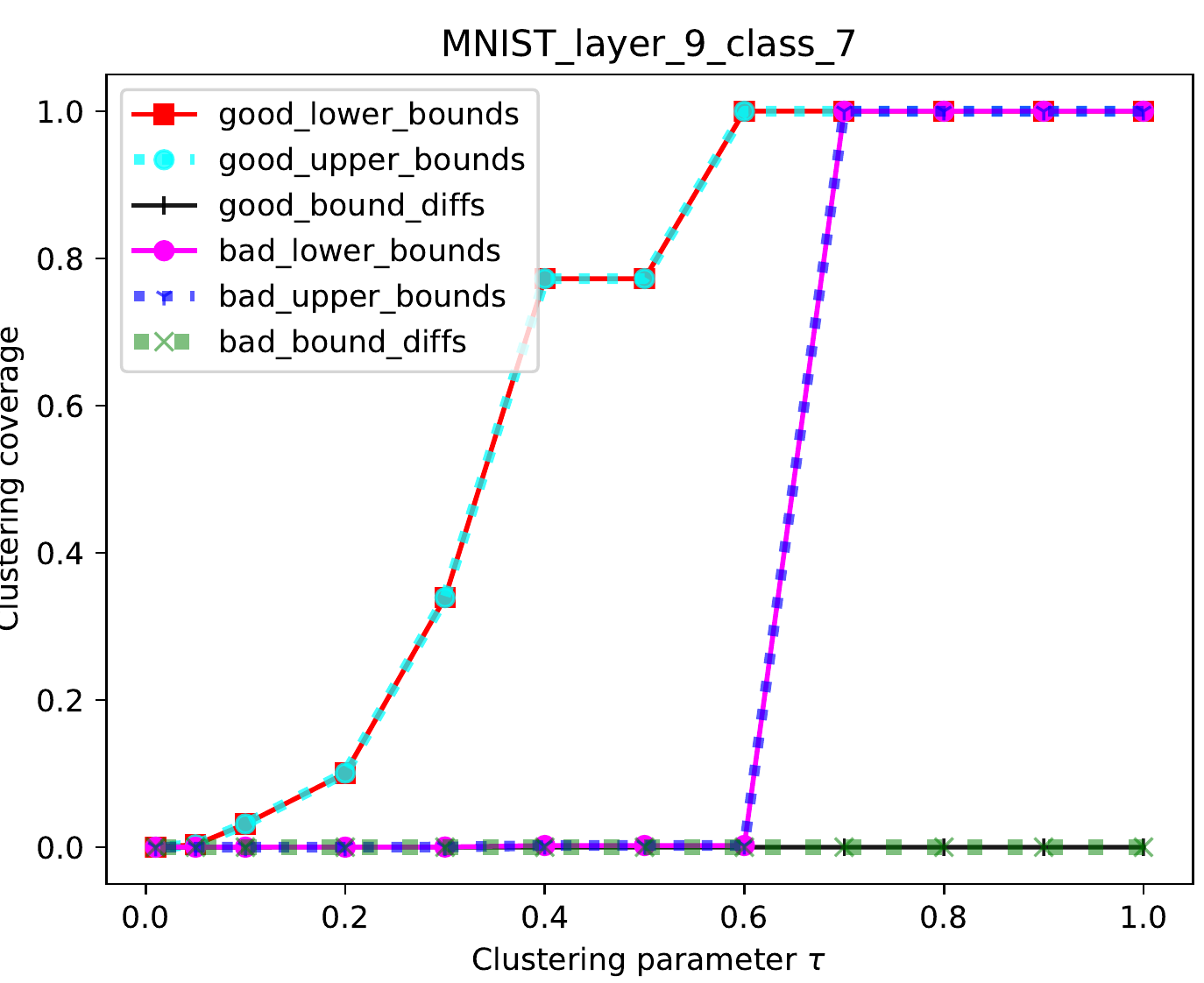}
    \end{subfigure}

    \begin{subfigure}[htbp]{0.23\textwidth}
        \centering
        \includegraphics[width=\textwidth]{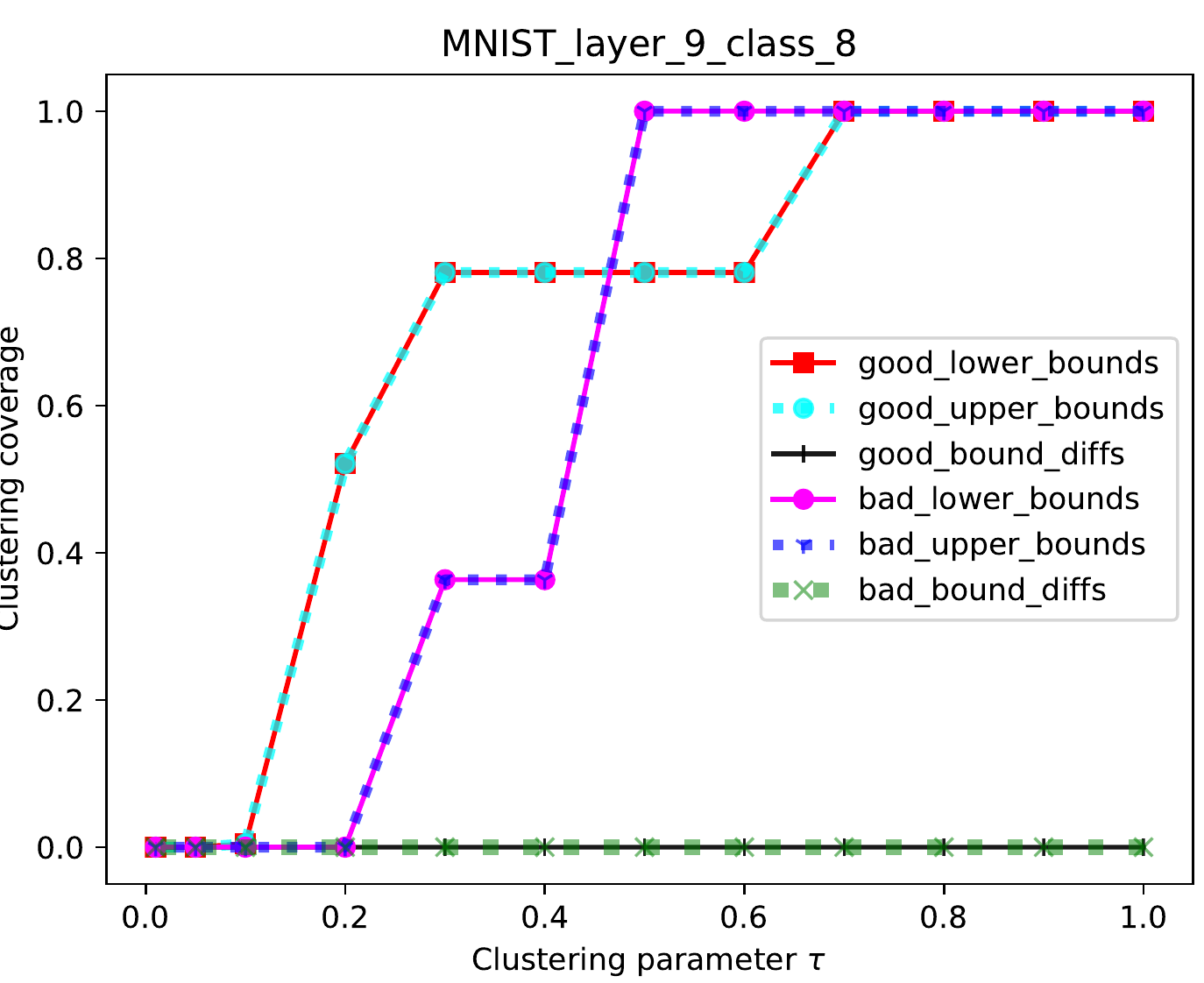}
    \end{subfigure}
    \hfill
    \begin{subfigure}[htbp]{0.23\textwidth}
        \centering
        \includegraphics[width=\textwidth]{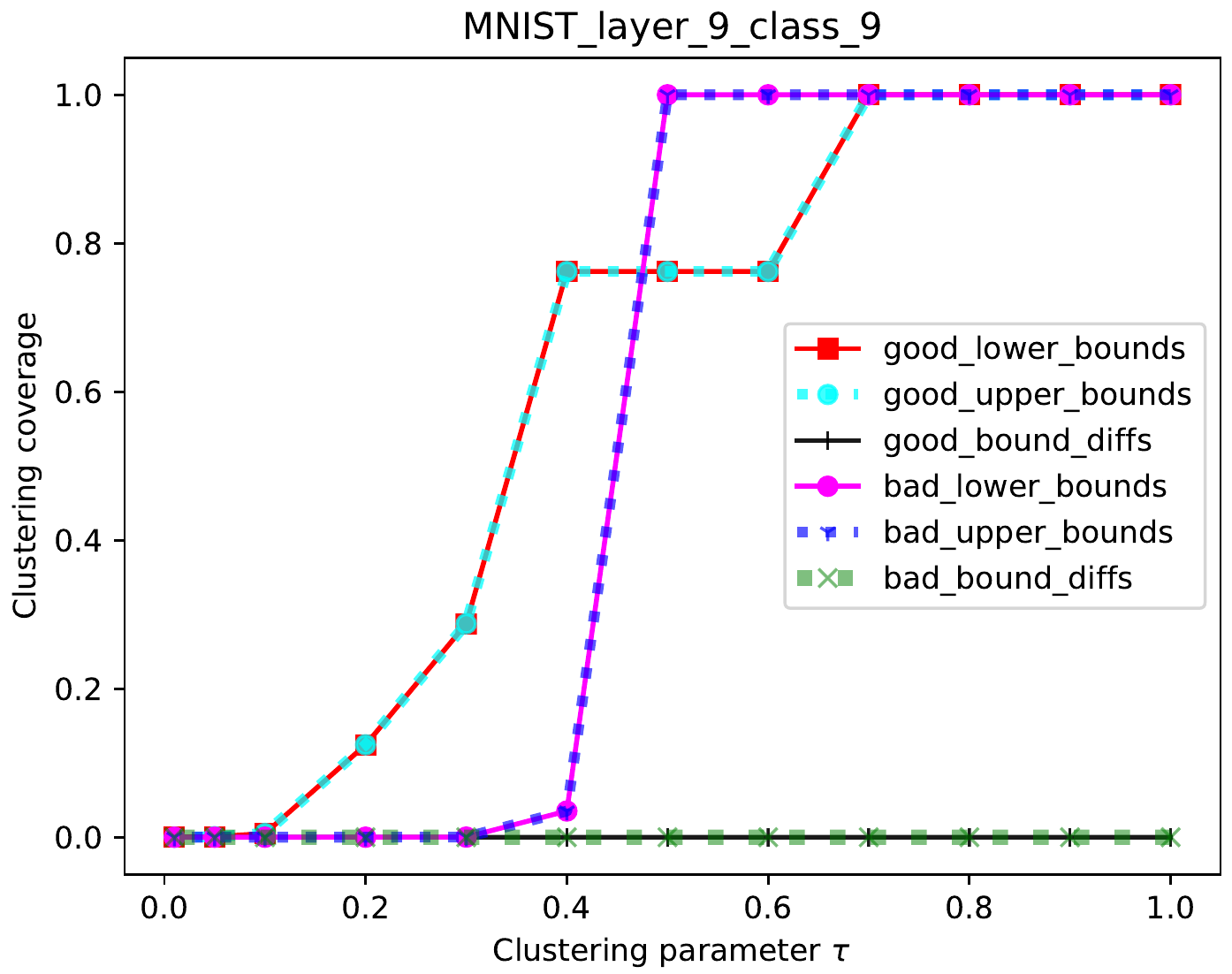}
    \end{subfigure}
    \caption{Clustering coverage estimations for the high-level features obtained at the output layer for benchmark MNIST.}
    \label{fig:clusteringMNIST}
\end{figure*}

Fig.~\ref{fig:clusteringMNIST} contains $10$ graphs which result from the clustering coverage estimation for partitioning the high-level features at the output layer by using the following clustering parameters $[1.0, 0.9, 0.8, 0.7, 0.6, 0.5, 0.4, 0.3, 0.2, 0.1, 0.05, 0.01]$.
Each graph contains six curves which represent the lower bounds, the upper bounds, the bound differences of clustering coverage for the  partitions corresponding to the good and bad features used to construct the corresponding monitor reference.

We observed that the relative difference between estimated lower and upper bounds are zeros or extremely small -- less than $1$\textperthousand, as shown in Fig.~\ref{fig:clusteringMNIST}.
\todocwyf{Review after finishing sect4.1}
Moreover, as we stated in previous section, the clustering coverage lines close to two endpoint are flat, which indicates that the difference of clustering coverage between the parameters $\tau$ in such regions is zero or very small.
%
\subsection{Assessing Monitor Precision}\label{sec:experiment:monitor_precision}
%
We discuss first how to evaluate the precision of a monitor for a classification system and then the relationship between the performance, the clustering parameter, and the monitored layer.

\paragraph{Evaluating monitors for classification systems.}
Since we create monitors for image classification systems, we use two sorts of images to construct the test dataset.
The first sort is referred to as known inputs; these are the images belonging to one of classes of the system (i.e., from MNIST).
The second sort is referred to as unknown inputs; these are the images not belonging to any class of the system (i.e., from F\_MNIST).
In testing a monitor, one can choose the ratio between known and unknown inputs depending on the criticality and the purpose of the classification system.
We choose to have the same number of known and unknown inputs, i.e., we took $10,000$ samples from each dataset.
Once the test data is prepared, we feed it to the network and evaluate the performance of every monitor by considering different kinds of outcomes as indicated in the confusion matrix shown in Table~\ref{table:monitorPerformance}.
We note that a false negative can correspond to two situations: a known input from a different class or an unknown input.

\begin{figure}[t]
    \centering
    \begin{subfigure}[htbp]{0.4\textwidth}
        \centering
        \includegraphics[width=\textwidth]{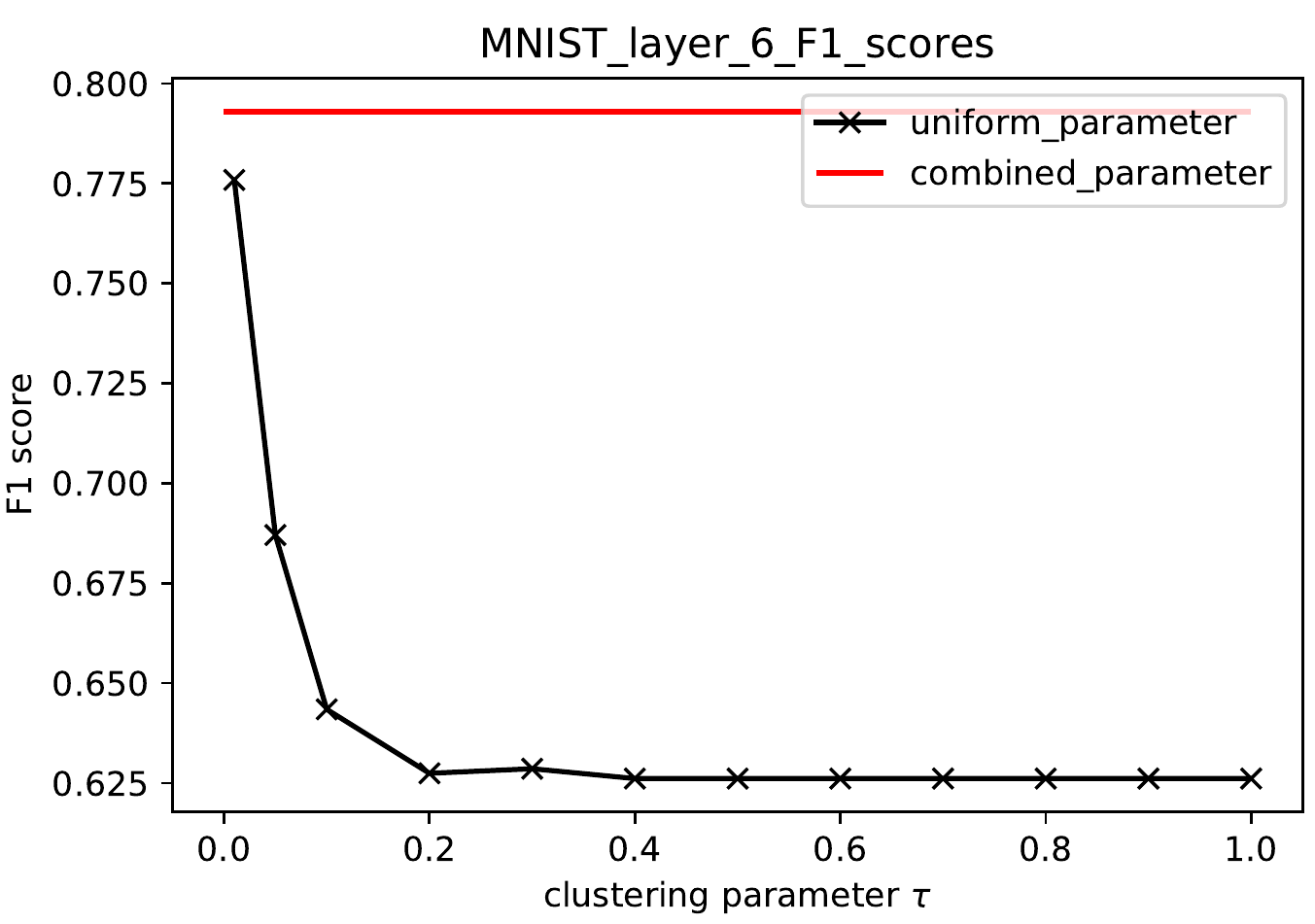}
    \end{subfigure}
    \hfill
    \begin{subfigure}[htbp]{0.4\textwidth}
        \centering
        \includegraphics[width=\textwidth]{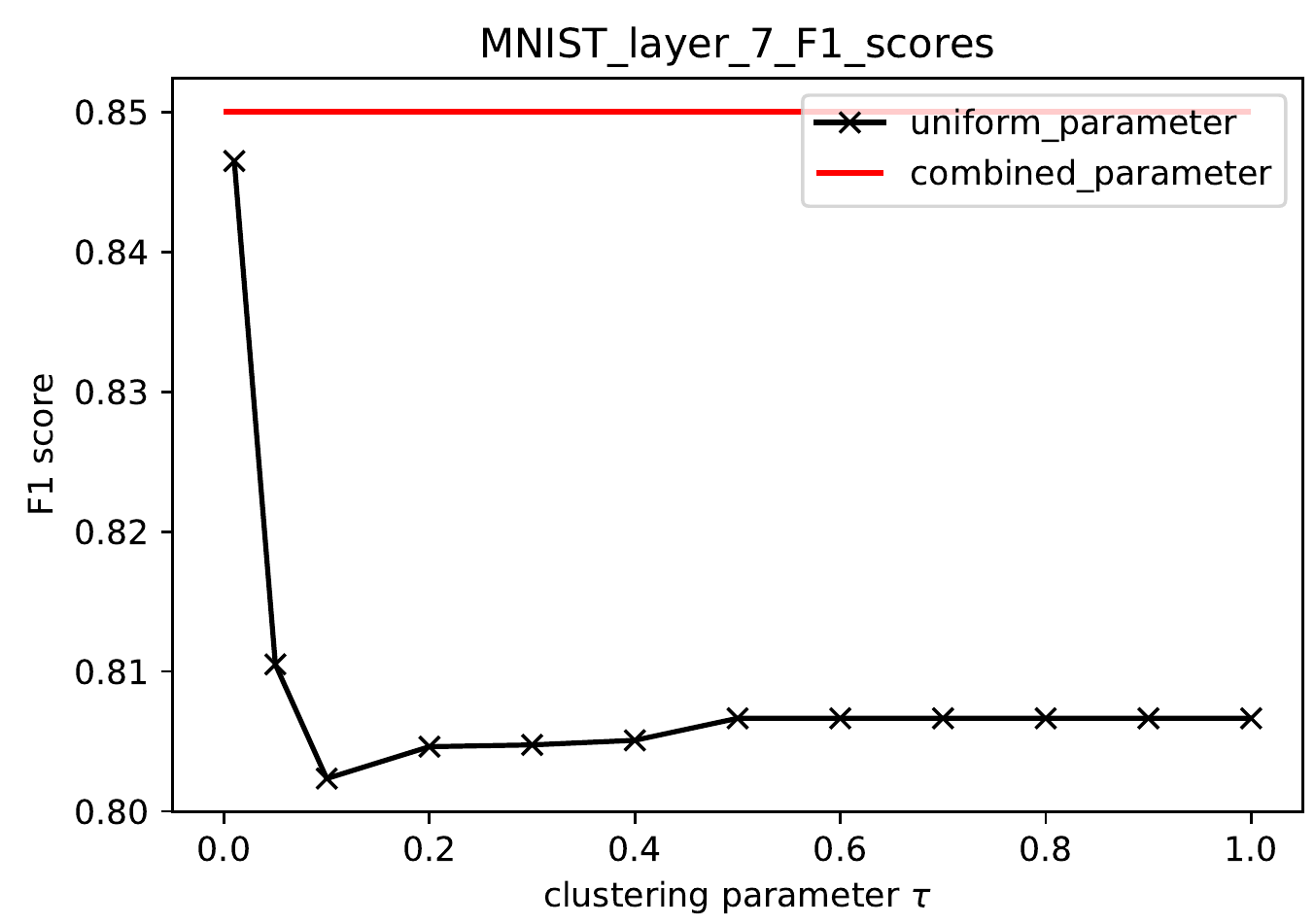}
    \end{subfigure}

	\begin{subfigure}[htbp]{0.4\textwidth}
        \centering
        \includegraphics[width=\textwidth]{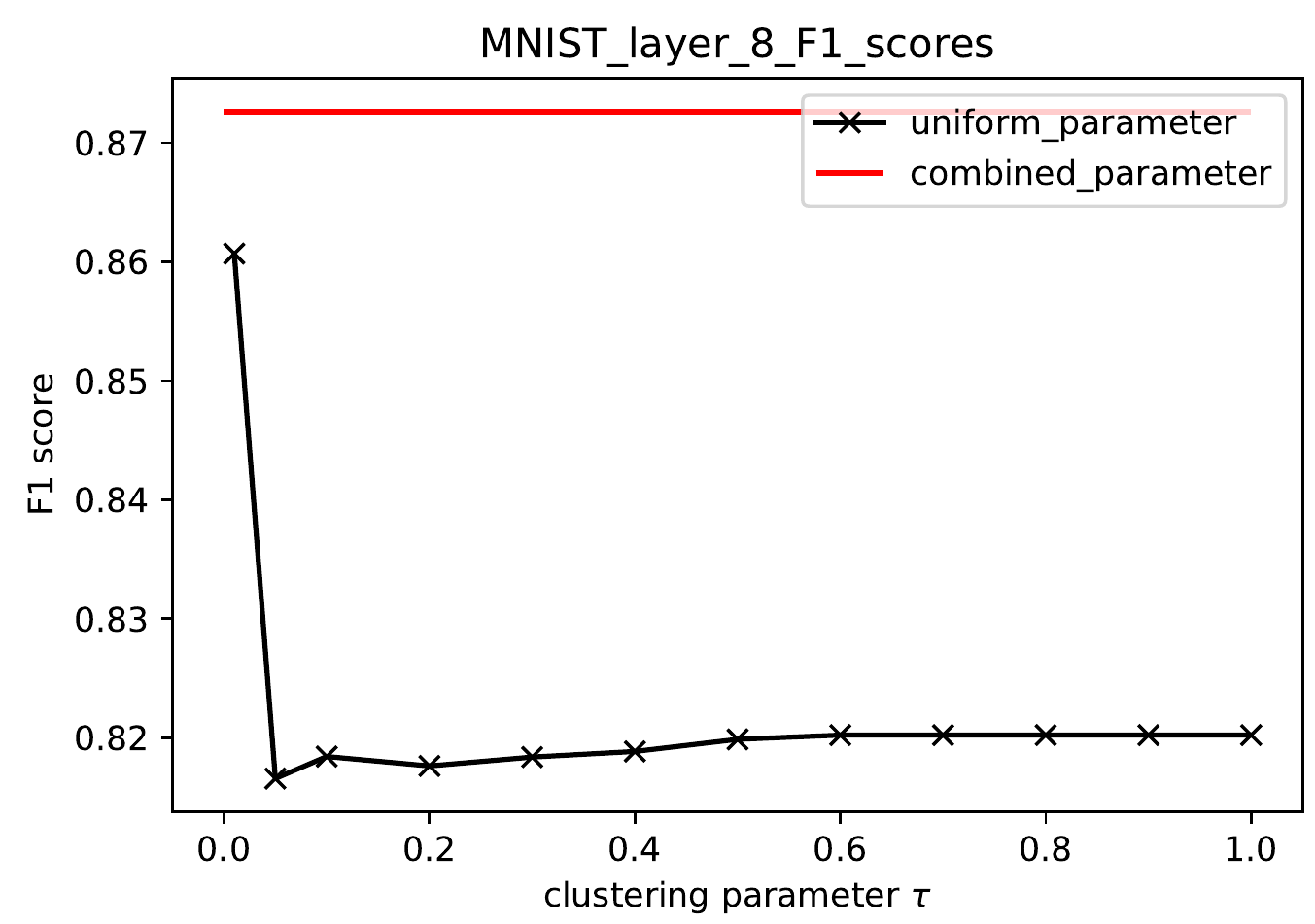}
    \end{subfigure}
    \hfill
    \begin{subfigure}[htbp]{0.4\textwidth}
        \centering
        \includegraphics[width=\textwidth]{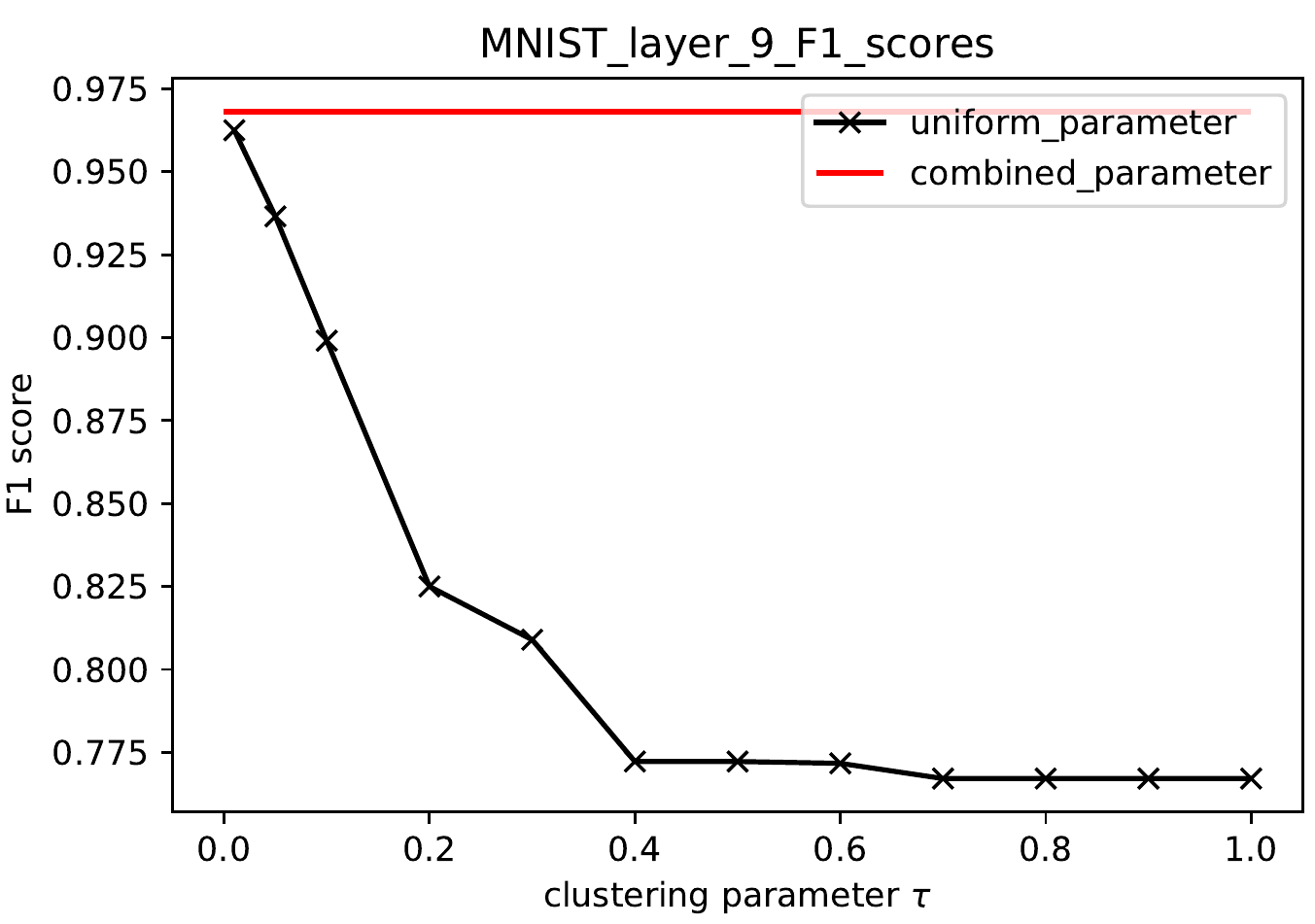}
    \end{subfigure}
    \caption{Comparison of F1 scores obtained via setting uniform (as in \cite{henzinger2019outside}) and combined parameters on benchmark MNIST.}
    \label{fig:F1ScoresMNIST}
\end{figure}

\paragraph{Comparison with \cite{henzinger2019outside}.~~}
Controlling the coarseness of built abstractions with regard to effectiveness is crucial.
However, how to control the size of sub-box abstractions via clustering parameter is ignored in~ \cite{henzinger2019outside}.
This paper addresses it as follows.
First, we introduce the notion of clustering coverage to measure, in terms of covered space, the relative size of sub-box abstractions w.r.t the global one.
Second, we leverage the network bad behaviours, which introduces a new monitor outcome of uncertainty.
The number of uncertainty outcomes (MN and MP in Table~\ref{table:monitorPerformance}) is an hint on the coarseness of built abstractions.
Furthermore, we provide the following improvements:
1) the object of study, novelty detection, is refined to each output class of a given network, since the original definition of novelty, \emph{true positive} in~\cite{henzinger2019outside}, includes not only unknown inputs, but also known inputs which belong to one of output classes but misclassified and whose feature is outsize abstraction;
2) we use a clustering parameter that is specific to each output class at each monitored layer, and possibly to each set of good and bad features; while~\cite{henzinger2019outside} uses a uniform clustering parameter for all output classes and monitored layers.
This enables precise control on the numbers of FN and FP, e.g., see Fig.~\ref{fig:F1ScoresMNIST}: a higher F1 score can be always achieved via selecting a group of clustering parameters customized to each output class.

In the sequel, we discuss in details three topics: i) relationship between monitor effectiveness and clustering parameter; ii) how to tune the clustering parameter; iii) how to select the best layers to monitor.

\begin{table}[t]
\centering
\caption{Confusion matrix giving the different outcomes when evaluating the verdict of a monitor according a network input.}\label{table:monitorPerformance}
\resizebox{0.85\linewidth}{!}{
    \begin{tabular}{cccc}\hline
        \diagbox{\makecell{real nature}}{\makecell{verdicts}} & \makecell{\textbf{negative}\\(accept)} & \makecell{\textbf{positive}\\(reject)} & \textbf{uncertainty} \\\hline
        \makecell{\textbf{negative}\\(labelled $y$)} & true negative (TN) & false positive (FP) & missed negative (MN) \\\hline
        \makecell{\textbf{positive}\\ (labelled non-$y$)} & false negative (FN) & true positive (TP) & missed positive (MP) \\\hline
    \end{tabular}
}
\end{table}

\begin{figure*}[htbp]
    \centering
    \begin{subfigure}[htbp]{0.245\textwidth}
        \centering
        \includegraphics[width=\textwidth]{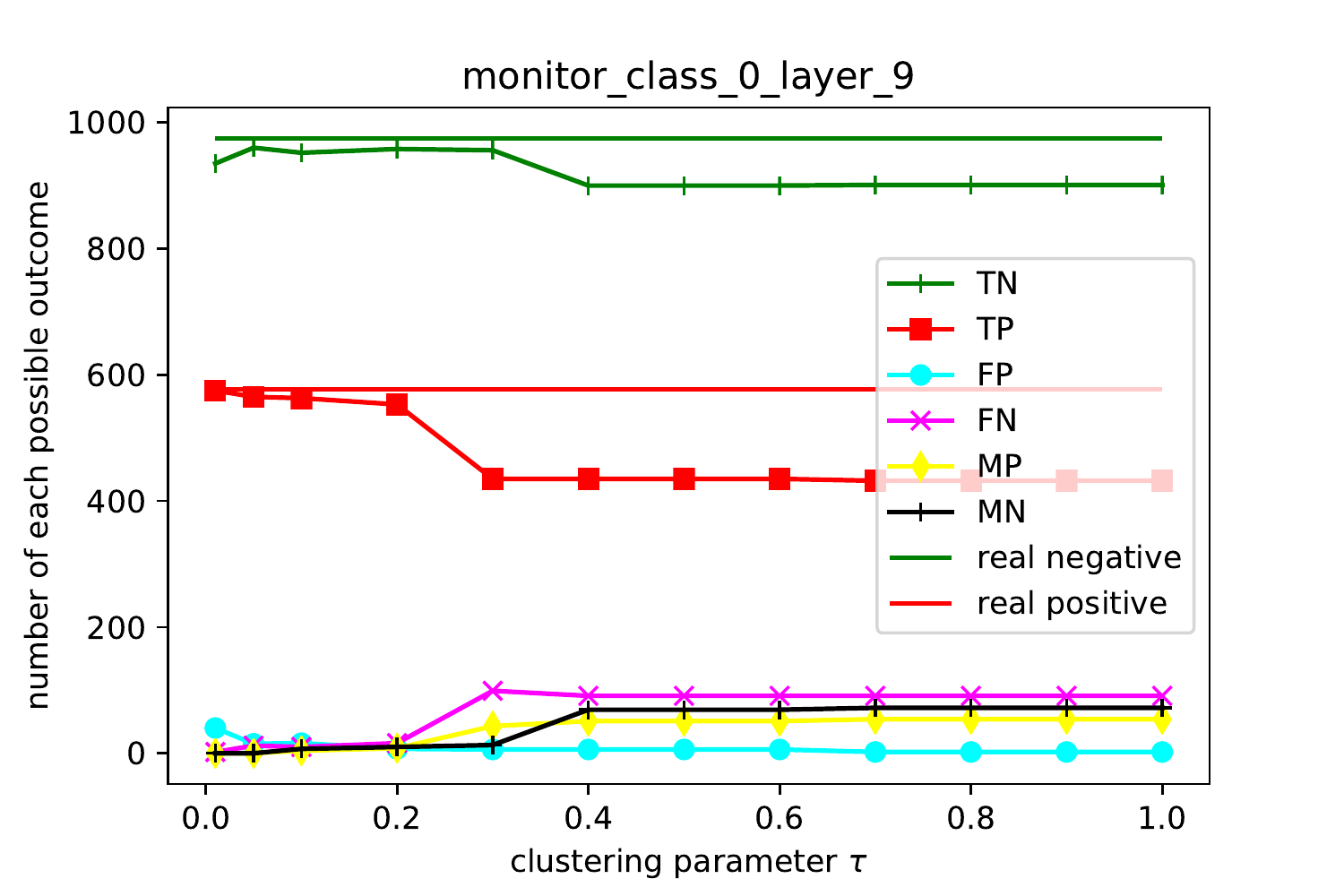}
    \end{subfigure}
    \hfill
    \begin{subfigure}[htbp]{0.245\textwidth}
        \centering
        \includegraphics[width=\textwidth]{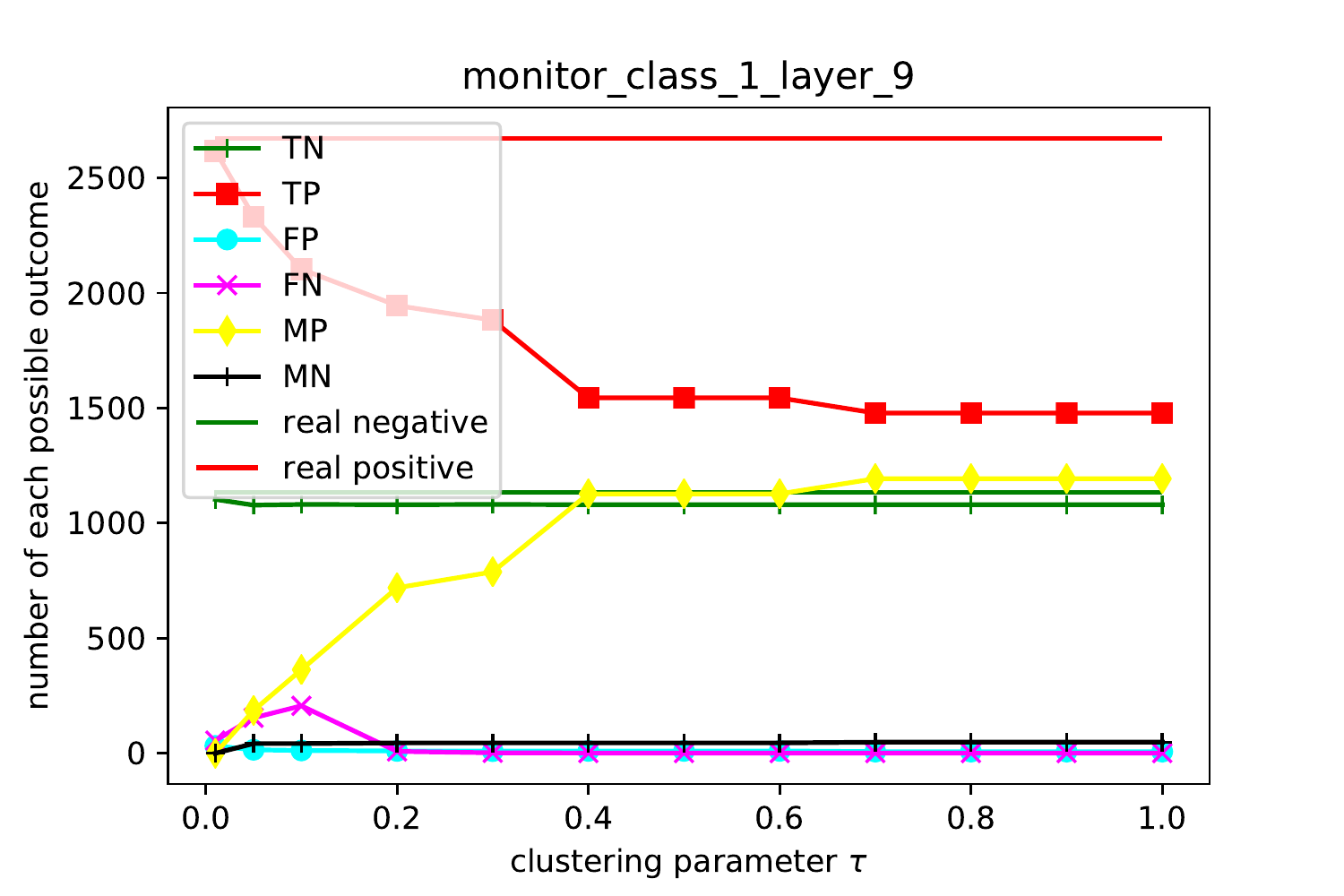}
    \end{subfigure}
    \hfill
	\begin{subfigure}[htbp]{0.245\textwidth}
        \centering
        \includegraphics[width=\textwidth]{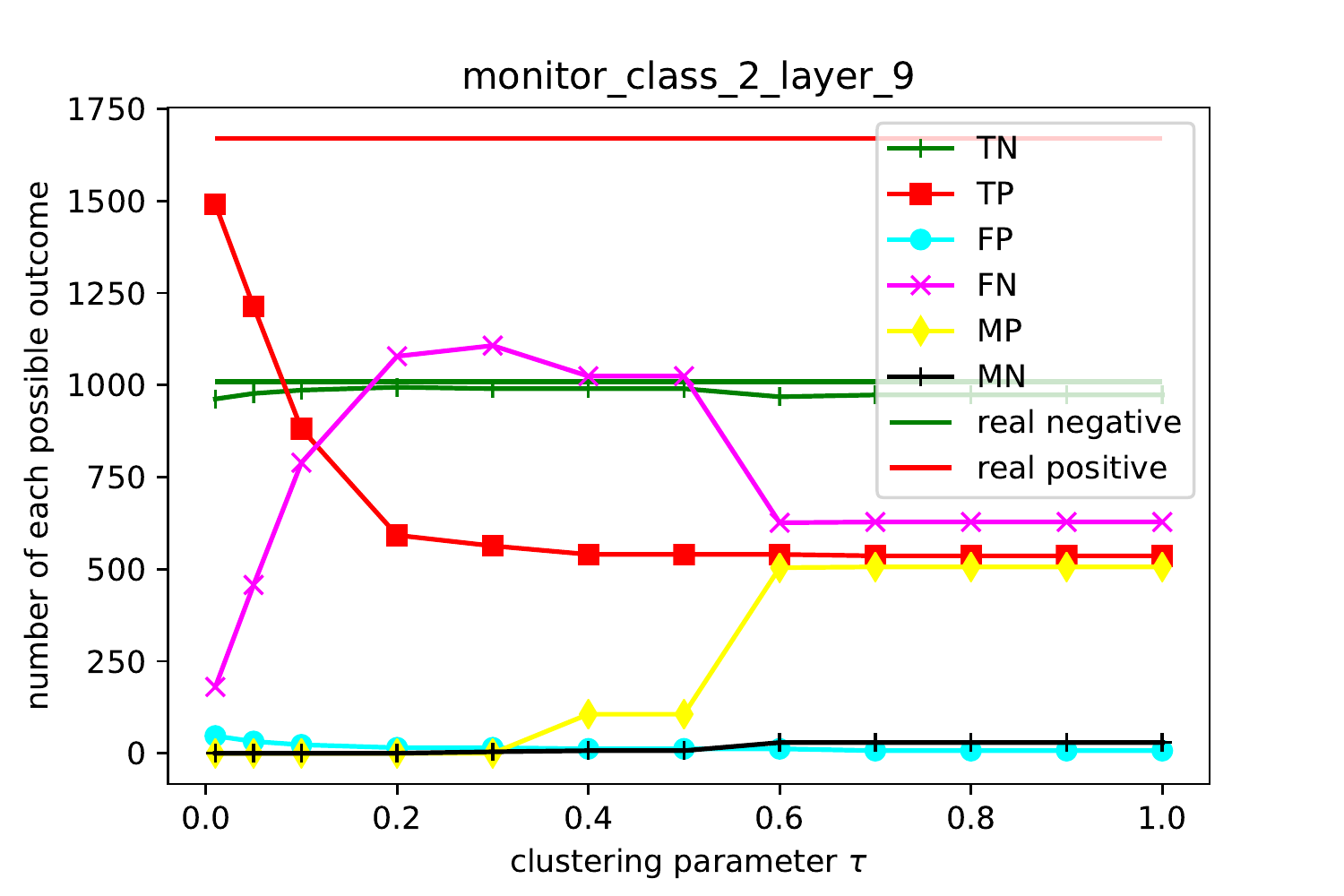}
    \end{subfigure}
    \hfill
    \begin{subfigure}[htbp]{0.245\textwidth}
        \centering
        \includegraphics[width=\textwidth]{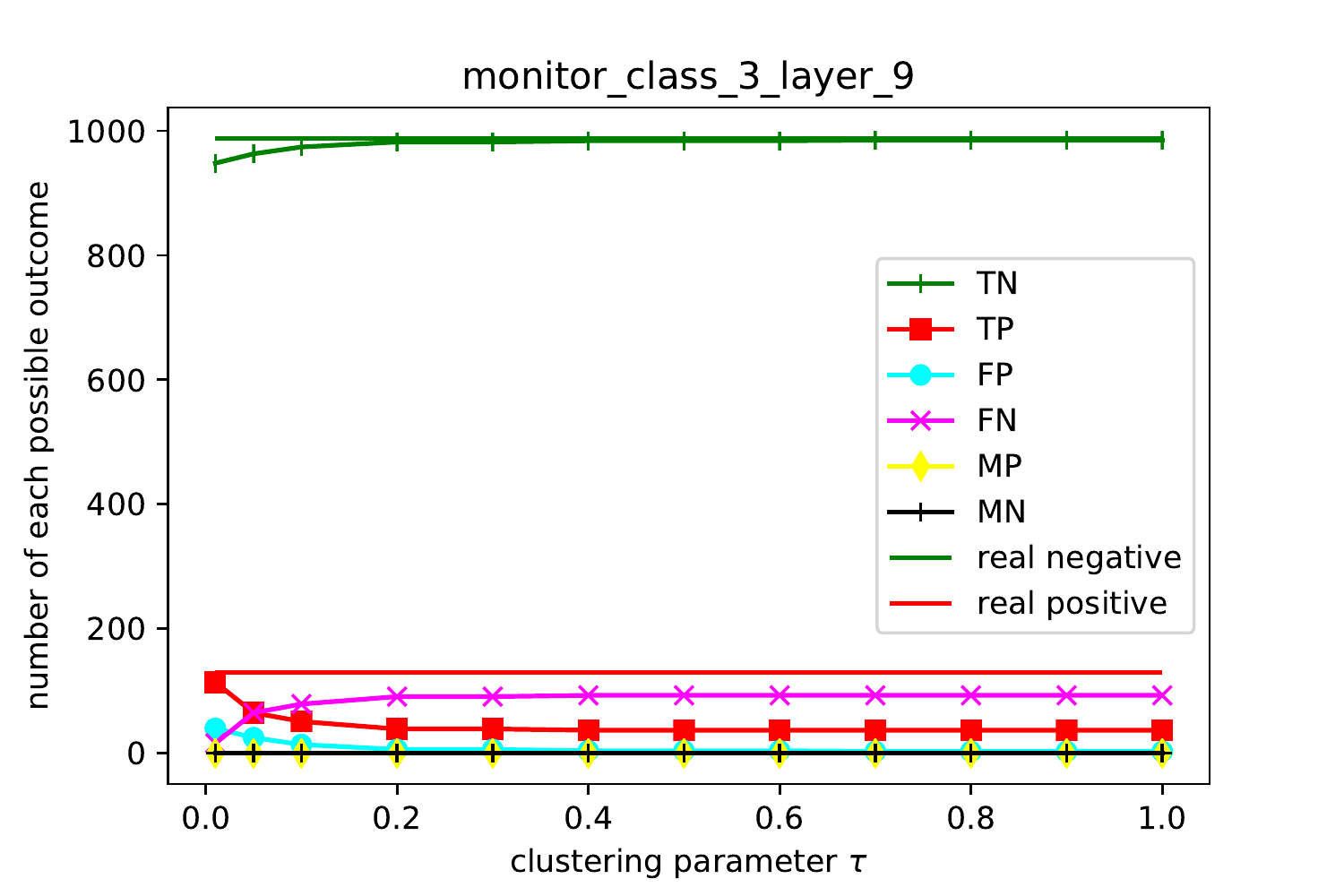}
    \end{subfigure}

    \begin{subfigure}[htbp]{0.245\textwidth}
        \centering
        \includegraphics[width=\textwidth]{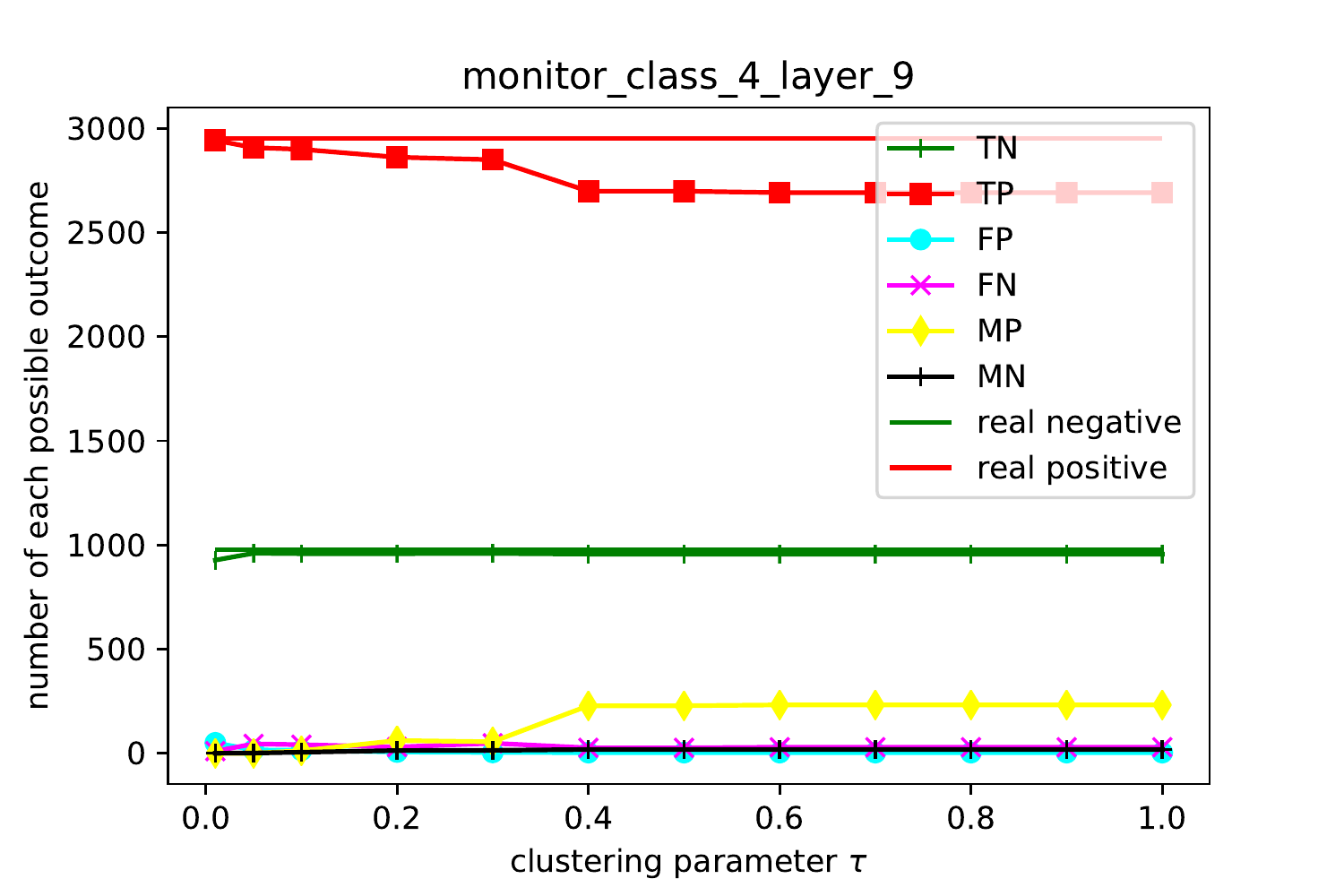}
    \end{subfigure}
    \hfill
    \begin{subfigure}[htbp]{0.245\textwidth}
        \centering
        \includegraphics[width=\textwidth]{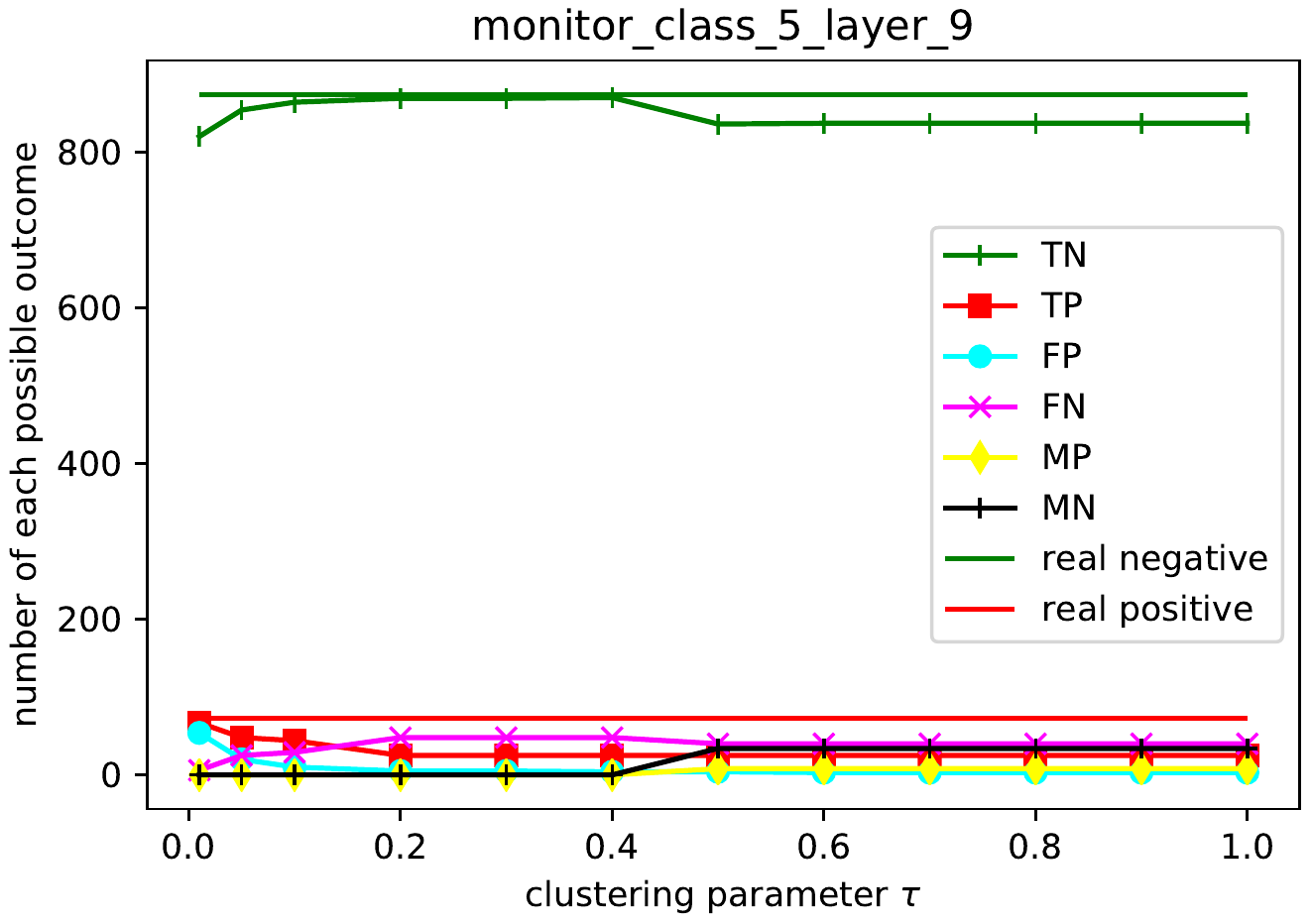}
    \end{subfigure}
    \hfill
    \begin{subfigure}[htbp]{0.245\textwidth}
        \centering
        \includegraphics[width=\textwidth]{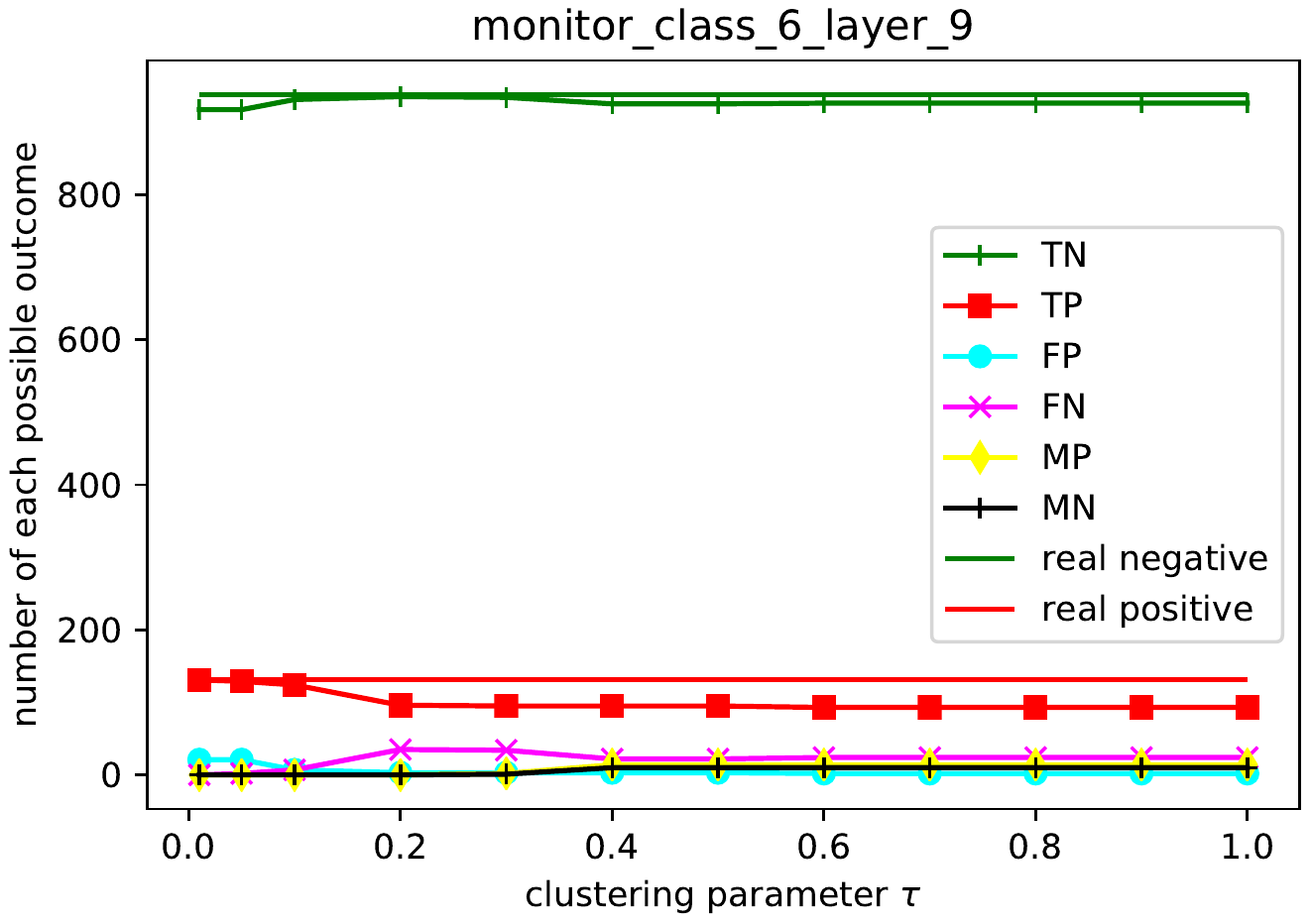}
    \end{subfigure}
    \hfill
    \begin{subfigure}[htbp]{0.245\textwidth}
        \centering
        \includegraphics[width=\textwidth]{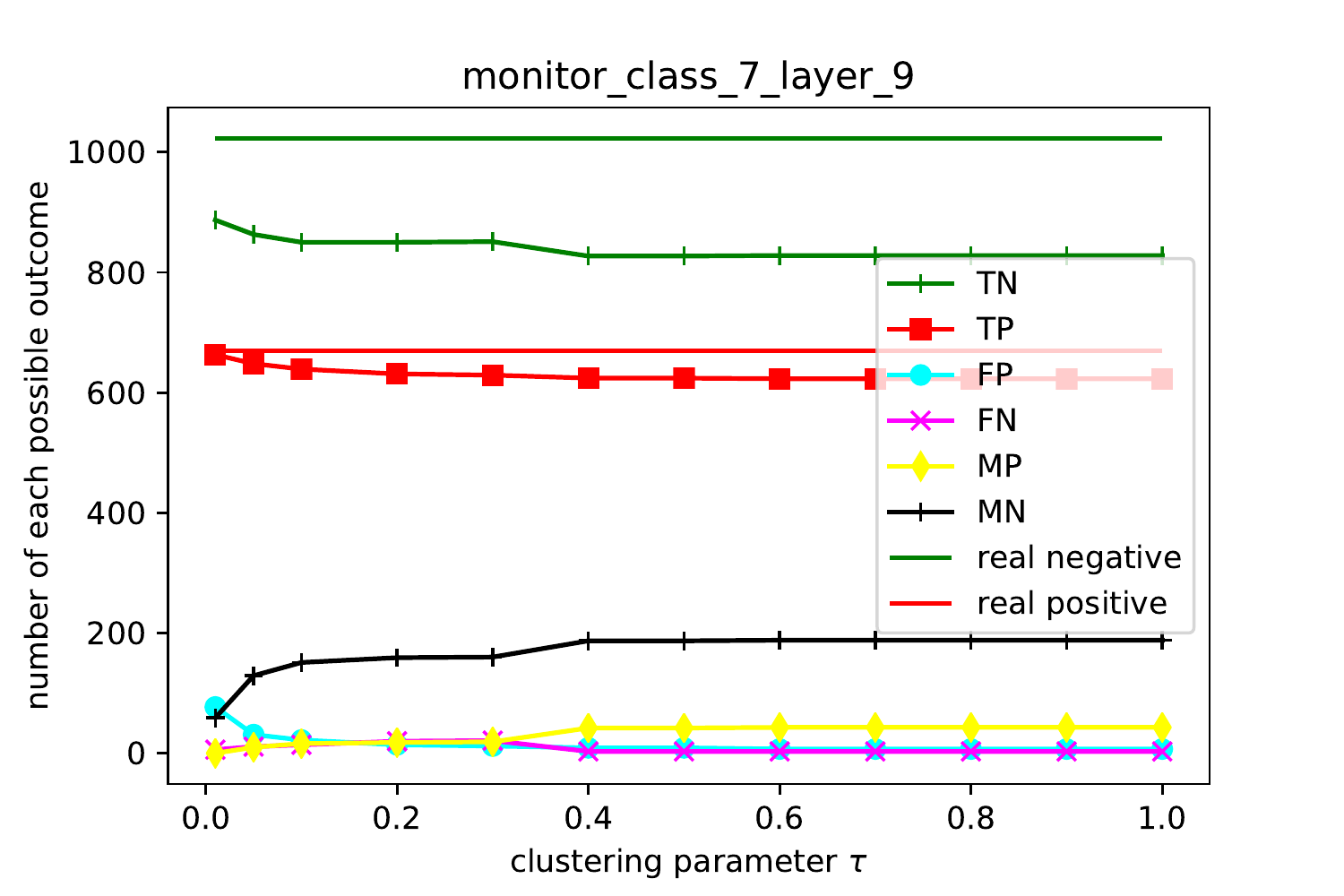}
    \end{subfigure}

    \begin{subfigure}[htbp]{0.245\textwidth}
        \centering
        \includegraphics[width=\textwidth]{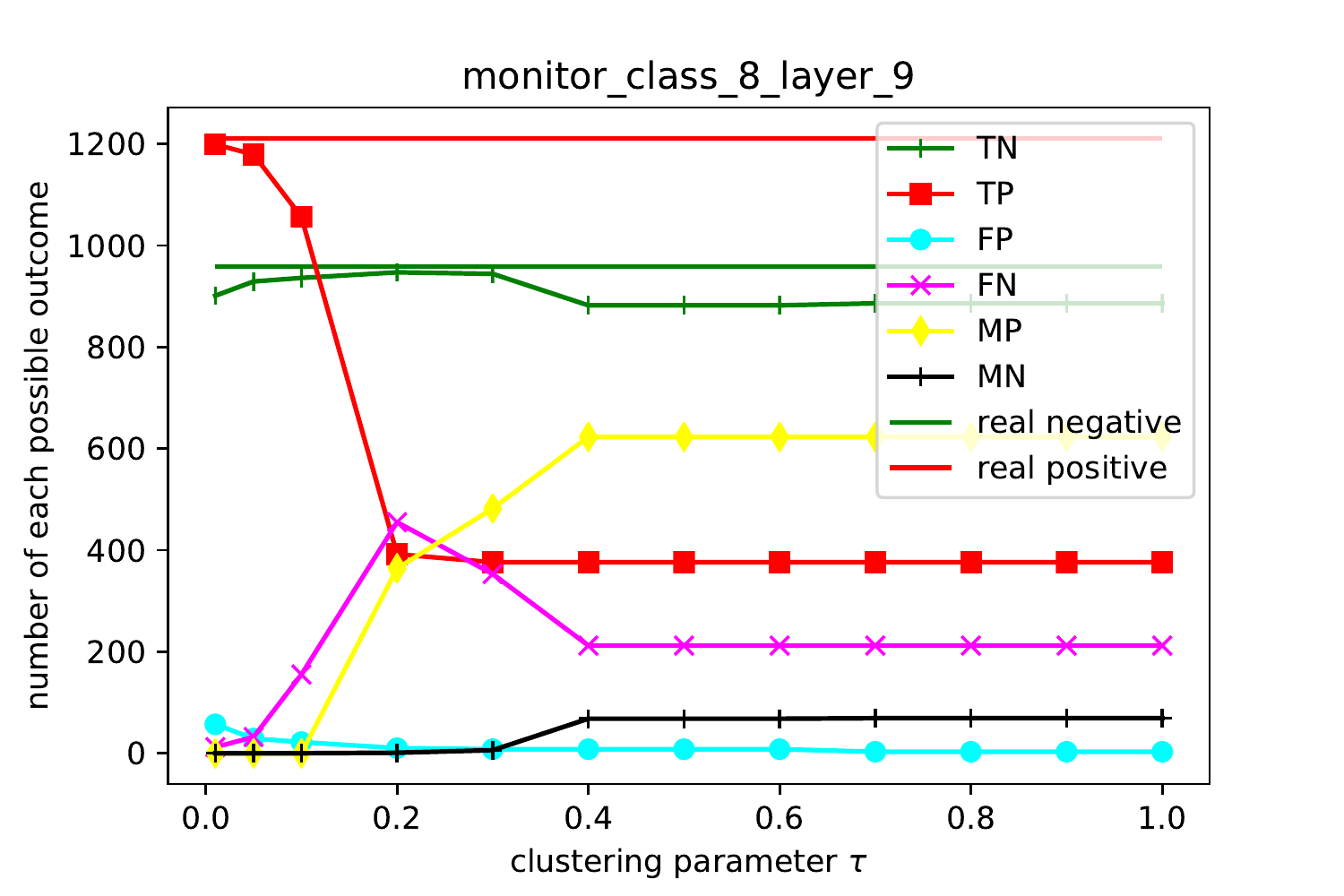}
    \end{subfigure}
    \hfill
    \begin{subfigure}[htbp]{0.245\textwidth}
        \centering
        \includegraphics[width=\textwidth]{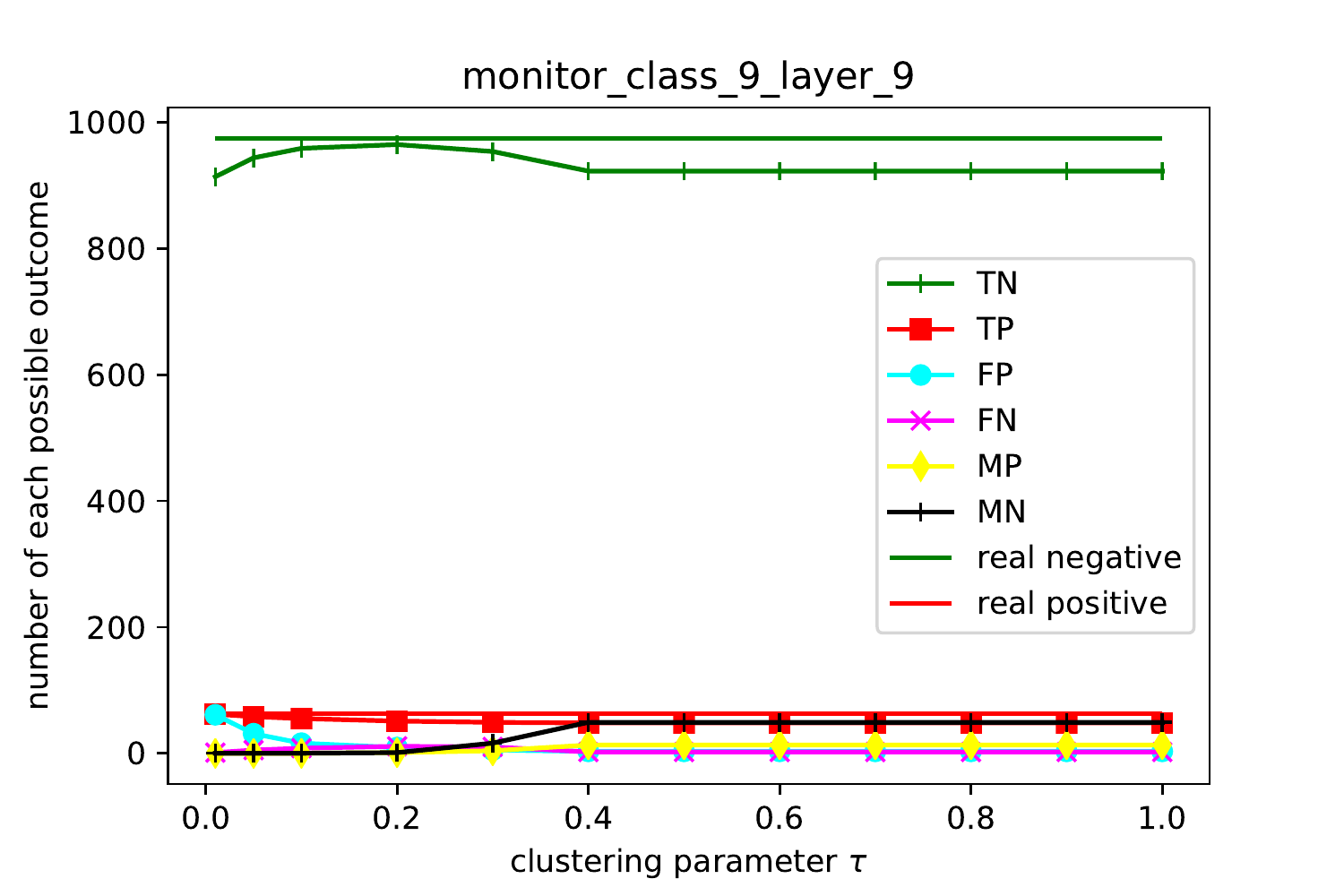}
    \end{subfigure}

    \caption{The numbers of different outcomes in Table~\ref{table:monitorPerformance} for $10$ monitors built at the output layer for benchmark MNIST.}
    \label{fig:verdictsMNIST}
\end{figure*}

\paragraph{Monitor effectiveness vs clustering parameter.}
In this experiment, we evaluate the effectiveness of the monitor according to different values of the clustering parameter $\tau$.
Due to space limitation, in Fig.~\ref{fig:verdictsMNIST}, we only present the results of $120$ out of $480$ monitors, representing $12$ configurations of the clustering parameter for each of the $10$ output classes.
Monitors watch the last layer of the network.
Each graph contains a curve depicting the number of occurrences for each of the six possible outcomes defined in Table~\ref{table:monitorPerformance} and the number of real positive and negative samples, according to the values of the clustering parameter.
As can be seen, by changing the value of the clustering parameter, one can directly control the sensitivity of the monitor to report an abnormal input.
This supports both the proposal of clustering the high-level features into smaller clusters before computing a global abstraction for them and also the idea of selecting a clustering parameter specific for each output-class monitor.

Observing the graphs more closely, we can make the following observations.
\begin{enumerate}
  \item The number of TN is only affected when the value of $\tau$ is very small, in most cases less than $0.1$.
        Consequently, the abstractions for good high-level features are too coarse in most cases.
  \item In most cases, the number of uncertainties is zero or reduced to zero by setting a small value of $\tau$.
        This indicates that our used network has a good separability of classification and also that the abstractions for good and bad features are precise enough.
        However, we note that the capability of reducing uncertainty to zero may stem from over-precise abstractions, i.e., a feature cluster containing only one feature making the abstractions pointless.
  \item In terms of errors of the monitors, the number of FP is very small, meaning that the probability of mis-classifying negative samples is very low, while the number of FN is very high.
        However, the number of FN can be greatly reduced.
        This indicates that the detection of positive samples is not good enough due to the coarseness of abstraction.
  \item The number of TP can be always augmented by shrinking the abstraction.
  \item The monitors for different output classes have different degrees of sensitivity to parameter $\tau$.
        For example, the performance of monitors for classes $3$, $5$, $6$, and $9$ does not depend on the value of $\tau$.
\end{enumerate}

\paragraph{Clustering parameter tuning.}
Clearly, the monitor effectiveness depends the clustering parameter.
One can adjust parameter $\tau$ to improve the monitor detection capability of positive samples, i.e., abnormal inputs.
When adjusting $\tau$, one can observe the following elements:
i) the clustering coverage estimation to decide whether the ``white space" has been sufficiently removed;
ii) the numbers of uncertainties (MN and MP in Table~\ref{table:monitorPerformance}) to determine if they can be reduced into zeros, which determines whether the good and bad features can be well separated by box abstractions;
iii) the number changes of TN and TP when $\tau$ is decreased after reducing the numbers of uncertainties into zeros (i.e., abstractions for good and bad features have been non-overlapping).
For instance, if one observes that the number of TN decreases while the number of TP increases in this case, it means that the box abstractions are not able to distinguish the real negative and positive samples.
This can be a signal to investigate the appropriateness of the box shape to abstract the features or the network separability, since there exists some region where real negative and positives samples are entangled.
Furthermore, the sensitivity of the dependence to the clustering parameter differs among output classes, so it is desirable to fine-tune the clustering parameter for each class separately.
This also applies to the monitors built in different layers because the features used to construct a monitor are totally different.
%


\begin{figure*}[htbp]
    \centering
    \begin{subfigure}[htbp]{0.245\textwidth}
        \centering
        \includegraphics[width=\textwidth]{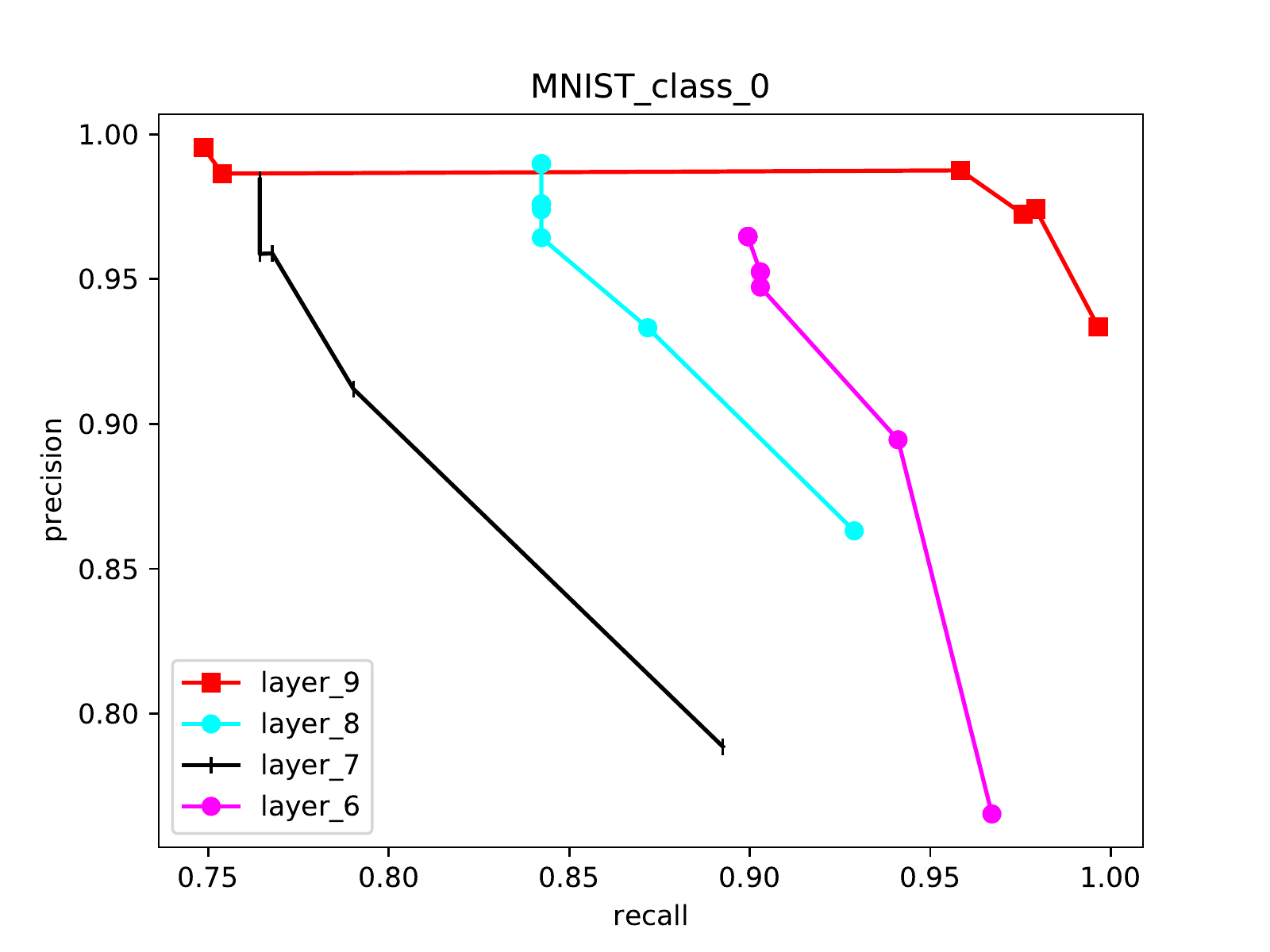}
    \end{subfigure}
    \hfill
    \begin{subfigure}[htbp]{0.245\textwidth}
        \centering
        \includegraphics[width=\textwidth]{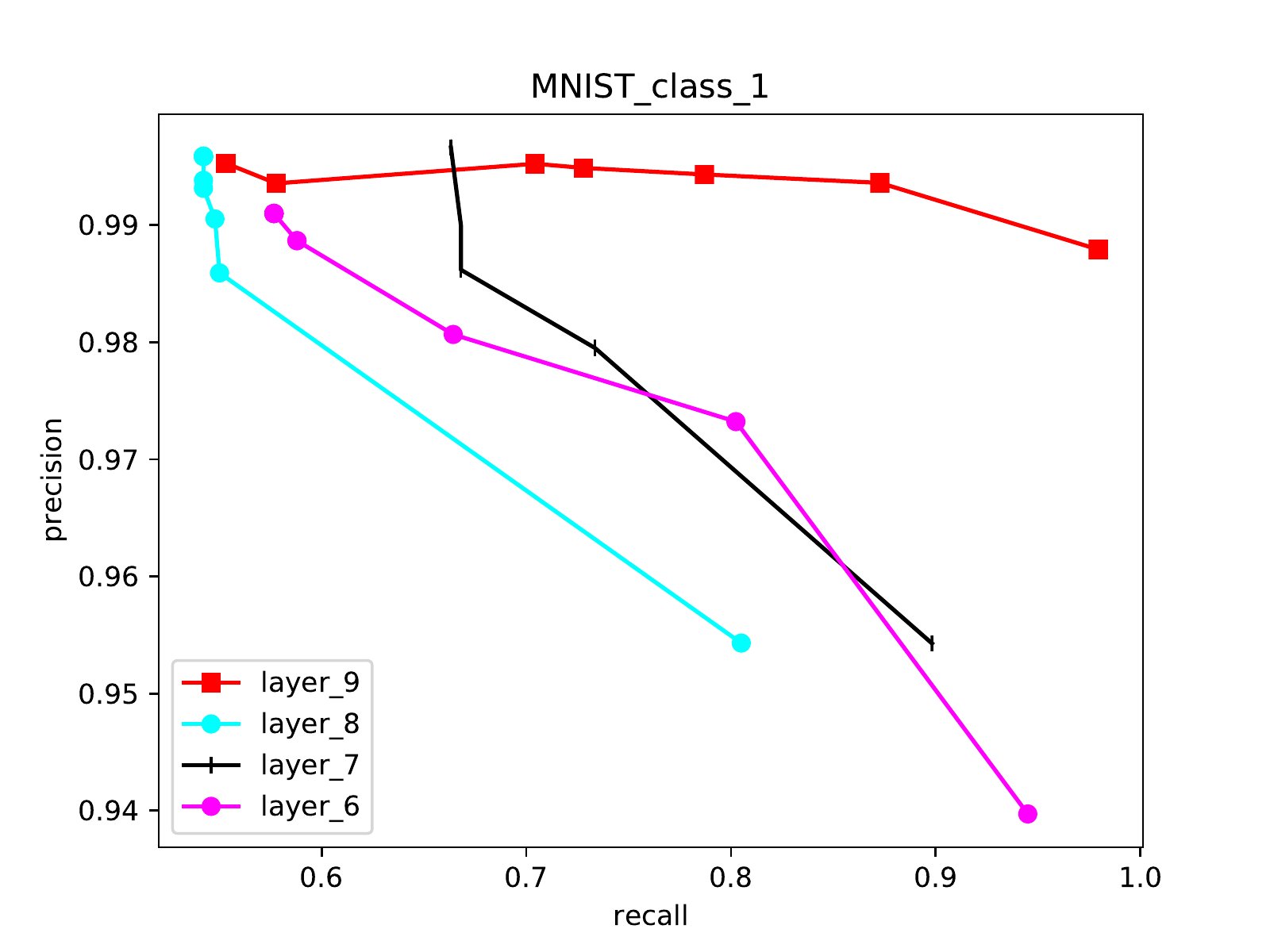}
    \end{subfigure}
    \hfill
    \begin{subfigure}[htbp]{0.245\textwidth}
        \centering
        \includegraphics[width=\textwidth]{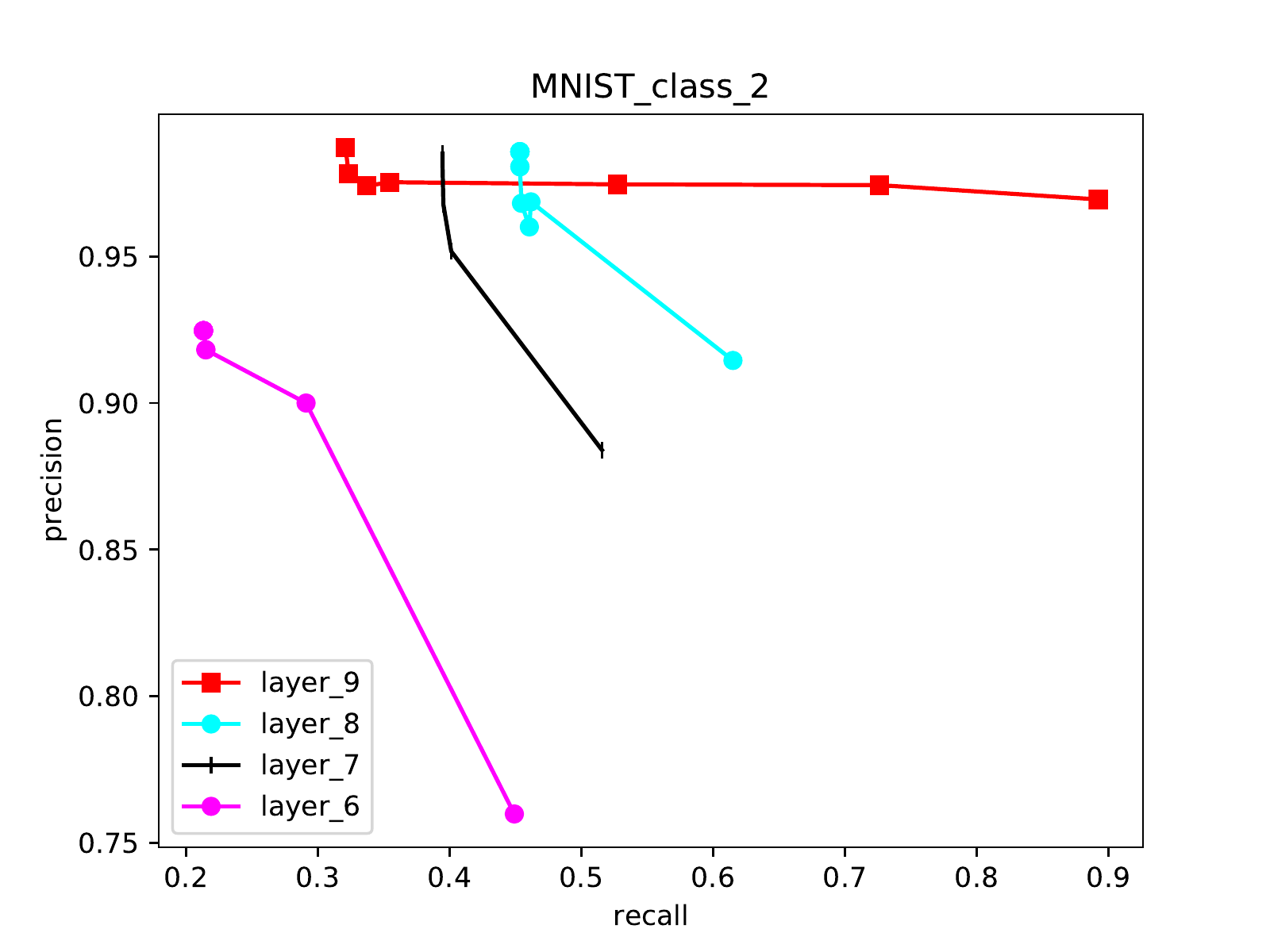}
    \end{subfigure}
    \hfill
    \begin{subfigure}[htbp]{0.245\textwidth}
        \centering
        \includegraphics[width=\textwidth]{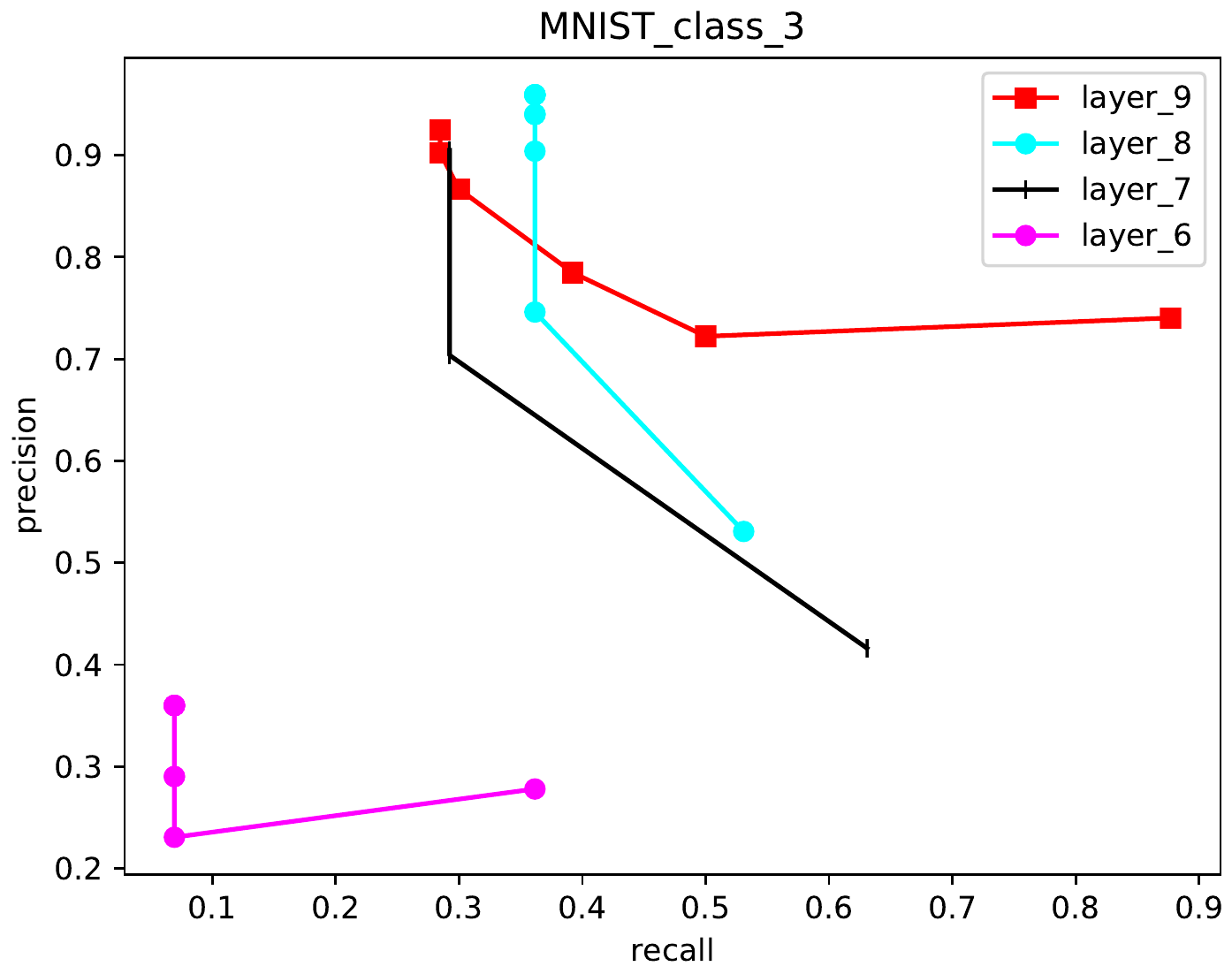}
    \end{subfigure}

    \begin{subfigure}[htbp]{0.245\textwidth}
        \centering
        \includegraphics[width=\textwidth]{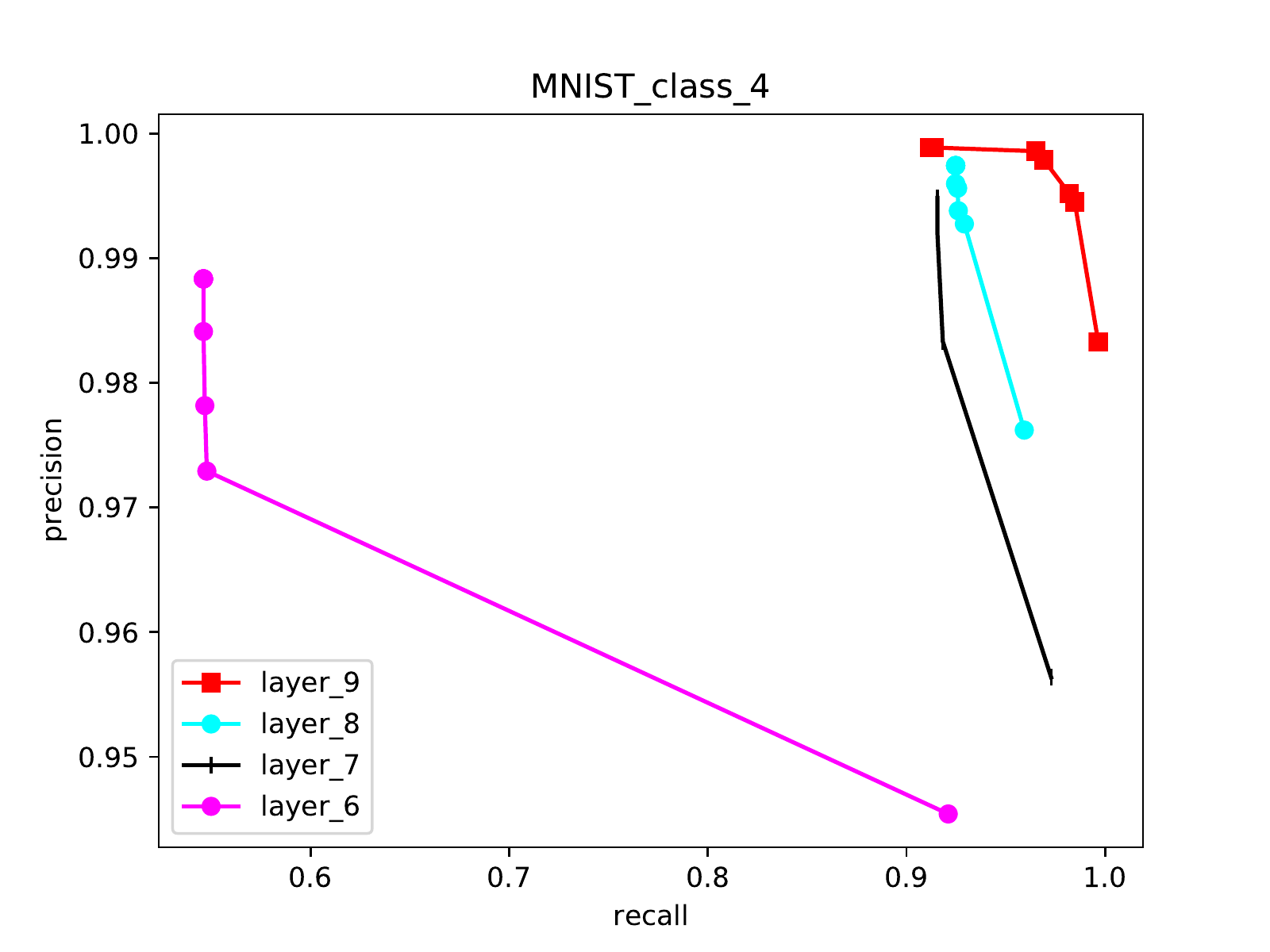}
    \end{subfigure}
    \hfill
    \begin{subfigure}[htbp]{0.245\textwidth}
        \centering
        \includegraphics[width=\textwidth]{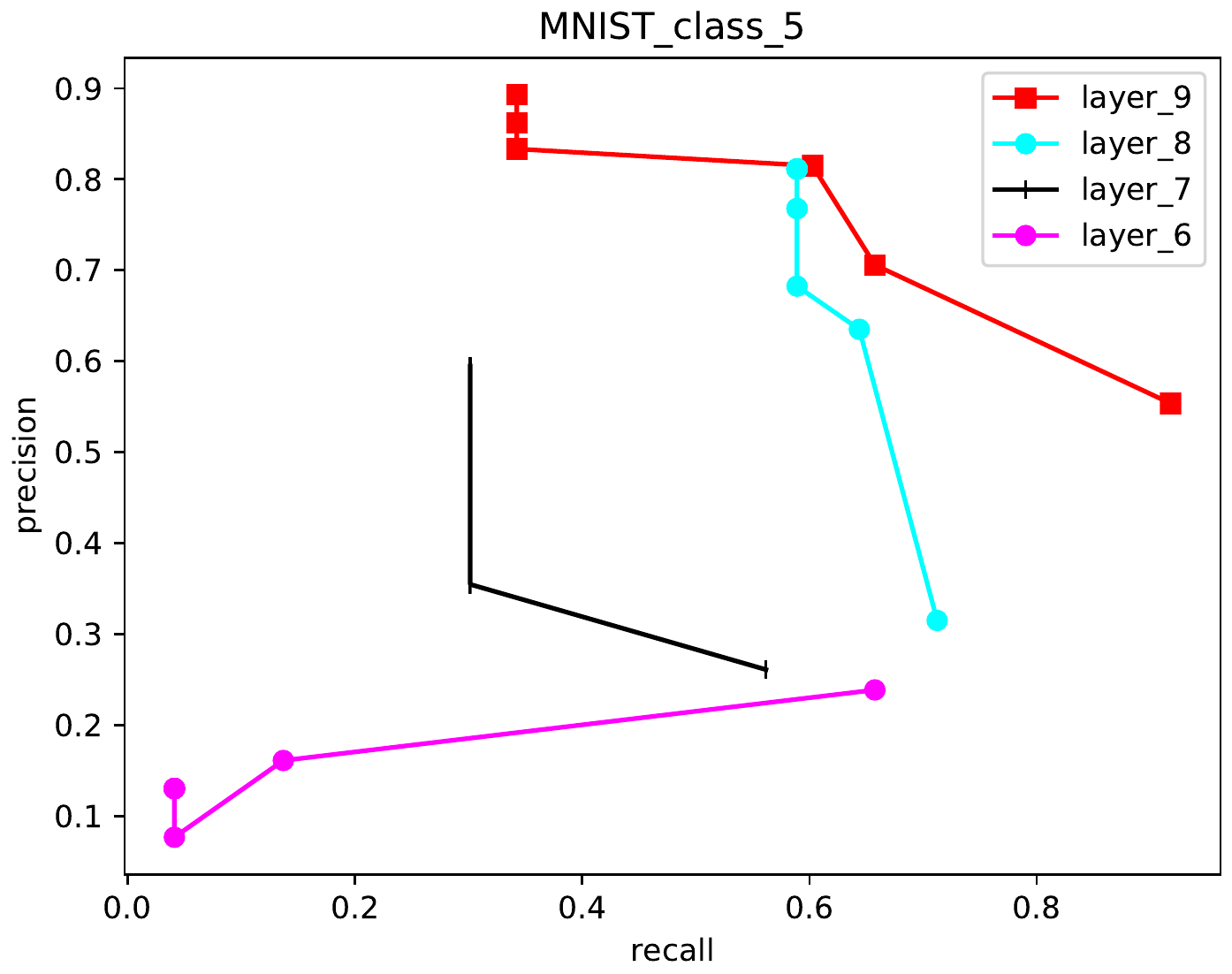}
    \end{subfigure}
    \hfill
    \begin{subfigure}[htbp]{0.245\textwidth}
        \centering
        \includegraphics[width=\textwidth]{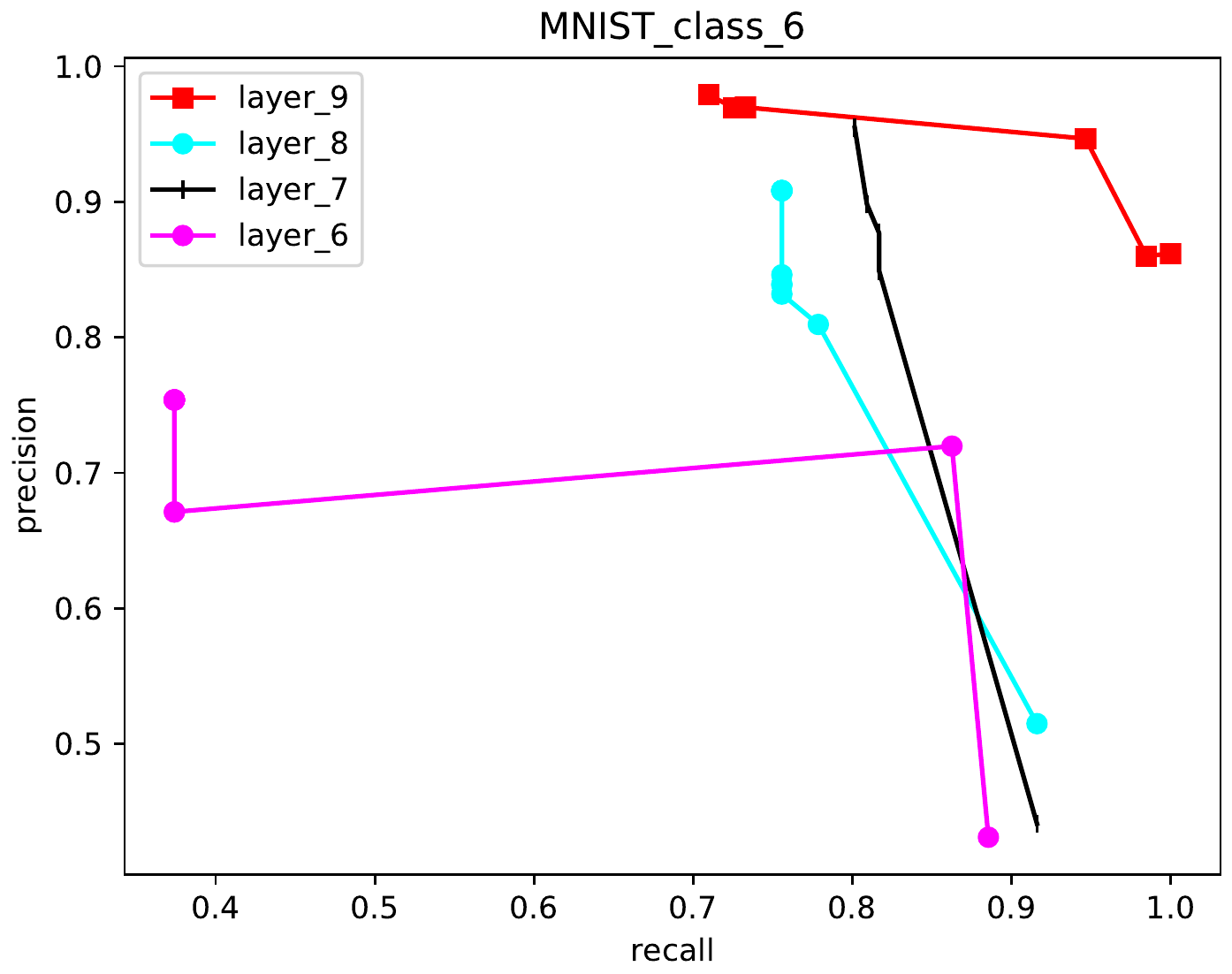}
    \end{subfigure}
    \hfill
    \begin{subfigure}[htbp]{0.245\textwidth}
        \centering
        \includegraphics[width=\textwidth]{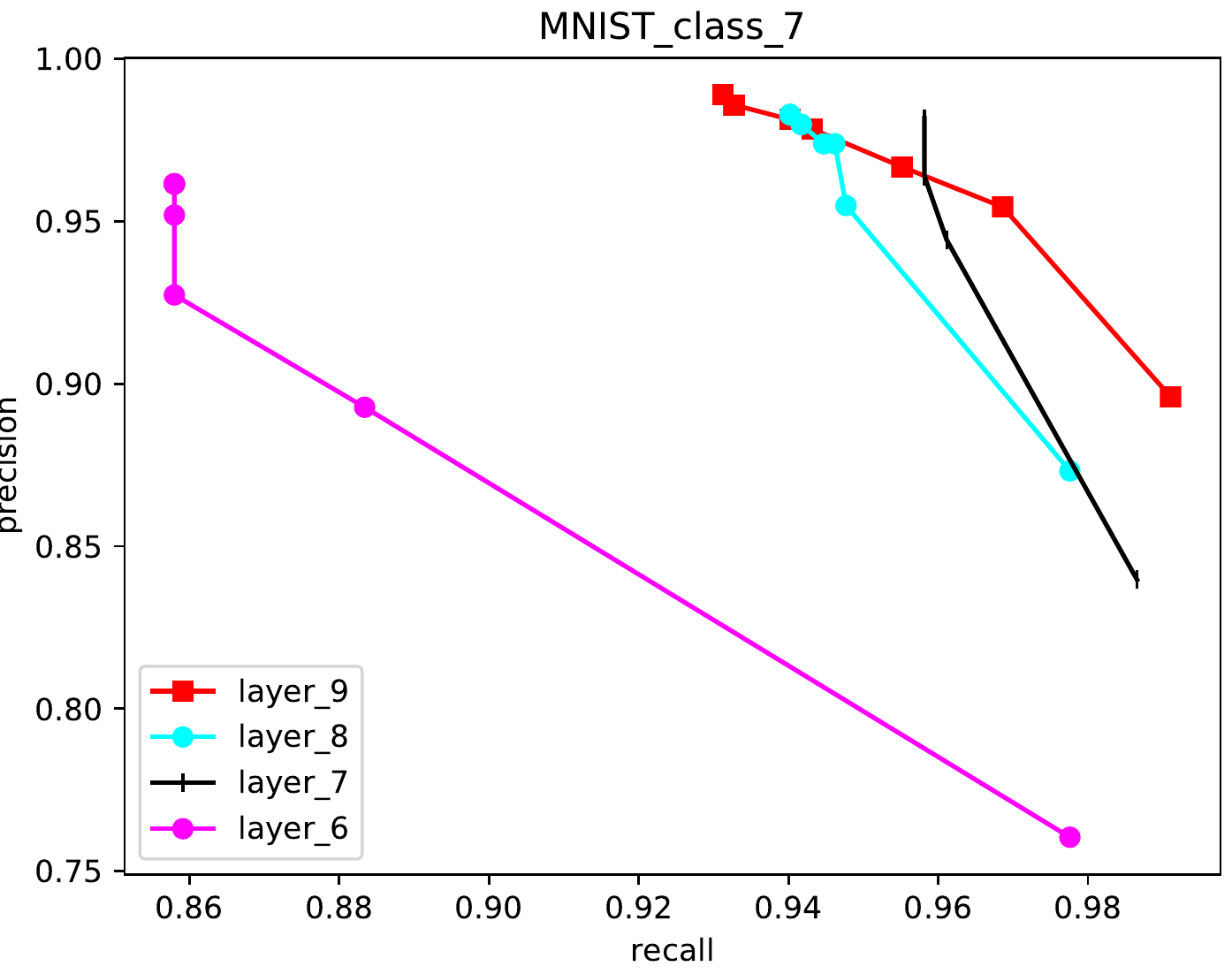}
    \end{subfigure}

    \begin{subfigure}[htbp]{0.245\textwidth}
        \centering
        \includegraphics[width=\textwidth]{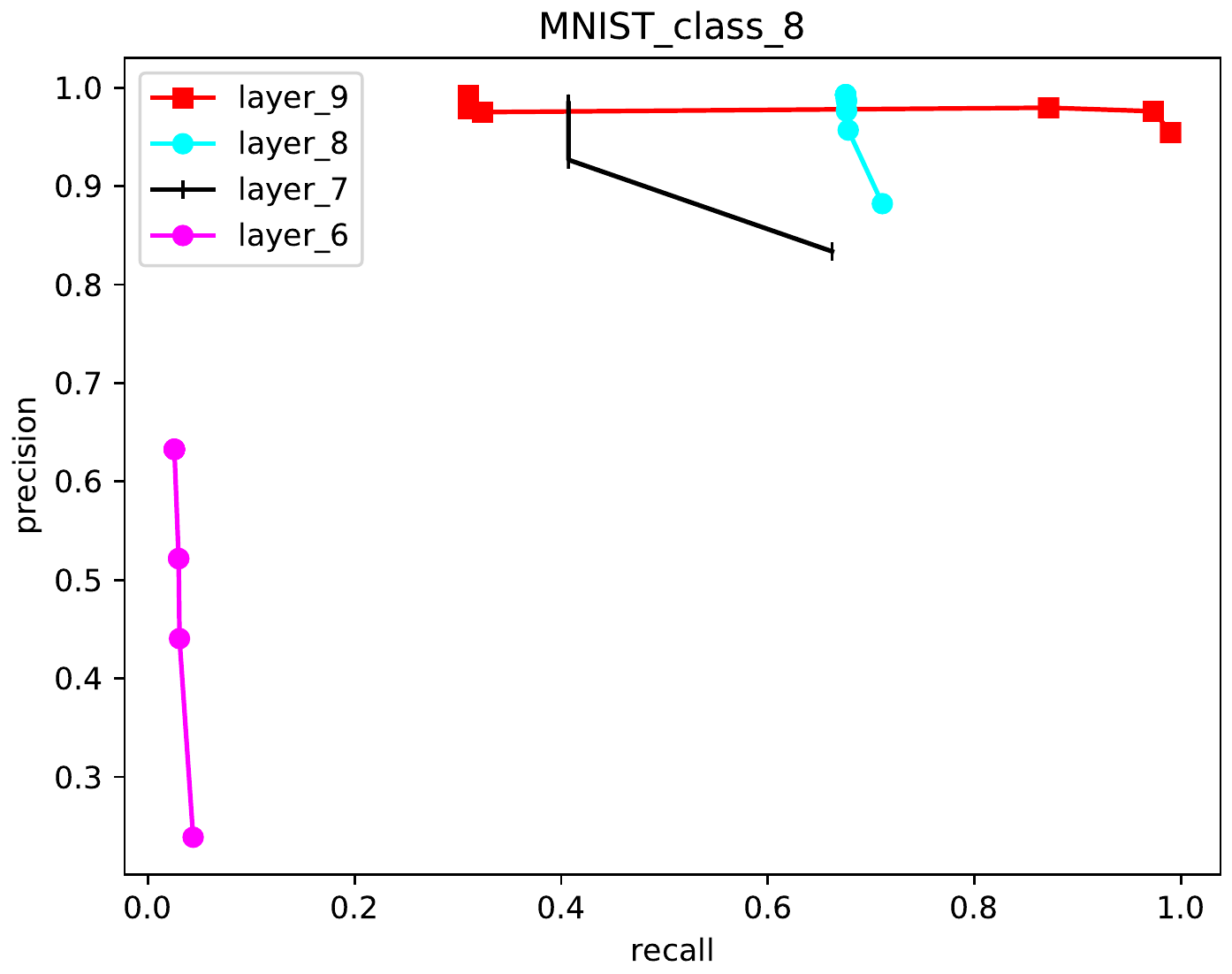}
    \end{subfigure}
    \hfill
    \begin{subfigure}[htbp]{0.245\textwidth}
        \centering
        \includegraphics[width=\textwidth]{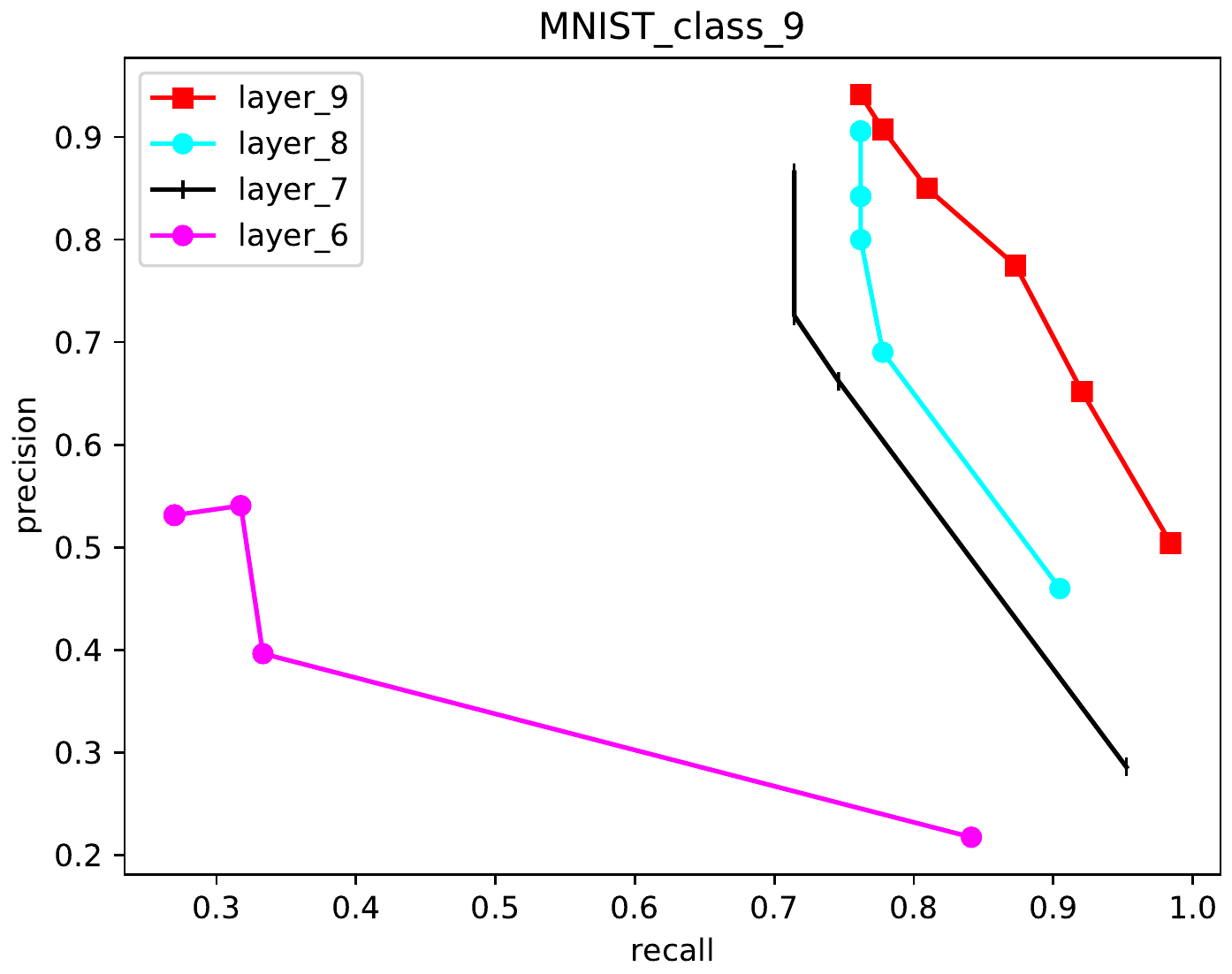}
    \end{subfigure}
    \hfil

    \caption{Precision-Recall curves for monitors built on benchmark MNIST.}
    \label{fig:PRcurveMNIST}
\end{figure*}

\paragraph{Monitor precision vs monitored layer.}
We use precision and recall to study the relationship between the monitor effectiveness and the monitored layer.
We represent precision-recall curves which show the tradeoff between precision and recall for different values of the clustering parameter $\tau$.
\emph{Precision or positive predictive value (PPV)} is defined as the number of TP over the total number of predictive positive samples, i.e., $\mathit{precision} = \frac{TP}{TP+FP}$, while \emph{recall or true positive rate (TPR)} is defined as the number of TP over the total number of real positive samples, i.e., $\mathit{recall} = \frac{TP}{TP+FN+MP}$.
A large area under the curve indicates both a high recall and high precision.
High precision indicates a low false positive rate, and high recall indicates a low false negative rate.
High scores for both show that the classifier is returning accurate results (high precision), as well as returning a majority of all positive results (high recall).
Based on this, by examining the precision-recall curves shown in \figref{fig:PRcurveMNIST}, we can see that for benchmark MNIST, it is better to monitor the output layer, as one can achieve high precision and recall in most cases.
\subsection{Discussion and Lessons Learned}
Our experiments were conducted on a Windows PC (Intel(R) Core(TM) i$7$-7600U CPU @2.80 GHz, with $8$ GB RAM) and the implementation is available \footnote{\url{https://gricad-gitlab.univ-grenoble-alpes.fr/rvai-public/abstraction_based_monitors}}.
The construction and test of $480$ monitors on benchmark MNIST took, respectively, $2,473$ and $85$ seconds.
Apart from MNIST, we have also tried out the same experiments on F\_MNIST and CIFAR10~\cite{krizhevsky2009learning} and report the results in the appendix.

Thanks to the overall results on different benchmarks, we are more convinced that it is necessary to partition (e.g., clustering here) the high-level used features before computing a global abstraction for them, since on all tried benchmarks there exist many uncertainties (missed positives and negatives).
These uncertainties are due to the overlapping between the abstractions built from the good and bad features when these features are not well partitioned, and can be removed when well partitioned, e.g., tuning the clustering parameter into a small value in this paper.
In addition to partitioning the features via the clustering method, we suggest that the clustering parameter should be customized for each output-class of a network at different layers, even for the good and bad features used to construct a monitor for same output-class.

\begin{figure*}[htbp]
    \centering
    \begin{subfigure}[htbp]{0.245\textwidth}
        \centering
        \includegraphics[width=\textwidth]{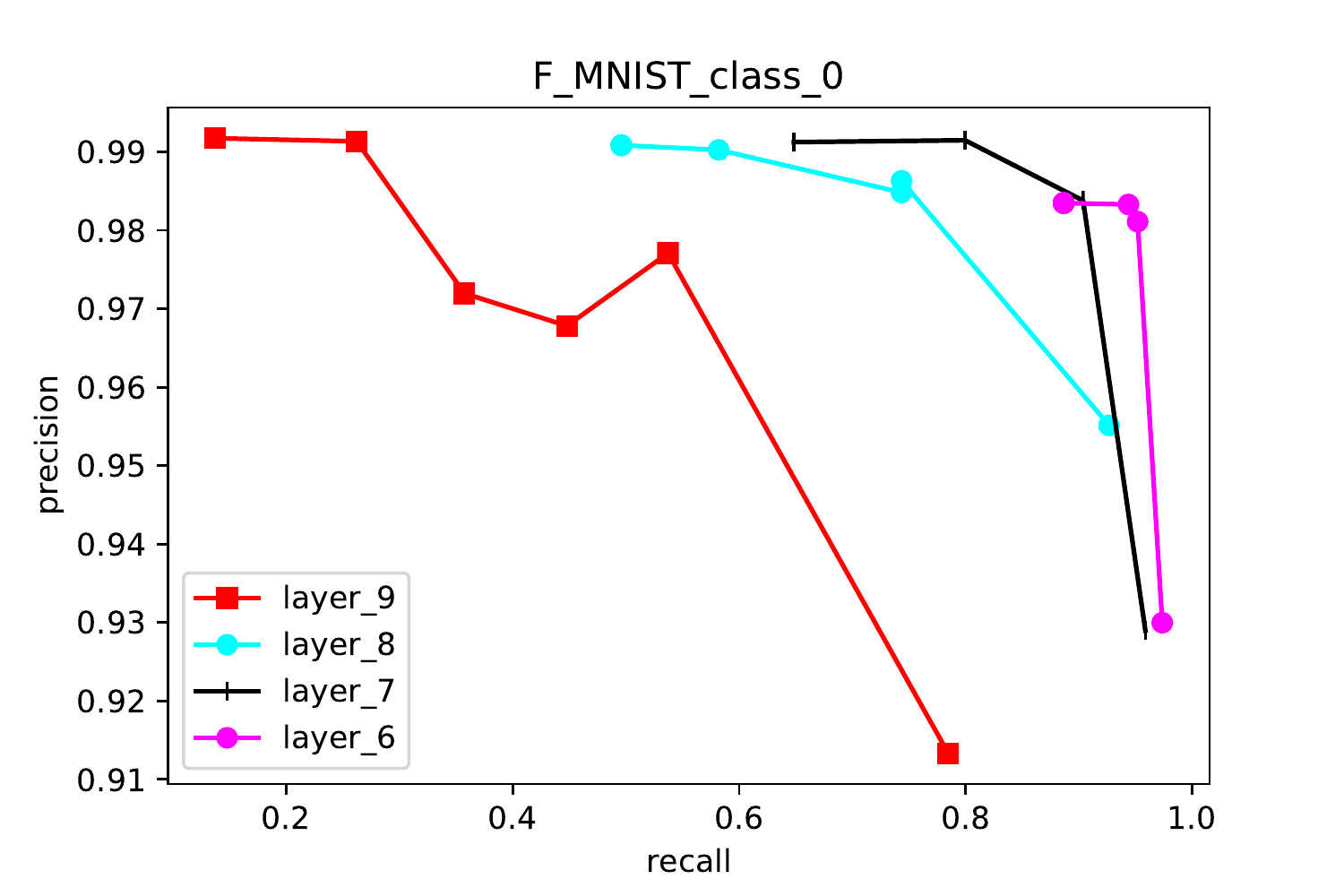}
    \end{subfigure}
    \hfill
    \begin{subfigure}[htbp]{0.245\textwidth}
        \centering
        \includegraphics[width=\textwidth]{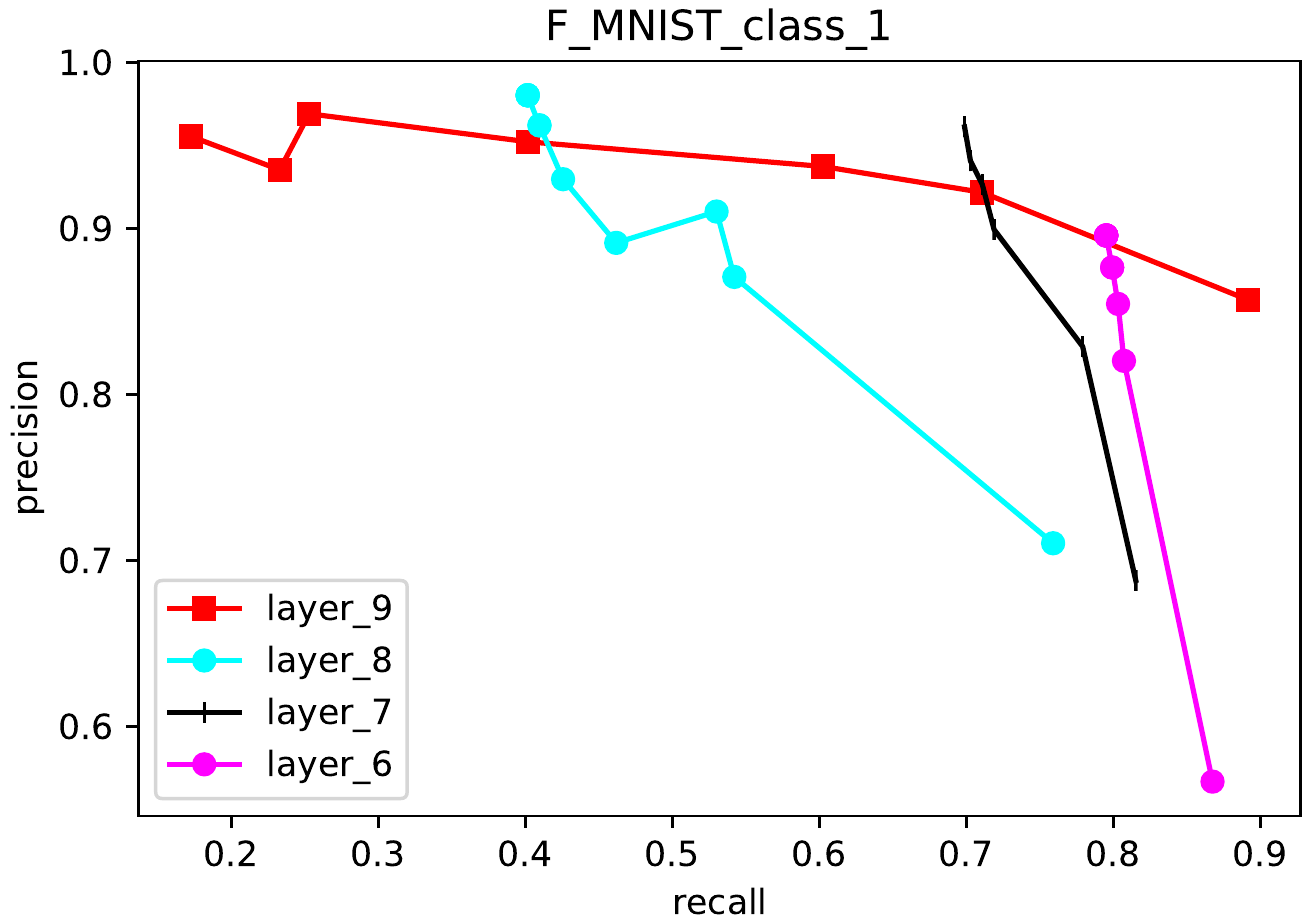}
    \end{subfigure}
    \hfill
    \begin{subfigure}[htbp]{0.245\textwidth}
        \centering
        \includegraphics[width=\textwidth]{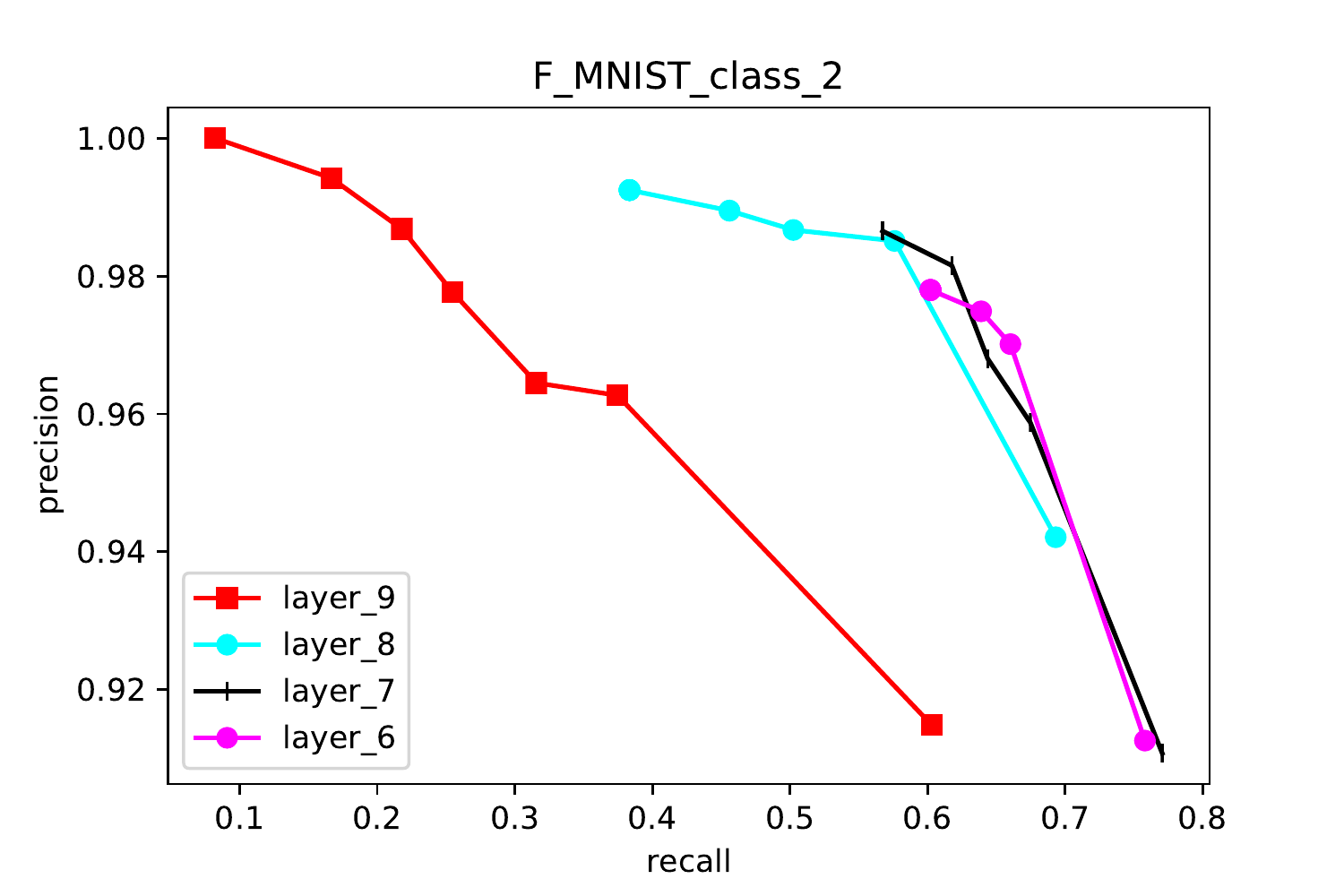}
    \end{subfigure}
    \hfill
    \begin{subfigure}[htbp]{0.245\textwidth}
        \centering
        \includegraphics[width=\textwidth]{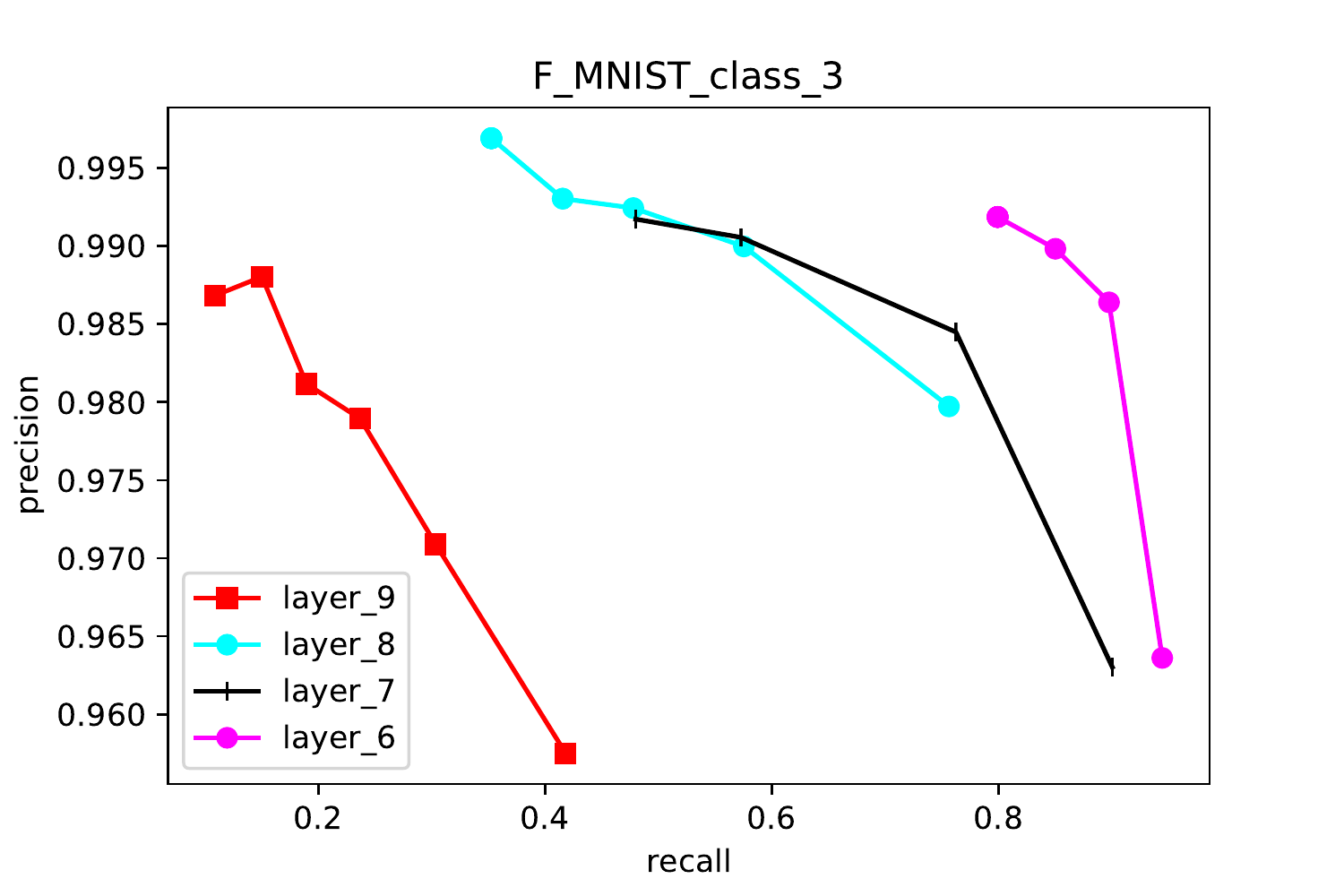}
    \end{subfigure}

    \begin{subfigure}[htbp]{0.245\textwidth}
        \centering
        \includegraphics[width=\textwidth]{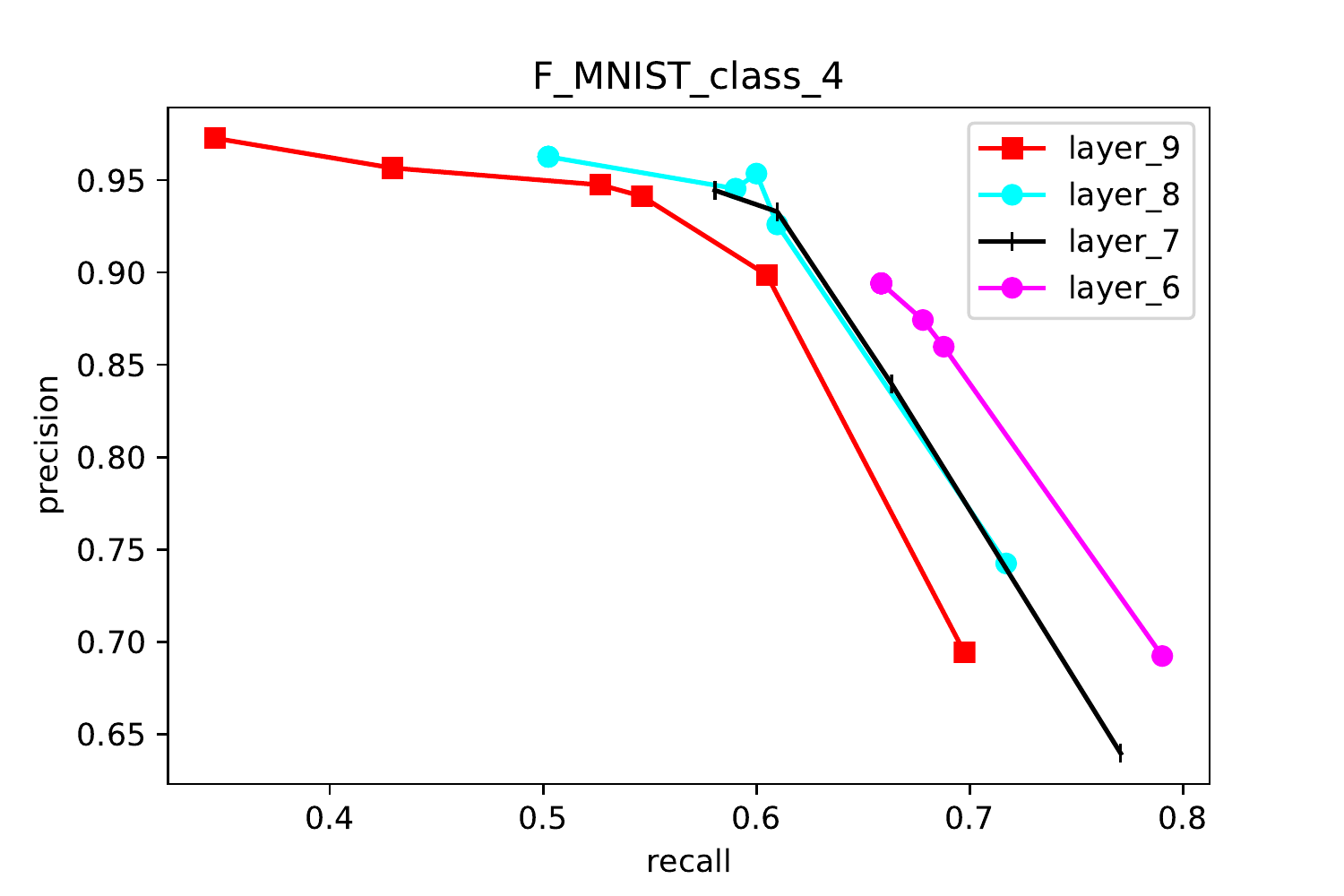}
    \end{subfigure}
    \hfill
    \begin{subfigure}[htbp]{0.245\textwidth}
        \centering
        \includegraphics[width=\textwidth]{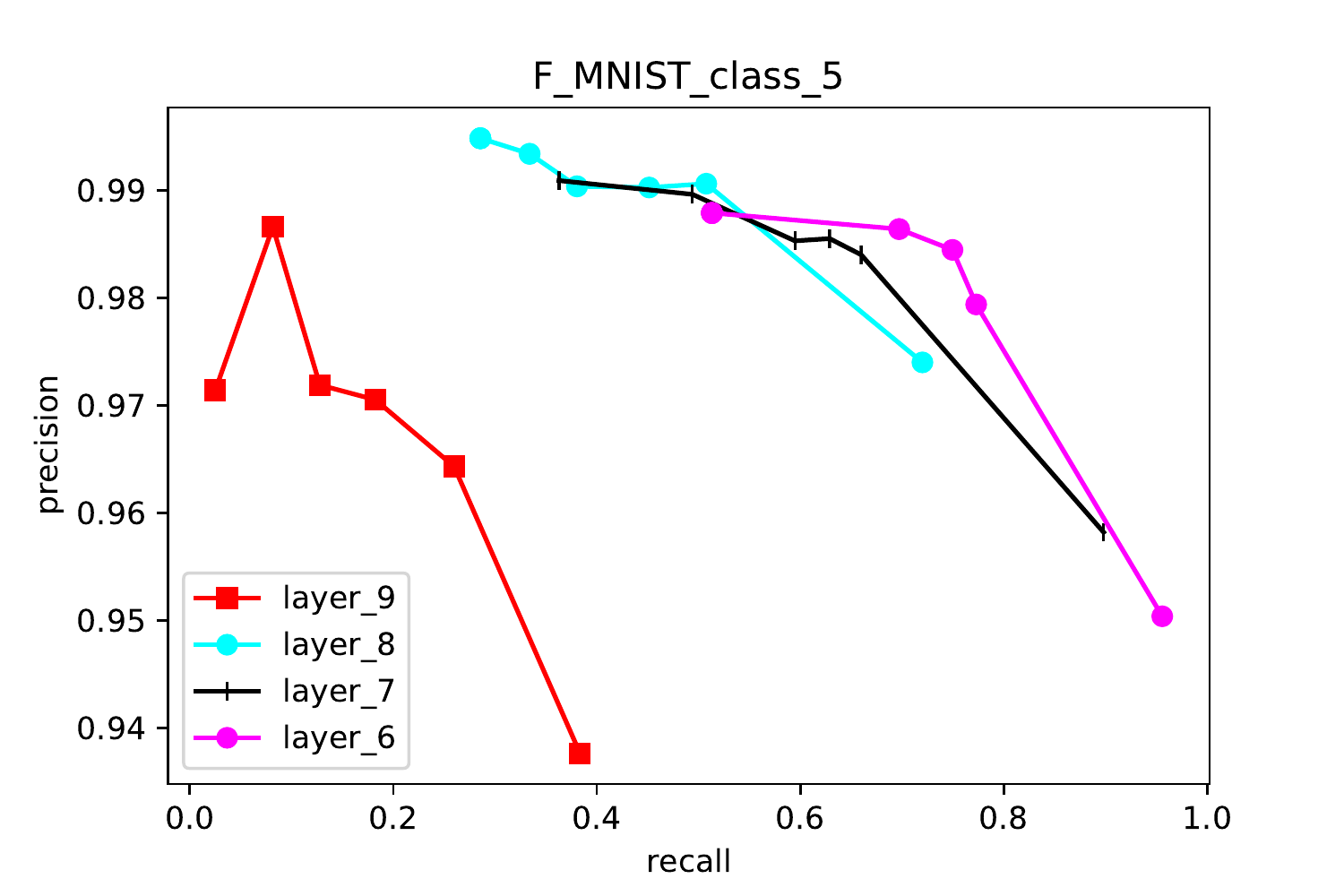}
    \end{subfigure}
    \hfill
    \begin{subfigure}[htbp]{0.245\textwidth}
        \centering
        \includegraphics[width=\textwidth]{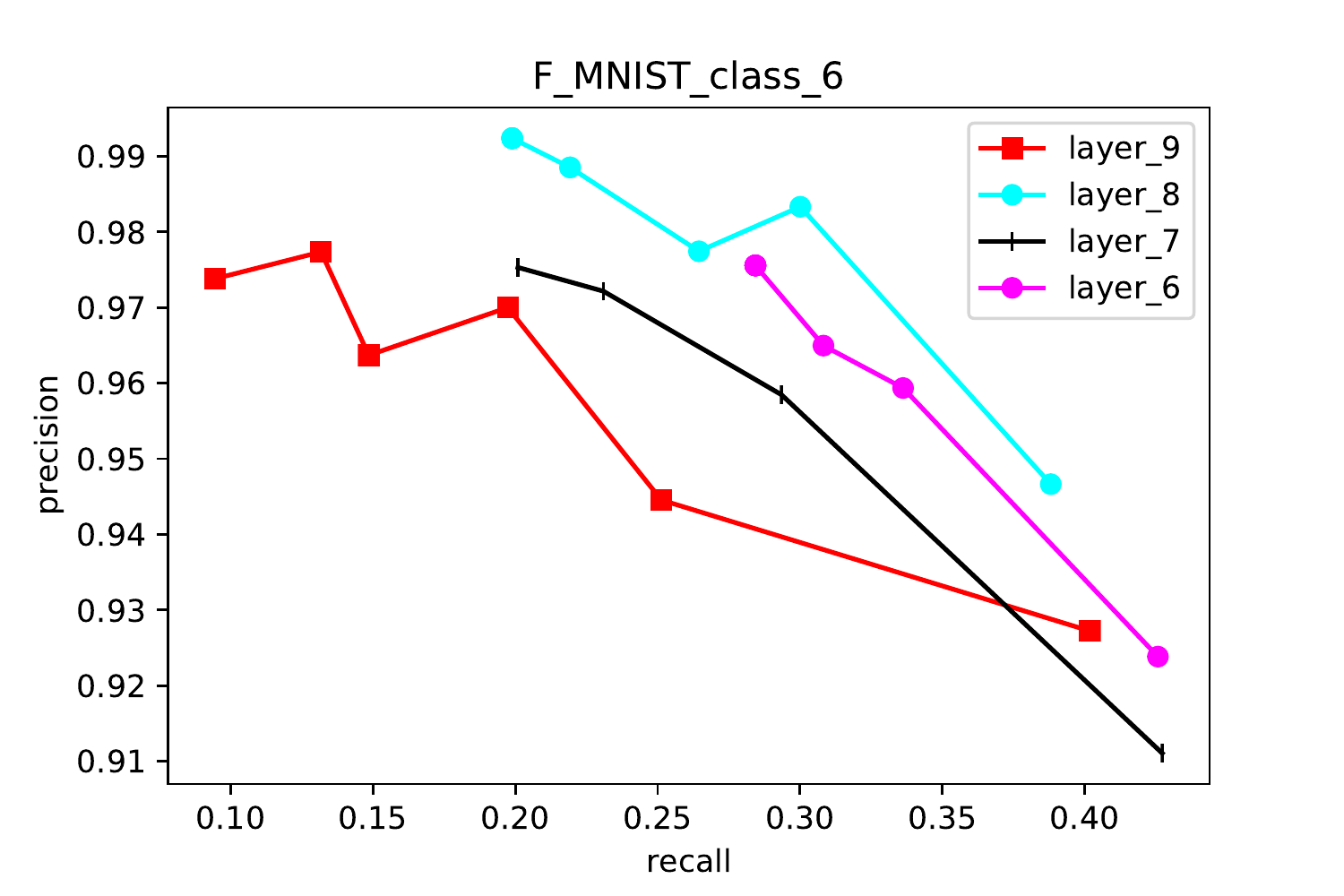}
    \end{subfigure}
    \hfill
    \begin{subfigure}[htbp]{0.245\textwidth}
        \centering
        \includegraphics[width=\textwidth]{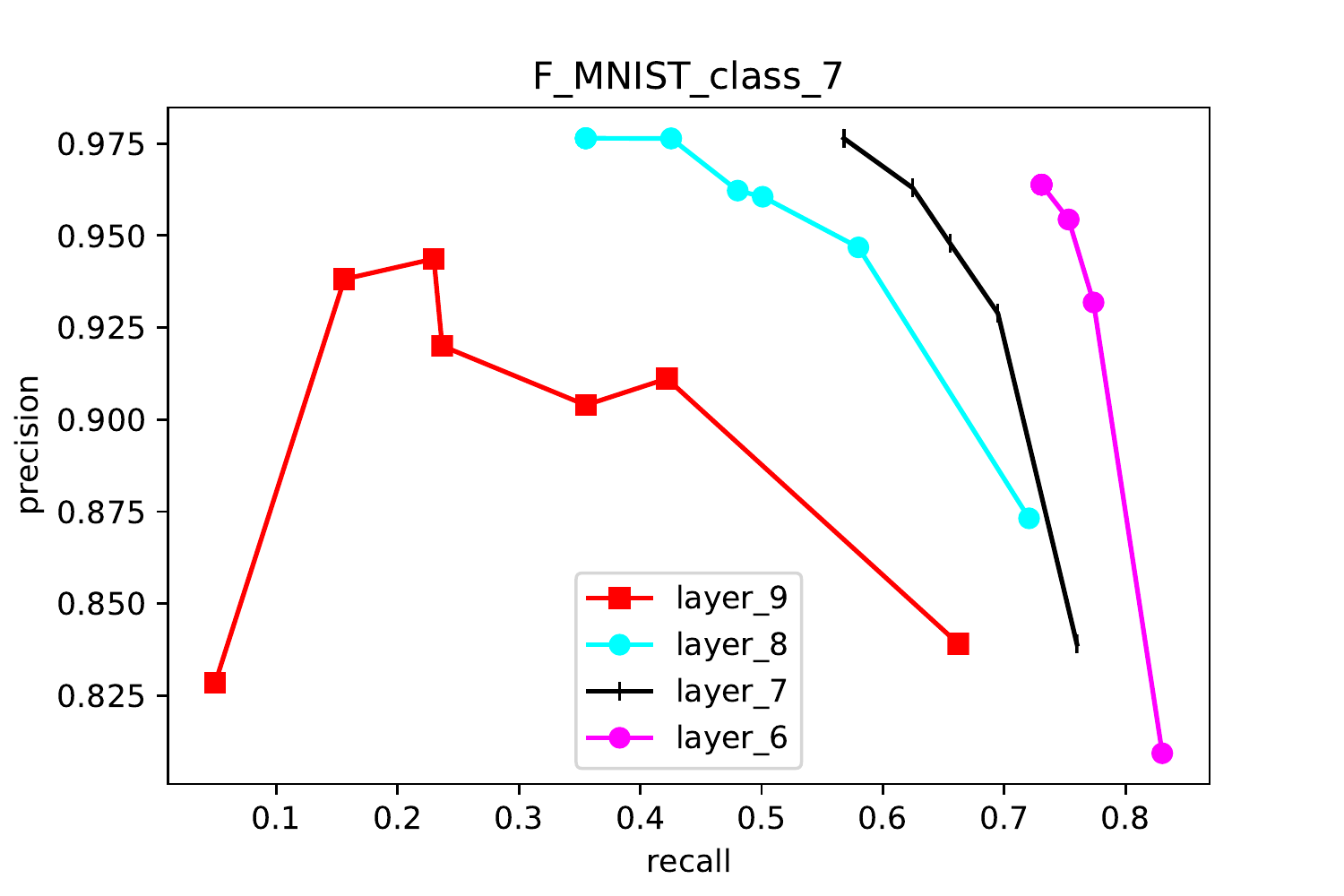}
    \end{subfigure}

    \begin{subfigure}[htbp]{0.245\textwidth}
        \centering
        \includegraphics[width=\textwidth]{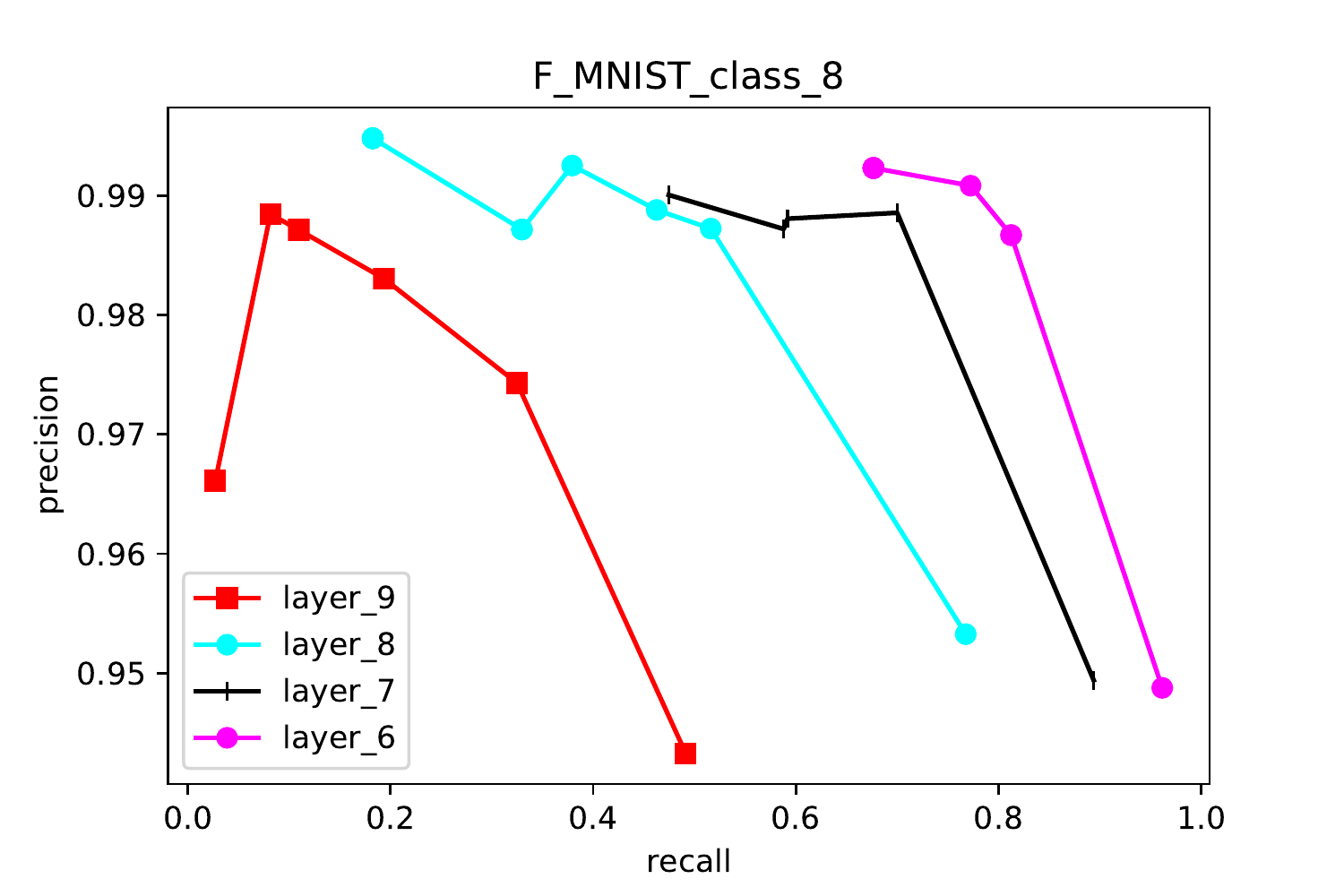}
    \end{subfigure}
    \hfill
    \begin{subfigure}[htbp]{0.245\textwidth}
        \centering
        \includegraphics[width=\textwidth]{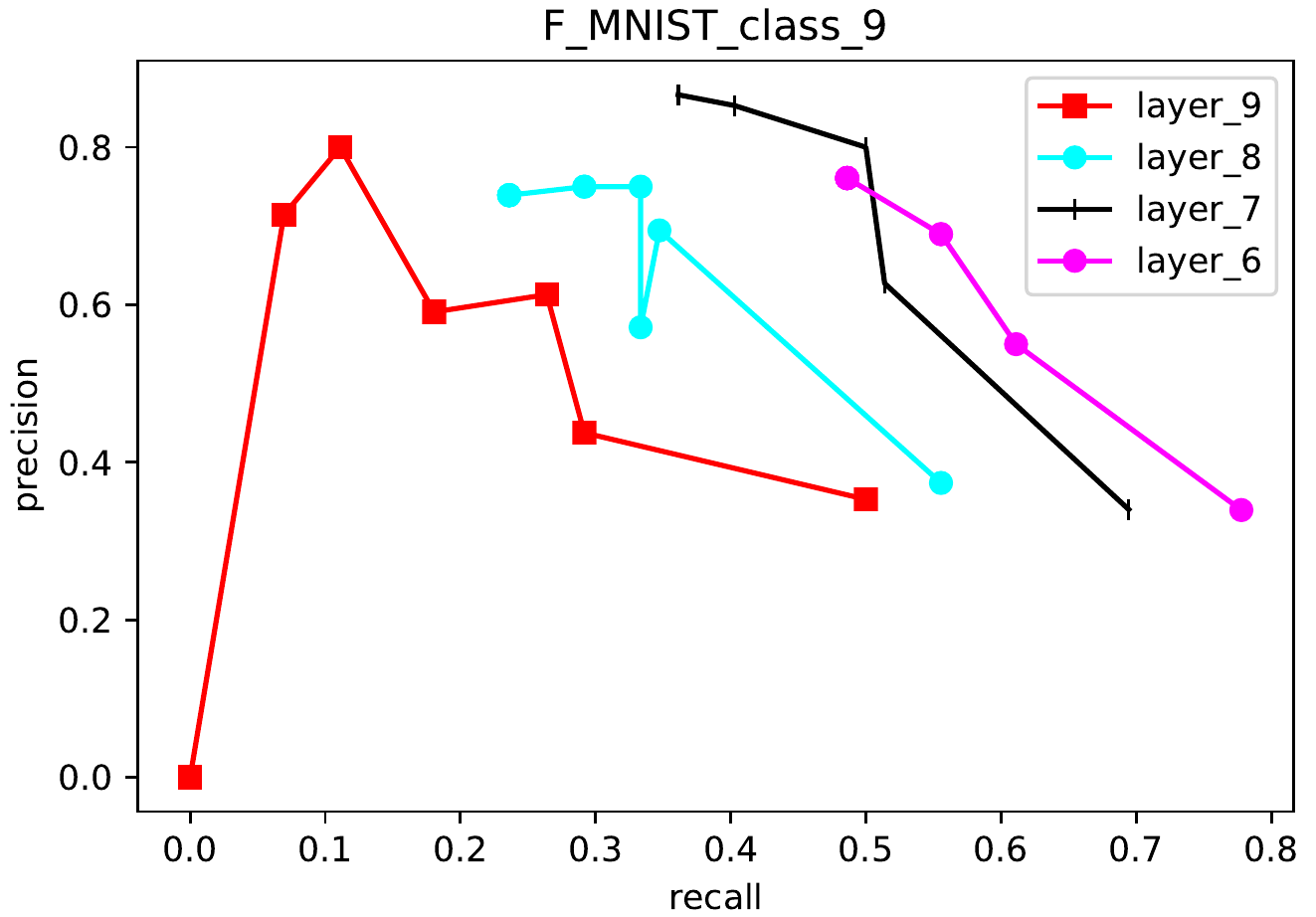}
    \end{subfigure}

    \caption{Precision-Recall curves for monitors built on benchmark F\_MNIST.}
    \label{fig:PRcurveFMNIST}
\end{figure*}

However, we point out that the optimal monitored layer is not always the output layer, since we observed on benchmark F\_MNIST that in most cases the monitor's performance at the fourth to the last layer is superior than that ones at the last three layers by observing their precision-recall curves in Fig.~\ref{fig:PRcurveFMNIST}.
Until now, no general law can be given to predict such optimal monitored layer and we believe that it makes the method presented in this paper important to best configure monitors for a given classification system.

Last but not least, the success of using abstraction-based monitors for networks relies on three pivotal factors:
i) accuracy and separability of classification of the monitored network, which determines the reliability of the system itself and can not be improved by added monitors;
ii) sufficiency of data for constructing a monitor, else the abstraction will be not representative so that the monitor is not reliable;
iii) implementation error of network and monitors since the construction and membership query of box abstractions demand precise computation.

%

\label{sec:experiments}


\section{Related Work}
\label{sec:rw}

There is a myriad of approaches and formalisms for handling data in runtime verification such as QEA~\cite{BarringerFHRR12}, parametric trace slicing~\cite{abs-1112-5761}, see~\cite{HavelundRTZ18} for an overview.
However, these approaches essentially consist in augmenting a behavior-based specification formalism such as finite-state machines to handle data rather than verifying a data-oriented system

Hence, in this section, we compare with the approaches aiming at assessing the decisions of neural networks to improve the confidence (in safety-critical scenarios).

One of the first technique for this purpose is anomaly detection, which has been extensively studied in the areas of statistics and traditional machine learning; see~\cite{hodge2004survey,chandola2009anomaly,chalapathy2019deep} for surveys.
All recent anomaly detection approaches~\cite{geifman2017selective,guo2017calibration,liang2017enhancing,devries2018learning,jiang2018trust,thulasidasan2019combating} consist essentially in computing a confidence score over the network decisions.
Whenever a decision has a score below the required threshold, the decision is rejected and the input declared abnormal.
For instance, in the context of deep learning, \cite{hendrycks2016baseline} is a well-known method that calculates a confidence score for the network decision in terms of sample distribution, where the confidence score is the softmax prediction probability of decisions.

The above referred methods are based on a statistical approach.
Some more recent work~\cite{cheng2019runtime,henzinger2019outside,cheng2020provably,lukina2020into}, inspired from formal methods, follow the same purposes by building runtime monitors to supervise the decision of a network.
Such runtime monitoring approaches fundamentally differ from the traditional runtime verification and monitoring approaches for software and hardware systems (cf.~\cite{HavelundG05,LeuckerS09,FalconeHR13,series/lncs/10457}) in that (1) the monitors are not obtained from formal specifications and (2) monitors focus on the data output of the network instead of some property about the ordering of actions.
In these approaches, an input into a network is declared as an anomaly when the network decision is rejected by the monitor according to some references.
The verdict of a monitor on a new input is based on the membership test of the induced neuron activation pattern in a pre-established sound over-approximation of neuron activation patterns recorded from network correct decisions constructed from re-applying the training dataset on a well trained network.
The approach in~\cite{cheng2019runtime} uses boolean abstraction to approximate and represent neuron activation patterns from correctly classified training data and is effective on reporting network misclassifications.
However, the construction and membership test of boolean formula are computationally expensive, especially when dealing with patterns at layers with many neurons, e.g, running out of 8 GB memory when building such formula for a layer of 84 neurons.
To reduce the complexity of abstraction methods, the approach in~\cite{henzinger2019outside} introduces \emph{box abstraction}, which can be easily computed and membership tested.
The paper introduces the idea of partitioning the obtained patterns into smaller clusters first and then constructing abstractions on these smaller clusters.
Furthermore, \cite{lukina2020into} extends~\cite{henzinger2019outside} by defining a framework of active monitoring of networks based on box abstractions that detects unknown classes of inputs and adapts to them at runtime by virtue of human interactions and retraining the network.

The approach in this paper is in the line of the ones in~\cite{cheng2019runtime,henzinger2019outside} complements the framework of runtime monitors in~\cite{cheng2019runtime} and~\cite{henzinger2019outside}.
Our approach generalizes these approach by allowing to observe and record the neuron activation patterns at hidden and output layers from both correct and incorrect classifications of known classes of inputs.
Additional patterns from misclassified known classes of inputs introduces ``uncertainty'' verdicts, which provides insights to the precision of the abstraction-based monitor and the separability of network.
We extended the idea of applying clustering before constructing abstractions by introducing \emph{boxes with a resolution} which allow defining clustering coverage as a metric to quantify the precision improvement in terms of the spaces covered by the box abstraction constructed with and without clustering.
Consequently, one can compare the effectiveness of different clustering parameters and tune the parameter of the monitoring approach.

\section{Conclusion and Future Research Directions}
\label{sec:conclusion}

We introduce a framework for the runtime monitoring of LECs with forward neural networks.
Our framework relies on boxes with resolution which are used to abstract the data seen during training and allows a monitor to state an accept/reject/uncertainty verdict based on the membership test of a new input to the built abstractions.
Resolution of boxes allows quantifying the coverage of the abstractions used by monitors and thus assessing their precision.
We showed experiments showing how an expert can fine-tune the parameters of the monitoring framework (clustering parameter and watched layers).
The notion of uncertainty allows understanding the uncertainty of networks decisions from a formal perspective.

As we believe, we are at the beginning of the search for defining methods and tools for verifying and monitoring learning-enabled components, a lot of research questions and perspectives remain to be addressed.
We chose to follow~\cite{henzinger2019outside,lukina2020into,cheng2020provably} and build upon the notion of box as such abstraction seems to provide a good tradeoff between precision and performance but a lot of other candidate abstractions (defined in the verification and compilation communities for instance) remain to be studied for monitoring purposes.
In terms of monitoring, being able to update at operation time the build abstraction would permit monitors that continuously learn even operation-time inputs.
Moreover, the question of reaction to abnormal inputs remains to be studied.
More generally, we believe that LECs will need different interacting and complementary methods and toolsets to provide comprehensive solutions affording the needed confidence in the implemented systems, e.g., testing and monitoring for design time, monitoring and enforcement for operation time.

\bibliographystyle{splncs04}
\bibliography{reference}

\appendix
\section{Additional Experimental Results on Benchmarks F\_MNIST and CIFAR10}
\label{sec:appendix}

We present the results referred to in the experimental section related to clustering coverage estimation, numbers of outcomes as per Table~\ref{table:monitorPerformance}, and precision-recall curves obtained on benchmark F\_MNIST and CIFAR10.


\begin{figure*}[htbp]
    \centering
    \begin{subfigure}[htbp]{0.245\textwidth}
        \centering
        \includegraphics[width=\textwidth]{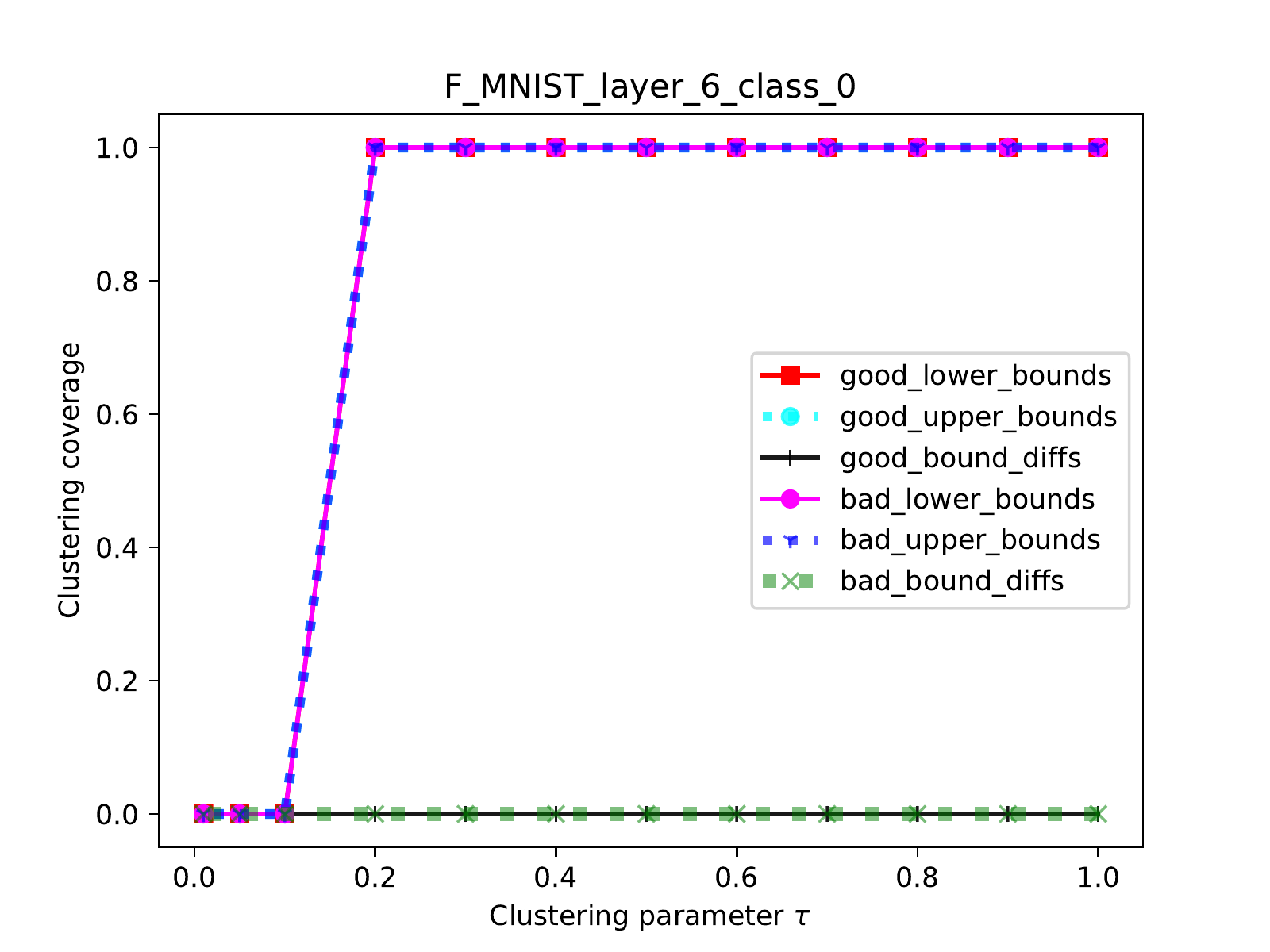}
    \end{subfigure}
    \hfill
    \begin{subfigure}[htbp]{0.245\textwidth}
        \centering
        \includegraphics[width=\textwidth]{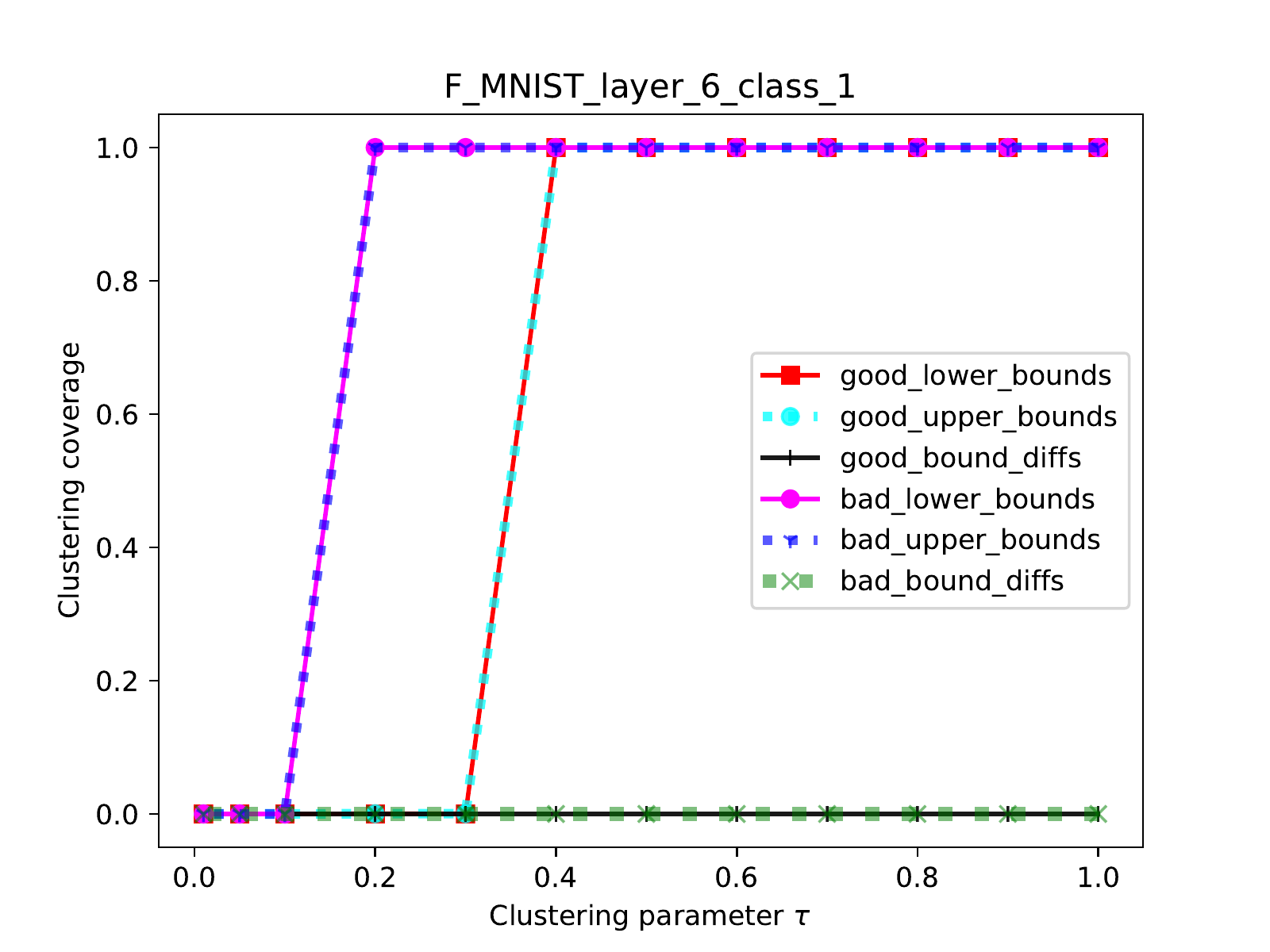}
    \end{subfigure}
    \hfill
    \begin{subfigure}[htbp]{0.245\textwidth}
        \centering
        \includegraphics[width=\textwidth]{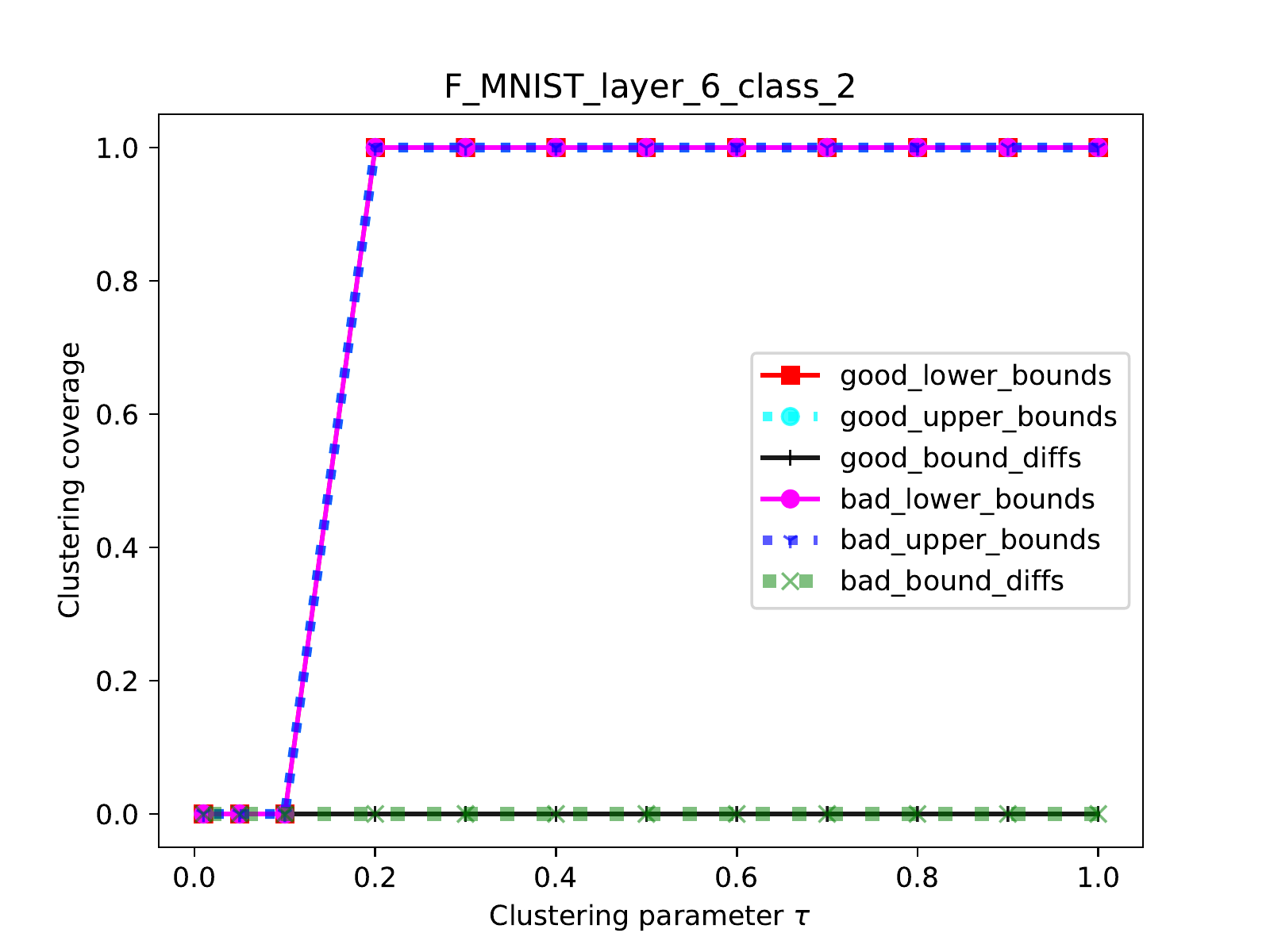}
    \end{subfigure}
    \hfill
    \begin{subfigure}[htbp]{0.245\textwidth}
        \centering
        \includegraphics[width=\textwidth]{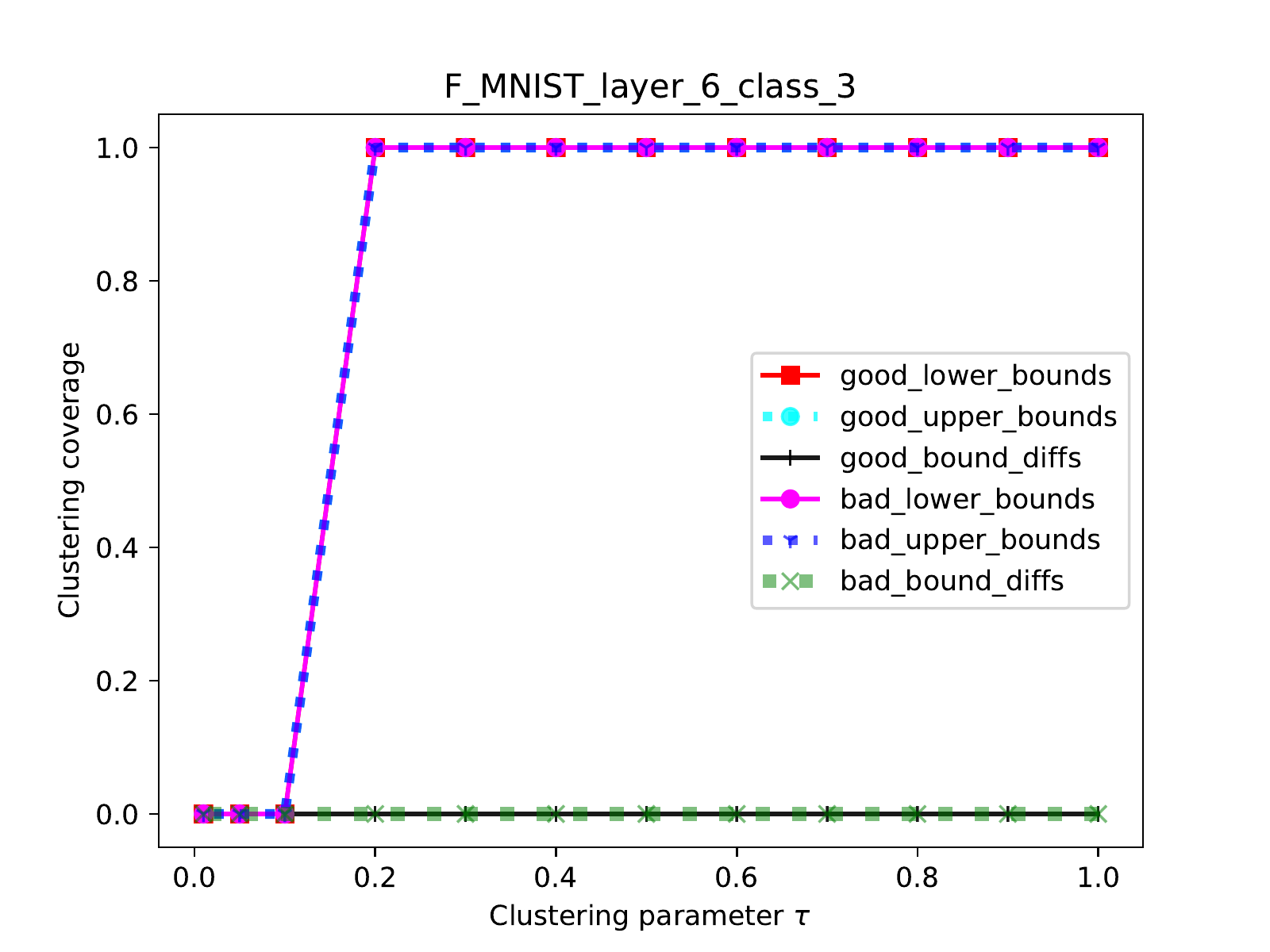}
    \end{subfigure}

    \begin{subfigure}[htbp]{0.245\textwidth}
        \centering
        \includegraphics[width=\textwidth]{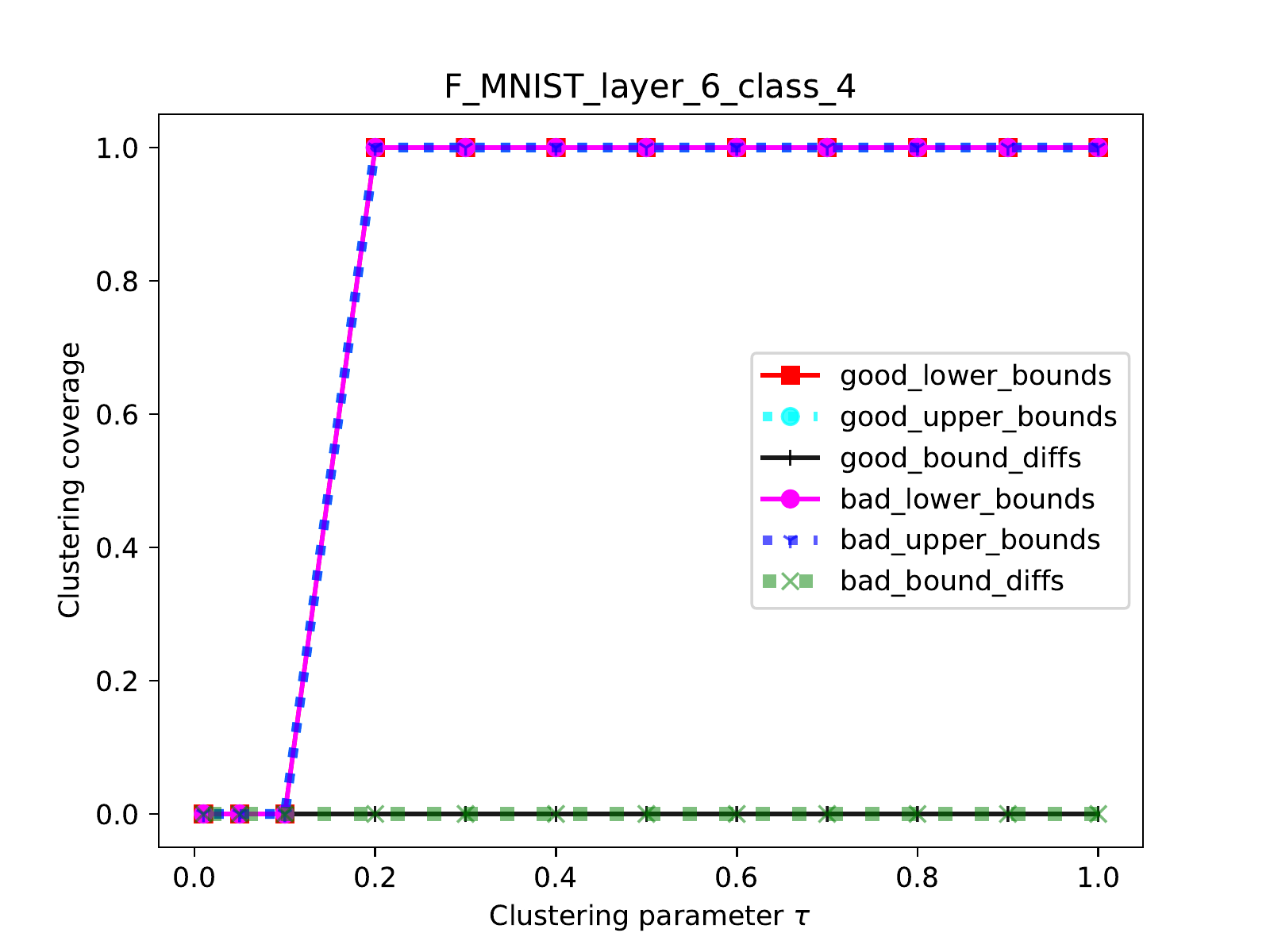}
    \end{subfigure}
    \hfill
    \begin{subfigure}[htbp]{0.245\textwidth}
        \centering
        \includegraphics[width=\textwidth]{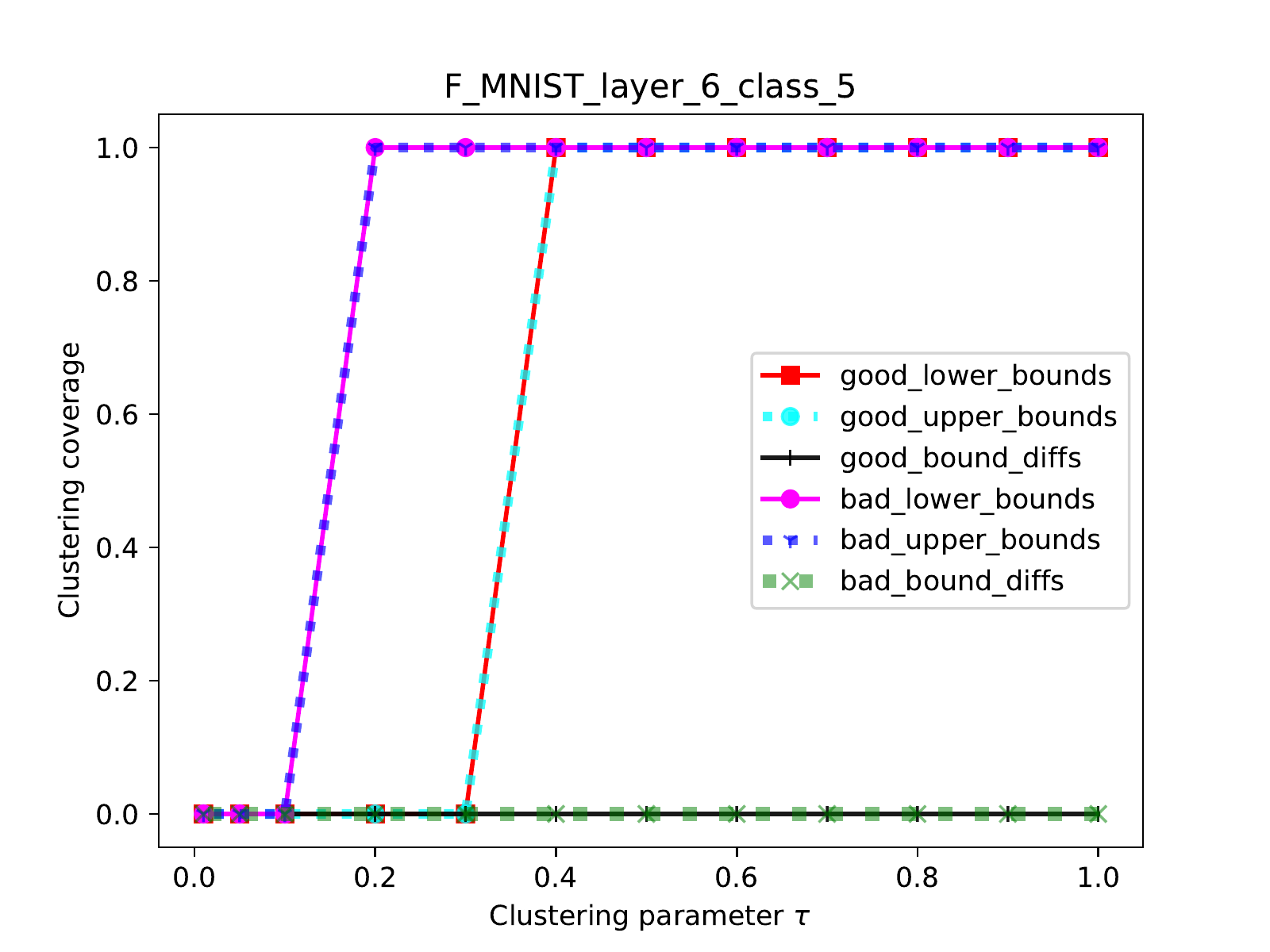}
    \end{subfigure}
    \hfill
    \begin{subfigure}[htbp]{0.245\textwidth}
        \centering
        \includegraphics[width=\textwidth]{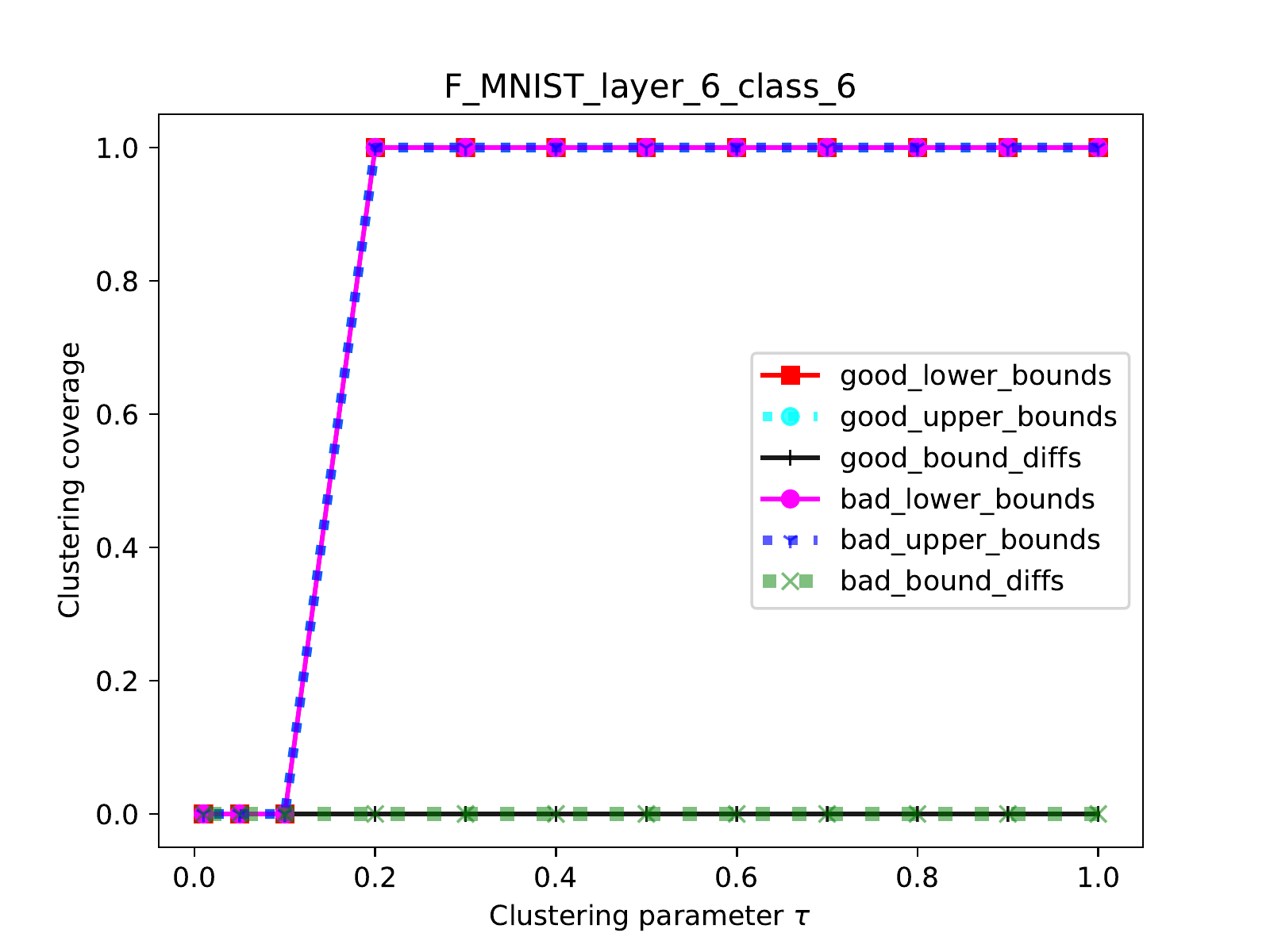}
    \end{subfigure}
    \hfill
	\begin{subfigure}[htbp]{0.245\textwidth}
        \centering
        \includegraphics[width=\textwidth]{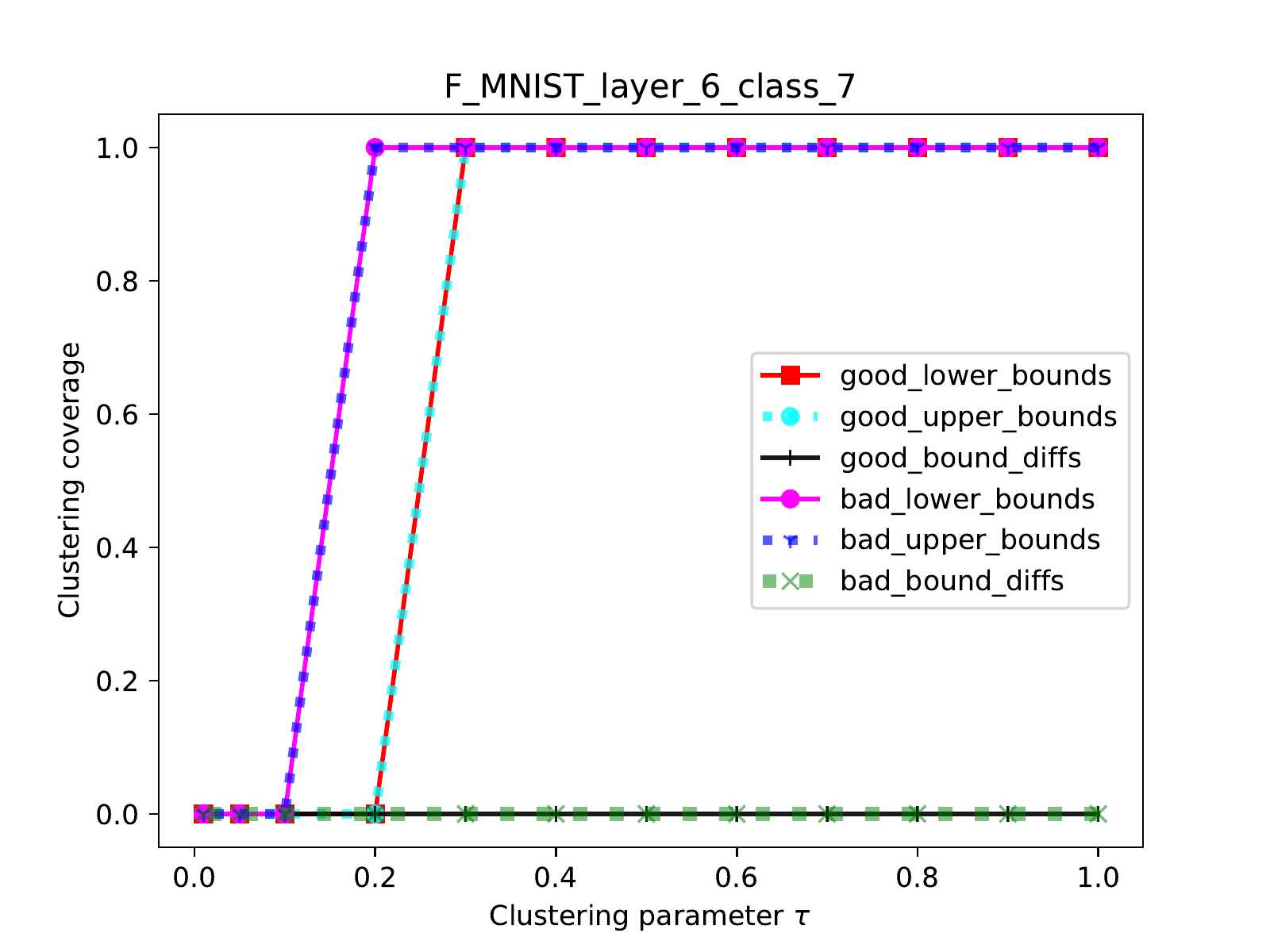}
    \end{subfigure}

    \begin{subfigure}[htbp]{0.245\textwidth}
        \centering
        \includegraphics[width=\textwidth]{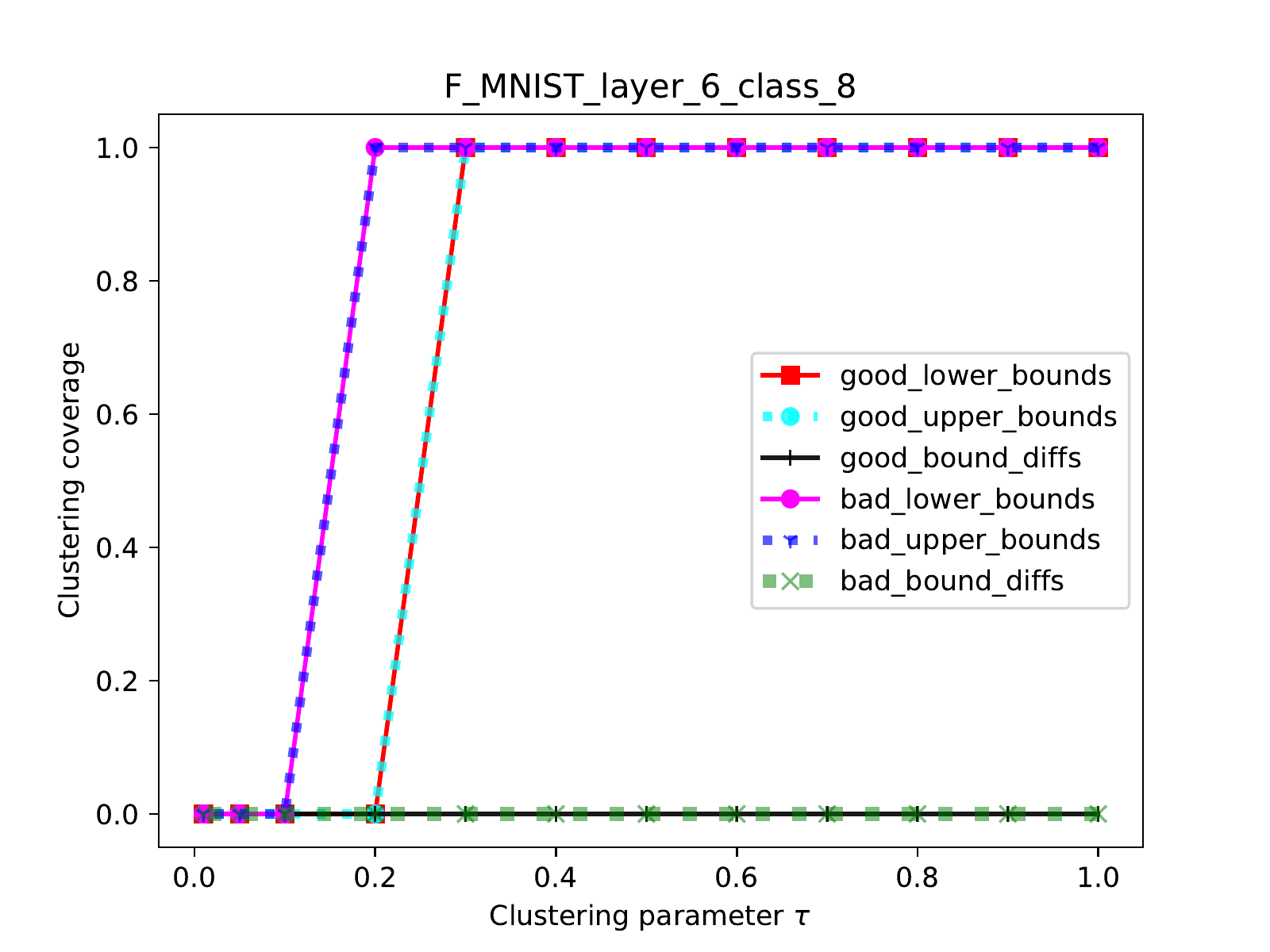}
    \end{subfigure}
    \hfill
    \begin{subfigure}[htbp]{0.245\textwidth}
        \centering
        \includegraphics[width=\textwidth]{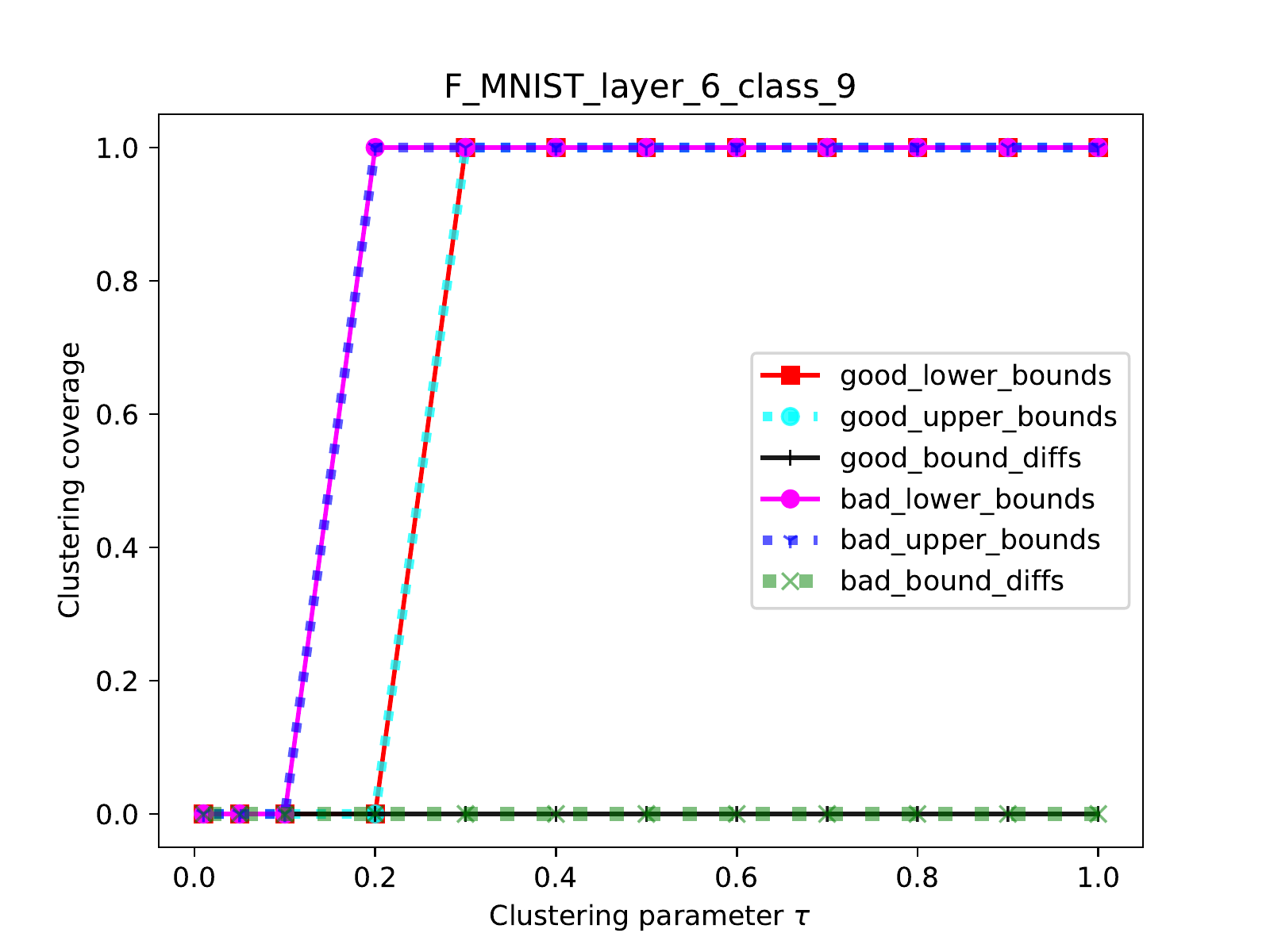}
    \end{subfigure}
    \caption{Clustering coverage estimations for the high-level features obtained at the output layer for benchmark F\_MNIST.}
    \label{fig:clusteringFMNIST}
\end{figure*}

\begin{figure*}[htbp]
    \centering
    \begin{subfigure}[htbp]{0.23\textwidth}
        \centering
        \includegraphics[width=\textwidth]{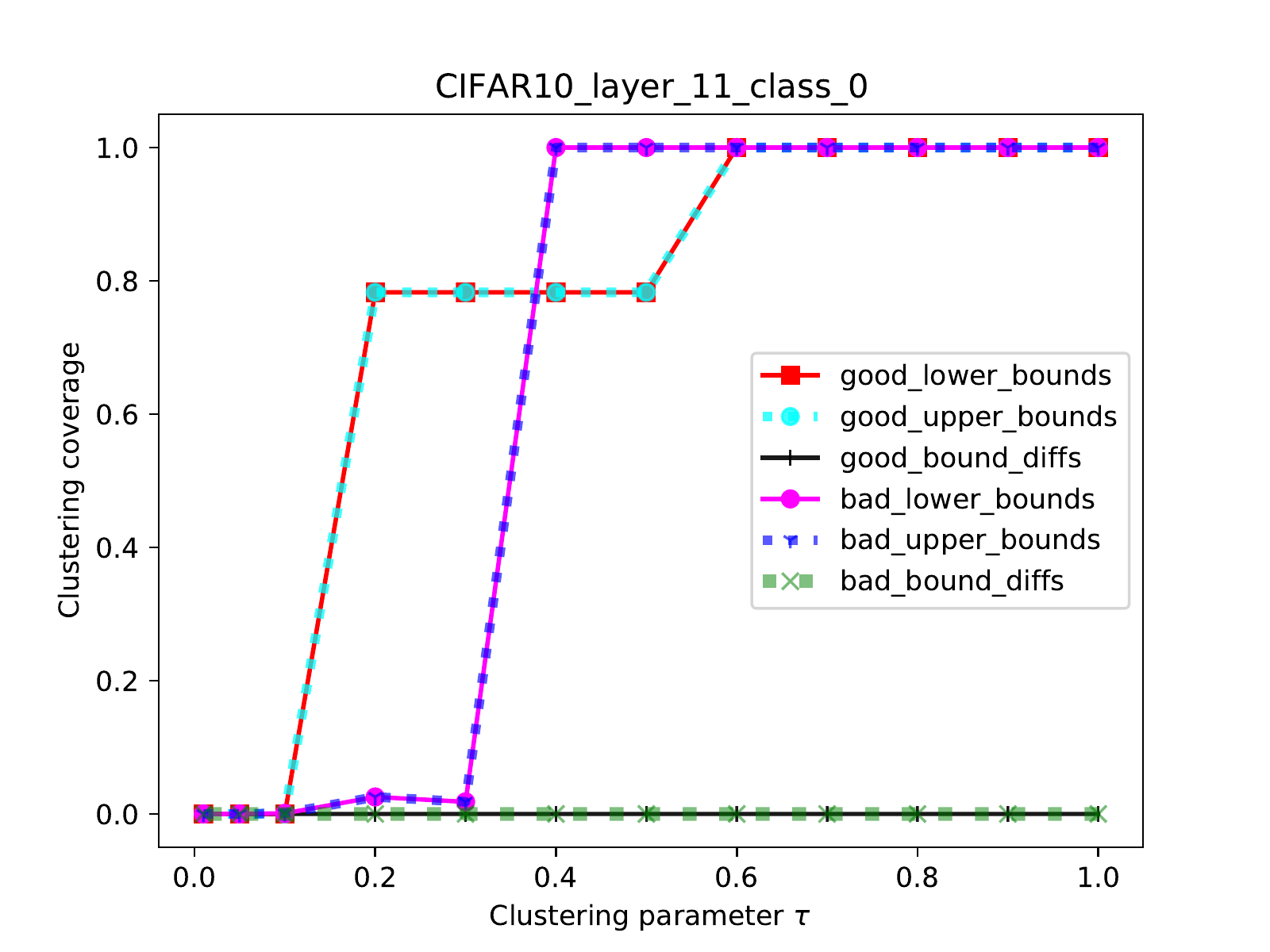}
    \end{subfigure}
    \hfill
    \begin{subfigure}[htbp]{0.23\textwidth}
        \centering
        \includegraphics[width=\textwidth]{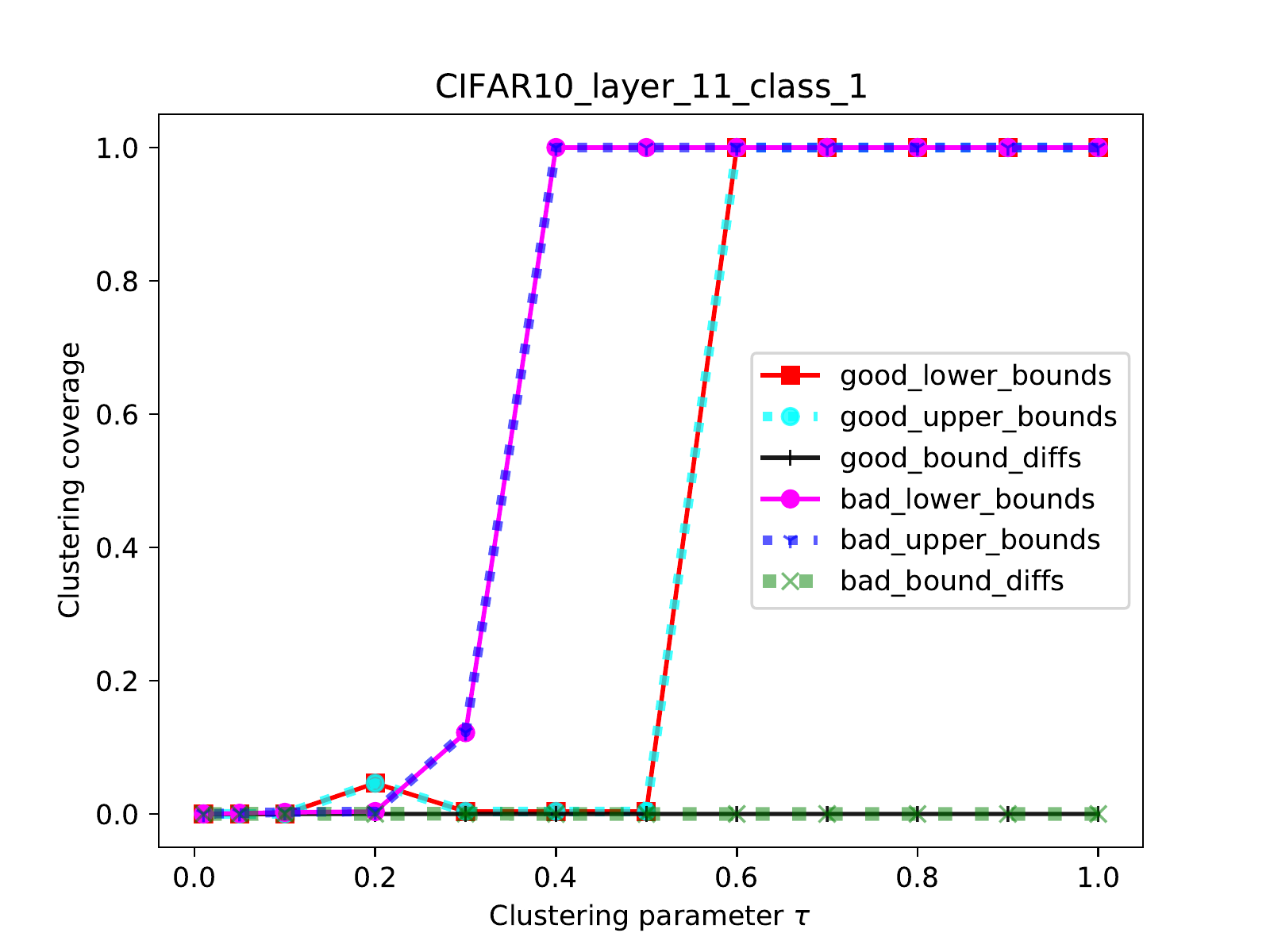}
    \end{subfigure}
    \hfill
    \begin{subfigure}[htbp]{0.23\textwidth}
        \centering
        \includegraphics[width=\textwidth]{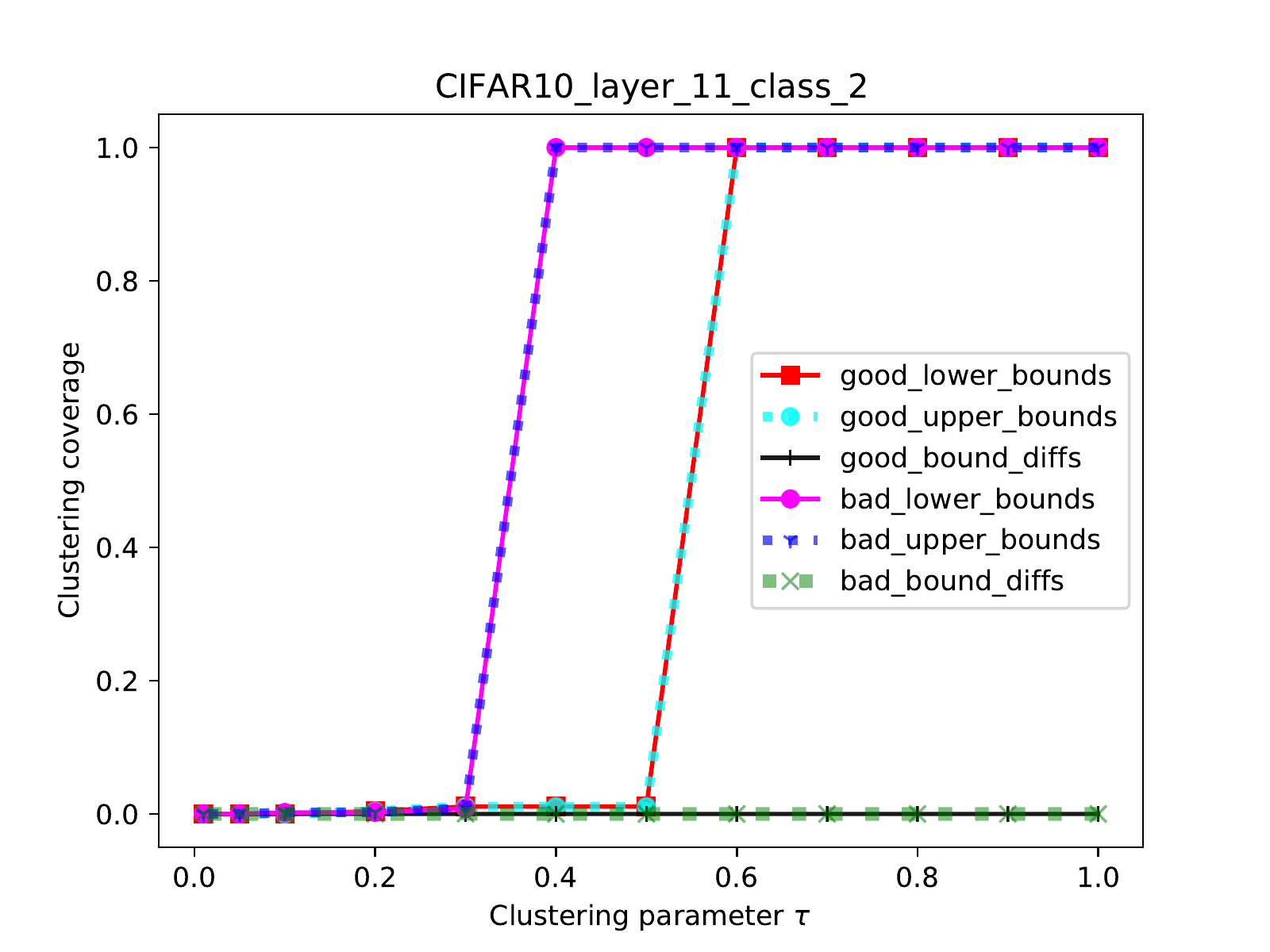}
    \end{subfigure}
    \hfill
    \begin{subfigure}[htbp]{0.23\textwidth}
        \centering
        \includegraphics[width=\textwidth]{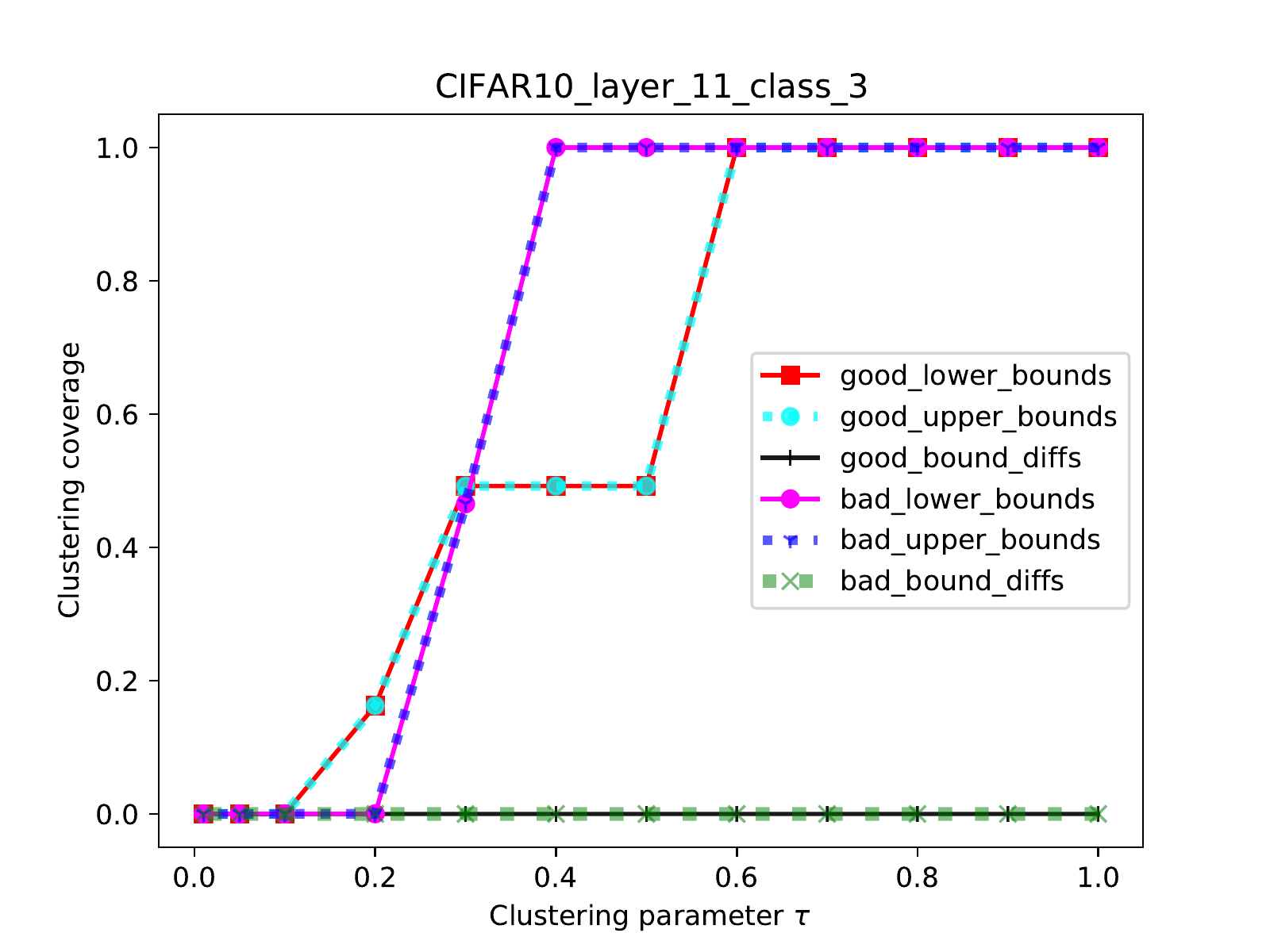}
    \end{subfigure}

    \begin{subfigure}[htbp]{0.23\textwidth}
        \centering
        \includegraphics[width=\textwidth]{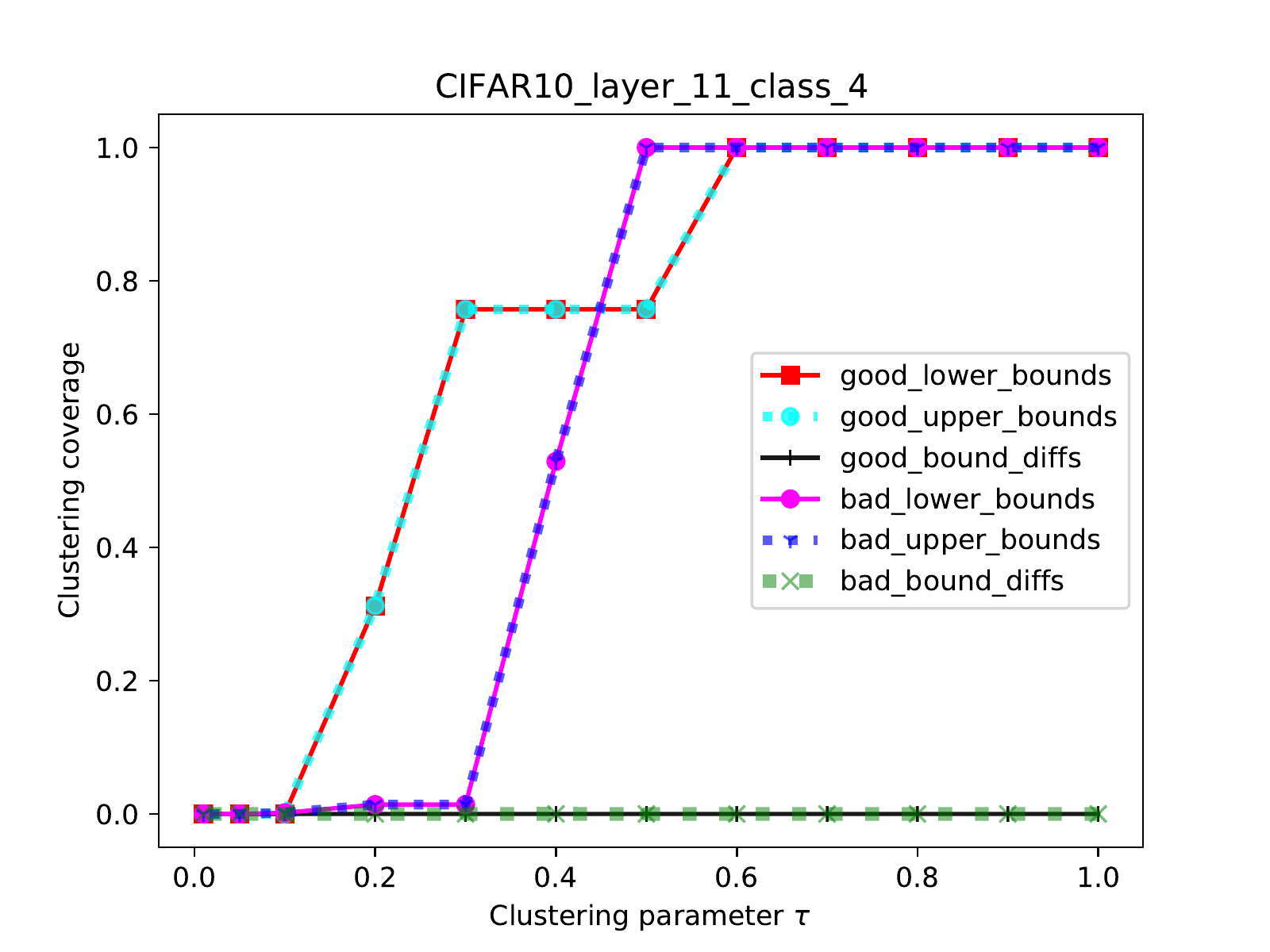}
    \end{subfigure}
    \hfill
    \begin{subfigure}[htbp]{0.23\textwidth}
        \centering
        \includegraphics[width=\textwidth]{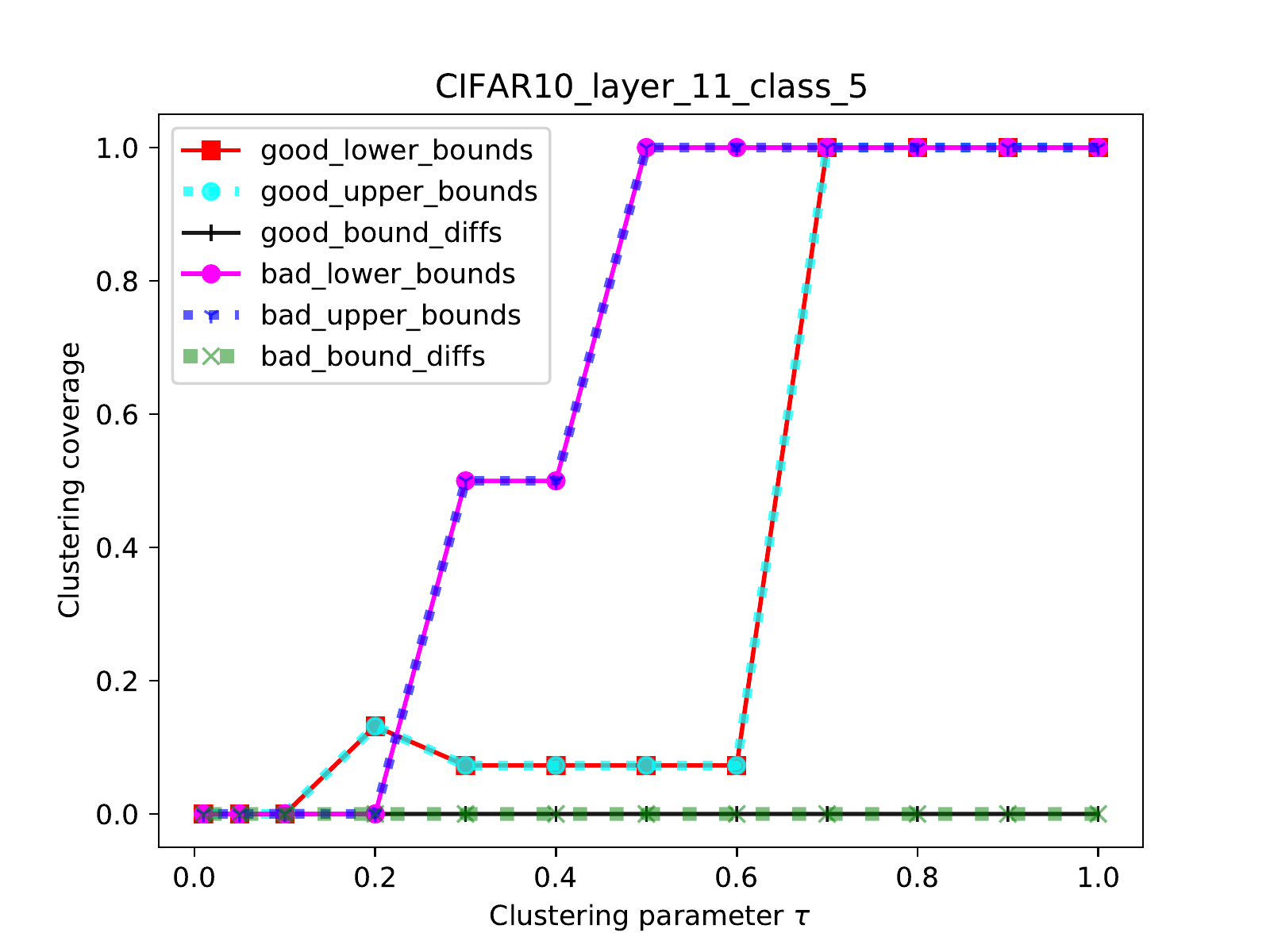}
    \end{subfigure}
    \hfill
    \begin{subfigure}[htbp]{0.23\textwidth}
        \centering
        \includegraphics[width=\textwidth]{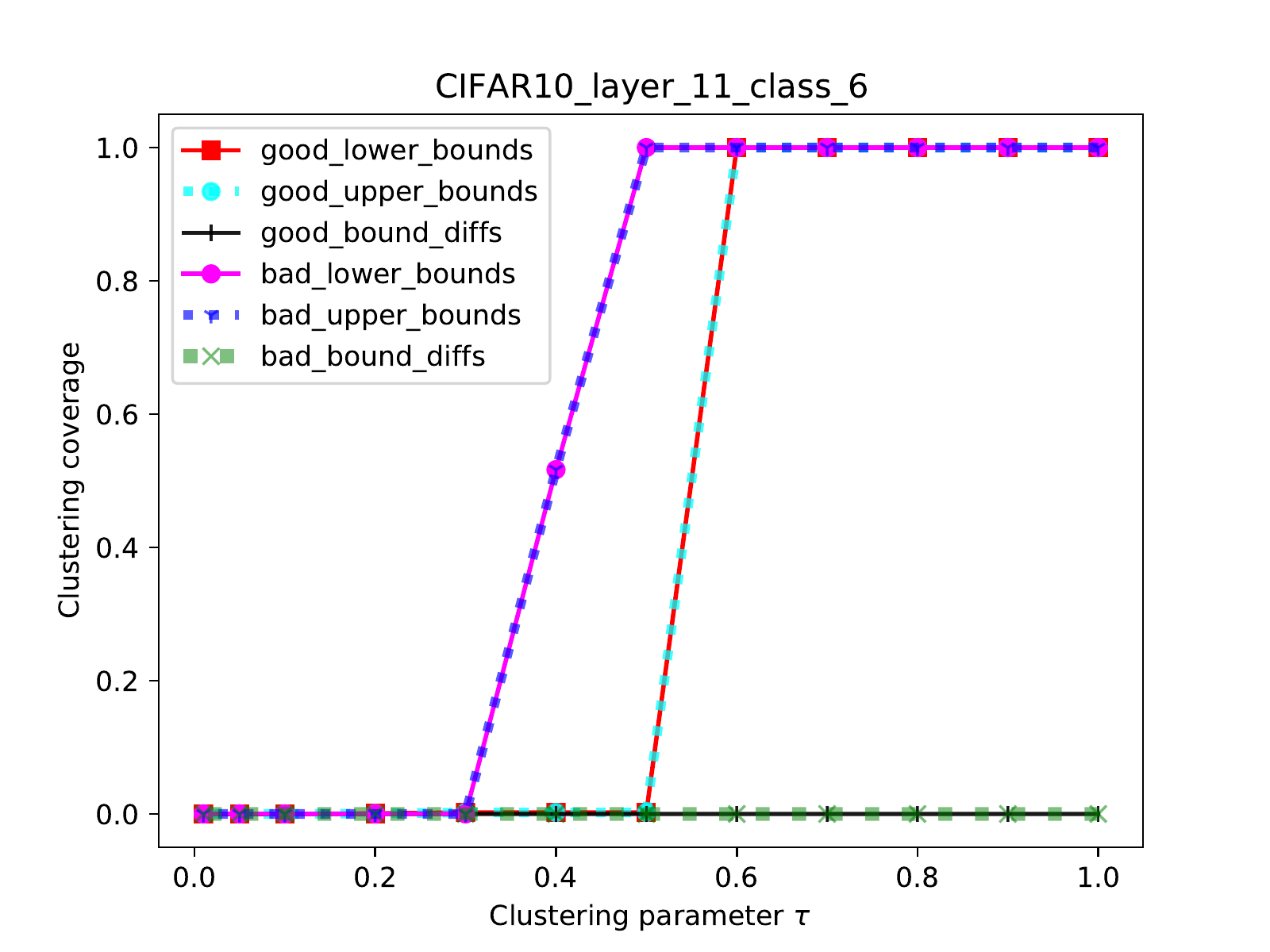}
    \end{subfigure}
    \hfill
	\begin{subfigure}[htbp]{0.23\textwidth}
        \centering
        \includegraphics[width=\textwidth]{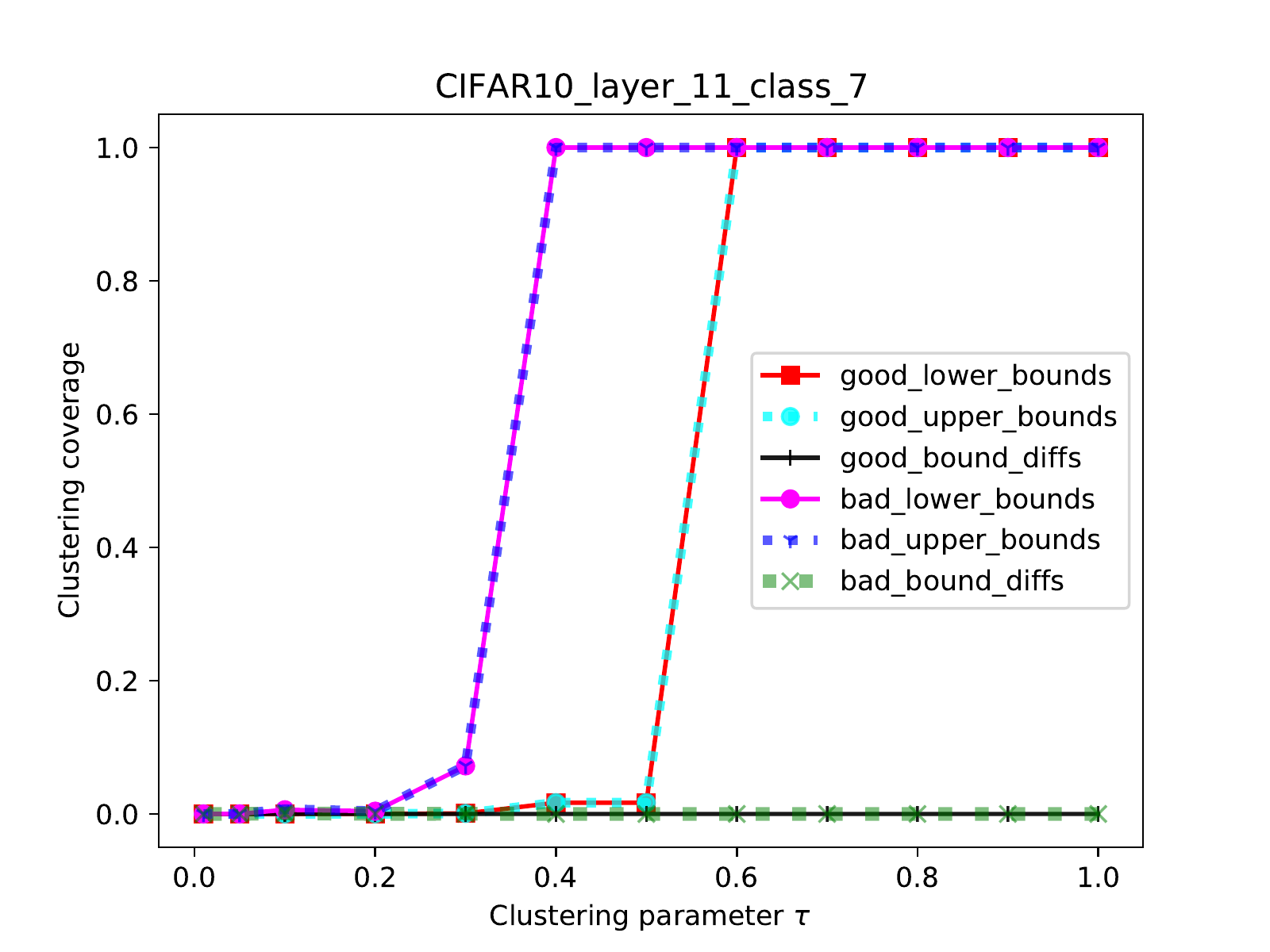}
    \end{subfigure}

    \begin{subfigure}[htbp]{0.23\textwidth}
        \centering
        \includegraphics[width=\textwidth]{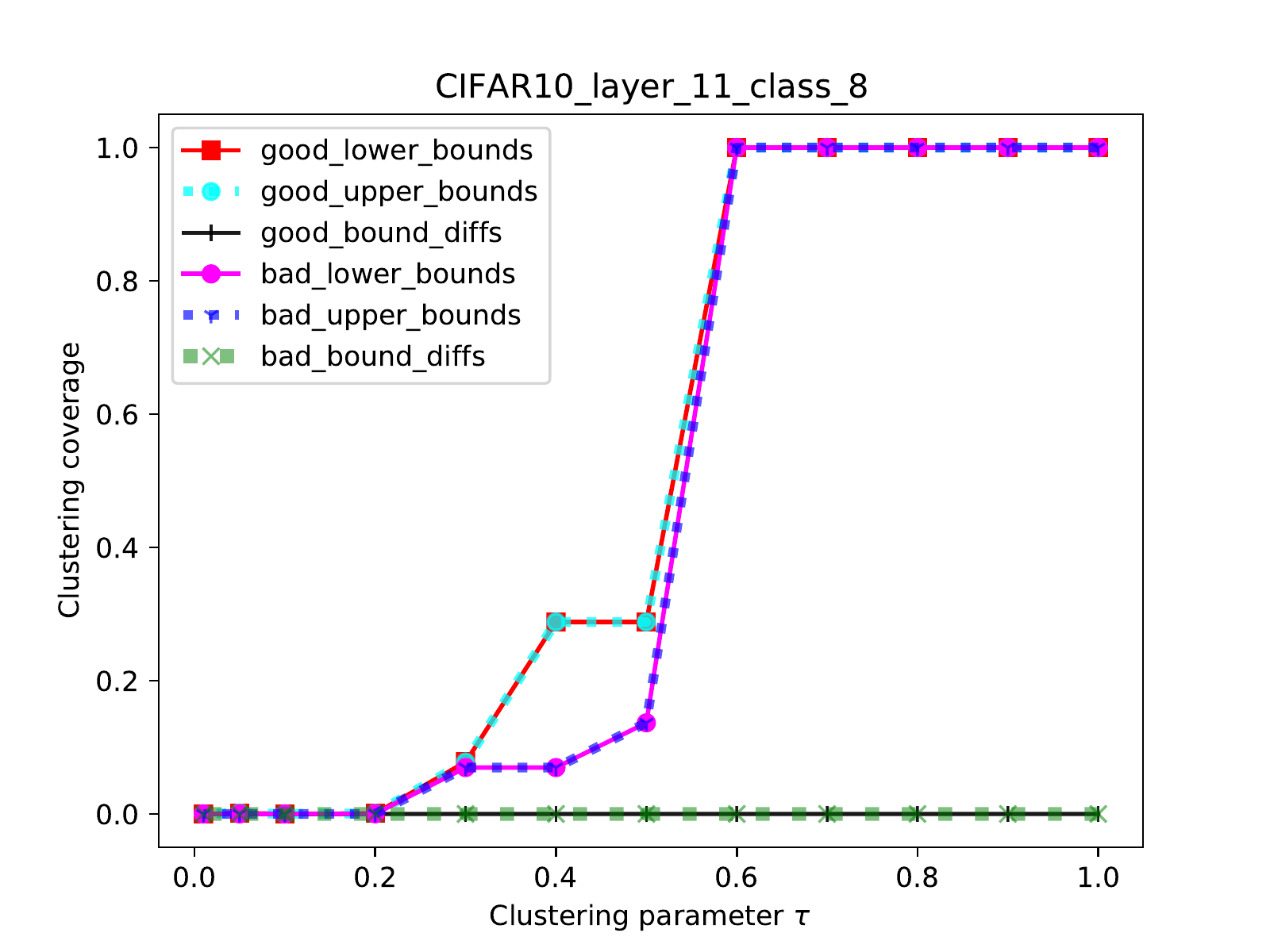}
    \end{subfigure}
    \hfill
    \begin{subfigure}[htbp]{0.23\textwidth}
        \centering
        \includegraphics[width=\textwidth]{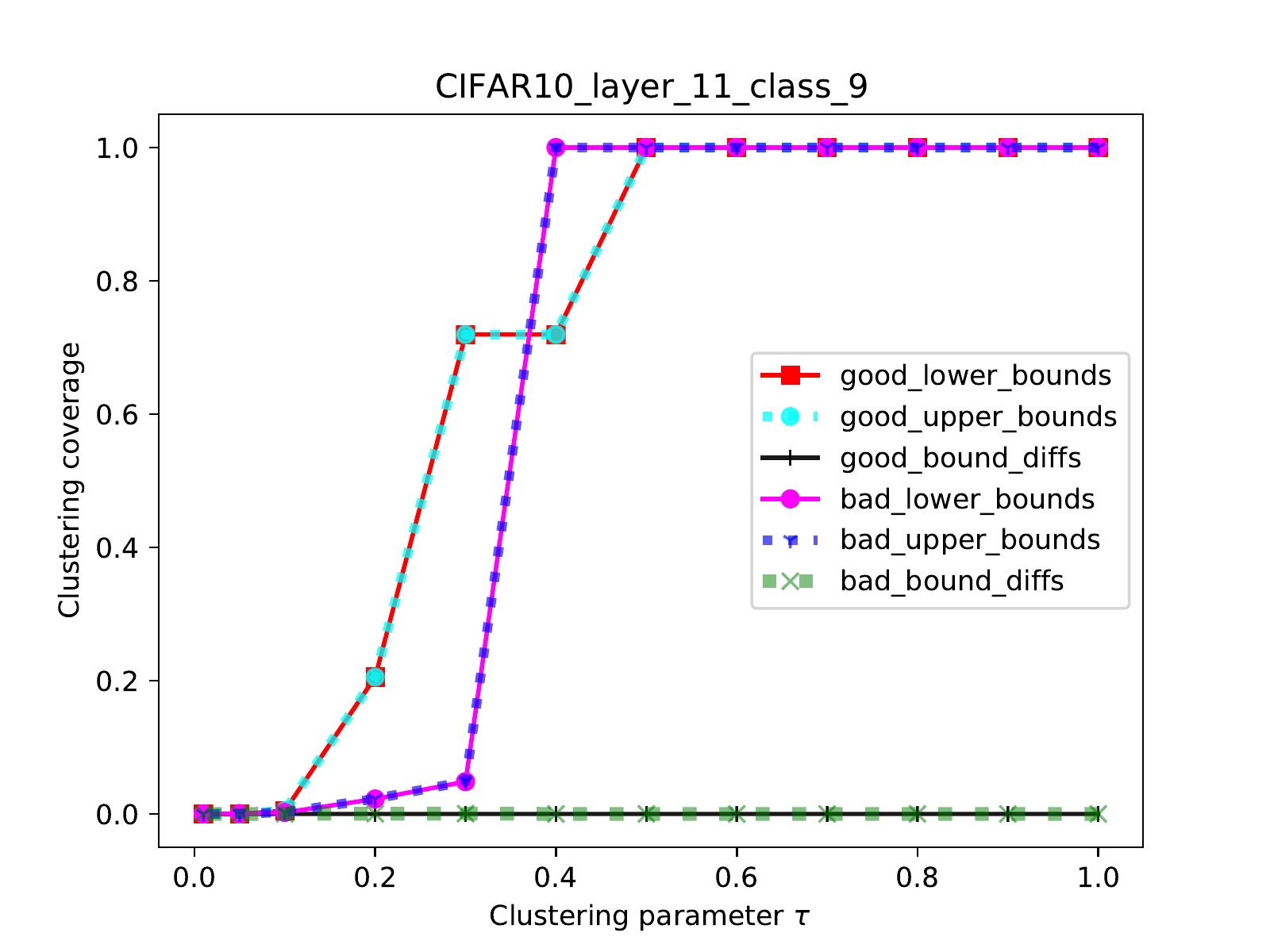}
    \end{subfigure}
    \caption{Clustering coverage estimations for the high-level features obtained at the output layer for benchmark CIFAR10.}
    \label{fig:clusteringCIFAR10}
\end{figure*}

\begin{figure*}[htbp]
    \centering
    \begin{subfigure}[htbp]{0.245\textwidth}
        \centering
        \includegraphics[width=\textwidth]{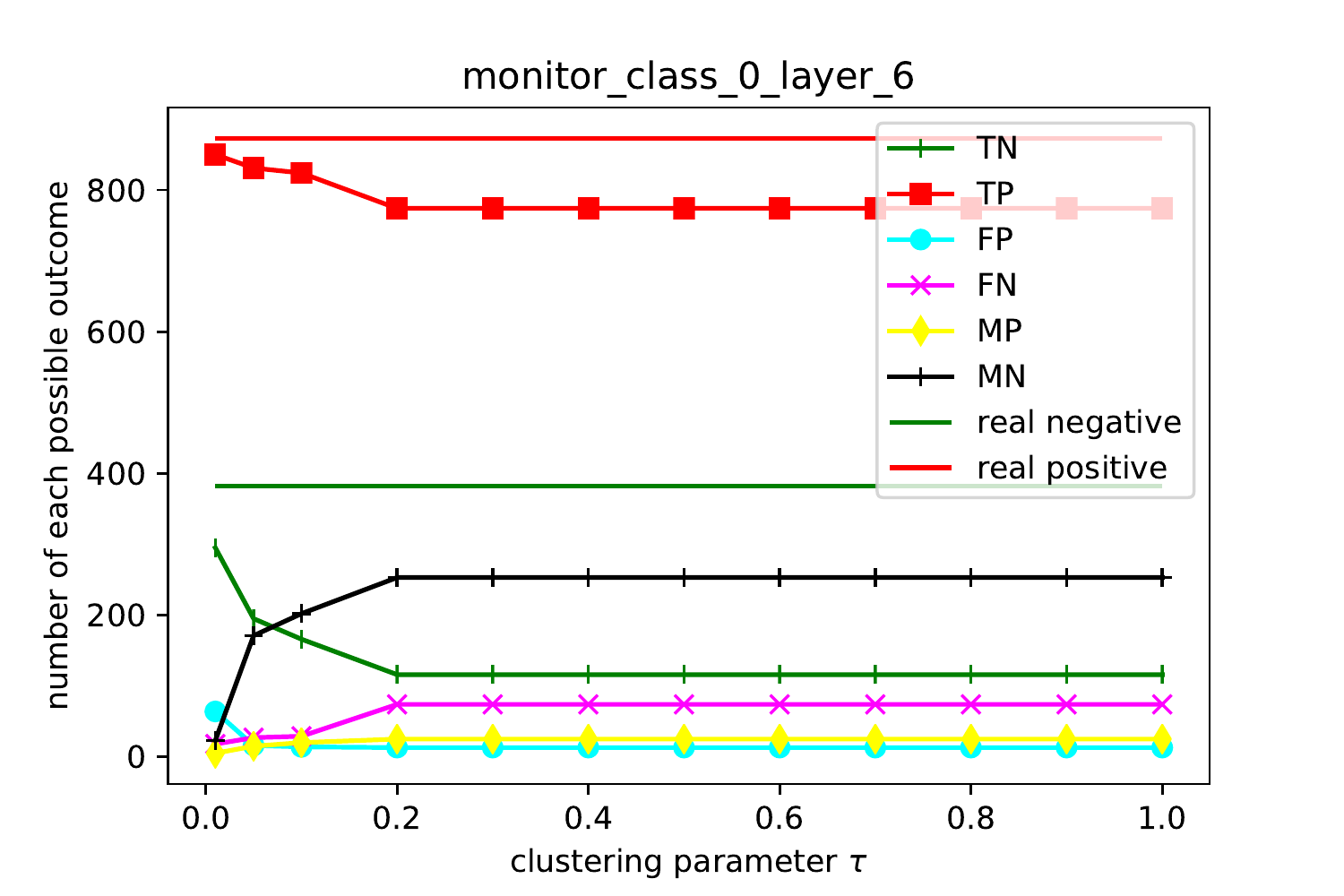}
    \end{subfigure}
    \hfill
    \begin{subfigure}[htbp]{0.245\textwidth}
        \centering
        \includegraphics[width=\textwidth]{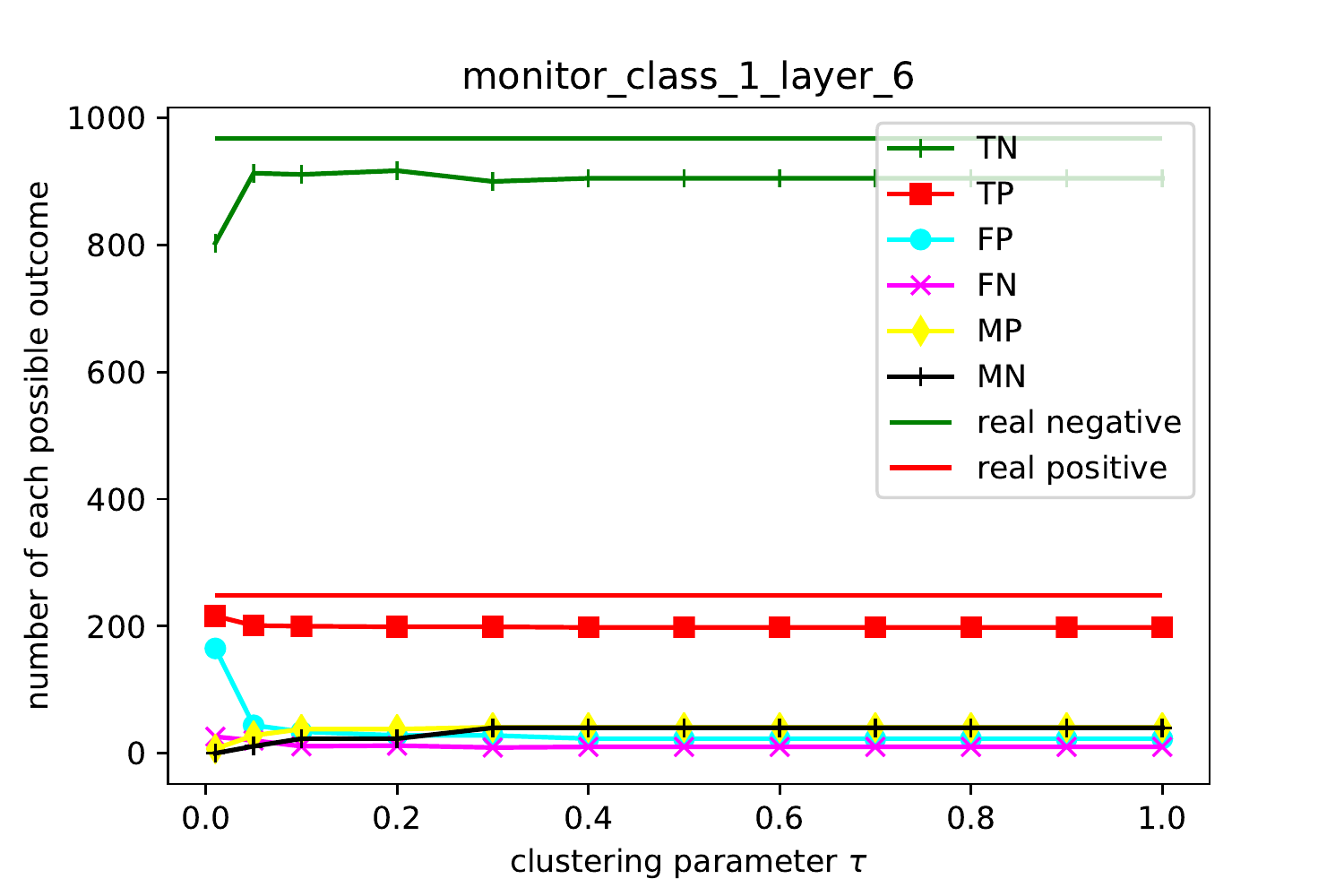}
    \end{subfigure}
    \hfill
    \begin{subfigure}[htbp]{0.245\textwidth}
        \centering
        \includegraphics[width=\textwidth]{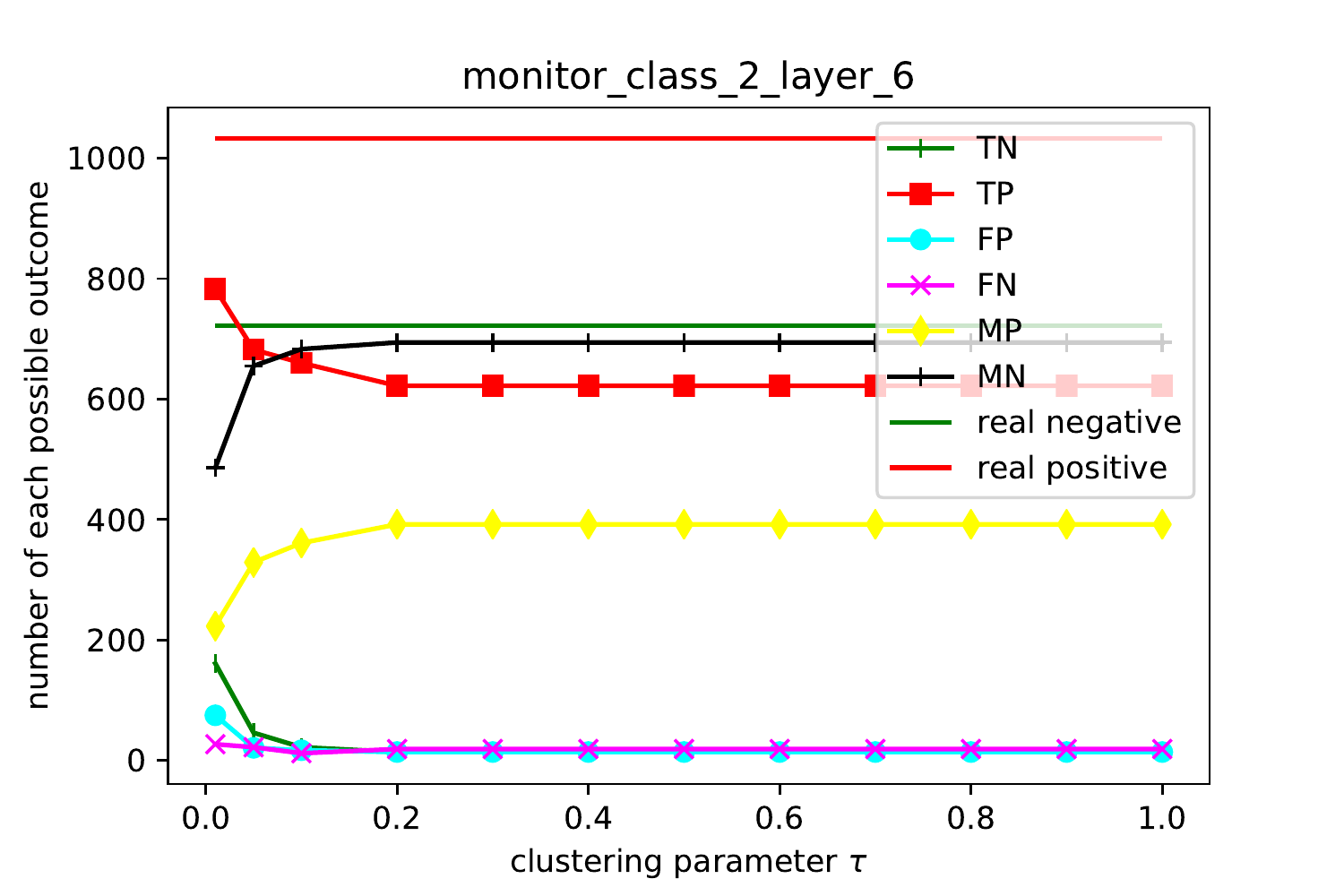}
    \end{subfigure}
    \hfill
    \begin{subfigure}[htbp]{0.23\textwidth}
        \centering
        \includegraphics[width=\textwidth]{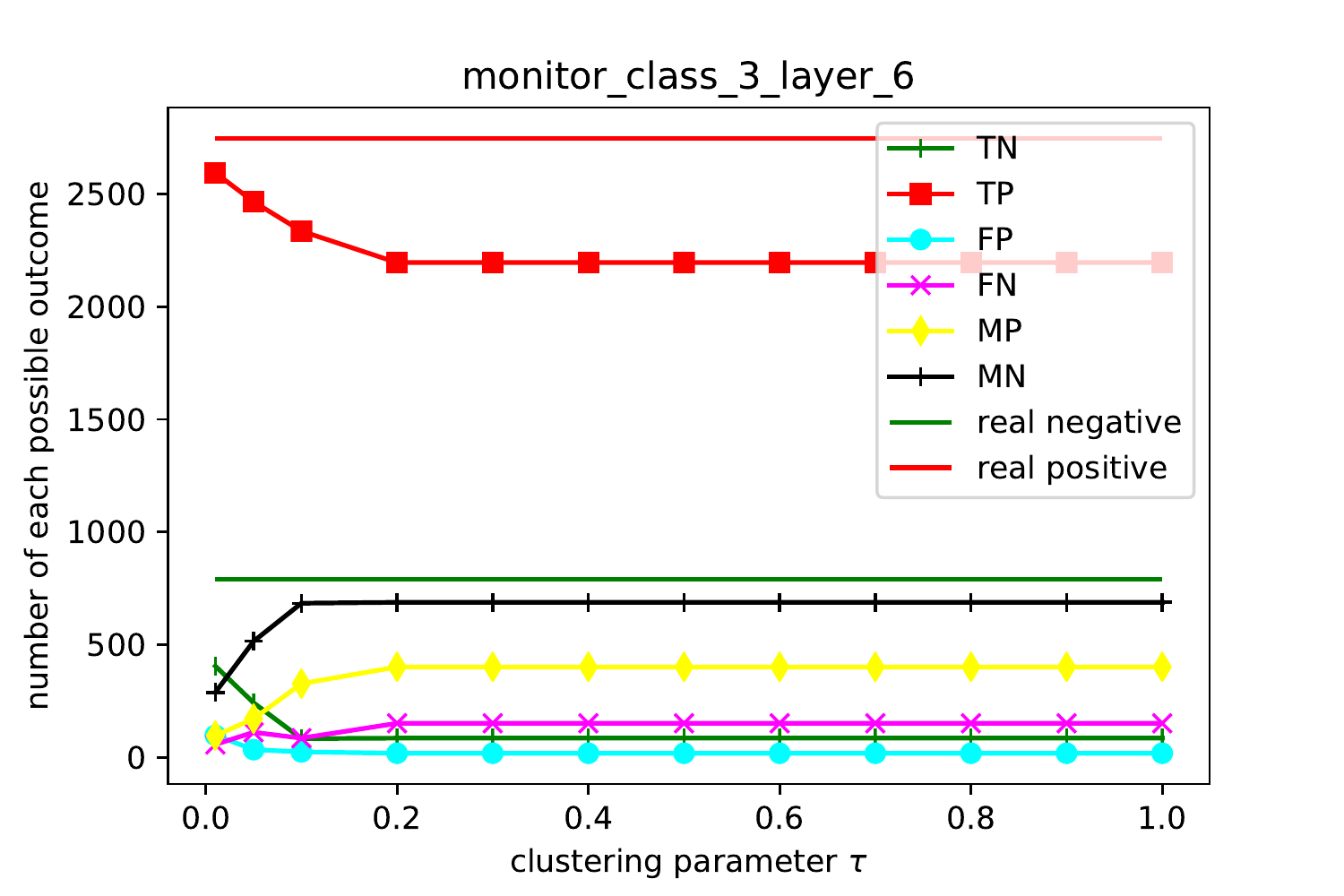}
    \end{subfigure}

    \begin{subfigure}[htbp]{0.245\textwidth}
        \centering
        \includegraphics[width=\textwidth]{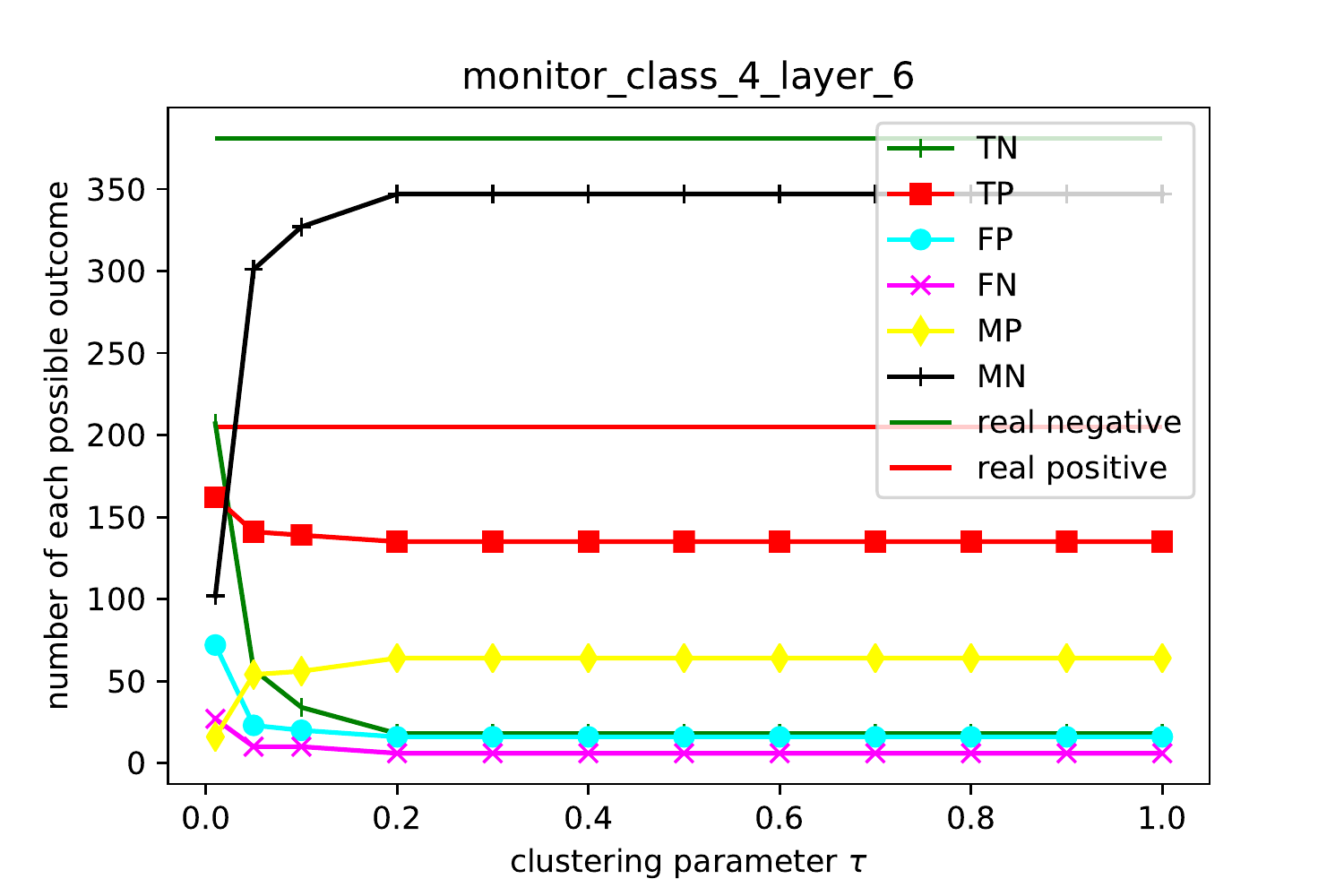}
    \end{subfigure}
    \hfill
    \begin{subfigure}[htbp]{0.245\textwidth}
        \centering
        \includegraphics[width=\textwidth]{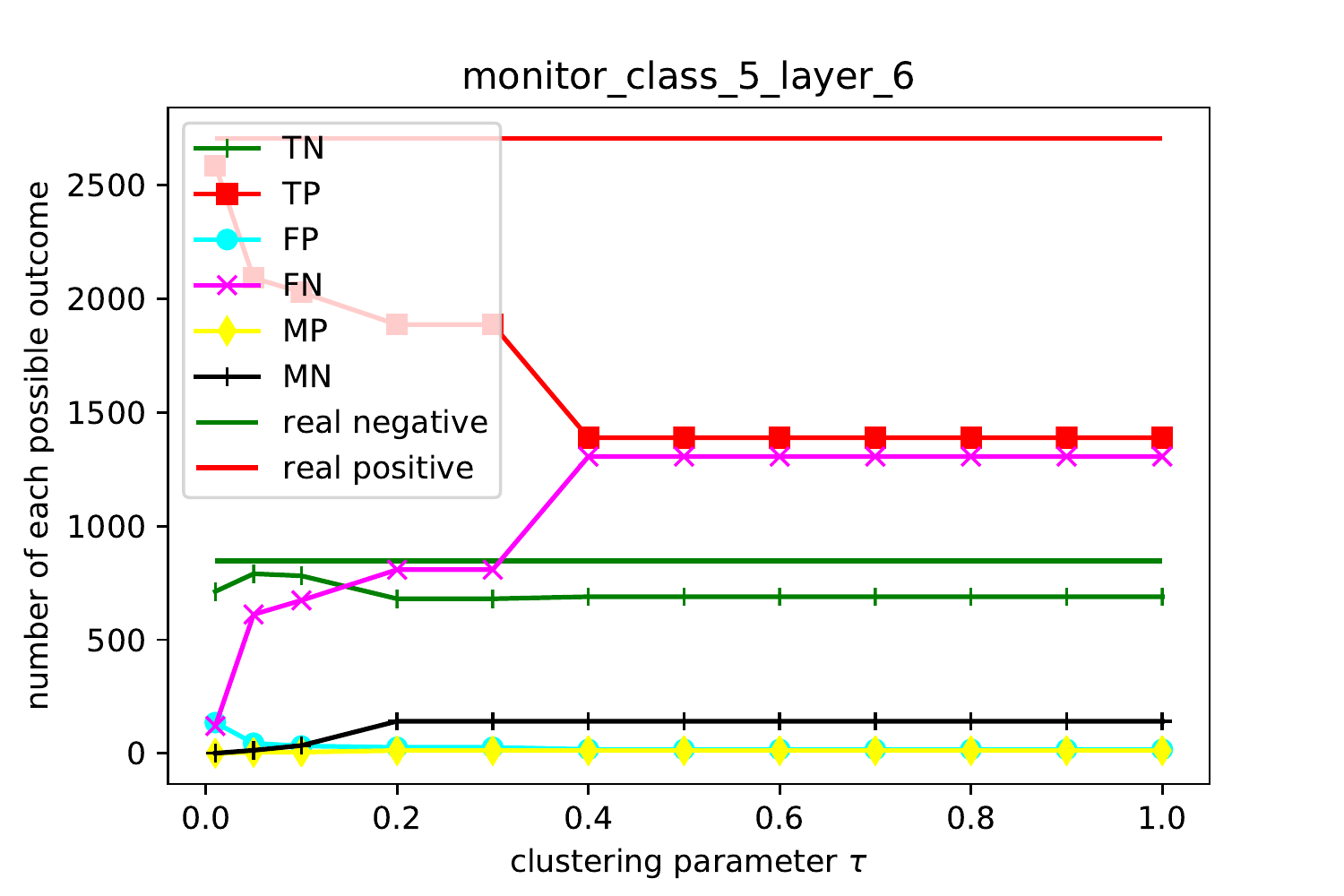}
    \end{subfigure}
    \hfill
    \begin{subfigure}[htbp]{0.245\textwidth}
        \centering
        \includegraphics[width=\textwidth]{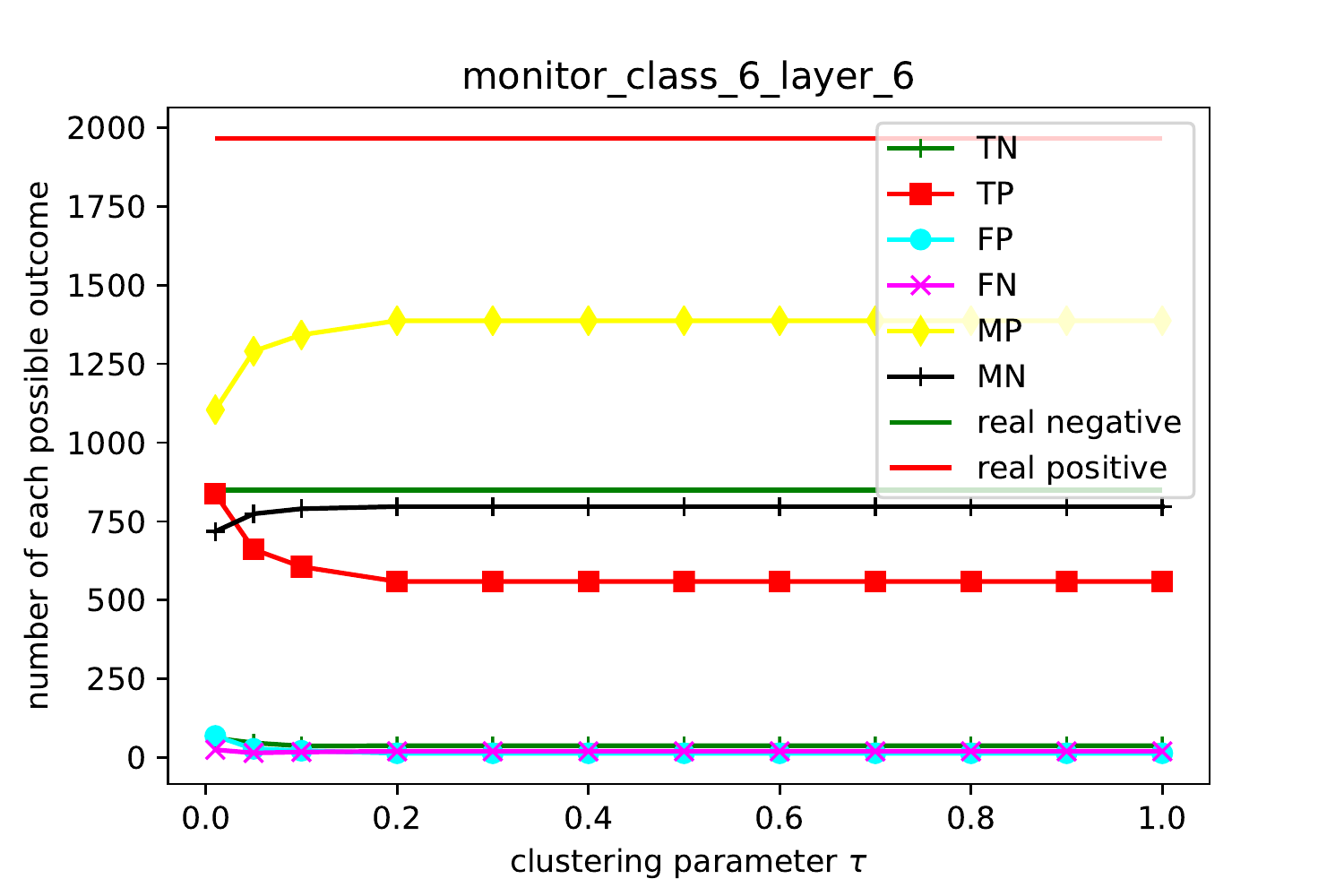}
    \end{subfigure}
    \hfill
	\begin{subfigure}[htbp]{0.245\textwidth}
        \centering
        \includegraphics[width=\textwidth]{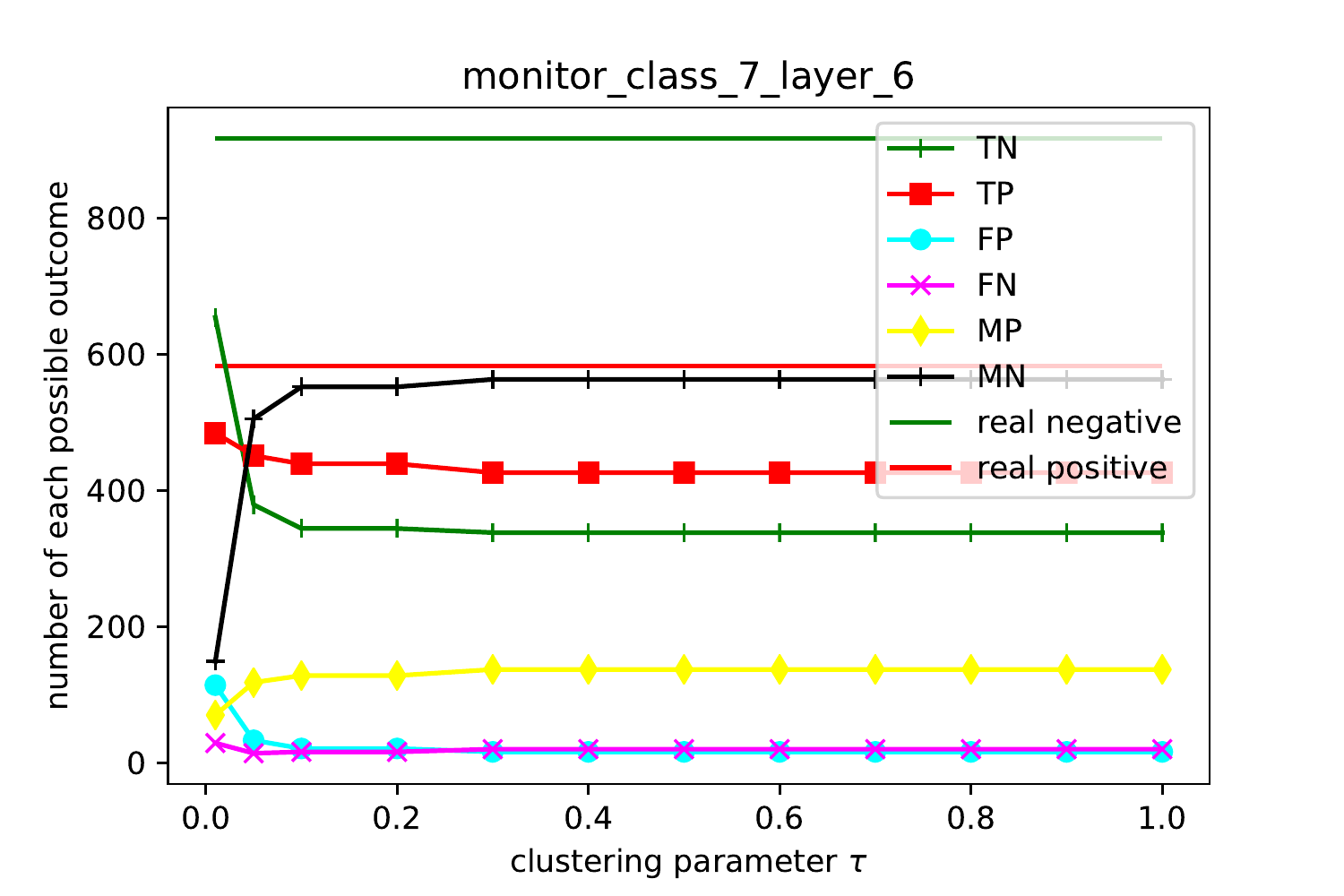}
    \end{subfigure}

    \begin{subfigure}[htbp]{0.245\textwidth}
        \centering
        \includegraphics[width=\textwidth]{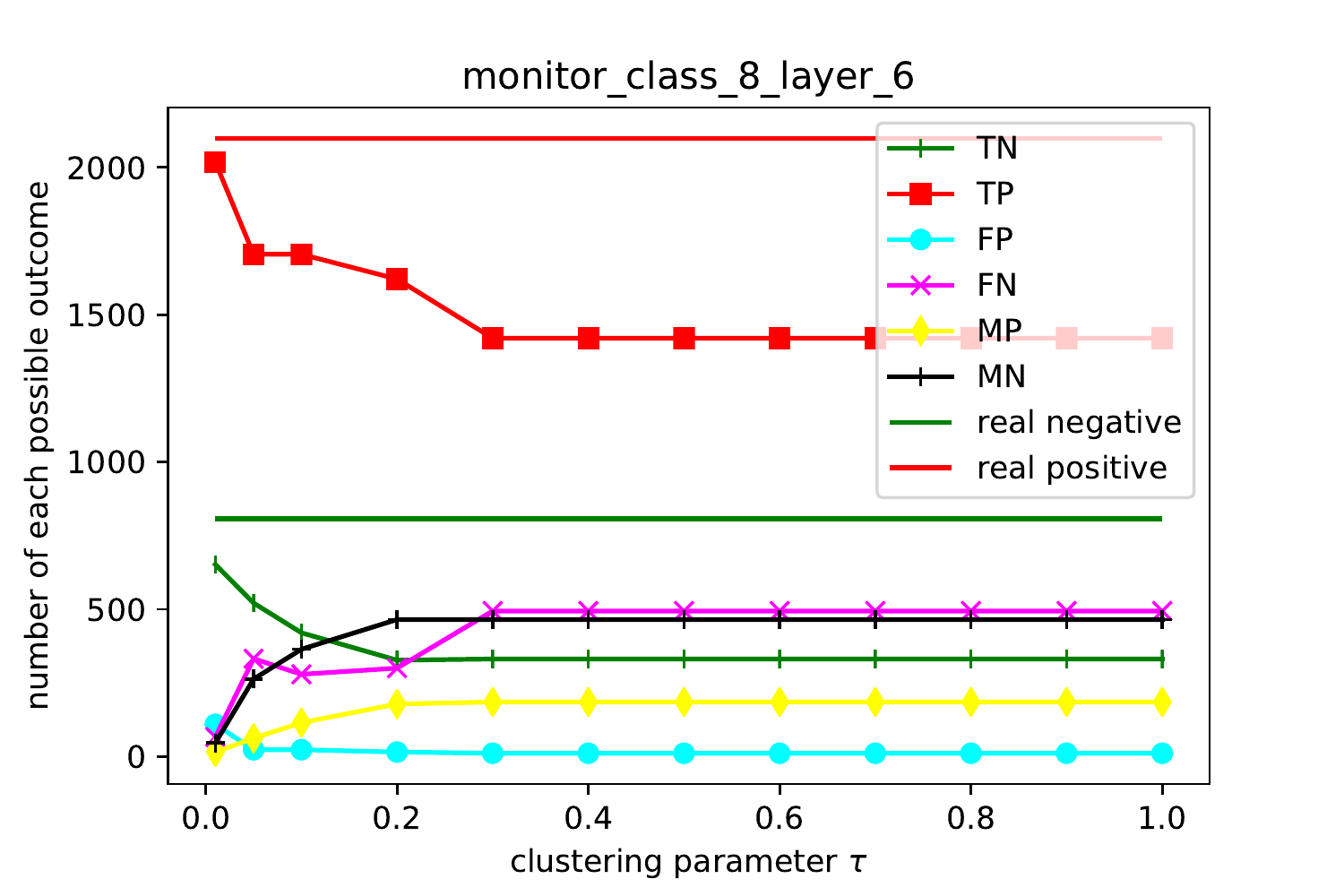}
    \end{subfigure}
    \hfill
    \begin{subfigure}[htbp]{0.245\textwidth}
        \centering
        \includegraphics[width=\textwidth]{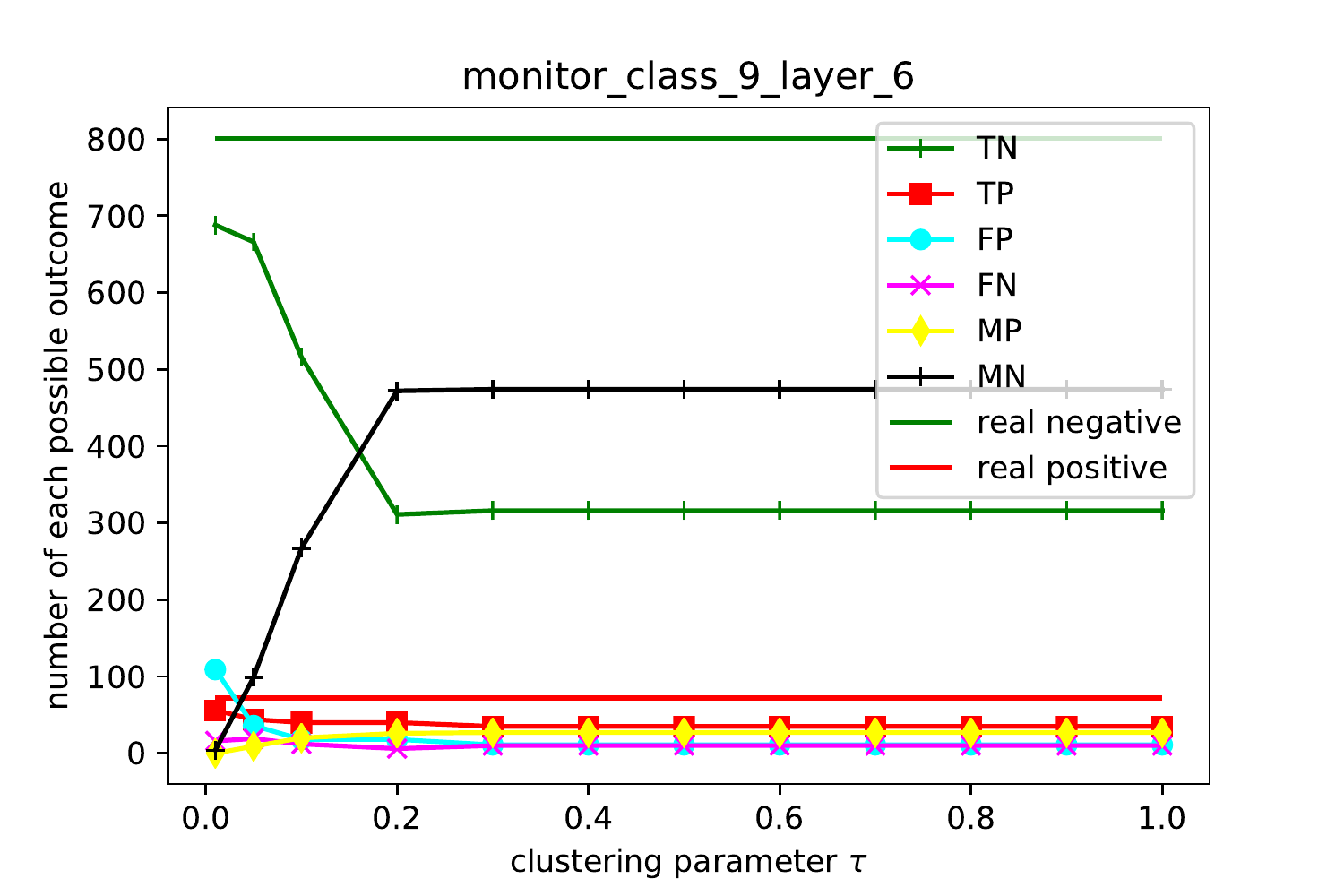}
    \end{subfigure}
    \caption{The numbers of different outcomes in Table~\ref{table:monitorPerformance} for $10$ monitors built at layer $6$ for benchmark F\_MNIST.}
    \label{fig:verdictsFMNIST}
\end{figure*}

\begin{figure*}[htbp]
    \centering
    \begin{subfigure}[htbp]{0.245\textwidth}
        \centering
        \includegraphics[width=\textwidth]{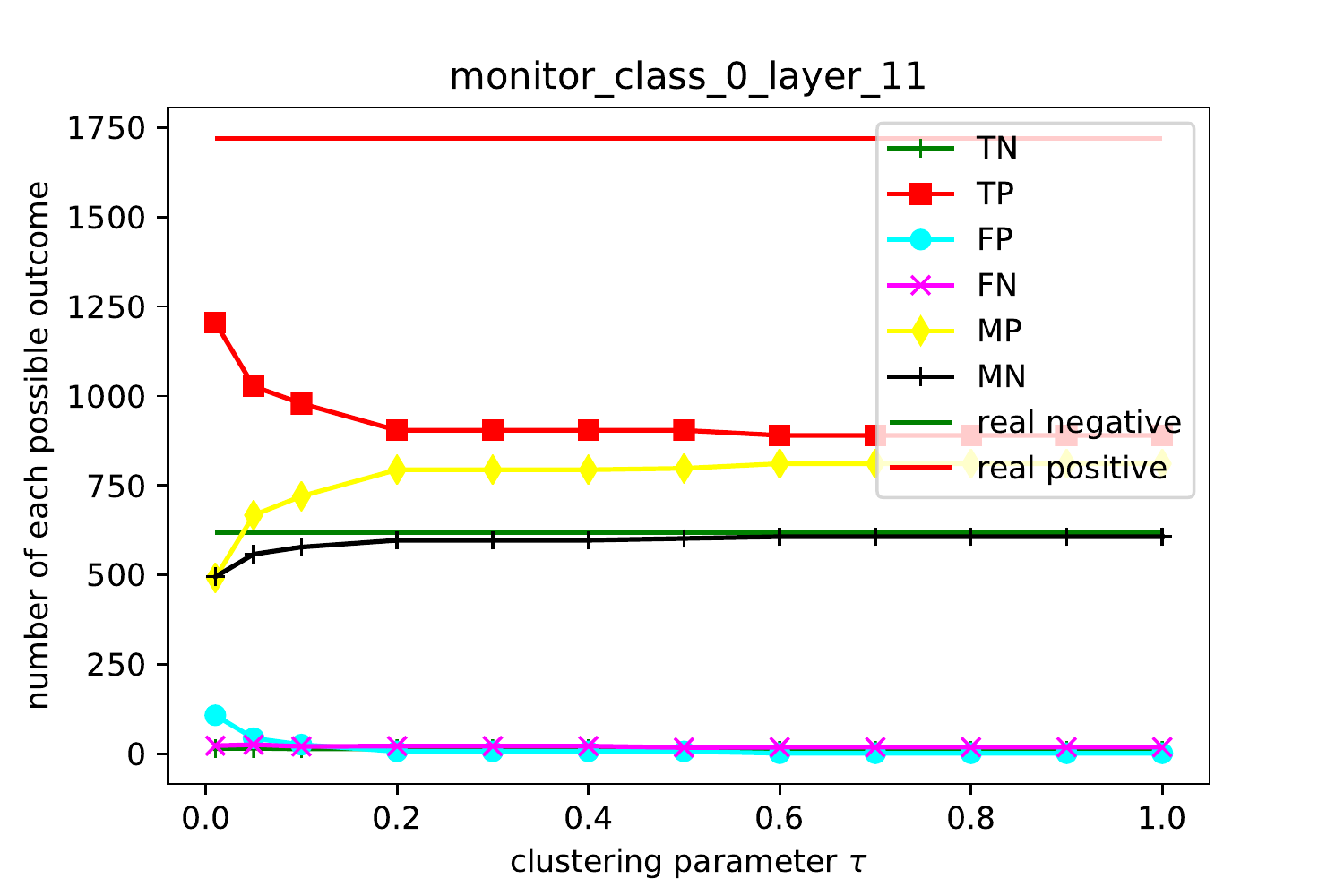}
    \end{subfigure}
    \hfill
    \begin{subfigure}[htbp]{0.245\textwidth}
        \centering
        \includegraphics[width=\textwidth]{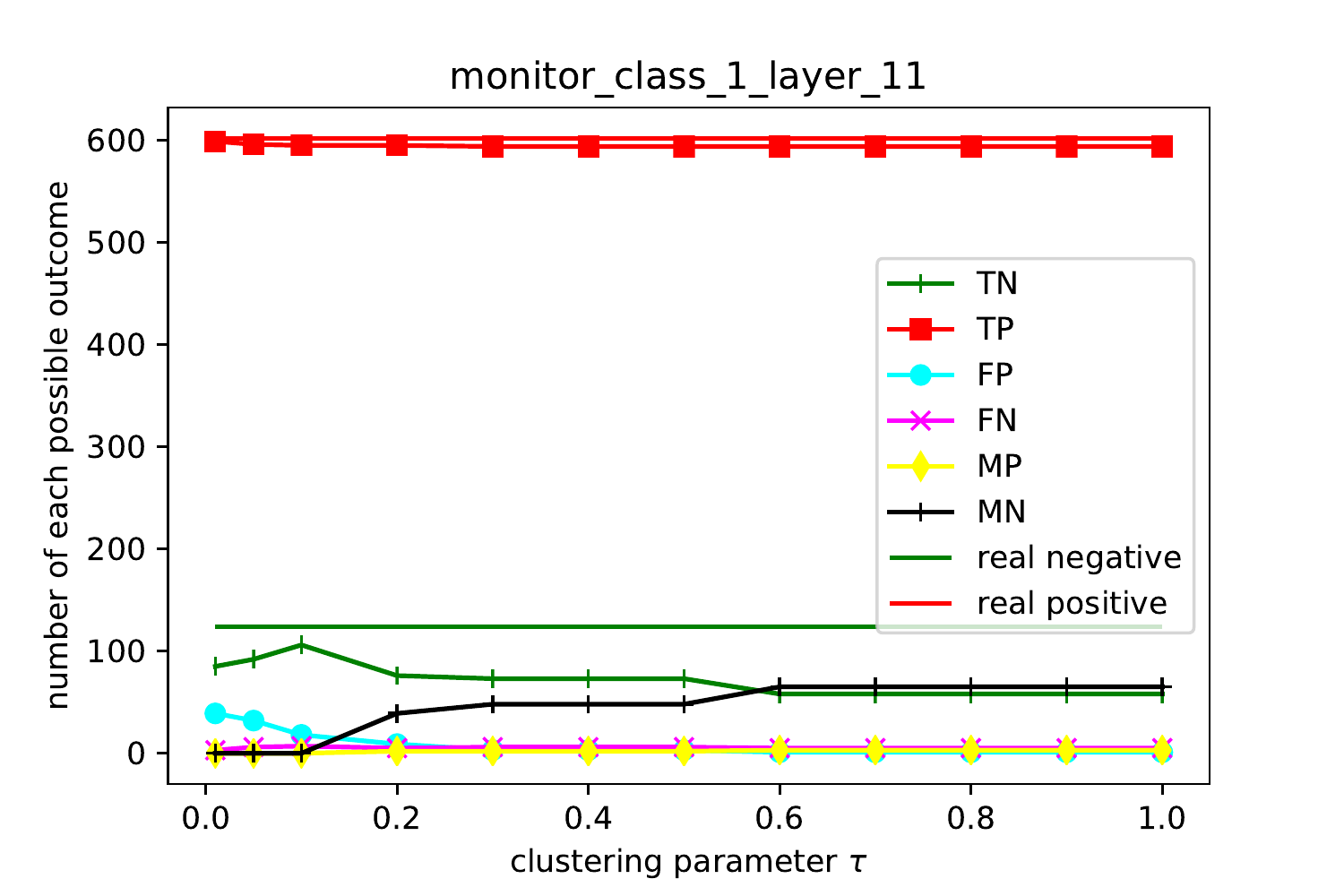}
    \end{subfigure}
    \hfill
    \begin{subfigure}[htbp]{0.245\textwidth}
        \centering
        \includegraphics[width=\textwidth]{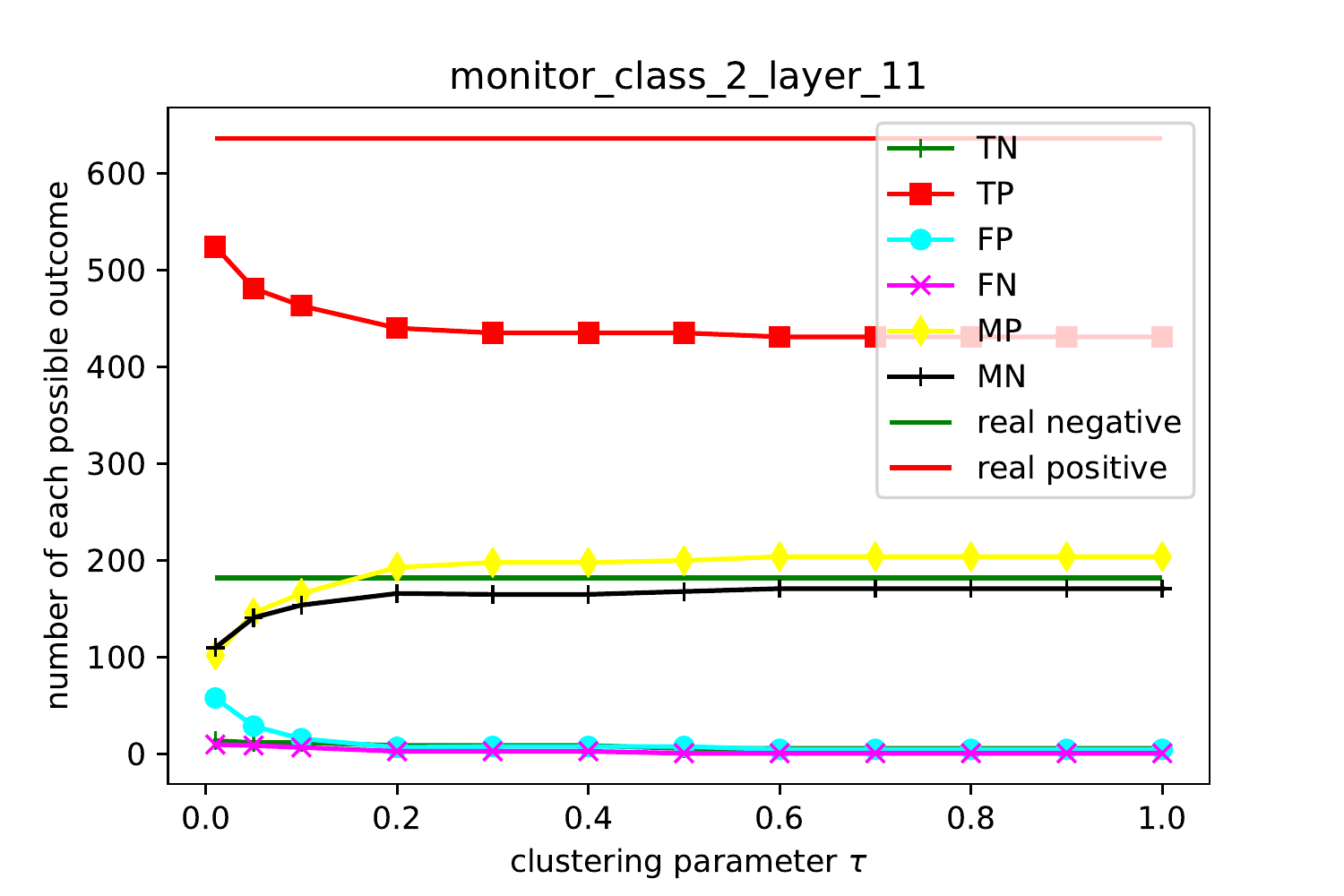}
    \end{subfigure}
    \hfill
    \begin{subfigure}[htbp]{0.245\textwidth}
        \centering
        \includegraphics[width=\textwidth]{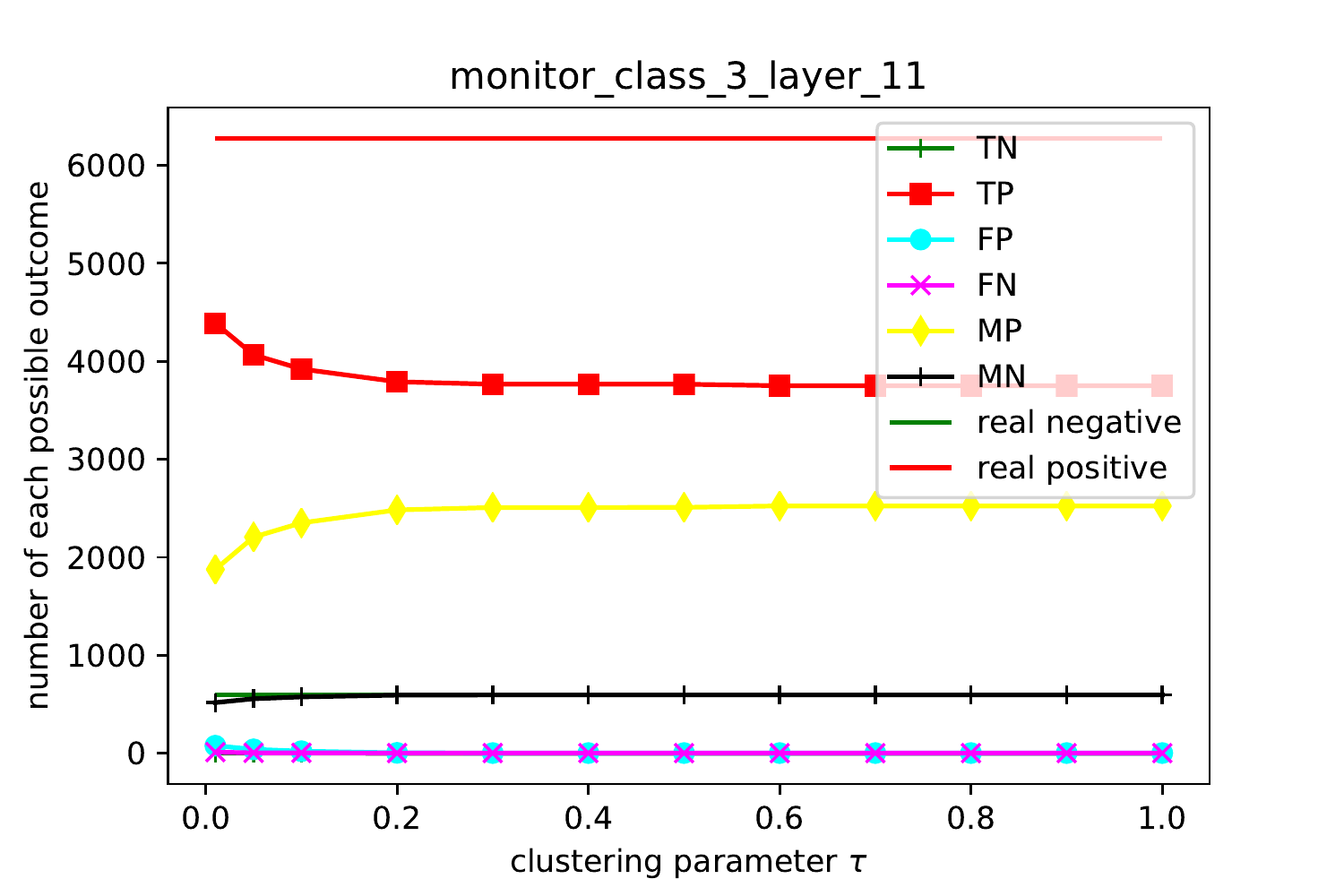}
    \end{subfigure}

    \begin{subfigure}[htbp]{0.245\textwidth}
        \centering
        \includegraphics[width=\textwidth]{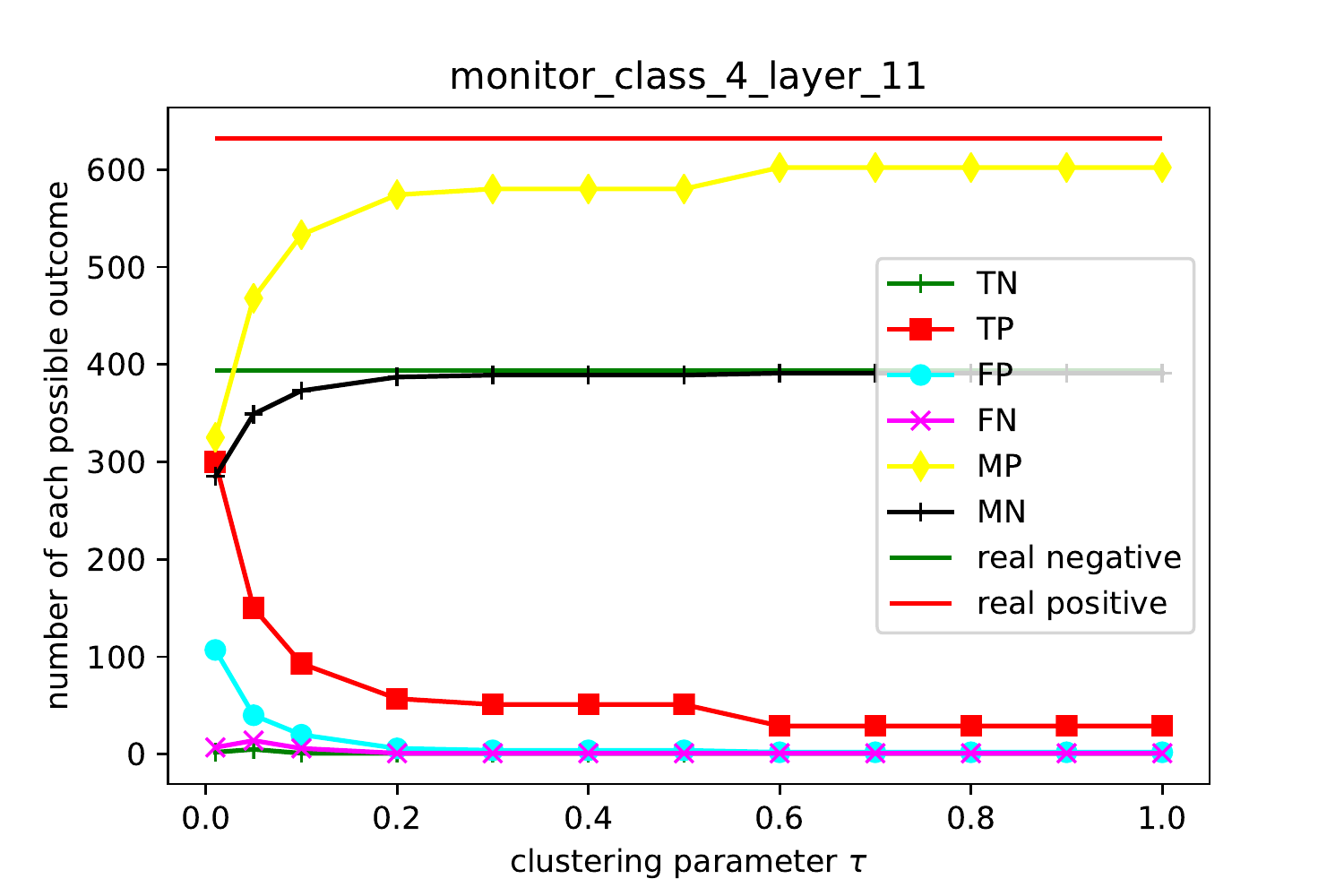}
    \end{subfigure}
    \hfill
    \begin{subfigure}[htbp]{0.245\textwidth}
        \centering
        \includegraphics[width=\textwidth]{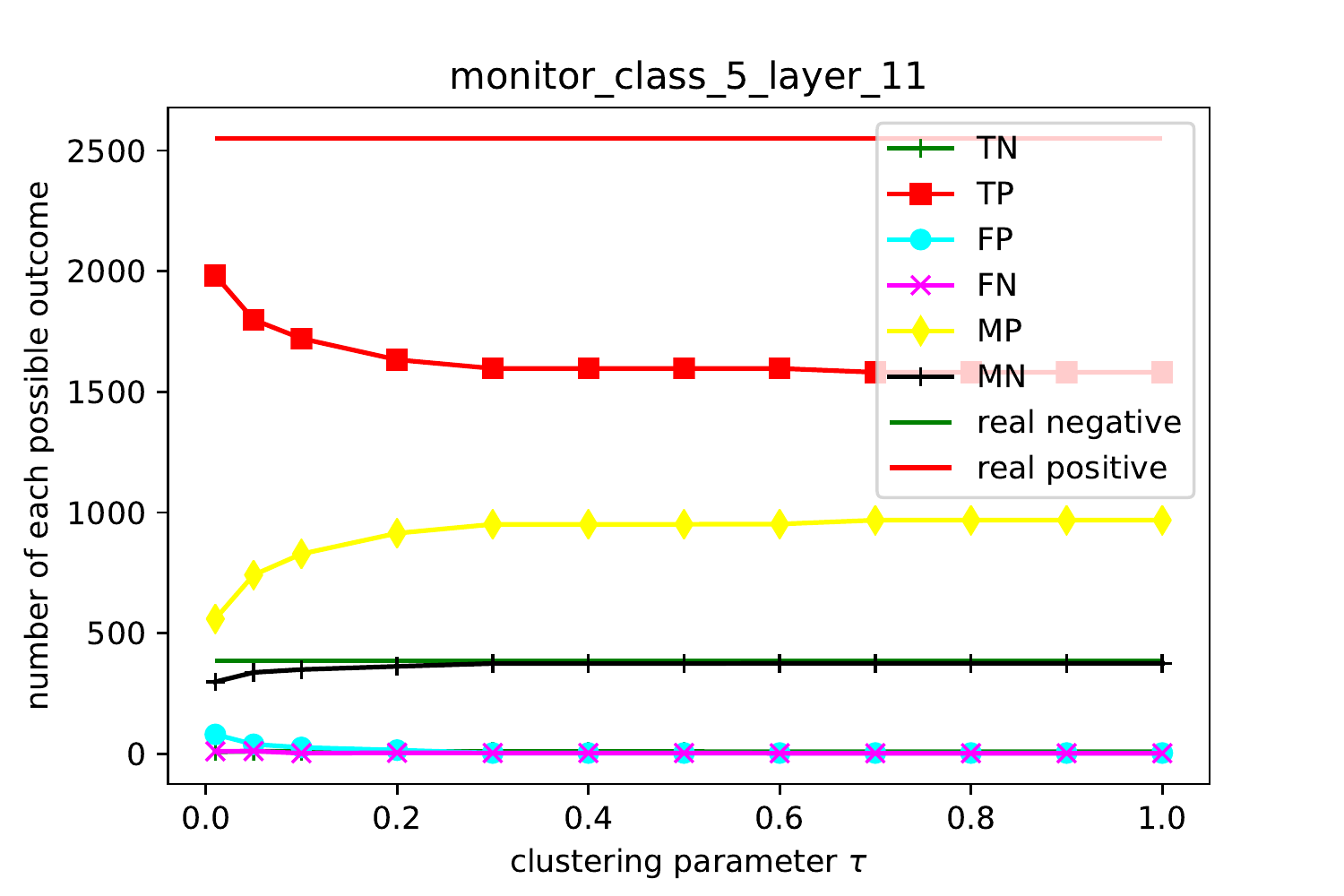}
    \end{subfigure}
    \hfill
    \begin{subfigure}[htbp]{0.245\textwidth}
        \centering
        \includegraphics[width=\textwidth]{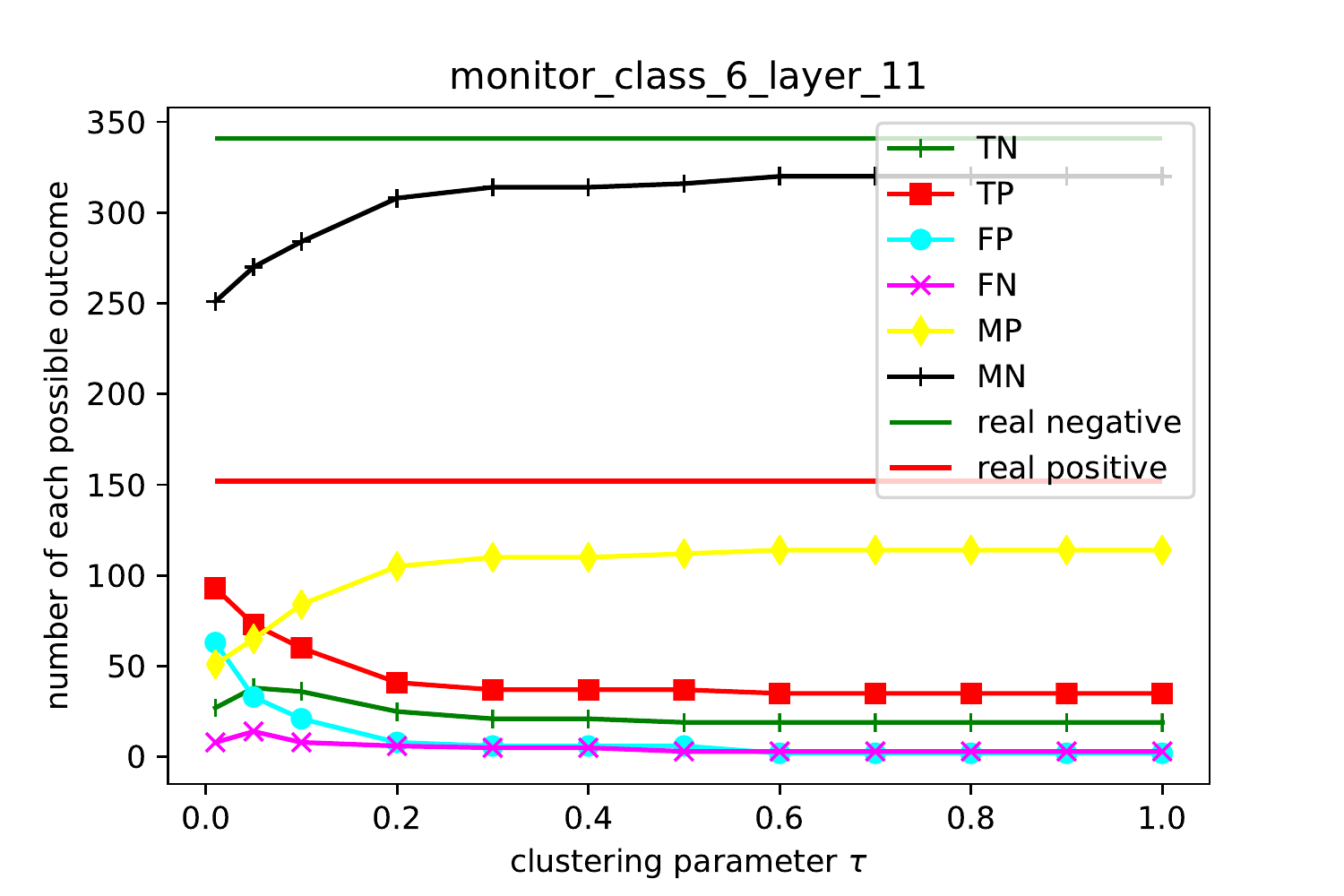}
    \end{subfigure}
    \hfill
	\begin{subfigure}[htbp]{0.245\textwidth}
        \centering
        \includegraphics[width=\textwidth]{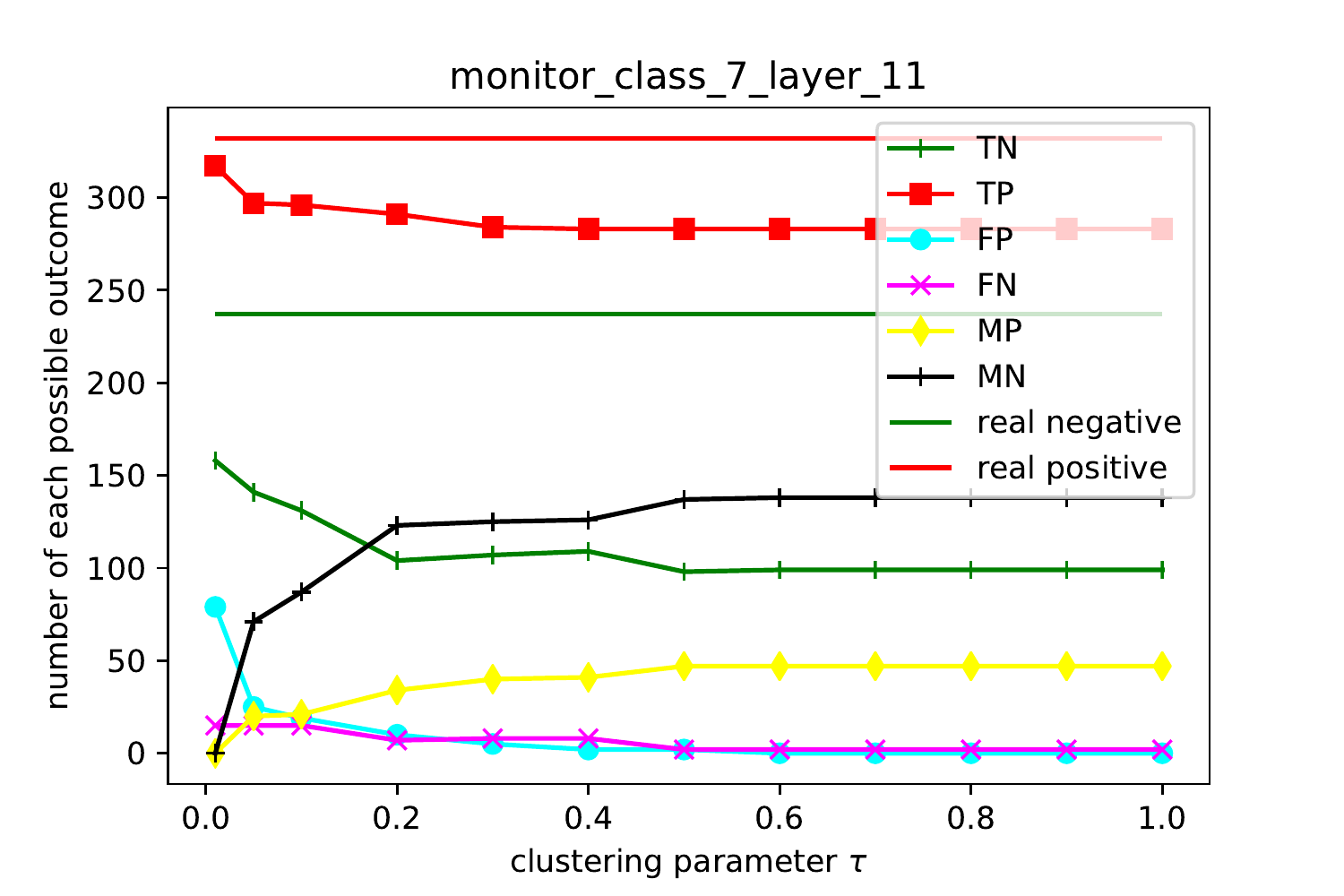}
    \end{subfigure}

    \begin{subfigure}[htbp]{0.245\textwidth}
        \centering
        \includegraphics[width=\textwidth]{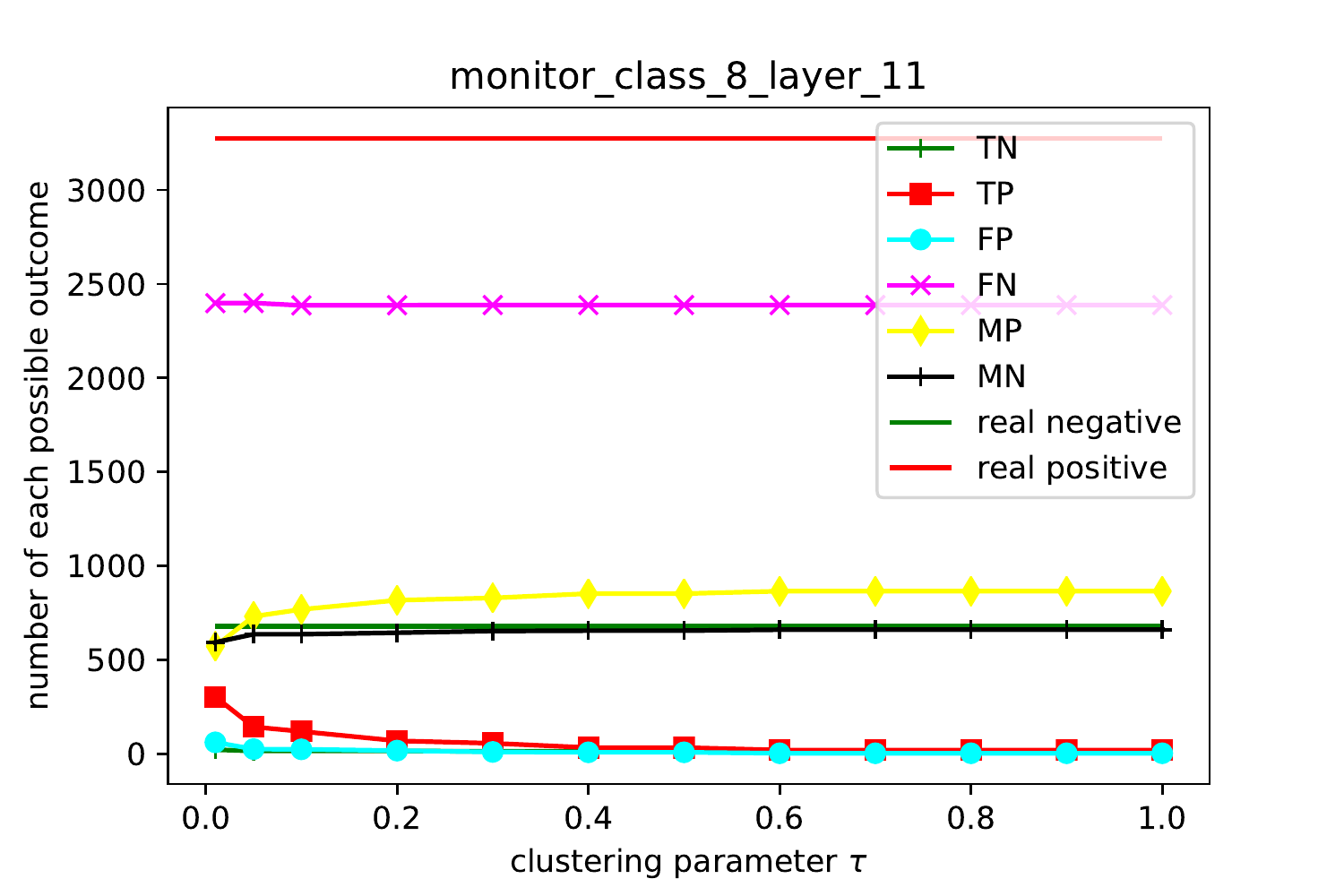}
    \end{subfigure}
    \hfill
    \begin{subfigure}[htbp]{0.245\textwidth}
        \centering
        \includegraphics[width=\textwidth]{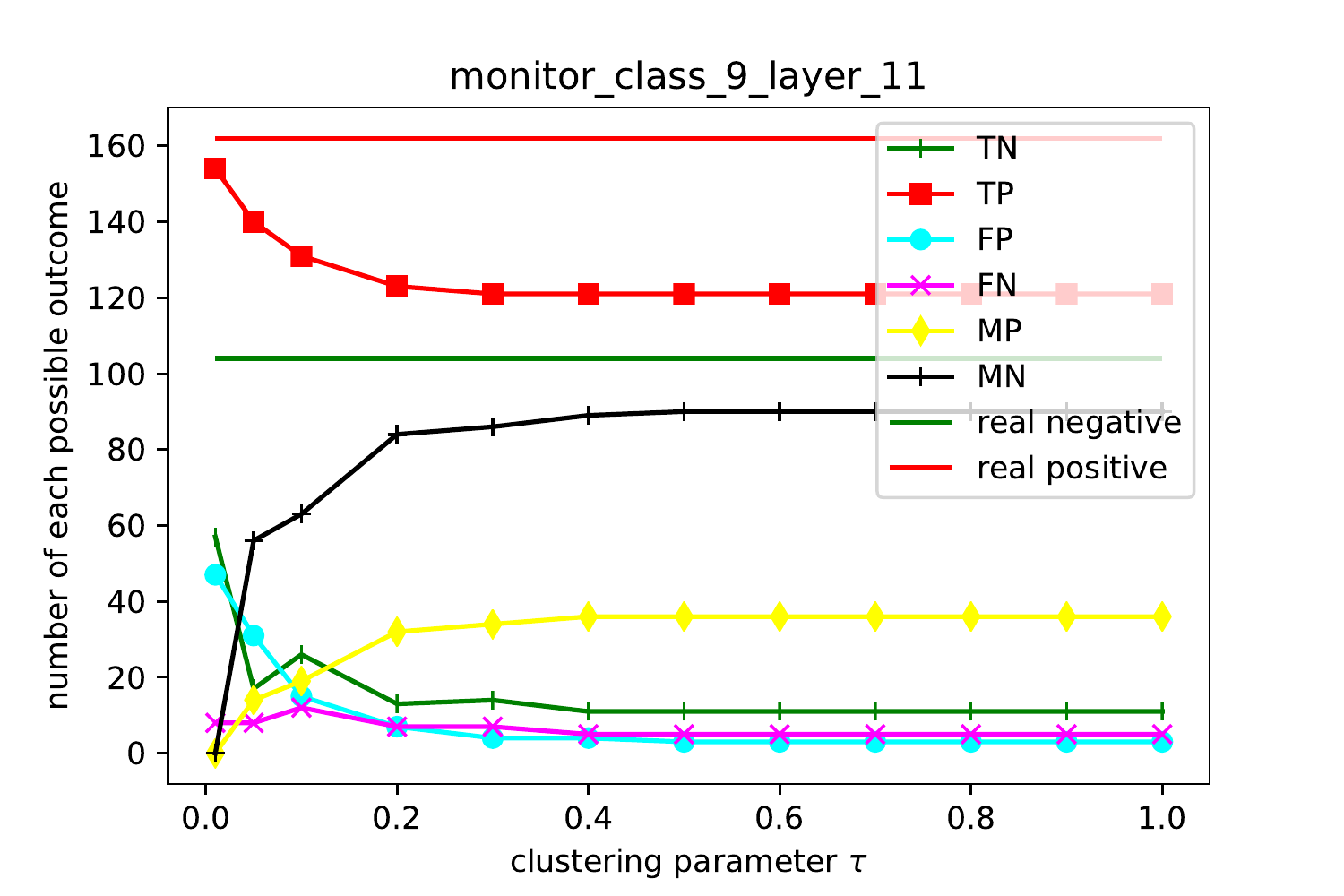}
    \end{subfigure}
    \caption{The numbers of different outcomes in Table~\ref{table:monitorPerformance} for $10$ monitors built at the output layer for benchmark CIFAR10.}
    \label{fig:verdictsCIFAR10}
\end{figure*}

\begin{figure*}[htbp]
    \centering
    \begin{subfigure}[htbp]{0.245\textwidth}
        \centering
        \includegraphics[width=\textwidth]{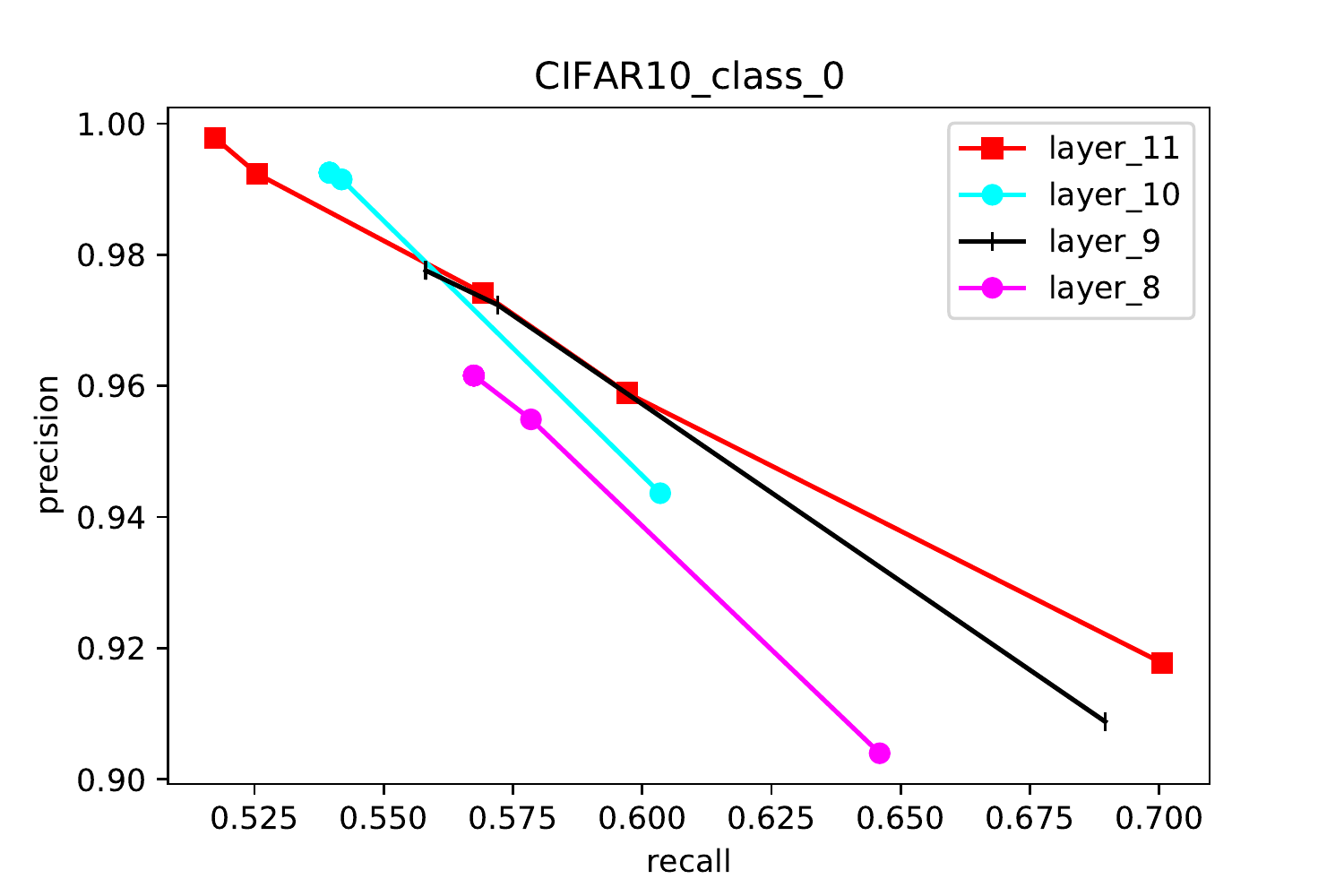}
    \end{subfigure}
    \hfill
    \begin{subfigure}[htbp]{0.245\textwidth}
        \centering
        \includegraphics[width=\textwidth]{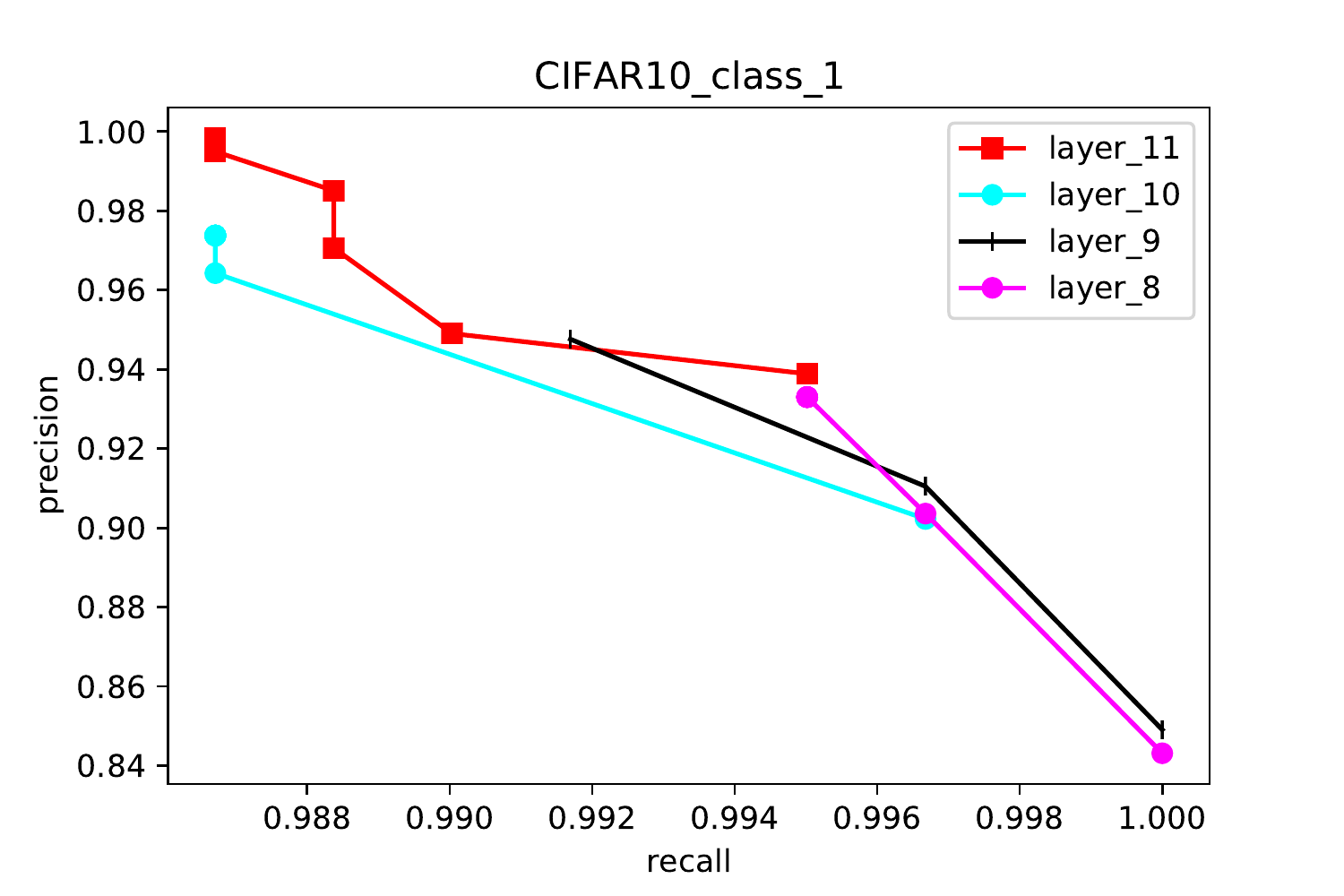}
    \end{subfigure}
    \hfill
    \begin{subfigure}[htbp]{0.245\textwidth}
        \centering
        \includegraphics[width=\textwidth]{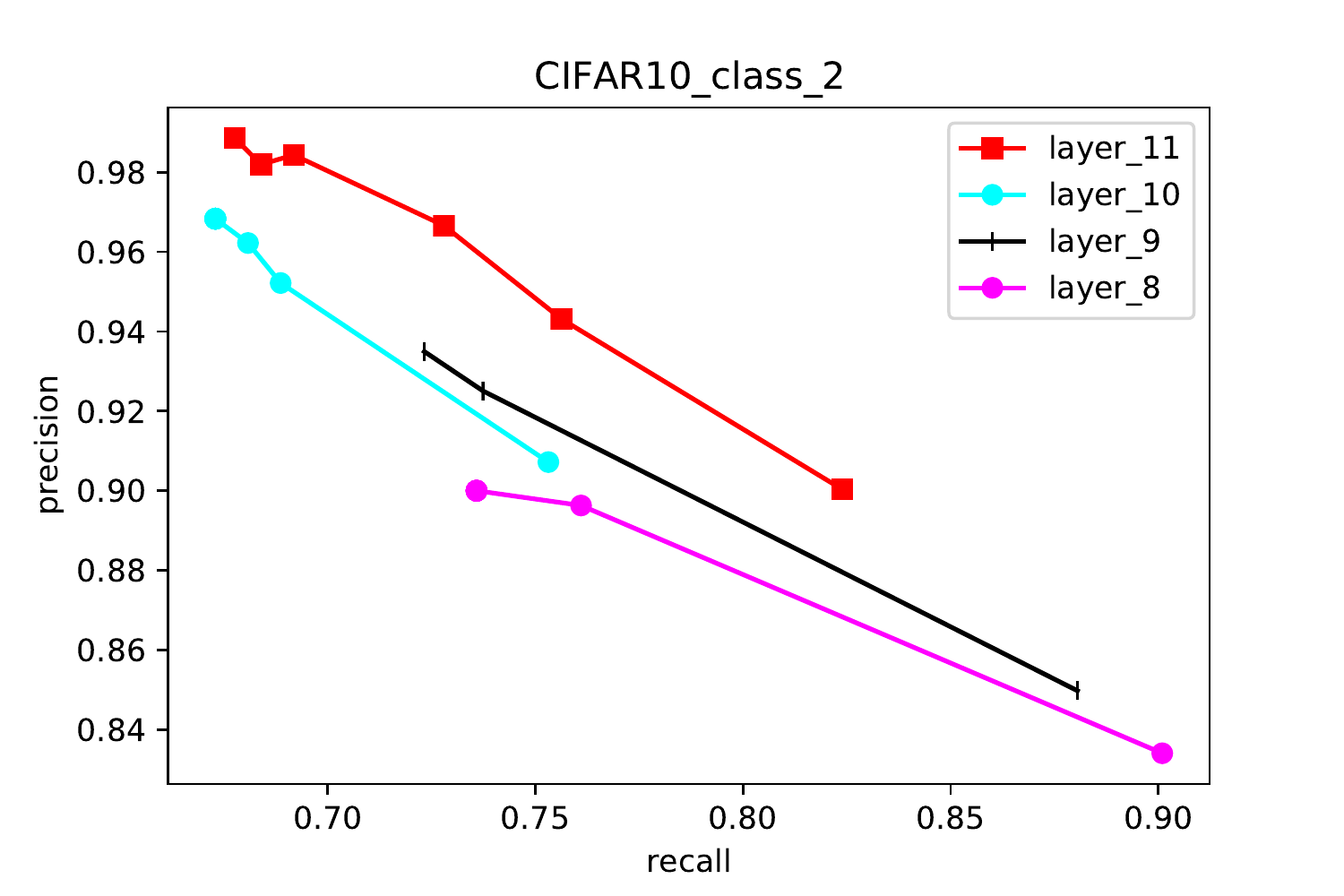}
    \end{subfigure}
    \hfill
    \begin{subfigure}[htbp]{0.245\textwidth}
        \centering
        \includegraphics[width=\textwidth]{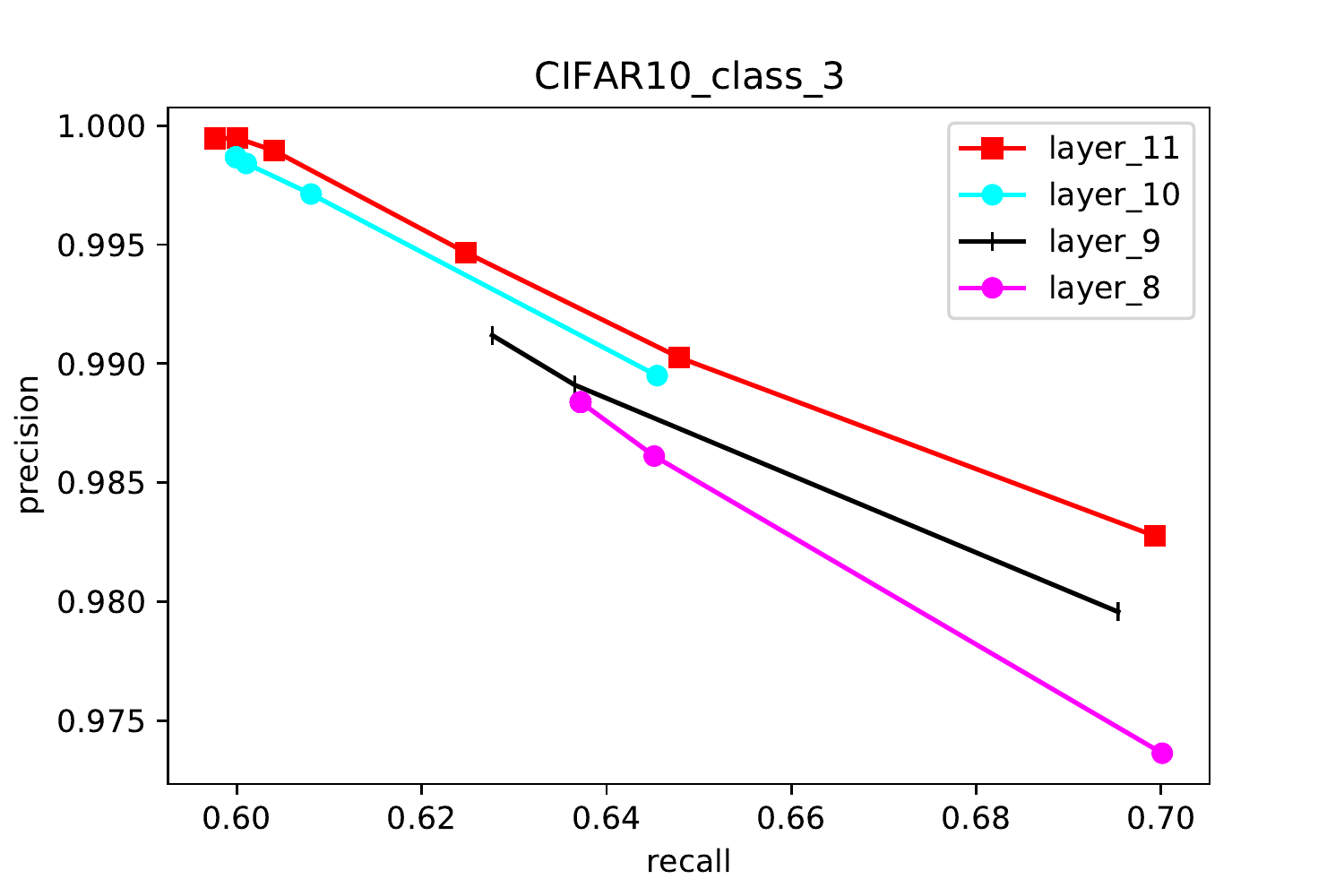}
    \end{subfigure}

    \begin{subfigure}[htbp]{0.245\textwidth}
        \centering
        \includegraphics[width=\textwidth]{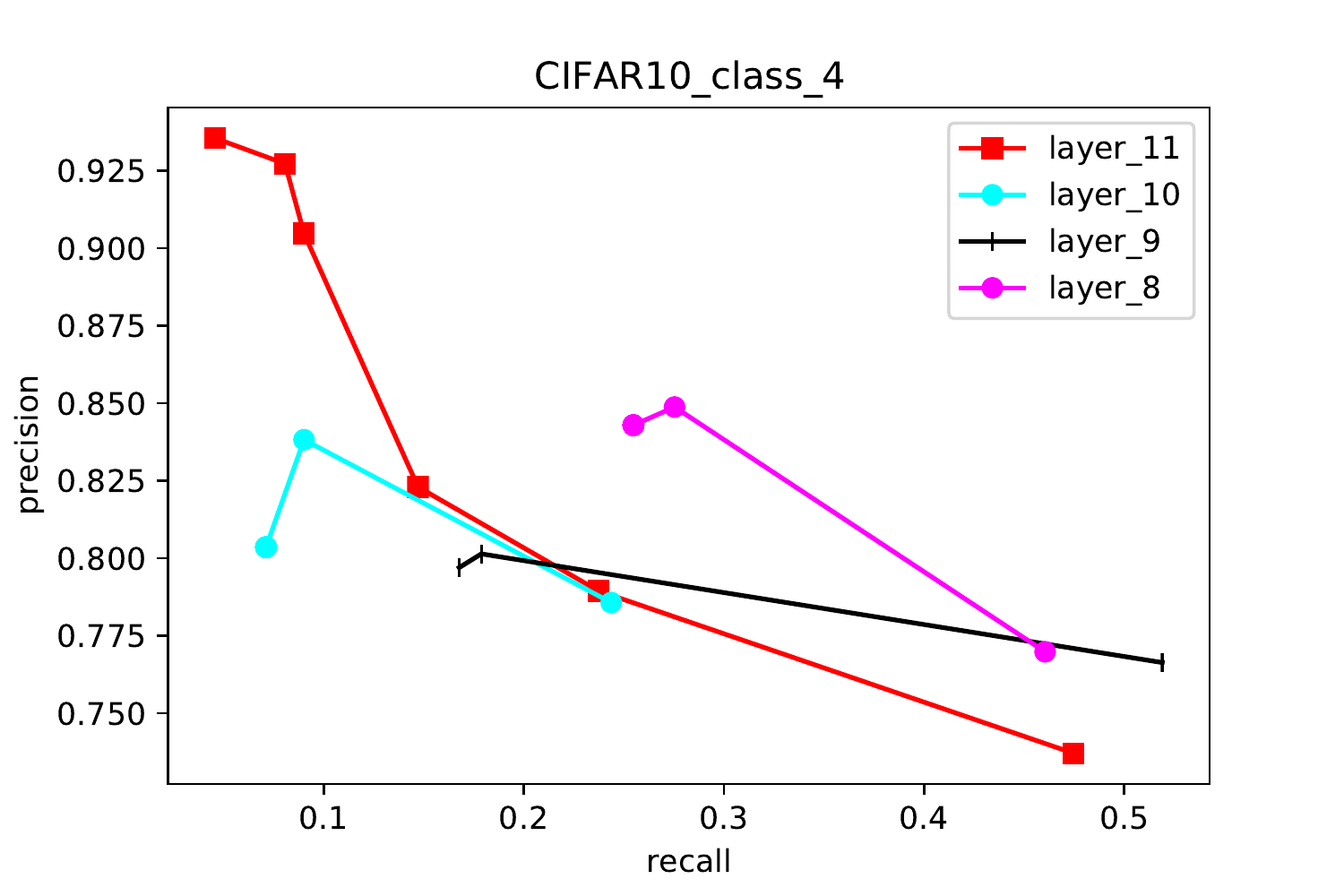}
    \end{subfigure}
    \hfill
    \begin{subfigure}[htbp]{0.245\textwidth}
        \centering
        \includegraphics[width=\textwidth]{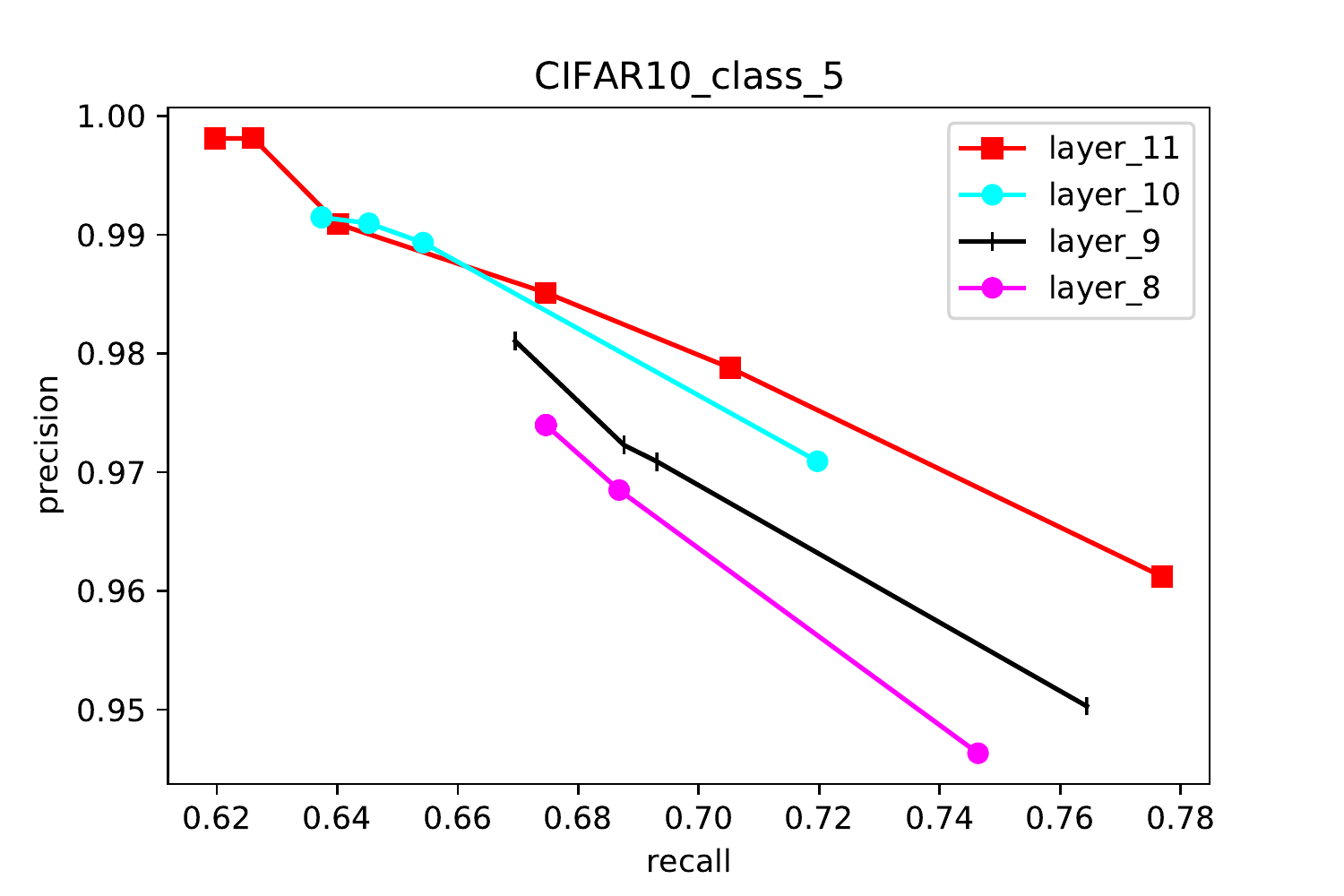}
    \end{subfigure}
    \hfill
    \begin{subfigure}[htbp]{0.245\textwidth}
        \centering
        \includegraphics[width=\textwidth]{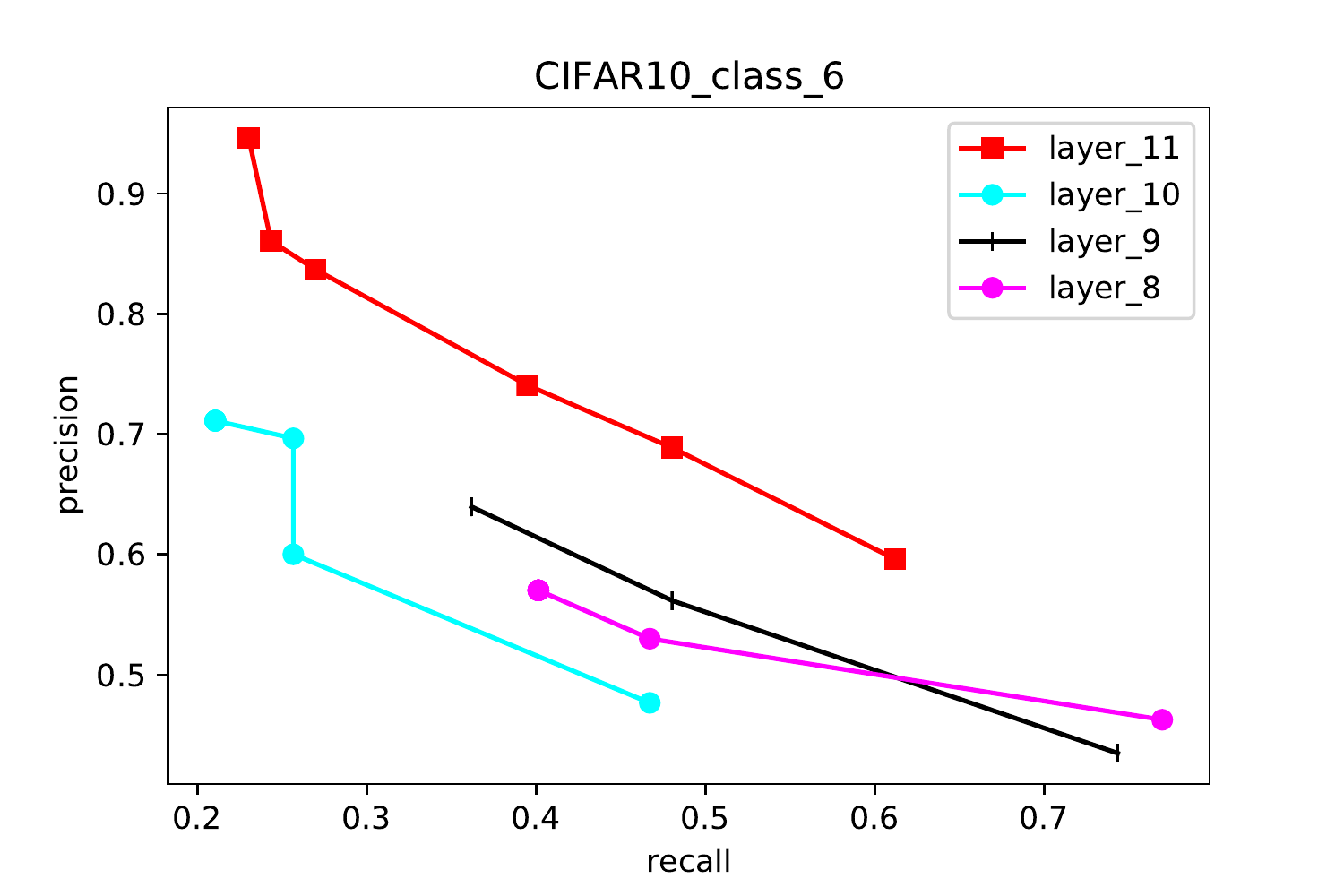}
    \end{subfigure}
    \hfill
    \begin{subfigure}[htbp]{0.245\textwidth}
        \centering
        \includegraphics[width=\textwidth]{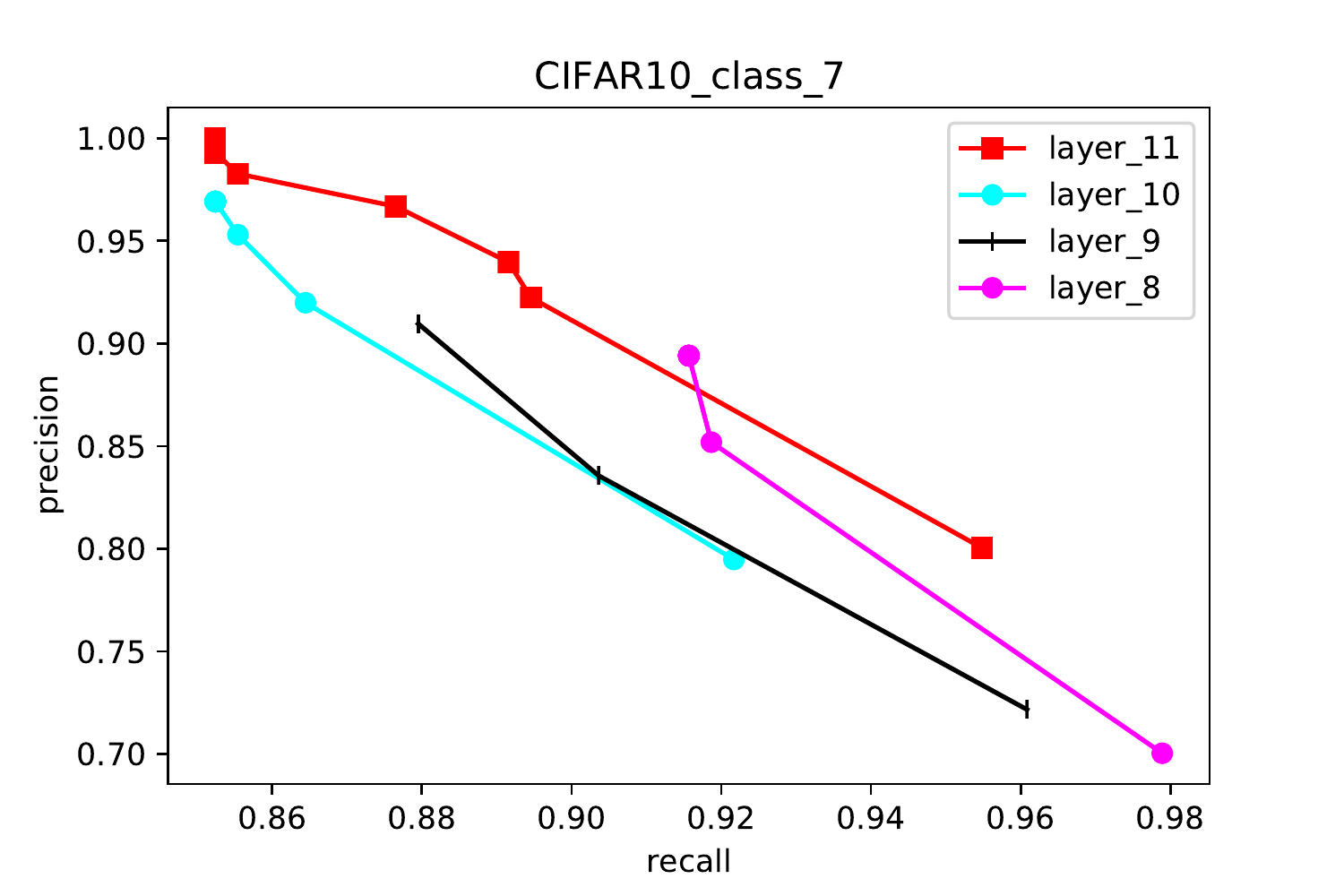}
    \end{subfigure}

    \begin{subfigure}[htbp]{0.245\textwidth}
        \centering
        \includegraphics[width=\textwidth]{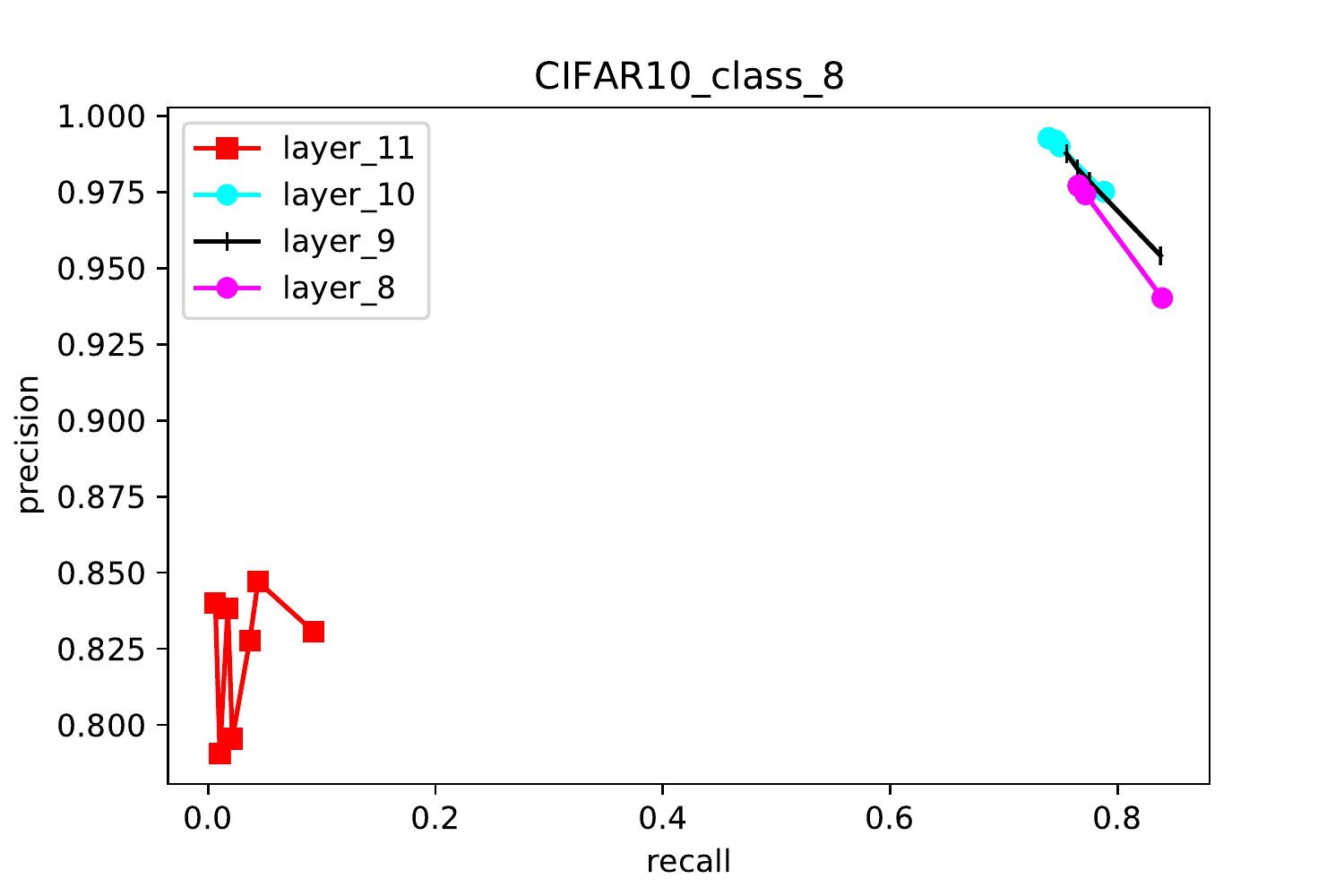}
    \end{subfigure}
    \hfill
    \begin{subfigure}[htbp]{0.245\textwidth}
        \centering
        \includegraphics[width=\textwidth]{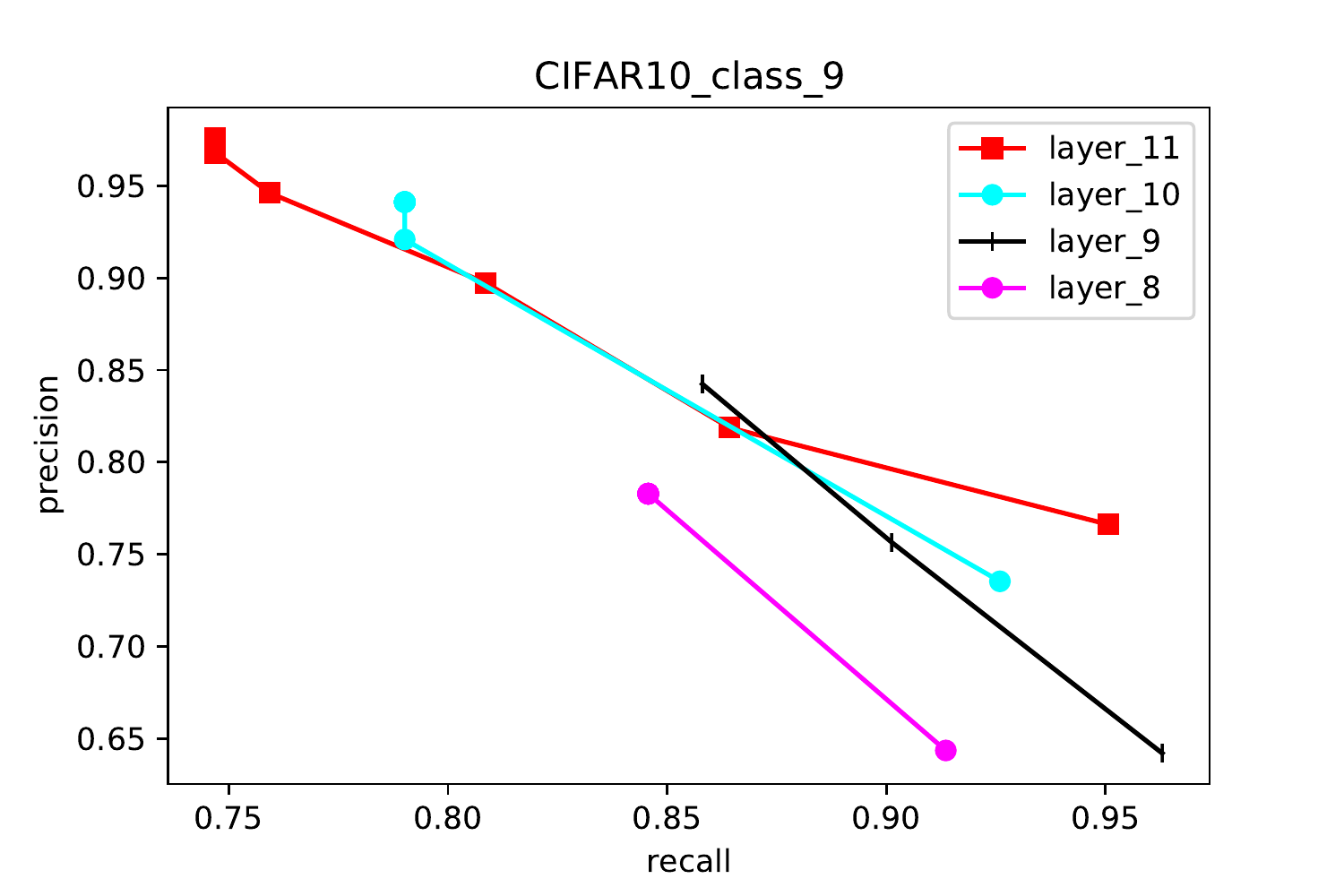}
    \end{subfigure}

    \caption{Precision-recall curves for monitors built on benchmark CIFAR10.}
    \label{fig:PRcurveCIFAR10}
\end{figure*}

\end{document}